\pdfoutput=1
\documentclass[11pt]{article}

\usepackage[preprint]{acl}

\usepackage{times}
\usepackage{latexsym}

\usepackage[T1]{fontenc}
\usepackage[utf8]{inputenc}

\usepackage{microtype}

\usepackage{inconsolata}

\usepackage{graphicx}
\usepackage{amsmath}
\usepackage{amsfonts}
\usepackage{booktabs}
\usepackage{multirow}
\usepackage{subcaption}
\usepackage{bm}
\usepackage{array} 
\usepackage{multirow}
\usepackage{placeins}
\usepackage{float}
\usepackage{CJKutf8}
\usepackage{hyperref}

\newcommand{\mypar}[1]{\vspace{2mm}\noindent\textbf{#1. }}

\newcommand{\steer}[1]{\textbf{\texttt{#1}}}
\newcommand{\DD}{\phantom{0}}

\title{Unveiling the Influence of Amplifying Language-Specific Neurons}

\author{%
Inaya Rahmanisa\textsuperscript{1}, Lyzander Marciano Andrylie\textsuperscript{1}, \bf Mahardika Krisna Ihsani\textsuperscript{2}, \\ \bf Alfan Farizki Wicaksono\textsuperscript{1}, Haryo Akbarianto Wibowo\textsuperscript{2}, \bf Alham Fikri Aji\textsuperscript{2}
 \\[1ex]
\textsuperscript{1}Faculty of Computer Science, Universitas Indonesia \\
\textsuperscript{2}Department of Natural Language Processing, MBZUAI \\
\texttt{\{inaya.rahmanisa, lyzander.marciano\}@ui.ac.id}, \texttt{alfan@cs.ui.ac.id} \\ \texttt{\{mahardika.ihsani,haryo.wibowo,alham.fikri\}@mbzuai.ac.ae}
}

\begin{document}
\maketitle
\begin{abstract}

Language-specific neurons, units in LLMs that strongly correlate with individual languages have been shown to influence model behavior by deactivating them. However, their role in amplification remains underexplored.
This work investigates the effect of amplifying language-specific neurons through interventions across 18 languages, including low-resource ones, using three models primarily trained in different languages. We compare amplification factors by their effectiveness in steering to the target language using a proposed Language Steering Shift (LSS) evaluation score, then evaluate it on downstream tasks: commonsense reasoning (XCOPA, XWinograd), knowledge (Include), and translation (FLORES). The optimal amplification steering factors effectively steer output toward nearly all tested languages. Intervention using this factor on downstream tasks improves self-language performance in some cases but generally degrades cross-language results. These findings highlight the effect of language-specific neurons in multilingual behavior, where amplification can be beneficial especially for low-resource languages, but provides limited advantage for cross-lingual transfer. \footnote{The code and dataset are made available at https://github.com/tauimbz/lang-task-neuron.}

\end{abstract}

\section{Introduction}

\begin{figure}[htbp]
    \centering
    \includegraphics[width=1.05\linewidth]{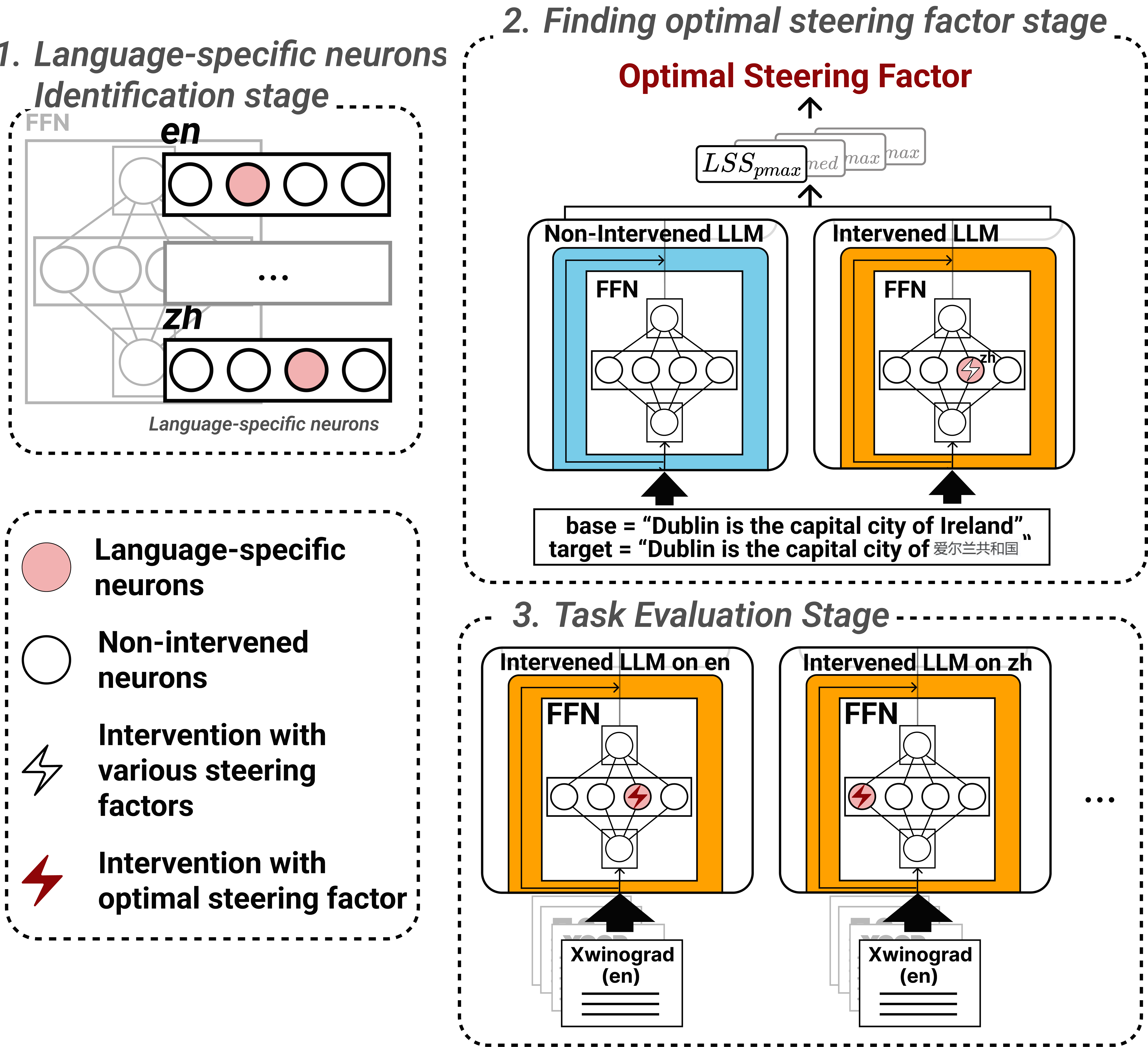}
    \caption{Illustration of our methodology. There are three sequential stages involving: 1). Identification of language-specific neurons, 2). Finding the optimal amplifying steering factor, and 3). Evaluation of the optimal amplifying steering factor on downstream tasks to understand its impact on the models' behavior.}
    \label{fig:method}
    \vspace{-15pt}
\end{figure}

Multilingual large language models (LLMs) have shown remarkable capabilities. However, they still continue to exhibit inconsistencies in downstream tasks (i.e., often producing biased or inaccurate outputs across different linguistic contexts), especially involving low-resource languages \cite{dang2024explainableinterpretablemultimodallarge}. Understanding the root cause of these inconsistencies remains challenging due to limited transparency in how LLMs represent language-specific features and perform cross-lingual transfer.

In elucidating this issue, a few recent studies have begun to interpret model inner workings in multilingual contexts, revealing neuron-level correlations with language features and suggesting the existence of language-specific neurons \cite{kojima-etal-2024-multilingual,tang2024languagespecificneuronskeymultilingual,zhao2024largelanguagemodelshandle}. 
These neurons have been shown to affect multilingual capability following deactivation and signs of language-steering capabilities in generation. Interventions are typically done to \textit{steer} outputs by replacing the neuron values with fixed activation values, called \textit{steering factor}, to control behavior. Most prior works use zero as the steering factor to deactivate targeted neurons. However, we argue that amplifying these neurons can offer deeper insight into their functional capabilities, particularly regarding performance improvements or task and language transfer. 

% These aspects remain unclear as existing studies have largely focused on self-language interventions \cite{mondal-etal-2025-language, wang2024sharingmattersanalysingneurons}, which have shown only limited gains.

% , though we argue that amplifying the neurons would provide better insights on these neurons' capabilities, particularly related to improving performance or enabling task and language transfer, which remains unclear as studies have focused only on self-language interventions \cite{mondal-etal-2025-language,wang2024sharingmattersanalysingneurons}, which showed limited gains. 

Building on the success of cross-lingual transfer methods such as training and fine tuning \cite{conneau-etal-2020-unsupervised,kim-etal-2019-effective}, we aim to investigate whether neuron overlap reflects shared linguistic representations. To this end, we extend existing studies that have been largely focusing on self-language interventions \cite{mondal-etal-2025-language, wang2024sharingmattersanalysingneurons} which showed limited gains, with cross-language interventions. We examine cross-lingual neuron behavior in closely related language pairs, distant language pairs, and low resource languages, to further explore whether neuron-level interventions can similarly provide insights on cross-lingual transfer, particularly revealing where and why such interventions succeed or fail.

This work is conducted by identifying and analyzing two types of neurons: language-specific neurons and language-activated (less-specific) neurons. We then compare various steering factors derived from different settings to evaluate their effect on the steerability of these neurons and analyze their impact on downstream task performance using neurons specific to the target language (self-intervention) and to every other languages (cross-intervention). We evaluate 18 languages including low-resource ones like Quechua, Haitian Creole, and Swahili, across models primarily trained in English and Chinese (Qwen), English (Gemma), and Southeast Asian languages (SeaLLMs) using XCOPA, XWinograd, and Include-lite for reasoning and knowledge tasks, and FLORES-200 for translation. 
% Interventions are performed using neurons specific to the target language (self-intervention) and to other languages (cross-intervention).
% Interventions are performed on target language neurons (self-intervention) and non-target languages neurons (cross-intervention). 

Several significant findings from this study are as follows:
\vspace{-10pt}
\begin{itemize}
    \item Comparison between different steering factors using the proposed Language Steering Shift (LSS) reveals that patched steering factors are the most effective and statistically outperform test-time intervention factors. Amplifying language-specific neurons with these factors consistently directs outputs toward all target languages, achieving over 90\% average success in language shifting.
    \vspace{-10pt}
    \item Optimal amplification of language-specific neurons in self-intervention can reduce perplexity and improve task performance for certain languages, with gains reaching up to $1\%$ in some models, showing improved results for low-resource languages compared to \citet{mondal-etal-2025-language}.
    \vspace{-10pt}
    \item Cross-lingual amplification can slightly reduce perplexity and enhance performance between closely related languages, but it generally leads to performance degradation in others, indicating limited effectiveness in cross-lingual transfer.
\end{itemize}

\section{Related Work}

Recent studies have explored neuron activations in language \cite{kojima-etal-2024-multilingual,tang2024languagespecificneuronskeymultilingual,zhao2024largelanguagemodelshandle}, task \cite{leng2025understandingmultitasklearninggeneralization,song-etal-2024-large}, and mixed contexts \cite{wang2024sharingmattersanalysingneurons}, often by inhibiting neurons to assess their roles. \citet{kojima-etal-2024-multilingual} showed that intervening on language-specific neurons can steer output language. Common intervention methods include activation patching \cite{zhang2024bestpracticesactivationpatching}, model editing  \cite{meng2023locatingeditingfactualassociations}, and suppression via zeroing or percentile thresholds \cite{mondal-etal-2025-language}. While \citet{mondal-etal-2025-language} has explored self-language intervention on downstream tasks, to the best of our knowledge, no study has yet focused on comparing amplification factors and provided improving insights into self- and cross-lingual interventions in downstream tasks.

\section{Methodology}

This section describes our method for identifying language-specific and language-activated neurons, introduces the Language Steering Shift (LSS) score to quantify language steering, and details our approach for determining the steering factors.
\subsection{Neuron Identification}
As illustrated in Figure \ref{fig:method}, initially, we define neurons as the outputs of the activation function within the FFN modules in LLMs as they have been found to store knowledge or features \cite{dai-etal-2022-knowledge, tang2024languagespecificneuronskeymultilingual}. 
% After the important neurons are identified, we search for the optimal steering factors and assess their impact on downstream tasks.

\paragraph{Identifying Language-Activated Neurons.}

As a baseline method, we adapt a method from \citet{wang2024sharingmattersanalysingneurons} to define language-activated neurons, i.e., neurons whose activation value exceeds 0 throughout every sentence $s$ in the dataset of language $k$, denoted $S_k$, allowing a single neuron to be language-activated for multiple languages. The set of language-activated neurons in layer $i$ for language $k$, denoted by $A_k(i)$, is described by:
\[
A_k(i) = \left\{ j \;\middle|\; \texttt{act}(i, j; s) > 0, \;\; \forall s \in S_k \right\} \, .
\]
For neuron $j$, $\texttt{act}(i,j;s)$ is the activation value of the $j$-th neuron in the $i$-th layer when processing sentence $s$. These neurons are not further filtered, resulting in relatively large number of neurons and overlapping activations across languages. We refer to these as \textit{Baseline neurons}.

\paragraph{Language-Specific Neurons.}

\textit{language activation probablity entropy} (LAPE) method~\citep{tang2024languagespecificneuronskeymultilingual} is adopted to identify language-specific neurons.
The probability of activation for the $j$-th neuron within the $i$-th layer when processing language $k$, denoted by $p^{k}_{i,j}$, is estimated according to:
\begin{equation*} \label{eq-4} 
\hat{p}^{k}_{i,j} = \mathbb{E} \left[ \mathbb{I}(\texttt{act}({i,j}) > 0) \mid \text{language } k \text{ tokens} \right] \, .
\end{equation*}
\noindent
The expectation is taken over all tokens from sentences in language $k$, where $\mathbb{I}(.)$ is an indicator function that returns $1$ if the condition is true and $0$ otherwise; and $\texttt{act}(i,j)$ denotes the activation value of the $j$-th neuron in the $i$-th layer.
Let $\mathbf{p}_{i,j} = [p^{1}_{i,j}, \dots, p^{k}_{i,j}, \dots, p^{|L|}_{i,j}]$ represent a list of activation probabilities for a set of languages $L$. These probabilities are then normalized to yield the L1-norm version, denoted by $p'_{i,j}$.
Using this normalized version, the LAPE score of the $j$-th neuron in the $i$-th layer is formally defined as:
\begin{equation*} \label{eq-5}
    \text{LAPE}_{i,j} = -\sum_{k=1}^{|L|} p'^{k}_{i,j} \log (p'^{k}_{i,j}) \,.
\end{equation*}
We filter neurons whose language activation probability exceeds the $m$-th percentile. Finally, we select neurons that fall within the bottom $n\%$ of LAPE scores.

\subsection{Language Steering Shift (LSS)} 

We propose \textbf{Language Steering Shift (LSS)} score to quantify the extent to which same-meaning answers are steered to the target language.
Let $L$ be the set of languages used. For every language \( k \in L \), we evaluate the dataset for language $k$
% in the reconstructed MLAMA 
by performing: 1) a no intervention scenario, and 2) an intervention scenario where neurons corresponding to language $l$ are intervened where $l \in L \setminus \{k\}$. The LSS score for evaluating language $k$ when intervened upon by neurons specific to language $l$, denoted by $\texttt{LSS}(k, l)$, is calculated as follows:
\begin{align*}
\texttt{LSS}(k, l) &= \mathbb{E}\left[\mathbb{I}\left(\delta(k, l)|_{\texttt{int}}-\delta(k, l)|_{\texttt{non}} > 0\right)\right]  \, ,
\\
\delta(k, l) &= \log p(\texttt{ans}_l) - \log p (\texttt{ans}_k) \, ,
\end{align*}
where $\log p(\texttt{ans}_x)$ denotes the log-probability assigned by the model to the answer in language $x$; the symbols $|_{\texttt{non}}$ and $|_{\texttt{int}}$ represent the no intervention and intervention scenario, respectively; $\mathbb{E}[.]$ is the expectation value across all sentences in the dataset.\begin{CJK}{UTF8}{gbsn} Intuitively, this metric measures the tendency of a model to produce the correct answer in target language $k$ given an incomplete sentence written in source language $l$ after intervening neurons corresponding to language $k$. For example, given the English sentence ``Dublin is the capital city of \underline{Ireland}'', the LSS score measures how much it shifts toward the Chinese translation, ``Dublin is the capital city of \underline{爱尔兰共和国}'', when intervened using \texttt{zh} neurons. The underlined words represent the answers being analyzed. \end{CJK}

\subsection{Determining steering factors} 

We steer language-specific neurons in each layer with amplifying steering factors, which were determined in the identification stage (patched) and test-time intervention stage.

\mypar{Patched Steering Factors} 
Let $v_{j,k}$ be the variable that contains the activation value of neuron $j$ that is specific to language $k$. In the identification stage involving the multilingual FLORES-200 dataset, the steering factors are computed from the max and median of activation values across all sentences within the dataset $S_k$ for language $k$. As a result, a steering factor is specific to neuron $j$ and language $k$, to which we patch the language-specific neurons.
This process is formally described as:
\[
v_{j, k} \leftarrow \texttt{agg} \{s \in S_k | a_{j,s}\} \, ,
\]
where $\texttt{agg} \in \{\texttt{max}, \texttt{median}\}$; and $a_{j,s}$  represents the activation value of neuron $j$ averaged across all tokens in sentence $s$ during the identification stage. We follow \citet{kojima-etal-2024-multilingual} in using the median and extend their approach by incorporating a stronger aggregation function, max. We refer to these steering factors as patch max (\steer{pmax}) and median (\steer{pmedian}).

\mypar{Test-Time Intervention Steering Factors}
Let $v_{j,k,m}$ be the variable containing the activation of neuron $j$ for language $k$ when processing sentence $m$, i.e., a sentence from the test dataset. We define the steering value as the maximum over activation values within the sentence. Using this value, we apply two operations to the language-specific neurons: addition (\steer{+max}) to assign with a value surpassing its activation range, and replacement  (\steer{=max}) to assign the maximum activation. This computation is formally represented as:
\[
\texttt{steer}_{j,m} \leftarrow \texttt{max} \{t \in T(m) | a_{j,m,t}\} \, ,
\]
\[
\steer{=max} \textrm{   :   } v_{j,k,m} \leftarrow   \texttt{steer}_{j,m}\, ,
\]
\[
\steer{+max} \textrm{   :   } v_{j,k,m} \leftarrow v_{j,k,m} + \texttt{steer}_{j,m} \, ,
\]
\noindent
where $\texttt{steer}_{j,m}$ is the steering factor for neuron $j$ during inference on sentence $m$; $T(m)$ is the set of tokens in sentence~$m$; 
and $a_{j,m,t}$ is the activation value of neuron $j$ associated with token $t$ on sentence $m$.
%\texttt{agg} is the \texttt{max} function, aggregating over tokens $t$ and neurons $m$. 
As a comparison to our amplifying factors, we also use deactivating factors $0$ and $10p$ (10th percentile) as a replacement of \texttt{max}, which has been similarly
demonstrated by \citet{mondal-etal-2025-language} to deactivate neurons. We refer to these as \steer{=0} and \steer{=10p}, respectively.

\begin{table*}[ht]
\centering
\small
\renewcommand{\arraystretch}{1.5}
\setlength{\tabcolsep}{2pt}

\resizebox{\textwidth}{!}{
\begin{tabular}{
|l
|*{6}{c}|
*{6}{c}|
*{6}{c}|
}
\hline
\multirow{2}{*}{} & 
\multicolumn{6}{c|}{\textbf{Qwen}} & 
\multicolumn{6}{c|}{\textbf{Gemma}} & 
\multicolumn{6}{c|}{\textbf{SeaLLM}} \\
\cline{2-19}
& \textbf{\scriptsize pmax\,$\uparrow$} & \textbf{\scriptsize pmed\,$\uparrow$} & \textbf{\scriptsize =max\,$\uparrow$}  & \textbf{\scriptsize +max\,$\uparrow$}  & \textbf{\scriptsize =0\,$\downarrow$}  & \textbf{\scriptsize =10p\,$\downarrow$}
& \textbf{\scriptsize pmax\,$\uparrow$} & \textbf{\scriptsize pmed\,$\uparrow$} & \textbf{\scriptsize =max\,$\uparrow$}  & \textbf{\scriptsize +max\,$\uparrow$}  & \textbf{\scriptsize =0\,$\downarrow$}  & \textbf{\scriptsize =10p\,$\downarrow$}
& \textbf{\scriptsize pmax\,$\uparrow$} & \textbf{\scriptsize pmed\,$\uparrow$} & \textbf{\scriptsize =max\,$\uparrow$}  & \textbf{\scriptsize +max\,$\uparrow$}  & \textbf{\scriptsize =0\,$\downarrow$}  & \textbf{\scriptsize =10p\,$\downarrow$} \\
\hline
en & \textbf{99.46} & 98.92 & 96.22 & 96.22 & 84.32 & 33.51 & \textbf{100.0} & \textbf{100.0} & \textbf{100.0} & \textbf{100.0} & \textbf{100.0} & \DD9.73 & \textbf{100.0} & \textbf{100.0} & 94.59 & 94.05 & 98.92 & \DD5.41 \\ nl & \textbf{100.0} & \textbf{100.0} & 92.97 & 90.81 & 67.57 & 32.97 & \textbf{100.0} & \textbf{100.0} & \textbf{100.0} & \textbf{100.0} & 68.65 & 15.14 & \textbf{100.0} & \textbf{100.0} & 93.51 & 89.73 & 61.08 & 16.22 \\ ms & 47.57 & \textbf{55.68} & 30.81 & 31.89 & 37.84 & 38.92 & \textbf{56.76} & 55.68 & \DD42.7 & 41.08 & 45.95 & 47.03 & \textbf{85.95} & 52.97 & 39.46 & 43.78 & 47.03 & 44.32 \\ vi & \textbf{100.0} & \textbf{100.0} & \DD80.0 & 75.14 & 23.78 & 18.92 & \textbf{100.0} & \textbf{100.0} & 96.76 & 96.22 & 14.05 & 10.81 & \textbf{100.0} & \textbf{100.0} & 87.57 & 76.76 & 14.59 & \DD8.65 \\ jp & \textbf{100.0} & \textbf{100.0} & 75.14 & 71.35 & 28.11 & \DD3.24 & \textbf{100.0} & \textbf{100.0} & 60.54 & \DD60.0 & \DD20.0 & \DD6.49 & \textbf{100.0} & \textbf{100.0} & 43.78 & 44.86 & 16.22 & \DD1.62 \\ zh & \textbf{100.0} & \textbf{100.0} & 48.11 & 43.78 & 14.05 & \DD5.41 & \textbf{100.0} & \textbf{100.0} & 87.03 & 81.08 & 33.51 & \DD9.73 & \textbf{100.0} & \textbf{100.0} & 56.76 & 56.76 & 49.19 & \DD1.08 \\ fr & \textbf{100.0} & \textbf{100.0} & 96.22 & 95.14 & 84.86 & \DD7.57 & \textbf{100.0} & \textbf{100.0} & 97.84 & 97.84 & \DD60.0 & 14.59 & \textbf{100.0} & \textbf{100.0} & 98.92 & 95.14 & 85.95 & \DD17.3 \\ pt & 99.46 & \textbf{100.0} & 96.76 & 94.05 & 72.43 & \DD9.73 & \textbf{100.0} & \textbf{100.0} & 98.92 & 98.92 & \DD77.3 & \DD9.19 & \textbf{100.0} & \textbf{100.0} & 96.76 & 91.35 & \DD82.7 & \DD4.32 \\ ru & \textbf{100.0} & \textbf{100.0} & 90.27 & 79.46 & \DD2.7 & \DD0.54 & \textbf{100.0} & \textbf{100.0} & 95.14 & 94.59 & 18.92 & \DD1.62 & \textbf{100.0} & \textbf{100.0} & 86.49 & 71.35 & \DD3.24 & \DD0.0 \\ et & \textbf{94.05} & \textbf{94.05} & 77.84 & 76.22 & 67.03 & 49.73 & \textbf{100.0} & 88.65 & 97.84 & 94.59 & 55.68 & 16.22 & \textbf{99.46} & 98.92 & 89.19 & 89.73 & 39.46 & 18.92 \\ it & \textbf{100.0} & \textbf{100.0} & 92.43 & 92.43 & \DD82.7 & 39.46 & \textbf{100.0} & \textbf{100.0} & 99.46 & 99.46 & 89.19 & 24.32 & \textbf{100.0} & 99.46 & 98.92 & 91.89 & 83.78 & 14.05 \\ ta & \textbf{100.0} & \textbf{100.0} & \DD1.08 & \DD1.08 & \DD0.0 & \DD0.0 & \textbf{100.0} & 98.38 & 50.81 & 45.95 & \DD0.0 & \DD0.0 & \textbf{100.0} & \textbf{100.0} & \DD1.08 & \DD1.08 & \DD0.0 & \DD0.0 \\ th & \textbf{100.0} & \textbf{100.0} & 41.62 & 41.62 & \DD0.54 & \DD0.54 & \textbf{100.0} & \textbf{100.0} & 65.41 & 64.32 & \DD1.08 & \DD0.0 & \textbf{100.0} & \textbf{100.0} & 25.41 & \DD22.7 & \DD0.54 & \DD0.0 \\ tr & 98.92 & \textbf{100.0} & 56.76 & 56.76 & 45.41 & 29.73 & 91.89 & 83.24 & \textbf{96.22} & 94.59 & 48.65 & \DD40.0 & \textbf{100.0} & \textbf{100.0} & 52.43 & 52.43 & 31.89 & 14.05 \\  \hline \textbf{avg} & 89.30 & \textbf{89.91} & 65.08 & 63.06 & 40.76 & 18.02 & \textbf{89.91} & 88.40 & 79.24 & 77.91 & 42.20 & 13.66 & \textbf{92.36} & 90.09 & 64.32 & 61.44 & 40.97 & 9.73 \\
\hline
\end{tabular}
}
\caption{Language Steering Shift (LSS) scores from \texttt{id} to languages listed for amplifying (\steer{pmax}, \steer{pmedian}, \steer{=max}, \steer{+max}) and deactivating (\steer{=0}, \steer{=10p}) LAPE neurons across models. Bolded texts denote the highest values within a model.}
\label{tab:id-lape}
\vspace{-10pt}
\end{table*}

\begin{figure}[htbp]
    \centering
    \includegraphics[width=1\linewidth]{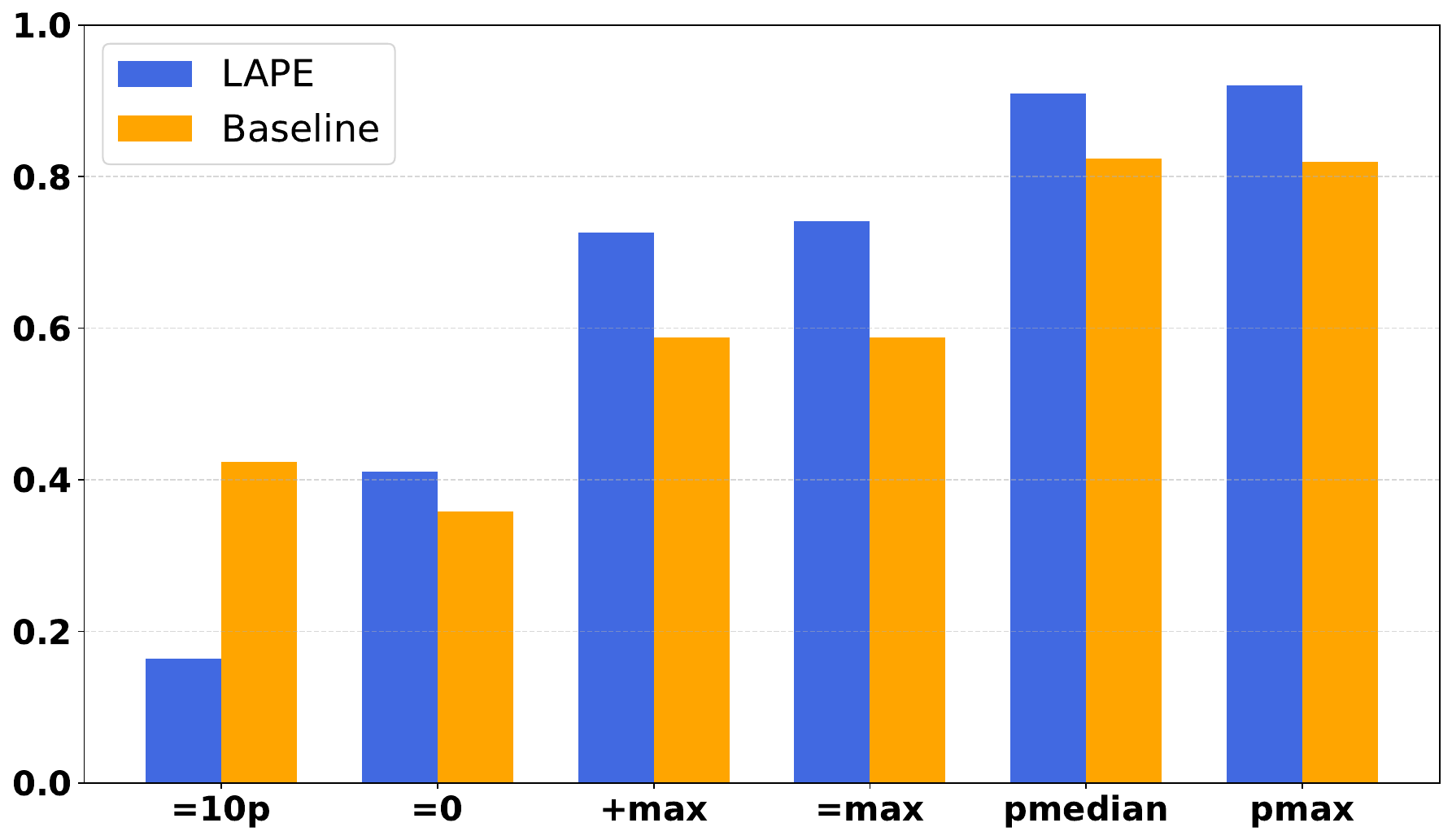}
    \caption{Comparison of averaged LSS scores of various steering factors across languages using LAPE and Baseline neurons.}
    \vspace{-10pt}
    \label{fig:mean-comparison}
\end{figure}
\section{Experimental Settings}

This section discusses the experimental settings used throughout the methodology. It consists of three core stages. First, the identification of neurons specifically activated by language and those exhibiting language-specific responses. Second, the process of finding the optimal amplifying steering factor for influencing these neural activities. Finally, the evaluation stage to determine the impact of this optimal amplifying steering factor on various downstream tasks to assess its efficacy.

\paragraph{Identification of Language-Activated Neurons and Language-Specific Neurons.}
We identify language-activated (Baseline) and language-specific (LAPE) neurons by feeding the LLMs sentences from the \texttt{dev} split of FLORES-200 dataset \cite{nllbteam2022languageleftbehindscaling} containing $30{,}000$ word-tokens per language. We assess 18 languages to ensure broader coverage of language specificity: \texttt{en}, \texttt{nl}, \texttt{id}, \texttt{ms}, \texttt{vi}, \texttt{jp}, \texttt{zh}, \texttt{fr}, \texttt{pt}, \texttt{ru}, \texttt{et}, \texttt{ht}, \texttt{it}, \texttt{qu}, \texttt{sw}, \texttt{ta}, \texttt{th}, and \texttt{tr} in small and large models, namely Qwen2.5 Instruct (0.5B and 7B), Gemma2 Instruct(2B and 9B), and SeaLLMs3 Chat (1.5B and 7B), totaling in 6 models (see Appendix \ref{sec:appendix} for details). For identifying LAPE neurons, we use a language activation probability threshold ($m$) set at the 95th percentile and a LAPE score threshold ($n$) of 1\%. All hyperparameter values are adopted from \citet{tang2024languagespecificneuronskeymultilingual} as they have been demonstrated to effectively capture language-specific neurons.

\paragraph{Finding the Optimal Steering Factor.}
We assess the determined steering factors on LSS score for the small models using a reconstructed version of MLAMA \cite{kassner2021multilingual}, a multilingual fact-probing dataset. MLAMA is reconstructed such that each language question includes answers from all other languages and is filtered to ensure that no answer in different languages are the same. For example, if both Indonesian and Malaysian answers are ``\texttt{Jerman}'' (i.e., Germany in English), that row is discarded. The dataset contains 185 instances per language, with 15 of the 18 available languages are used due to limitation in the dataset.
% (see Appendix \ref{sec:appendix-find} for more details). 

\paragraph{Evaluating the Optimal Steering Factor on Downstream Tasks.}
We evaluate the model’s confidence and downstream task performances by conducting interventions using the optimal amplifying steering factors. We provide the deltas between the intervened and non-intervened performances to quantify the extent to which the interventions shift task outcomes. Interventions are performed in every layer of the model, with layer-wise amplification result and analysis provided in Appendix \ref{sec:appendix-layerwise}. 

We evaluate on commonsense reasoning task, which is considered one of the most difficult tasks for LLMs \cite{li2022systematicinvestigationcommonsenseknowledge}, using XCOPA \cite{ponti2020xcopa} and XWinograd \cite{muennighoff2022crosslingual} dataset. For factual knowledge, we use Include-lite \cite{romanou2024include} dataset. Building on \citet{mondal-etal-2025-language}, who focused only on self-language intervention, we extend the analysis of these task performances to cross-lingual intervention. The \texttt{devtest} split of FLORES-200 is used to measure perplexity and BLEU for translation from English (\texttt{en}) to every other languages. 

Translation tasks are performed using 2 types of prompts: \texttt{targeted} and \texttt{non-targeted}. Targeted prompts explicitly mention the target language, whereas non-targeted prompts do not. For example, when translating to \texttt{id}, the targeted prompt begins with \textit{"Translate from English to Indonesian."}, whereas the non-targeted prompt begins with \textit{"Translate from English into the target language."}

\section{Result and Analysis}

This section presents the findings and analysis of the optimal amplifying steering factors and their evaluation on model confidence and downstream tasks.

\subsection{LSS Scores and Optimal Amplifying Steering Factor}

Result of the Language Steering Shift (LSS) score shows that LAPE neuron amplification yields high scores for many languages, often reaching 100\%, as shown in Table~\ref{tab:id-lape} for the intervention on \texttt{id}. These strong results are consistent across other languages, as detailed in Appendix~\ref{sec:appendix-lss}, indicating successful language shifting. However, interventions involving certain language pairs may result in lower LSS scores. For example, translation between \texttt{id} and \texttt{ms} in either direction tends to produce scores capped at 50\% in most models. This is likely due to the high similarity between the two languages, which makes it more difficult for the models to distinguish between them. Notably, \texttt{en} remains easy to steer to, suggesting a strong influence that is possibly due to its dominance and widespread presence in LLM training data.

The optimal amplifying factors, defined as those that result in the highest aggregated LSS scores, are found to be \steer{pmax} and \steer{pmedian}. A one-way ANOVA followed by Tukey’s HSD test for post-hoc pairwise comparisons reveals that the patched steering factors \steer{pmax} (mean = 0.9204) and \steer{pmedian} ($\texttt{mean} = 0.9095$) significantly outperform the other factors ($p < 0.001$); Further results are provided in Appendix \ref{sec:appendix-lss}. However, the difference between the two is not statistically significant ($p = 0.98$), indicating that both are equally effective as optimal amplifying factors. 

Test-time intervention factors, \steer{+max} and \steer{=max}, perform well mainly for high-resource languages such as \texttt{en}, \texttt{pt}, \texttt{it}, and \texttt{fr}, but show limited effectiveness for other languages, with the lowest performance observed for \texttt{ta} in Qwen and Gemma despite its substantial neuron presence. The deactivating factor \steer{=0} shows variation in results, with successful steering in some languages (e.g., \texttt{en}, \texttt{nl}, and \texttt{it}), while \steer{=10p} demonstrates stronger deactivation capability, often resulting in LSS scores close to $0\%$ across many languages, align with findings from \citet{mondal-etal-2025-language}.

Compared to Baseline neurons, LAPE neurons yield higher average LSS scores across languages as illustrated in Figure \ref{fig:mean-comparison}. This suggests that LAPE is more effective at steering language, likely due to its stronger language specificity. In contrast, Baseline neurons show greater variability (see Appendix \ref{sec:appendix-lss}); due to their lower specificity, deactivating these neurons with \steer{=10} and \steer{=10p} sometimes result in higher LSS scores than amplifying them. This contrast further supports LAPE neurons as more reliable and consistent for steering towards corresponding language.

\newcommand{\D}{\phantom{-}}
\begin{table}[htbp]
\centering
\small
\resizebox{\linewidth}{!}{%
\begin{tabular}{l|cc|cc|cc}
\toprule
\textbf{Model} 
& \multicolumn{2}{c|}{\textbf{Include-lite}} 
& \multicolumn{2}{c|}{\textbf{XWinograd}} 
& \multicolumn{2}{c}{\textbf{XCOPA}} \\
& \textbf{D} & \textbf{O} 
& \textbf{D} & \textbf{O} 
& \textbf{D} & \textbf{O} \\
\midrule
S-7B      
& $\mathbf{0.38}$ & $-0.11$ 
& $\mathbf{\D1.27}$ & $-6.71$ 
& $\mathbf{-0.44}$ & $-5.49$ \\
S-1.5B    
& $\mathbf{1.02}$ & $\D0.54$ 
& $\mathbf{-0.76}$ & $-6.83$ 
& $\mathbf{-0.18}$ & $-3.18$ \\
Q-7B      
& $\mathbf{0.26}$ & $-0.17$ 
& $\mathbf{-0.03}$ & $-7.19$ 
& $\mathbf{-0.22}$ & $-5.98$ \\
Q-0.5B    
& $\mathbf{0.16}$ & $-0.36$ 
& $\mathbf{\D0.17}$ & $-3.38$ 
& $\mathbf{-0.04}$ & $-1.69$ \\
G-9B      
& $\mathbf{0.87}$ & $-0.77$ 
& $\mathbf{\D0.29}$ & $-5.46$ 
& $\mathbf{\D0.13}$ & $-2.24$ \\
G-2B      
& $\mathbf{0.25}$ & $-1.03$ 
& $\mathbf{\D0.15}$ & $-3.38$ 
& $\mathbf{\D0.07}$ & $-1.43$ \\
\bottomrule
\end{tabular}
}
\caption{Average delta task accuracies after \steer{pmax} intervention on LAPE neurons for self-intervention/diagonal (D) and cross-intervention/off-diagonal (O) neuron steering for every models used. Negative and positive values respectively represent decreased and increased accuracy from the baseline (non-intervention scenario). S denotes SeaLLMs3, Q = Qwen2.5 and G = Gemma2.}
\label{tab:pmax-reasoning}
\vspace{-10pt}
\end{table}

% SeaLLMv3-7B   
% SeaLLMv3-1.5B   
% Qwen2.5-7B    
% Qwen2.5-0.5B   
% Gemma2-9B       
% Gemma2-2B    

\subsection{Impact of Neuron Amplification}

\begin{figure}[htbp]
    \centering
    \includegraphics[width=\linewidth]{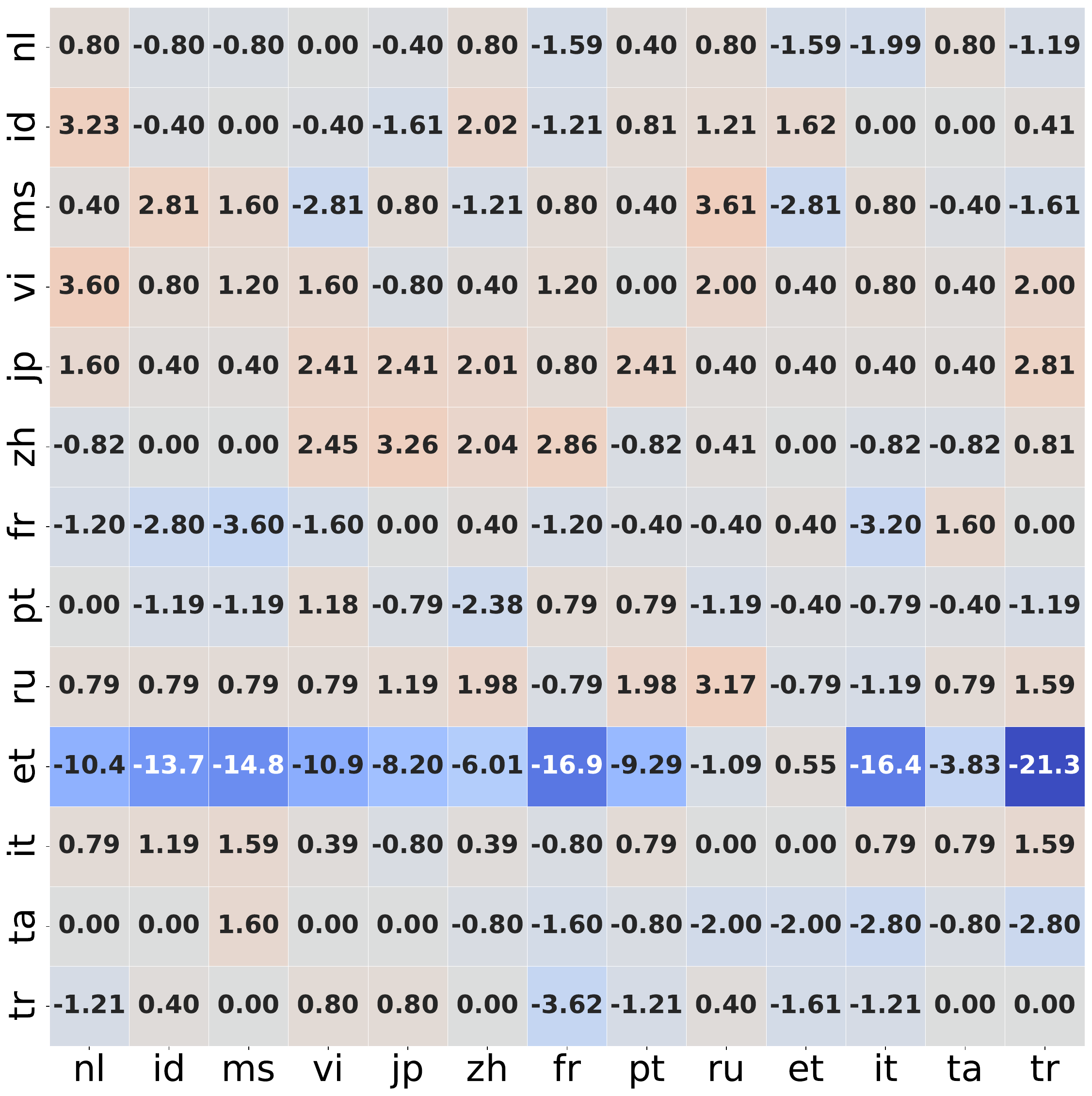}
    \caption{Delta Include-lite accuracy under LAPE neurons on Gemma2 9B. Blue highlights reductions, whereas red highlights increases.}
    \label{fig:lape-include}
    \vspace{-10pt}
\end{figure}

To further evaluate the optimal steering factors, we analyze their effectiveness in steering LAPE neurons using the most amplifying factor, \steer{pmax}, across multiple downstream tasks and language modeling performance via perplexity. The next best factor, \steer{pmedian}, shows comparable but less prominent results, while Baseline neuron amplification generally degrades performance in both self- and cross-intervention settings (see Appendix~\ref{sec:appendix} for further results).

\begin{figure}[htbp]
\centering

\begin{subfigure}{0.49\columnwidth}
    \includegraphics[width=\linewidth]{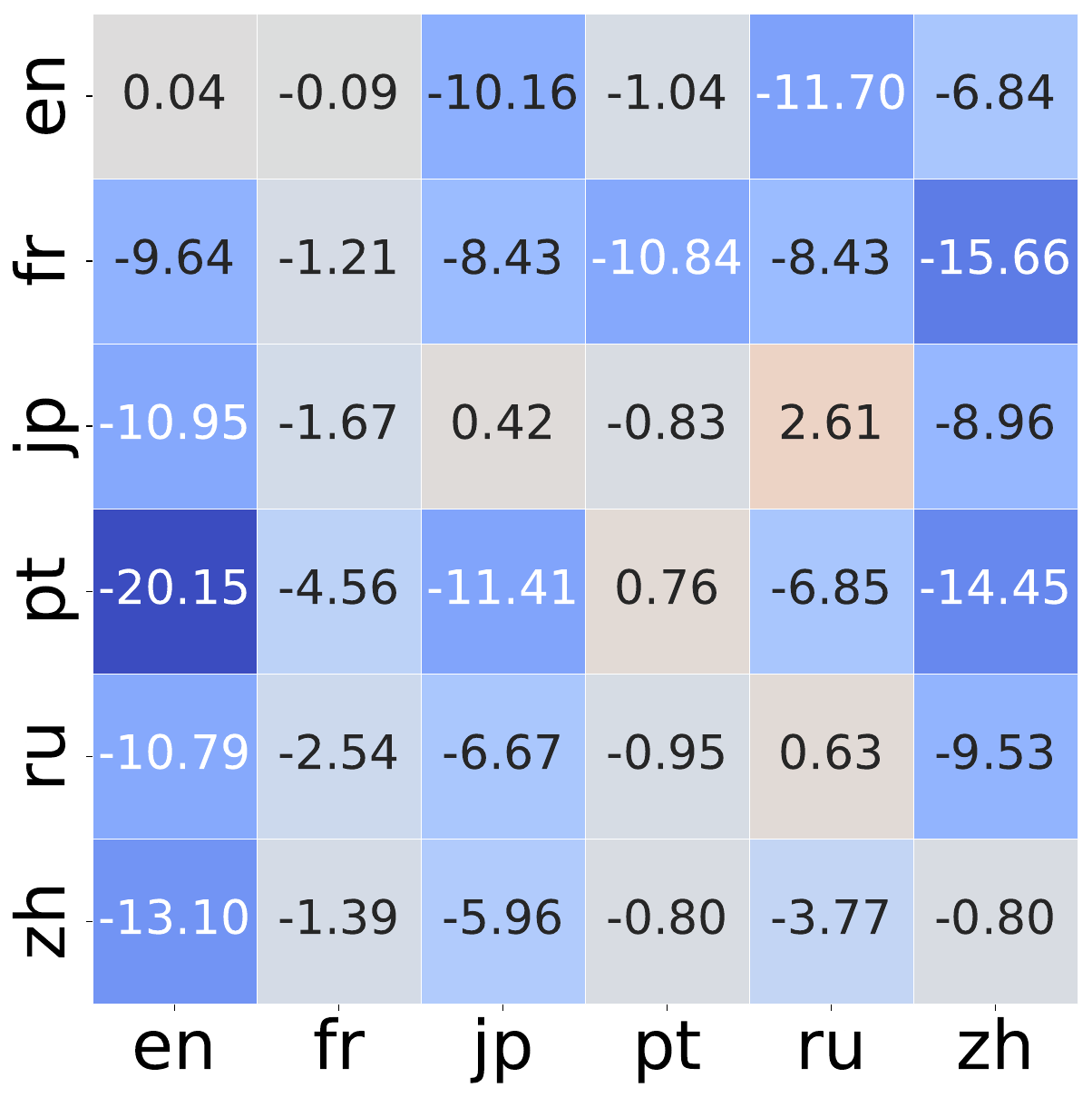}
    % \caption{Gemma2 9B}
\end{subfigure}
\hfill
\begin{subfigure}{0.49\columnwidth}
    \includegraphics[width=\linewidth]{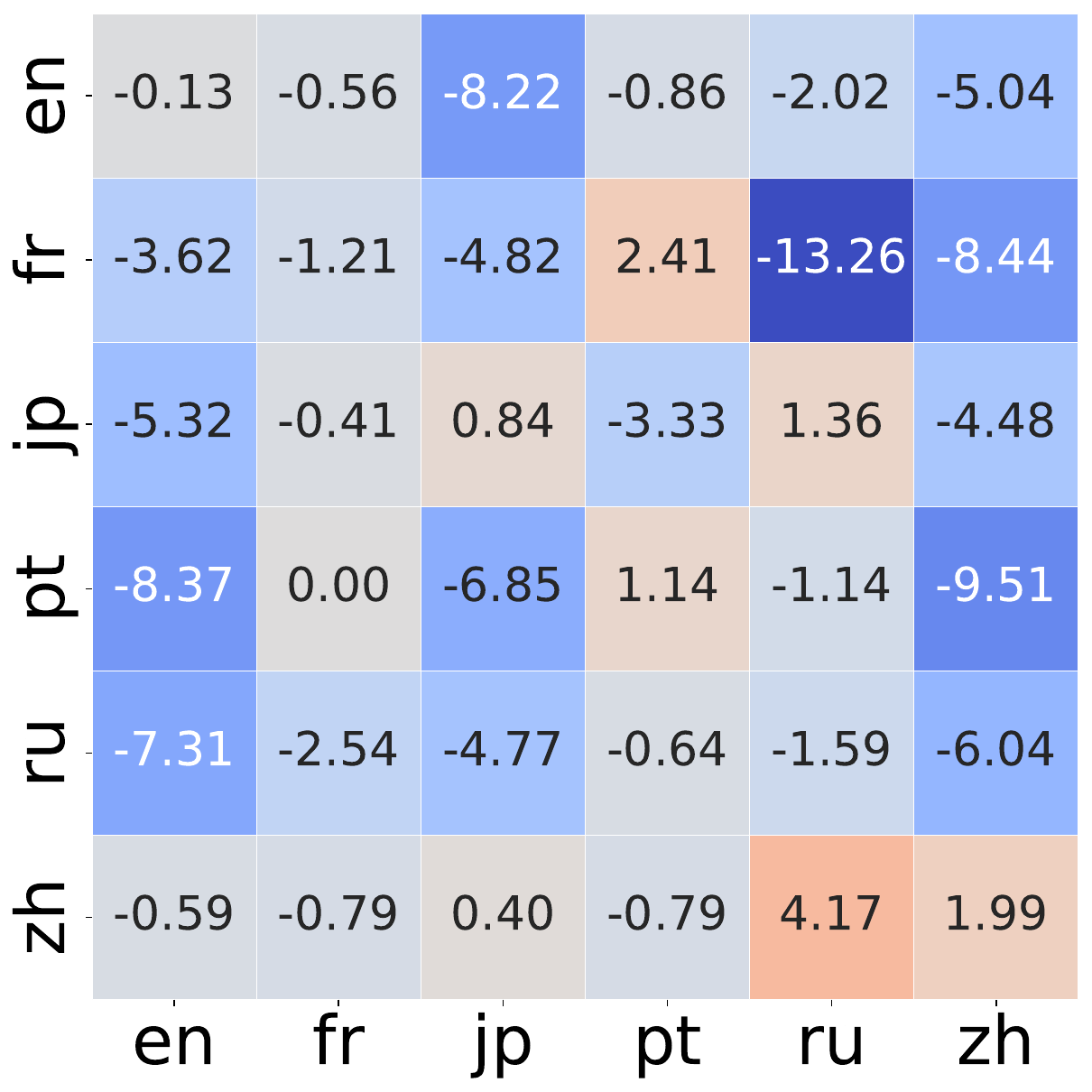}
    % \caption{Qwen2.5 7B}
\end{subfigure}

\caption{Delta XWinograd accuracy on Qwen2.5 7B (left) and Qwen2.5 0.5B (right) after amplification of LAPE neurons. Blue highlights reductions, whereas red highlights increases.}
\label{fig:lape-xwin}
\vspace{2mm}

\vspace{-2mm}
% \caption{Delta XWinograd accuracy across models.}
\vspace{-3mm}
% \label{fig:raw-xwin}
\end{figure}

\subsubsection{Knowledge and Reasoning Tasks}
Overall performance of knowledge and reasoning tasks intervention as presented in Table \ref{tab:pmax-reasoning} shows that \steer{pmax} self-intervention can improve accuracy up to 1\%, with consistent gains observed in Include-lite. XCOPA and XWinograd self-intervention may show less consistent improvements with only modest changes in some models. In contrast, cross-language interventions generally lead to performance drops on average. These results suggest that amplifying LAPE neurons in self-language on knowledge and reasoning tasks is potentially beneficial for enhancing performance, though their effectiveness in cross-lingual transfer appears limited. 
% \vspace{-10pt}
\paragraph{Include-lite Evaluation.}

The result in Figure \ref{fig:lape-include} reveals that self-intervention improves performance or modestly changes. As a comparison, cross-intervention shows stronger degradation, visibly captured in \texttt{et}. Some models capture increasing accuracy across similar languages such as \texttt{id} and \texttt{ms}, though the correlations are not always clear as they often show better performance than self-intervention.

% \vspace{-10pt}

\paragraph{XWinograd Evaluation.}

Figure \ref{fig:lape-xwin} shows that self-intervention tends to be non-destructive,  with some languages such as \texttt{jp} and \texttt{pt}, showing slight improvements across most models. However, the improved languages vary by models; for example, \texttt{zh} improves in Qwen2.5 0.5B but not in the 7B variant. Cross-lingual interventions degrade performance on average, particularly in larger models. Nevertheless, occasional improvements are observed in less related language pairs, such as \texttt{zh} when intervened by \texttt{ru} in Qwen2.5 0.5B (see Appendix~\ref{sec:appendix-acc} for additional results). A consistent trend across models is that amplifying \texttt{en} neurons reliably reduces accuracy, suggesting they may play a role in reasoning rather than language transfer. 

\begin{figure}[htbp]
    \centering
    \includegraphics[width=\linewidth]{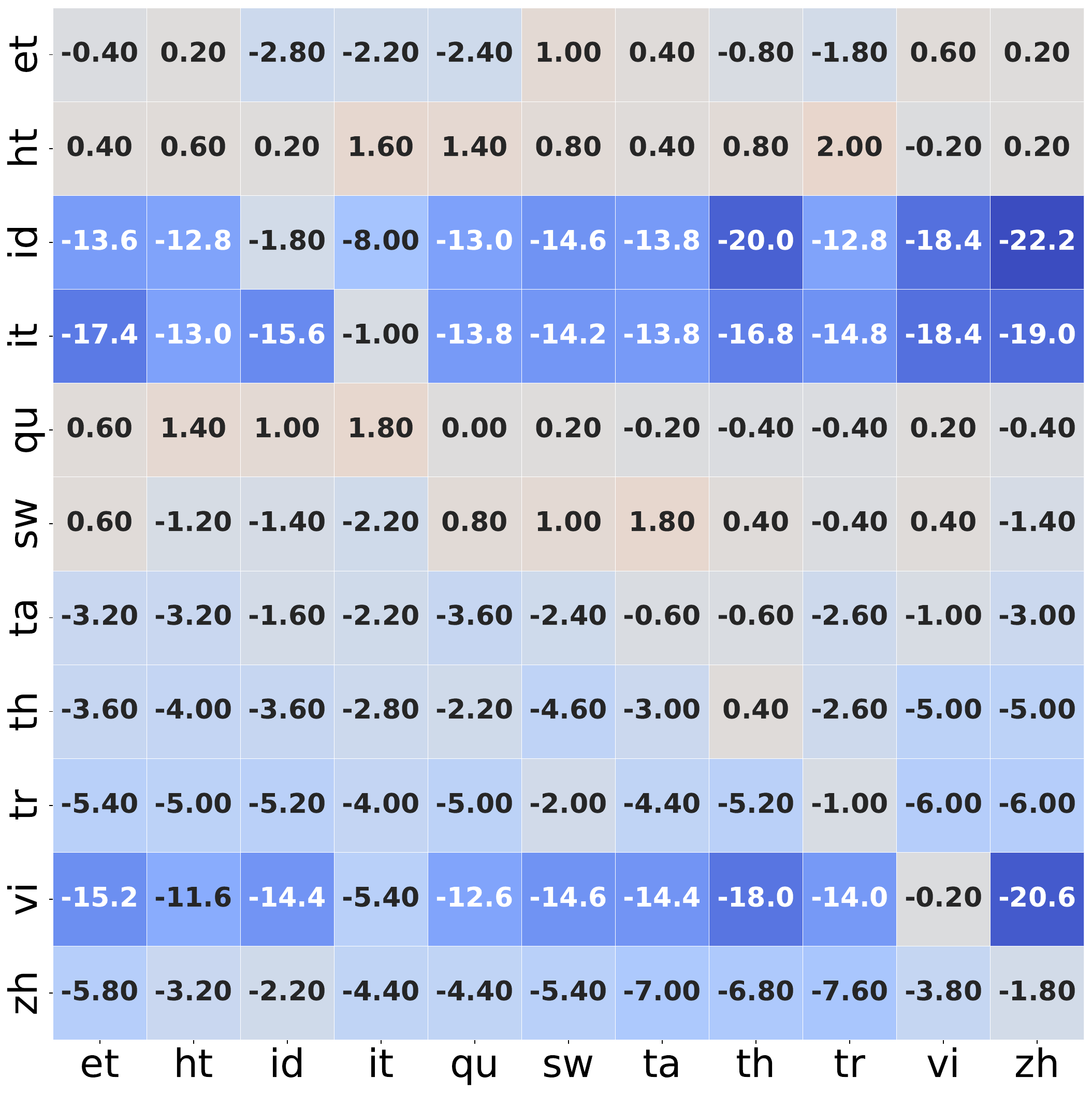}
    \caption{Delta XCOPA accuracy after intevention of LAPE neurons for SeaLLMs3 7B (right). Row \textit{i} depicts the initial language and column \textit{j} is the intervention language.}
    \label{fig:lape-xcopa}
    \vspace{-10pt}
\end{figure}

% \vspace{-10pt}

\paragraph{XCOPA Evaluation.}
Figure \ref{fig:lape-xcopa} shows that amplifying LAPE neurons does not consistently improve in self-intervention, but changes are generally reserved, indicating non-destructive effects. In contrast, cross-language amplification often leads to larger drops, visibly captured on the intervention of \texttt{it}, \texttt{id}, \texttt{th}, \texttt{tr}, and \texttt{vi} neurons, degrading other languages more than their own. Low-resource languages like \texttt{qu}, \texttt{et}, \texttt{sw}, and \texttt{ta} show self-intervention gains, particularly in SeaLLMs3. SeaLLMs3 and Qwen2.5, having shared model architecture \cite{zhang2024seallms3openfoundation}, display very similar patterns, suggesting architecture may contribute to neuron behavior. 
% Baseline neuron intervention and \steer{pmedian} factor yield lower deltas, reinforcing the importance of neuron specificity and \steer{pmax}. 
Overall, XCOPA deltas reveal clearer signs of degradation than improvement.
% which may indicate that supporting the view that these neurons are language-specific but not inherently beneficial for performance.

% \input{figs/bleu-targeted}.
% \input{figs/bleu/bleu-nontarget}

% \input{figs/stats_factor}
% \input{figs/bleu/bleu-nontarget}

\begin{table*}[htbp]
\centering

\renewcommand{\arraystretch}{1.3}
\resizebox{1.01\textwidth}{!}{%
\begin{tabular}{c|cccccc|cccccc}
\toprule
\textbf{Lang} 
& \multicolumn{6}{c|}{\large\textbf{Targeted Prompt}} 
& \multicolumn{6}{c}{\large\textbf{Non-targeted Prompt}} \\
\cmidrule(lr){2-7} \cmidrule(lr){8-13}
& \large\textbf{G-2B} & \large\textbf{G-9B} & \large\textbf{Q-0.5B} & \large\textbf{Q-7B} & \large\textbf{S-1.5B} & \large\textbf{S-7B}
& \large\textbf{G-2B} & \large\textbf{G-9B} & \large\textbf{Q-0.5B} & \large\textbf{Q-7B} & \large\textbf{S-1.5B} & \large\textbf{S-7B} \\
\midrule
\texttt{et} & \textcolor{Red}{\large\textbf{$-$0.44}} \large\textsubscript{(5.27)} & \textcolor{Green}{\large\textbf{$+$0.04}} \large\textsubscript{(5.06)} & \textcolor{Red}{\large\textbf{$-$0.04}} \large\textsubscript{(1.24)} & \textcolor{Red}{\large\textbf{$-$0.74}} \large\textsubscript{(4.09)} & \textcolor{Red}{\large\textbf{$-$0.15}} \large\textsubscript{(2.32)} & \textcolor{Red}{\large\textbf{$-$0.81}} \large\textsubscript{(3.76)} & \textcolor{Green}{\large\textbf{$+$0.07}} \large\textsubscript{(0.45)} & \textcolor{Red}{\large\textbf{$-$0.01}} \large\textsubscript{(0.24)} & \textcolor{Green}{\large\textbf{$+$0.40}} \large\textsubscript{(0.49)} & \textcolor{Green}{\large\textbf{$+$1.78}} \large\textsubscript{(0.76)} & \textcolor{Green}{\large\textbf{$+$1.01}} \large\textsubscript{(0.87)} & \textcolor{Green}{\large\textbf{$+$2.56}} \large\textsubscript{(0.43)} \\
\texttt{fr} & \textcolor{Red}{\large\textbf{$-$1.23}} \large\textsubscript{(26.49)} & \textcolor{Red}{\large\textbf{$-$4.19}} \large\textsubscript{(41.90)} & \textcolor{Green}{\large\textbf{$+$0.08}} \large\textsubscript{(13.93)} & \textcolor{Red}{\large\textbf{$-$0.16}} \large\textsubscript{(27.24)} & \textcolor{Green}{\large\textbf{$+$0.21}} \large\textsubscript{(18.23)} & \textcolor{Red}{\large\textbf{$-$1.36}} \large\textsubscript{(30.47)} & \textcolor{Green}{\large\textbf{$+$8.32}} \large\textsubscript{(0.76)} & \textcolor{Green}{\large\textbf{$+$3.97}} \large\textsubscript{(0.57)} & \textcolor{Green}{\large\textbf{$+$12.39}} \large\textsubscript{(0.52)} & \textcolor{Green}{\large\textbf{$+$14.40}} \large\textsubscript{(13.75)} & \textcolor{Green}{\large\textbf{$+$18.38}} \large\textsubscript{(1.43)} & \textcolor{Green}{\large\textbf{$+$24.69}} \large\textsubscript{(0.32)} \\
\texttt{ht} & \textcolor{Red}{\large\textbf{$-$0.94}} \large\textsubscript{(2.53)} & \textcolor{Red}{\large\textbf{$-$1.06}} \large\textsubscript{(2.21)} & \textcolor{Red}{\large\textbf{$-$0.08}} \large\textsubscript{(1.55)} & \textcolor{Red}{\large\textbf{$-$0.15}} \large\textsubscript{(3.35)} & \textcolor{Red}{\large\textbf{$-$0.19}} \large\textsubscript{(3.34)} & \textcolor{Red}{\large\textbf{$-$2.03}} \large\textsubscript{(5.12)} & \textcolor{Green}{\large\textbf{$+$0.01}} \large\textsubscript{(0.17)} & \textcolor{Red}{\large\textbf{$-$0.01}} \large\textsubscript{(0.13)} & \textcolor{Green}{\large\textbf{$+$0.78}} \large\textsubscript{(0.25)} & \textcolor{Green}{\large\textbf{$+$2.19}} \large\textsubscript{(0.35)} & \textcolor{Green}{\large\textbf{$+$1.53}} \large\textsubscript{(0.56)} & \textcolor{Green}{\large\textbf{$+$2.53}} \large\textsubscript{(0.43)} \\
\texttt{id} & \textcolor{Red}{\large\textbf{$-$0.53}} \large\textsubscript{(18.96)} & \textcolor{Red}{\large\textbf{$-$1.60}} \large\textsubscript{(17.89)} & \textcolor{Green}{\large\textbf{$+$0.49}} \large\textsubscript{(7.34)} & \textcolor{Green}{\large\textbf{$+$2.15}} \large\textsubscript{(21.29)} & \textcolor{Red}{\large\textbf{$-$0.64}} \large\textsubscript{(20.46)} & \textcolor{Green}{\large\textbf{$+$0.61}} \large\textsubscript{(30.02)} & \textcolor{Green}{\large\textbf{$+$1.09}} \large\textsubscript{(0.55)} & \textcolor{Green}{\large\textbf{$+$0.21}} \large\textsubscript{(0.29)} & \textcolor{Green}{\large\textbf{$+$5.72}} \large\textsubscript{(0.65)} & \textcolor{Green}{\large\textbf{$+$23.88}} \large\textsubscript{(1.13)} & \textcolor{Green}{\large\textbf{$+$14.07}} \large\textsubscript{(1.92)} & \textcolor{Green}{\large\textbf{$+$26.36}} \large\textsubscript{(0.52)} \\
\texttt{it} & \textcolor{Red}{\large\textbf{$-$0.67}} \large\textsubscript{(15.79)} & \textcolor{Red}{\large\textbf{$-$2.44}} \large\textsubscript{(25.30)} & \textcolor{Green}{\large\textbf{$+$0.03}} \large\textsubscript{(6.34)} & \textcolor{Red}{\large\textbf{$-$1.23}} \large\textsubscript{(15.90)} & \textcolor{Red}{\large\textbf{$-$1.04}} \large\textsubscript{(8.23)} & \textcolor{Red}{\large\textbf{$-$0.49}} \large\textsubscript{(17.42)} & \textcolor{Green}{\large\textbf{$+$3.66}} \large\textsubscript{(0.48)} & \textcolor{Green}{\large\textbf{$+$0.20}} \large\textsubscript{(0.25)} & \textcolor{Green}{\large\textbf{$+$1.74}} \large\textsubscript{(0.63)} & \textcolor{Green}{\large\textbf{$+$13.44}} \large\textsubscript{(1.16)} & \textcolor{Green}{\large\textbf{$+$9.62}} \large\textsubscript{(1.15)} & \textcolor{Green}{\large\textbf{$+$15.55}} \large\textsubscript{(0.44)} \\
\texttt{jp} & \textcolor{Green}{\large\textbf{$+$0.01}} \large\textsubscript{(0.11)} & \textcolor{Red}{\large\textbf{$-$0.38}} \large\textsubscript{(0.46)} & \textcolor{Green}{\large\textbf{$+$0.15}} \large\textsubscript{(0.05)} & \textcolor{Green}{\large\textbf{$+$0.20}} \large\textsubscript{(0.05)} & \textcolor{Green}{\large\textbf{$+$0.36}} \large\textsubscript{(0.14)} & \textcolor{Green}{\large\textbf{$+$0.06}} \large\textsubscript{(0.44)} & \textcolor{Green}{\large\textbf{$+$0.09}} \large\textsubscript{(0.02)} & \textcolor{Red}{\large\textbf{$-$0.02}} \large\textsubscript{(0.03)} & \textcolor{Green}{\large\textbf{$+$0.09}} \large\textsubscript{(0.06)} & \textcolor{Green}{\large\textbf{$+$0.06}} \large\textsubscript{(0.09)} & \textcolor{Green}{\large\textbf{$+$0.24}} \large\textsubscript{(0.11)} & \textcolor{Green}{\large\textbf{$+$0.06}} \large\textsubscript{(0.08)} \\
\texttt{ms} & \textcolor{Red}{\large\textbf{$-$0.38}} \large\textsubscript{(13.51)} & \textcolor{Red}{\large\textbf{$-$0.89}} \large\textsubscript{(35.14)} & \textcolor{Green}{\large\textbf{$+$0.08}} \large\textsubscript{(4.85)} & \textcolor{Red}{\large\textbf{$-$0.24}} \large\textsubscript{(16.13)} & \textcolor{Red}{\large\textbf{$-$0.25}} \large\textsubscript{(15.39)} & \textcolor{Red}{\large\textbf{$-$0.85}} \large\textsubscript{(24.82)} & \textcolor{Green}{\large\textbf{$+$0.69}} \large\textsubscript{(0.49)} & \textcolor{Green}{\large\textbf{$+$0.08}} \large\textsubscript{(0.31)} & \textcolor{Green}{\large\textbf{$+$3.32}} \large\textsubscript{(0.61)} & \textcolor{Red}{\large\textbf{$-$0.37}} \large\textsubscript{(1.01)} & \textcolor{Green}{\large\textbf{$+$9.05}} \large\textsubscript{(1.54)} & \textcolor{Green}{\large\textbf{$+$18.64}} \large\textsubscript{(0.50)} \\
\texttt{nl} & \textcolor{Green}{\large\textbf{$+$0.05}} \large\textsubscript{(13.08)} & \textcolor{Red}{\large\textbf{$-$2.72}} \large\textsubscript{(21.08)} & \textcolor{Green}{\large\textbf{$+$0.15}} \large\textsubscript{(4.52)} & \textcolor{Green}{\large\textbf{$+$1.01}} \large\textsubscript{(14.30)} & \textcolor{Green}{\large\textbf{$+$0.33}} \large\textsubscript{(8.57)} & \textcolor{Red}{\large\textbf{$-$0.66}} \large\textsubscript{(14.67)} & \textcolor{Green}{\large\textbf{$+$4.49}} \large\textsubscript{(0.45)} & \textcolor{Green}{\large\textbf{$+$0.06}} \large\textsubscript{(0.24)} & \textcolor{Green}{\large\textbf{$+$2.31}} \large\textsubscript{(0.55)} & \textcolor{Green}{\large\textbf{$+$9.84}} \large\textsubscript{(0.85)} & \textcolor{Green}{\large\textbf{$+$5.36}} \large\textsubscript{(1.20)} & \textcolor{Green}{\large\textbf{$+$11.01}} \large\textsubscript{(0.41)} \\
\texttt{pt} & \textcolor{Red}{\large\textbf{$-$2.42}} \large\textsubscript{(25.08)} & \textcolor{Red}{\large\textbf{$-$1.77}} \large\textsubscript{(44.80)} & \textcolor{Green}{\large\textbf{$+$0.41}} \large\textsubscript{(13.98)} & \textcolor{Red}{\large\textbf{$-$1.02}} \large\textsubscript{(36.48)} & \textcolor{Green}{\large\textbf{$+$0.39}} \large\textsubscript{(12.19)} & \textcolor{Green}{\large\textbf{$+$0.23}} \large\textsubscript{(30.99)} & \textcolor{Green}{\large\textbf{$+$10.11}} \large\textsubscript{(0.61)} & \textcolor{Green}{\large\textbf{$+$12.45}} \large\textsubscript{(0.44)} & \textcolor{Green}{\large\textbf{$+$9.45}} \large\textsubscript{(0.61)} & \textcolor{Green}{\large\textbf{$+$23.96}} \large\textsubscript{(1.33)} & \textcolor{Green}{\large\textbf{$+$15.32}} \large\textsubscript{(1.47)} & \textcolor{Green}{\large\textbf{$+$18.47}} \large\textsubscript{(0.58)} \\
\texttt{qu} & \textcolor{Red}{\large\textbf{$-$0.06}} \large\textsubscript{(0.34)} & \textcolor{Green}{\large\textbf{$+$0.39}} \large\textsubscript{(0.21)} & \textcolor{Green}{\large\textbf{$+$0.05}} \large\textsubscript{(0.46)} & \textcolor{Red}{\large\textbf{$-$0.12}} \large\textsubscript{(1.39)} & \textcolor{Red}{\large\textbf{$-$0.28}} \large\textsubscript{(1.05)} & \textcolor{Red}{\large\textbf{$-$0.30}} \large\textsubscript{(0.64)} & \textcolor{Green}{\large\textbf{$+$0.12}} \large\textsubscript{(0.44)} & \textcolor{Green}{\large\textbf{$+$0.16}} \large\textsubscript{(0.34)} & \textcolor{Green}{\large\textbf{$+$0.30}} \large\textsubscript{(0.46)} & \textcolor{Green}{\large\textbf{$+$0.38}} \large\textsubscript{(0.75)} & \textcolor{Red}{\large\textbf{$-$0.00}} \large\textsubscript{(0.93)} & \textcolor{Green}{\large\textbf{$+$0.11}} \large\textsubscript{(0.63)} \\
\texttt{ru} & \textcolor{Red}{\large\textbf{$-$0.33}} \large\textsubscript{(13.72)} & \textcolor{Green}{\large\textbf{$+$1.66}} \large\textsubscript{(16.16)} & \textcolor{Green}{\large\textbf{$+$0.26}} \large\textsubscript{(5.67)} & \textcolor{Red}{\large\textbf{$-$0.51}} \large\textsubscript{(16.88)} & \textcolor{Red}{\large\textbf{$-$0.99}} \large\textsubscript{(8.90)} & \textcolor{Red}{\large\textbf{$-$3.64}} \large\textsubscript{(17.02)} & \textcolor{Red}{\large\textbf{$-$0.02}} \large\textsubscript{(0.38)} & \textcolor{Green}{\large\textbf{$+$1.61}} \large\textsubscript{(0.24)} & \textcolor{Red}{\large\textbf{$-$0.93}} \large\textsubscript{(5.22)} & \textcolor{Green}{\large\textbf{$+$14.03}} \large\textsubscript{(2.20)} & \textcolor{Green}{\large\textbf{$+$9.62}} \large\textsubscript{(0.75)} & \textcolor{Green}{\large\textbf{$+$11.58}} \large\textsubscript{(4.77)} \\
\texttt{sw} & \textcolor{Red}{\large\textbf{$-$0.34}} \large\textsubscript{(7.62)} & \textcolor{Green}{\large\textbf{$+$2.71}} \large\textsubscript{(12.15)} & \textcolor{Red}{\large\textbf{$-$0.24}} \large\textsubscript{(0.58)} & \textcolor{Red}{\large\textbf{$-$0.48}} \large\textsubscript{(2.39)} & \textcolor{Red}{\large\textbf{$-$0.37}} \large\textsubscript{(1.56)} & \textcolor{Red}{\large\textbf{$-$0.06}} \large\textsubscript{(3.06)} & \textcolor{Green}{\large\textbf{$+$0.03}} \large\textsubscript{(0.45)} & \textcolor{Red}{\large\textbf{$-$0.05}} \large\textsubscript{(0.26)} & \textcolor{Green}{\large\textbf{$+$0.35}} \large\textsubscript{(0.51)} & \textcolor{Green}{\large\textbf{$+$1.01}} \large\textsubscript{(0.78)} & \textcolor{Red}{\large\textbf{$-$0.06}} \large\textsubscript{(1.07)} & \textcolor{Green}{\large\textbf{$+$1.59}} \large\textsubscript{(0.36)} \\
\texttt{ta} & \textcolor{Red}{\large\textbf{$-$0.66}} \large\textsubscript{(2.21)} & \textcolor{Red}{\large\textbf{$-$0.16}} \large\textsubscript{(5.58)} & \textcolor{Red}{\large\textbf{$-$0.01}} \large\textsubscript{(0.13)} & \textcolor{Red}{\large\textbf{$-$0.05}} \large\textsubscript{(0.28)} & \textcolor{Red}{\large\textbf{$-$0.05}} \large\textsubscript{(0.42)} & \textcolor{Red}{\large\textbf{$-$0.14}} \large\textsubscript{(0.59)} & \textcolor{Red}{\large\textbf{$-$0.01}} \large\textsubscript{(0.40)} & \textcolor{Red}{\large\textbf{$-$0.10}} \large\textsubscript{(0.23)} & \textcolor{Red}{\large\textbf{$-$0.17}} \large\textsubscript{(0.39)} & \textcolor{Red}{\large\textbf{$-$0.23}} \large\textsubscript{(0.45)} & \textcolor{Red}{\large\textbf{$-$0.21}} \large\textsubscript{(0.64)} & \textcolor{Green}{\large\textbf{$+$0.12}} \large\textsubscript{(0.29)} \\
\texttt{th} & \textcolor{Red}{\large\textbf{$-$0.45}} \large\textsubscript{(8.38)} & \textcolor{Red}{\large\textbf{$-$0.63}} \large\textsubscript{(19.93)} & \textcolor{Red}{\large\textbf{$-$0.49}} \large\textsubscript{(2.19)} & \textcolor{Red}{\large\textbf{$-$0.40}} \large\textsubscript{(8.08)} & \textcolor{Red}{\large\textbf{$-$0.10}} \large\textsubscript{(3.69)} & \textcolor{Red}{\large\textbf{$-$1.20}} \large\textsubscript{(6.61)} & \textcolor{Green}{\large\textbf{$+$0.06}} \large\textsubscript{(0.45)} & \textcolor{Green}{\large\textbf{$+$0.06}} \large\textsubscript{(0.28)} & \textcolor{Green}{\large\textbf{$+$0.71}} \large\textsubscript{(0.48)} & \textcolor{Green}{\large\textbf{$+$6.96}} \large\textsubscript{(0.77)} & \textcolor{Green}{\large\textbf{$+$2.36}} \large\textsubscript{(1.02)} & \textcolor{Green}{\large\textbf{$+$5.29}} \large\textsubscript{(0.43)} \\
\texttt{tr} & \textcolor{Red}{\large\textbf{$-$2.49}} \large\textsubscript{(22.69)} & \textcolor{Red}{\large\textbf{$-$5.95}} \large\textsubscript{(32.65)} & \textcolor{Green}{\large\textbf{$+$0.10}} \large\textsubscript{(10.71)} & \textcolor{Green}{\large\textbf{$+$4.73}} \large\textsubscript{(23.58)} & \textcolor{Green}{\large\textbf{$+$0.08}} \large\textsubscript{(18.12)} & \textcolor{Green}{\large\textbf{$+$0.89}} \large\textsubscript{(26.70)} & \textcolor{Green}{\large\textbf{$+$0.77}} \large\textsubscript{(0.33)} & \textcolor{Green}{\large\textbf{$+$4.42}} \large\textsubscript{(0.16)} & \textcolor{Green}{\large\textbf{$+$9.59}} \large\textsubscript{(0.52)} & \textcolor{Green}{\large\textbf{$+$25.26}} \large\textsubscript{(0.74)} & \textcolor{Green}{\large\textbf{$+$15.65}} \large\textsubscript{(1.65)} & \textcolor{Green}{\large\textbf{$+$21.40}} \large\textsubscript{(0.36)} \\
\texttt{vi} & \textcolor{Red}{\large\textbf{$-$0.35}} \large\textsubscript{(0.53)} & \textcolor{Red}{\large\textbf{$-$2.23}} \large\textsubscript{(2.89)} & \textcolor{Green}{\large\textbf{$+$0.10}} \large\textsubscript{(0.13)} & \textcolor{Red}{\large\textbf{$-$0.53}} \large\textsubscript{(1.53)} & \textcolor{Red}{\large\textbf{$-$0.17}} \large\textsubscript{(0.62)} & \textcolor{Red}{\large\textbf{$-$0.30}} \large\textsubscript{(0.56)} & \textcolor{Green}{\large\textbf{$+$0.00}} \large\textsubscript{(0.09)} & \textcolor{Green}{\large\textbf{$+$0.03}} \large\textsubscript{(0.09)} & \textcolor{Red}{\large\textbf{$-$0.02}} \large\textsubscript{(0.20)} & \textcolor{Green}{\large\textbf{$+$0.28}} \large\textsubscript{(0.19)} & \textcolor{Red}{\large\textbf{$-$0.10}} \large\textsubscript{(0.32)} & \textcolor{Green}{\large\textbf{$+$0.95}} \large\textsubscript{(0.24)} \\

\bottomrule
\end{tabular}
}
\caption{BLEU score changes of self-language \steer{pmax} intervention from \texttt{en} to target languages with Targeted  and Non-targeted prompts on the FLORES-200 dataset. 
Each values are changes with increases shown in green and decreases in red. The subscripted values to the right of each number indicate the BLEU score without intervention. Q denotes Qwen2.5, G denotes Gemma2 and S denotes SeaLLMs-v3.}
\label{tab:combined_bleu}
\vspace{-10pt}
\end{table*}

\begin{figure}[htbp]
    % \vspace{-5pt}
    \centering
    \includegraphics[width=1.0\linewidth, trim=10 10 0 15, clip]{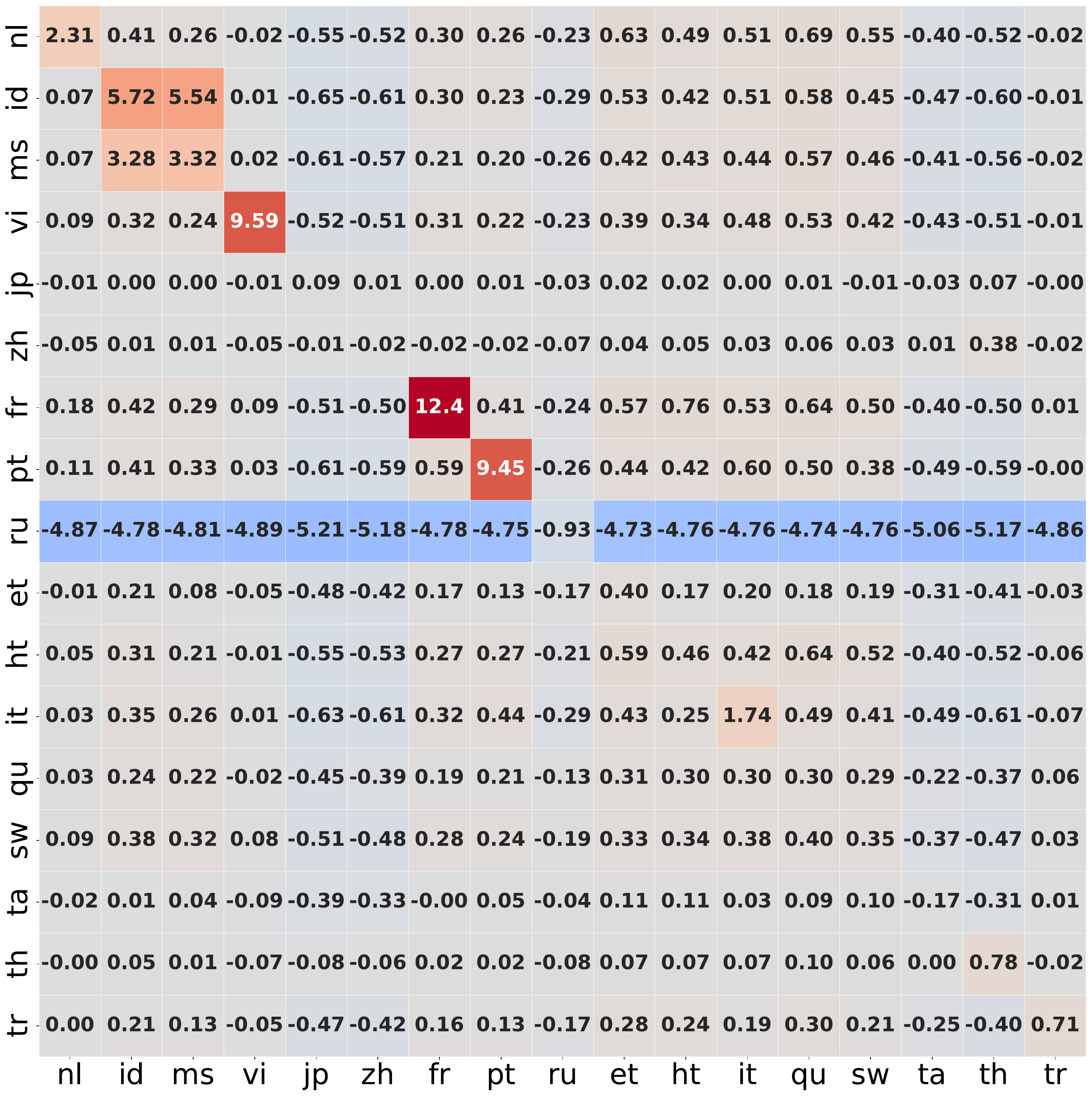}
    \caption{BLEU deltas for \textit{non-targeted} \steer{pmax} intervention in Qwen2.5 0.5B. Row $i$ is the target language and column $j$ is the intervention language. For example, row $x$ col $y$ represents BLEU score when translating English to language $x$ intervened by $y$ neurons.Blue highlights reductions, whereas red highlights increases.}
    \label{fig:bleu-cross}
    % \vspace{-5pt}
\end{figure}

\subsubsection{Translation Task}
% \vspace{-10pt}

Table \ref{tab:combined_bleu} shows that BLEU scores increase after amplifying LAPE neurons with \steer{pmax} for translating English to 17 other languages with non-targeted prompt. The results suggest that amplifying language-specific neurons enhances the model's ability to generate text in the target language, including low-resource languages such as \texttt{qu}, \texttt{ht}, and \texttt{sw}, across most models. In contrast, the improvements from targeted prompts are less pronounced. Smaller models like Qwen2.5 0.5B and SeaLLMs3 1.5B tend to exhibit more consistent gains, which may indicate a higher degree of neuron locality in these models. The difference in performance between targeted and non-targeted prompts may suggest that amplifying these neurons improves the generation of the target language, though not always with high accuracy for translation.
In contrast to amplification of Baseline neurons, which often degrades BLEU scores (Appendix \ref{sec:appendix-translate}), LAPE demonstrates the ability to improve performance by targeting language-specific neurons.
% \begin{figure}[htbp]
%     \centering
%     \includegraphics[width=1\linewidth, trim=10 10 10 20, clip]{figs/ppl-flores/T_max_pt_fixed_gemma-2-9b-it_flores200_ppl_full_csv.pdf}
%     \caption{PPL change with \steer{pmax} steering factor under LAPE neurons on Gemma2 9B. Row $i$ is the input language and column $j$ is the language of intervention.}
%     % \vspace{-10pt}
%     \label{fig:ppl-lape}
% \end{figure}

To support the increasing BLEU findings, particularly in non-targeted prompt, we compare the performance with the cross-intervention setting presented in Figure \ref{fig:bleu-cross}. Self-intervention results in increasing BLEU scores compared to cross-intervention. Notably, using \texttt{ru} as the intervention language reduces performance in the cross-intervention case, while other languages show more modest increase compared to self-intervention. The observed improvements across languages may relate to linguistic similarity. For example, some increases occur when translating to \texttt{id} with \texttt{ms} as the intervention language, and vice versa; or when translating to \texttt{pt} with \texttt{fr} or \texttt{it}, and to \texttt{fr} with \texttt{ht}.

% Other non-similar languages may show increasing BLEU, though less visible. 

% Non-targeted prompt as the benchmark prompt for translation task also show important BLEU improvements in self-intervention (Table \ref{tab:bleu-tgt}). Interestingly, smaller models improve more consistently, align with perplexity findings, possibly implying neuron locality on these models. Gemma2 2B demonstrates greater and more consistent improvement in most languages, including low-resource ones, highlighting the their importance on multilingual capabilities for underrepresented languages. In contrast to amplification of Baseline neurons, which often degrade BLEU scores (Appendix \ref{sec:appendix-translate}), LAPE demonstrates the ability to improve performance by targeting language-specific neurons.
\subsubsection{Perplexity Changes}

% \begin{figure}[htbp]
%     \centering
%     \includegraphics[width=1\linewidth, trim=10 10 10 20, clip]{figs/ppl-flores/T_max_pt_fixed_gemma-2-9b-it_flores200_ppl_full_csv.pdf}
%     \caption{PPL change with \steer{pmax} steering factor under LAPE neurons on Gemma2 9B. Row $i$ is the input language and column $j$ is the language of intervention.}
%     \vspace{-10pt}
%     \label{fig:ppl-lape}
% \end{figure}

\begin{figure}[htbp]
    \centering
    \includegraphics[width=1\linewidth, trim=10 10 10 20, clip]{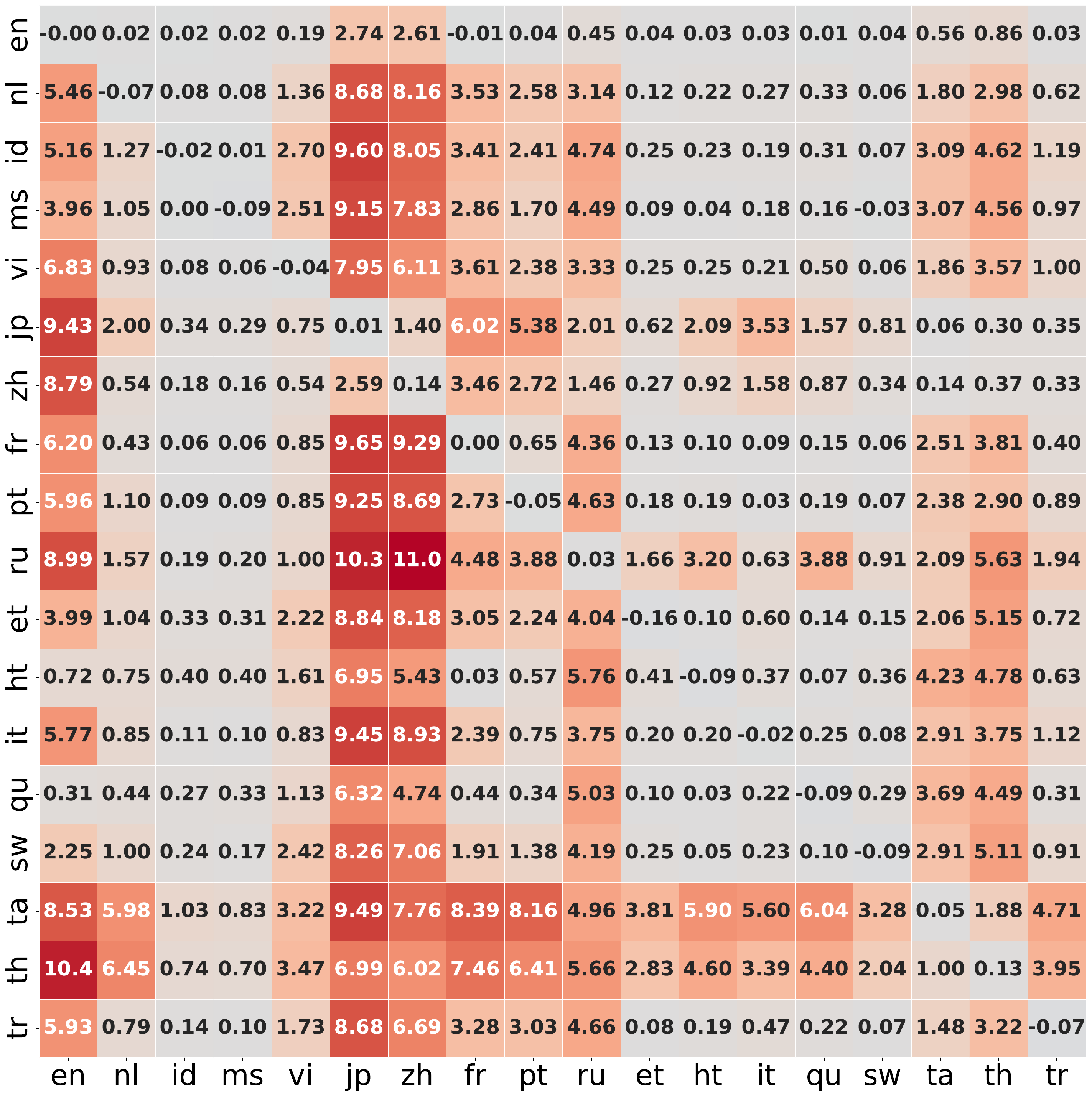}
    \caption{PPL change with \steer{pmax} steering factor under LAPE neurons on Gemma2 9B. Row $i$ is the input language and column $j$ is the language of intervention.}
    \vspace{-10pt}
    \label{fig:ppl-lape}
\end{figure}

Self-intervention of LAPE neurons using \steer{pmax} can modestly reduce perplexity in most languages, showing success in improving language modeling performance as illustrated in Figure \ref{fig:ppl-lape}. Perplexity tends to increase more in cross-intervention settings than in self-intervention. However, the extent of the increase is not always related to the similarity of the language pair; For example, intervention between highly similar and same-family languages like \texttt{id} and \texttt{ms} does not lower perplexity, while less-related pairs like \texttt{fr} on \texttt{en} can. Interestingly, smaller models exhibit larger increases in perplexity, a pattern consistent with findings from the translation task intervention. This may suggest that smaller models contain more localized neurons responsible for language identity (see Appendix \ref{sec:appendix-ppl} for additional results).

% \begin{figure}[htbp]
%     \centering
%     \includegraphics[width=1\linewidth, trim=10 10 10 20, clip]{figs/ppl-flores/T_max_pt_fixed_gemma-2-9b-it_flores200_ppl_full_csv.pdf}
%     \caption{PPL change with \steer{pmax} steering factor under LAPE neurons on Gemma2 9B. Row $i$ is the input language and column $j$ is the language of intervention.}
%     \vspace{-10pt}
%     \label{fig:ppl-lape}
% \end{figure}

The positive effects of perplexity reduction from LAPE neuron amplification are further supported by a comparison with Baseline neuron amplification. Intervention using Baseline neurons shows a similar pattern of increased perplexity in cross-intervention settings, although the increases are more pronounced. In self-intervention, amplification with Baseline neurons barely reduces perplexity, indicating no improvement in language modeling performance. These results highlight the advantage of LAPE neurons, whose specificity contributes to their ability to enhance language modeling performance. Detailed results for Baseline neuron interventions are provided in Appendix \ref{sec:appendix-ppl}.

% \begin{figure}[htbp]
%     \centering
%     \includegraphics[width=1\linewidth, trim=10 10 10 20, clip]{figs/ppl-flores/T_max_pt_fixed_gemma-2-9b-it_flores200_ppl_full_csv.pdf}
%     \caption{PPL change with \steer{pmax} steering factor under LAPE neurons on Gemma2 9B. Row $i$ is the input language and column $j$ is the language of intervention.}
%     \vspace{-10pt}
%     \label{fig:ppl-lape}
% \end{figure}
\subsection{Overlap Patterns in Language-Specific Neurons}

\begin{figure}[htbp]
\centering
    \includegraphics[width=\linewidth, trim=130 70 70 120, clip]{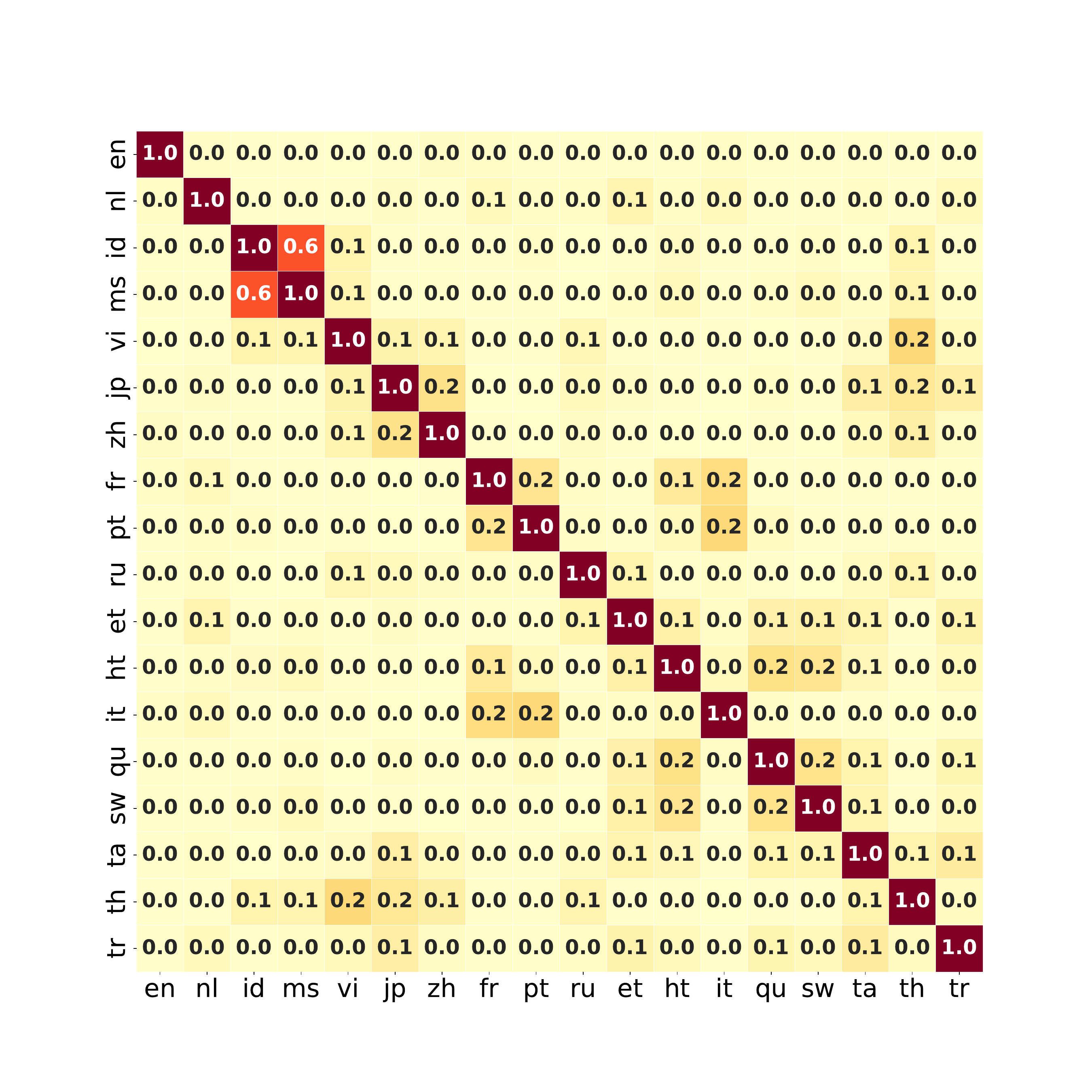}
\vspace{-10pt}
\caption{Symmetric heatmaps of overlap LAPE neuron proportions in SeaLLMs3 7B using jaccard distance.}
\vspace{-10pt}
\label{fig:overlaps}
\end{figure}

As illustrated in Figure \ref{fig:overlaps}, although LAPE neurons are inherently tailored to individual languages, overlaps still exist across languages, albeit at significantly lower proportions. Compared with Baseline neurons with higher overlap proportions (Appendix \ref{sec:overlap}), the stronger specificity of LAPE neurons significantly improve self-intervention and capture patterns in cross-intervention for perplexity and task performance, indicating that better specificity leads to better performance for neuron-level intervention. 
% \begin{figure}[htbp]
%     \centering
%     \includegraphics[width=1\linewidth, trim=10 10 10 20, clip]{figs/ppl-flores/T_max_pt_fixed_gemma-2-9b-it_flores200_ppl_full_csv.pdf}
%     \caption{PPL change with \steer{pmax} steering factor under LAPE neurons on Gemma2 9B. Row $i$ is the input language and column $j$ is the language of intervention.}
%     \vspace{-10pt}
%     \label{fig:ppl-lape}
% \end{figure}

A strong overlap is captured in LAPE neurons across mutually intelligible languages like Indonesian (\texttt{id}) and Malaysian (\texttt{ms}), likely due to their linguistic similarity \cite{ahmad2011kesenjangan}. Languages sharing writing systems, such as Chinese (\texttt{zh}) and Japanese (\texttt{jp}) \cite{taylor1995writing}, also show overlap, consistent with \citet{kojima-etal-2024-multilingual}. Romance languages, i.e., Italian (\texttt{it}), French (\texttt{fr}), and Portuguese (\texttt{pt}), share neurons, and Vietnamese (\texttt{vi}) overlaps moderately with Thai (\texttt{th}) and slightly with \texttt{id}, \texttt{ms}, \texttt{jp}, and \texttt{zh}, likely due to regional or historical ties. Haitian Creole (\texttt{ht}) overlaps with French, reflecting shared roots \cite{prou2012haitian}. Surprisingly, low-resource languages like Quechua (\texttt{qu}), Haitian Creole (\texttt{ht}), and Swahili (\texttt{sw}) also show overlap, possibly indicating more general or shared neurons.

\section{Conclusion}
This study proposes the Language Steering Shift (LSS) metric to compare steering factors of language-specific (LAPE) neuron intervention and identify the optimal one, then evaluates its impact in downstream tasks and language modeling. Optimal amplification steers outputs toward target languages with over $90\%$ success on average.
% Self-language intervention of knowledge and reasoning tasks, translation, and language model tends to modestly improve or maintain performance. Cross-intervention on the other hand, generally decreases performance, with some increases can be captured between highly related languages for perplexity and BLEU score for translation. Compared to language-activated (less specific) neurons, LAPE neurons show clearer and more consistent improvements, demonstrating that higher neuron specificity leads to greater potential for multilingual enhancement. 
Self-language intervention modestly improves or maintains performance in language modeling measured by perplexity, translation, knowledge, and reasoning tasks. Cross-language intervention generally lowers performance, though gains can appear between closely related languages in perplexity and BLEU scores for translation. Compared to less specific language-activated neurons, LAPE neurons yield clearer, more consistent improvements, suggesting that greater neuron specificity enhances multilingual performance when intervened. 
Overall, these findings suggest that amplifying  LAPE neurons in self-language can enhance a model’s language capabilities, although not always improving task accuracy. Furthermore, the consistent degradation from cross-language interventions highlights their limited effectiveness for cross-lingual transfer.

% using this amplification factor improves perplexity and translation performance, especially in smaller models; while cross-language intervention often degrades it, except between closely related languages, likely due to neuron overlap. For knowledge and reasoning tasks (XCOPA, XWinograd, and Include-lite), self-language intervention tends to modestly improve or maintain performance, while cross-language intervention generally reduces it. Compared to language-activated (less specific) neurons, LAPE neurons show clearer and more consistent improvements, demonstrating that higher neuron specificity leads to greater potential for multilingual enhancement. Overall, these findings suggest that amplifying self-language LAPE neurons can enhance a model’s language capabilities, although not always improving task accuracy. The consistent degradation from cross-language interventions underscores their limited effectiveness for cross-lingual transfer.

% LAPE neurons show partial overlap across similar languages which may influence intervention performance in cross-lingual setting.

\section*{Limitations}
This study is limited to decoder-only models in 0.5B to 9B parameters and limited sets of tasks. Methods to identify language-specific neurons are also limited to LAPE, there are still gaps on the effect of amplification on neurons obtained by fine-grained attribution methods such as Integrated Gradients. 

% \section*{Acknowledgments}

\bibliography{custom}

\begin{thebibliography}{25}
\providecommand{\natexlab}[1]{#1}

\bibitem[{Ahmad and Pustaka(2011)}]{ahmad2011kesenjangan}
I.H. Ahmad and Dewan~Bahasa Pustaka. 2011.
\newblock \href {https://books.google.co.id/books?id=ZpnlOCB8htQC} {\emph{Kesenjangan leksikal bahasa Melayu Malaysia dan bahasa Indonesia}}.
\newblock Dewan Bahasa dan Pustaka.

\bibitem[{Conneau et~al.(2020)Conneau, Khandelwal, Goyal, Chaudhary, Wenzek, Guzm{\'a}n, Grave, Ott, Zettlemoyer, and Stoyanov}]{conneau-etal-2020-unsupervised}
Alexis Conneau, Kartikay Khandelwal, Naman Goyal, Vishrav Chaudhary, Guillaume Wenzek, Francisco Guzm{\'a}n, Edouard Grave, Myle Ott, Luke Zettlemoyer, and Veselin Stoyanov. 2020.
\newblock \href {https://doi.org/10.18653/v1/2020.acl-main.747} {Unsupervised cross-lingual representation learning at scale}.
\newblock In \emph{Proceedings of the 58th Annual Meeting of the Association for Computational Linguistics}, pages 8440--8451, Online. Association for Computational Linguistics.

\bibitem[{Dai et~al.(2022)Dai, Dong, Hao, Sui, Chang, and Wei}]{dai-etal-2022-knowledge}
Damai Dai, Li~Dong, Yaru Hao, Zhifang Sui, Baobao Chang, and Furu Wei. 2022.
\newblock \href {https://doi.org/10.18653/v1/2022.acl-long.581} {Knowledge neurons in pretrained transformers}.
\newblock In \emph{Proceedings of the 60th Annual Meeting of the Association for Computational Linguistics (Volume 1: Long Papers)}, pages 8493--8502, Dublin, Ireland. Association for Computational Linguistics.

\bibitem[{Dang et~al.(2024)Dang, Huang, Huo, Yan, Huang, Liu, Gao, Zhang, Qian, Wang, Liu, Shao, Xiong, and Hu}]{dang2024explainableinterpretablemultimodallarge}
Yunkai Dang, Kaichen Huang, Jiahao Huo, Yibo Yan, Sirui Huang, Dongrui Liu, Mengxi Gao, Jie Zhang, Chen Qian, Kun Wang, Yong Liu, Jing Shao, Hui Xiong, and Xuming Hu. 2024.
\newblock \href {https://arxiv.org/abs/2412.02104} {Explainable and interpretable multimodal large language models: A comprehensive survey}.
\newblock \emph{Preprint}, arXiv:2412.02104.

\bibitem[{Edoardo M.~Ponti and Korhonen(2020)}]{ponti2020xcopa}
Olga Majewska Qianchu Liu Ivan~Vuli'{c} Edoardo M.~Ponti, Goran Glava~{s} and Anna Korhonen. 2020.
\newblock \href {https://ducdauge.github.io/files/xcopa.pdf} {{XCOPA: A} multilingual dataset for causal commonsense reasoning}.
\newblock \emph{arXiv preprint}.

\bibitem[{Kassner et~al.(2021)Kassner, Dufter, and Sch{\"{u}}tze}]{kassner2021multilingual}
Nora Kassner, Philipp Dufter, and Hinrich Sch{\"{u}}tze. 2021.
\newblock \href {https://arxiv.org/abs/2102.00894} {Multilingual {LAMA:} investigating knowledge in multilingual pretrained language models}.
\newblock \emph{CoRR}, abs/2102.00894.
\newblock To appear in EACL2021.

\bibitem[{Kim et~al.(2019)Kim, Gao, and Ney}]{kim-etal-2019-effective}
Yunsu Kim, Yingbo Gao, and Hermann Ney. 2019.
\newblock \href {https://doi.org/10.18653/v1/P19-1120} {Effective cross-lingual transfer of neural machine translation models without shared vocabularies}.
\newblock In \emph{Proceedings of the 57th Annual Meeting of the Association for Computational Linguistics}, pages 1246--1257, Florence, Italy. Association for Computational Linguistics.

\bibitem[{Kojima et~al.(2024)Kojima, Okimura, Iwasawa, Yanaka, and Matsuo}]{kojima-etal-2024-multilingual}
Takeshi Kojima, Itsuki Okimura, Yusuke Iwasawa, Hitomi Yanaka, and Yutaka Matsuo. 2024.
\newblock \href {https://doi.org/10.18653/v1/2024.naacl-long.384} {On the multilingual ability of decoder-based pre-trained language models: Finding and controlling language-specific neurons}.
\newblock In \emph{Proceedings of the 2024 Conference of the North American Chapter of the Association for Computational Linguistics: Human Language Technologies (Volume 1: Long Papers)}, pages 6919--6971, Mexico City, Mexico. Association for Computational Linguistics.

\bibitem[{Leng and Xiong(2025)}]{leng2025understandingmultitasklearninggeneralization}
Yongqi Leng and Deyi Xiong. 2025.
\newblock \href {https://arxiv.org/abs/2407.06488} {Towards understanding multi-task learning (generalization) of llms via detecting and exploring task-specific neurons}.
\newblock \emph{Preprint}, arXiv:2407.06488.

\bibitem[{Li et~al.(2022)Li, Kuncoro, Hoffmann, de~Masson~d'Autume, Blunsom, and Nematzadeh}]{li2022systematicinvestigationcommonsenseknowledge}
Xiang~Lorraine Li, Adhiguna Kuncoro, Jordan Hoffmann, Cyprien de~Masson~d'Autume, Phil Blunsom, and Aida Nematzadeh. 2022.
\newblock \href {https://arxiv.org/abs/2111.00607} {A systematic investigation of commonsense knowledge in large language models}.
\newblock \emph{Preprint}, arXiv:2111.00607.

\bibitem[{Meng et~al.(2023)Meng, Bau, Andonian, and Belinkov}]{meng2023locatingeditingfactualassociations}
Kevin Meng, David Bau, Alex Andonian, and Yonatan Belinkov. 2023.
\newblock \href {https://arxiv.org/abs/2202.05262} {Locating and editing factual associations in gpt}.
\newblock \emph{Preprint}, arXiv:2202.05262.

\bibitem[{Mondal et~al.(2025)Mondal, Sen, Singhania, and Jyothi}]{mondal-etal-2025-language}
Soumen~Kumar Mondal, Sayambhu Sen, Abhishek Singhania, and Preethi Jyothi. 2025.
\newblock \href {https://aclanthology.org/2025.insights-1.6/} {Language-specific neurons do not facilitate cross-lingual transfer}.
\newblock In \emph{The Sixth Workshop on Insights from Negative Results in NLP}, pages 46--62, Albuquerque, New Mexico. Association for Computational Linguistics.

\bibitem[{Muennighoff et~al.(2022)Muennighoff, Wang, Sutawika, Roberts, Biderman, Scao, Bari, Shen, Yong, Schoelkopf, Tang, Radev, Aji, Almubarak, Albanie, Alyafeai, Webson, Raff, and Raffel}]{muennighoff2022crosslingual}
Niklas Muennighoff, Thomas Wang, Lintang Sutawika, Adam Roberts, Stella Biderman, Teven~Le Scao, M~Saiful Bari, Sheng Shen, Zheng-Xin Yong, Hailey Schoelkopf, Xiangru Tang, Dragomir Radev, Alham~Fikri Aji, Khalid Almubarak, Samuel Albanie, Zaid Alyafeai, Albert Webson, Edward Raff, and Colin Raffel. 2022.
\newblock \href {https://arxiv.org/abs/2211.01786} {Crosslingual generalization through multitask finetuning}.
\newblock \emph{Preprint}, arXiv:2211.01786.

\bibitem[{Prou et~al.(2012)Prou, Spears, and Joseph}]{prou2012haitian}
M.~Prou, A.K. Spears, and C.M.B. Joseph. 2012.
\newblock \href {https://books.google.co.id/books?id=4xbGzLuBvWwC} {\emph{The Haitian Creole Language: History, Structure, Use, and Education}}.
\newblock Caribbean Studies. Bloomsbury Academic.

\bibitem[{Qwen et~al.(2025)Qwen, :, Yang, Yang, Zhang, Hui, Zheng, Yu, Li, Liu, Huang, Wei, Lin, Yang, Tu, Zhang, Yang, Yang, Zhou, Lin, Dang, Lu, Bao, Yang, Yu, Li, Xue, Zhang, Zhu, Men, Lin, Li, Tang, Xia, Ren, Ren, Fan, Su, Zhang, Wan, Liu, Cui, Zhang, and Qiu}]{qwen2025qwen25technicalreport}
Qwen, :, An~Yang, Baosong Yang, Beichen Zhang, Binyuan Hui, Bo~Zheng, Bowen Yu, Chengyuan Li, Dayiheng Liu, Fei Huang, Haoran Wei, Huan Lin, Jian Yang, Jianhong Tu, Jianwei Zhang, Jianxin Yang, Jiaxi Yang, Jingren Zhou, and 25 others. 2025.
\newblock \href {https://arxiv.org/abs/2412.15115} {Qwen2.5 technical report}.
\newblock \emph{Preprint}, arXiv:2412.15115.

\bibitem[{Romanou et~al.(2024)Romanou, Foroutan, Sotnikova, Chen, Nelaturu, Singh, Maheshwary, Altomare, Haggag, Amayuelas et~al.}]{romanou2024include}
Angelika Romanou, Negar Foroutan, Anna Sotnikova, Zeming Chen, Sree~Harsha Nelaturu, Shivalika Singh, Rishabh Maheshwary, Micol Altomare, Mohamed~A Haggag, Alfonso Amayuelas, and 1 others. 2024.
\newblock Include: Evaluating multilingual language understanding with regional knowledge.
\newblock \emph{arXiv preprint arXiv:2411.19799}.

\bibitem[{Song et~al.(2024)Song, He, Jiang, Xian, Gao, Liu, and Yu}]{song-etal-2024-large}
Ran Song, Shizhu He, Shuting Jiang, Yantuan Xian, Shengxiang Gao, Kang Liu, and Zhengtao Yu. 2024.
\newblock \href {https://doi.org/10.18653/v1/2024.emnlp-main.403} {Does large language model contain task-specific neurons?}
\newblock In \emph{Proceedings of the 2024 Conference on Empirical Methods in Natural Language Processing}, pages 7101--7113, Miami, Florida, USA. Association for Computational Linguistics.

\bibitem[{Tang et~al.(2024)Tang, Luo, Huang, Zhang, Wang, Zhao, Wei, and Wen}]{tang2024languagespecificneuronskeymultilingual}
Tianyi Tang, Wenyang Luo, Haoyang Huang, Dongdong Zhang, Xiaolei Wang, Xin Zhao, Furu Wei, and Ji-Rong Wen. 2024.
\newblock \href {https://arxiv.org/abs/2402.16438} {Language-specific neurons: The key to multilingual capabilities in large language models}.
\newblock \emph{Preprint}, arXiv:2402.16438.

\bibitem[{Taylor et~al.(1995)Taylor, Taylor, and Taylor}]{taylor1995writing}
I.~Taylor, M.M. Taylor, and M.M. Taylor. 1995.
\newblock \href {https://books.google.co.id/books?id=WDw4gBaPjZgC} {\emph{Writing and Literacy in Chinese, Korean and Japanese}}.
\newblock Studies in written language and literacy. John Benjamins Publishing Company.

\bibitem[{Team et~al.(2024)Team, Riviere, Pathak, Sessa, Hardin, Bhupatiraju, Hussenot, Mesnard, Shahriari, Ramé, Ferret, Liu, Tafti, Friesen, Casbon, Ramos, Kumar, Lan, Jerome, Tsitsulin, Vieillard, Stanczyk, Girgin, Momchev, Hoffman, Thakoor, Grill, Neyshabur, Bachem, Walton, Severyn, Parrish, Ahmad, Hutchison, Abdagic, Carl, Shen, Brock, Coenen, Laforge, Paterson, Bastian, Piot, Wu, Royal, Chen, Kumar, Perry, Welty, Choquette-Choo, Sinopalnikov, Weinberger, Vijaykumar, Rogozińska, Herbison, Bandy, Wang, Noland, Moreira, Senter, Eltyshev, Visin, Rasskin, Wei, Cameron, Martins, Hashemi, Klimczak-Plucińska, Batra, Dhand, Nardini, Mein, Zhou, Svensson, Stanway, Chan, Zhou, Carrasqueira, Iljazi, Becker, Fernandez, van Amersfoort, Gordon, Lipschultz, Newlan, yeong Ji, Mohamed, Badola, Black, Millican, McDonell, Nguyen, Sodhia, Greene, Sjoesund, Usui, Sifre, Heuermann, Lago, McNealus, Soares, Kilpatrick, Dixon, Martins, Reid, Singh, Iverson, Görner, Velloso, Wirth, Davidow, Miller, Rahtz, Watson, Risdal,
  Kazemi, Moynihan, Zhang, Kahng, Park, Rahman, Khatwani, Dao, Bardoliwalla, Devanathan, Dumai, Chauhan, Wahltinez, Botarda, Barnes, Barham, Michel, Jin, Georgiev, Culliton, Kuppala, Comanescu, Merhej, Jana, Rokni, Agarwal, Mullins, Saadat, Carthy, Cogan, Perrin, Arnold, Krause, Dai, Garg, Sheth, Ronstrom, Chan, Jordan, Yu, Eccles, Hennigan, Kocisky, Doshi, Jain, Yadav, Meshram, Dharmadhikari, Barkley, Wei, Ye, Han, Kwon, Xu, Shen, Gong, Wei, Cotruta, Kirk, Rao, Giang, Peran, Warkentin, Collins, Barral, Ghahramani, Hadsell, Sculley, Banks, Dragan, Petrov, Vinyals, Dean, Hassabis, Kavukcuoglu, Farabet, Buchatskaya, Borgeaud, Fiedel, Joulin, Kenealy, Dadashi, and Andreev}]{gemmateam2024gemma2improvingopen}
Gemma Team, Morgane Riviere, Shreya Pathak, Pier~Giuseppe Sessa, Cassidy Hardin, Surya Bhupatiraju, Léonard Hussenot, Thomas Mesnard, Bobak Shahriari, Alexandre Ramé, Johan Ferret, Peter Liu, Pouya Tafti, Abe Friesen, Michelle Casbon, Sabela Ramos, Ravin Kumar, Charline~Le Lan, Sammy Jerome, and 179 others. 2024.
\newblock \href {https://arxiv.org/abs/2408.00118} {Gemma 2: Improving open language models at a practical size}.
\newblock \emph{Preprint}, arXiv:2408.00118.

\bibitem[{Team et~al.(2022)Team, Costa-jussà, Cross, Çelebi, Elbayad, Heafield, Heffernan, Kalbassi, Lam, Licht, Maillard, Sun, Wang, Wenzek, Youngblood, Akula, Barrault, Gonzalez, Hansanti, Hoffman, Jarrett, Sadagopan, Rowe, Spruit, Tran, Andrews, Ayan, Bhosale, Edunov, Fan, Gao, Goswami, Guzmán, Koehn, Mourachko, Ropers, Saleem, Schwenk, and Wang}]{nllbteam2022languageleftbehindscaling}
NLLB Team, Marta~R. Costa-jussà, James Cross, Onur Çelebi, Maha Elbayad, Kenneth Heafield, Kevin Heffernan, Elahe Kalbassi, Janice Lam, Daniel Licht, Jean Maillard, Anna Sun, Skyler Wang, Guillaume Wenzek, Al~Youngblood, Bapi Akula, Loic Barrault, Gabriel~Mejia Gonzalez, Prangthip Hansanti, and 20 others. 2022.
\newblock \href {https://arxiv.org/abs/2207.04672} {No language left behind: Scaling human-centered machine translation}.
\newblock \emph{Preprint}, arXiv:2207.04672.

\bibitem[{Wang et~al.(2024)Wang, Haddow, Wu, Peng, and Birch}]{wang2024sharingmattersanalysingneurons}
Weixuan Wang, Barry Haddow, Minghao Wu, Wei Peng, and Alexandra Birch. 2024.
\newblock \href {https://arxiv.org/abs/2406.09265} {Sharing matters: Analysing neurons across languages and tasks in llms}.
\newblock \emph{Preprint}, arXiv:2406.09265.

\bibitem[{Zhang and Nanda(2024)}]{zhang2024bestpracticesactivationpatching}
Fred Zhang and Neel Nanda. 2024.
\newblock \href {https://arxiv.org/abs/2309.16042} {Towards best practices of activation patching in language models: Metrics and methods}.
\newblock \emph{Preprint}, arXiv:2309.16042.

\bibitem[{Zhang et~al.(2024)Zhang, Chan, Zhao, Aljunied, Wang, Liu, Deng, Hu, Xu, Chia, Li, and Bing}]{zhang2024seallms3openfoundation}
Wenxuan Zhang, Hou~Pong Chan, Yiran Zhao, Mahani Aljunied, Jianyu Wang, Chaoqun Liu, Yue Deng, Zhiqiang Hu, Weiwen Xu, Yew~Ken Chia, Xin Li, and Lidong Bing. 2024.
\newblock \href {https://arxiv.org/abs/2407.19672} {Seallms 3: Open foundation and chat multilingual large language models for southeast asian languages}.
\newblock \emph{Preprint}, arXiv:2407.19672.

\bibitem[{Zhao et~al.(2024)Zhao, Zhang, Chen, Kawaguchi, and Bing}]{zhao2024largelanguagemodelshandle}
Yiran Zhao, Wenxuan Zhang, Guizhen Chen, Kenji Kawaguchi, and Lidong Bing. 2024.
\newblock \href {https://arxiv.org/abs/2402.18815} {How do large language models handle multilingualism?}
\newblock \emph{Preprint}, arXiv:2402.18815.

\end{thebibliography}

\appendix

\section{Ablation Study}

\subsection{Layer-Wise Amplification on LSS Score}
\label{sec:appendix-layerwise}
\begin{figure}[t]
    \centering
    \includegraphics[width=1\linewidth]{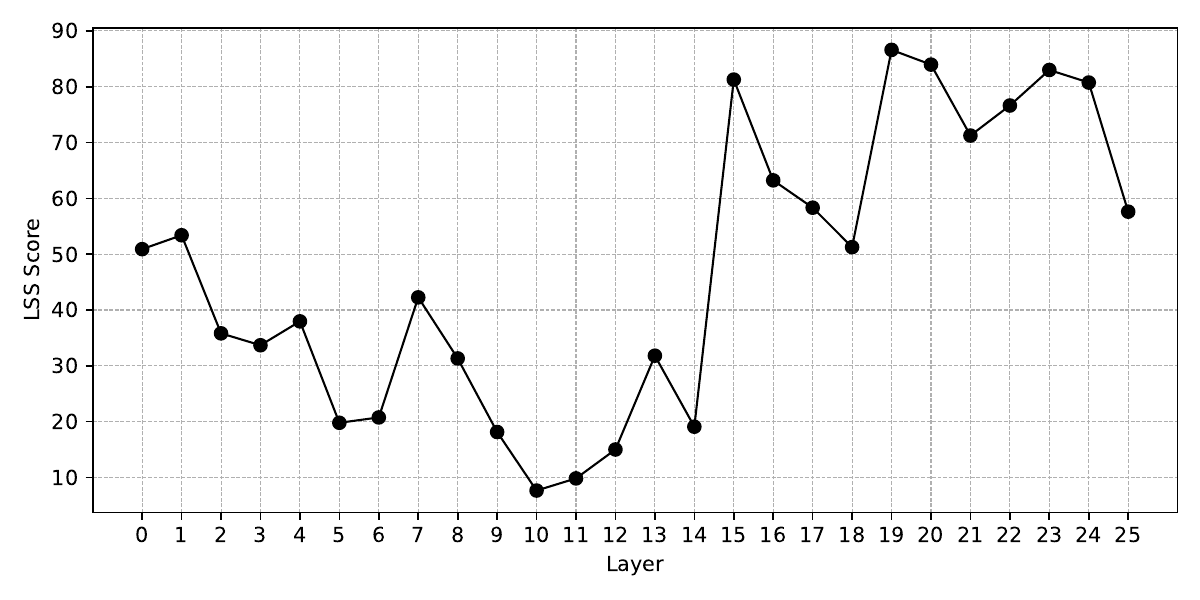}
    \caption{Layer-wise average LSS scores across all languages for Gemma2 2B.}
    \label{fig:perlayer}
\end{figure}

We perform layer-wise amplification on Gemma2 2B model (26 layers) on LSS score using the reconstructed MLAMA dataset. Consistent throughout each language of intervention as captured by the average LSS scores across all languages in Figure \ref{fig:perlayer}, the highest average is found in the 19th layer, and the lowest is in the 10th layer.

Later layers are more likely to result in higher LSS score, especially from the 15th layer until just before the last layer. First few layers may also result in high LSS scores, though not as high. The middle layers result in relatively low LSS scores, often reaching 0 in most languages, with the exception of \texttt{tr} whose the lowest LSS is found in the 18th layer. We observe that some languages from the same family share similar layer-wise LSS scores, such as \texttt{id}, \texttt{ms}, \texttt{vi} and \texttt{pt}, \texttt{it}, \texttt{fr}.

\begin{figure}[t]
    \centering
    \includegraphics[width=1\linewidth]{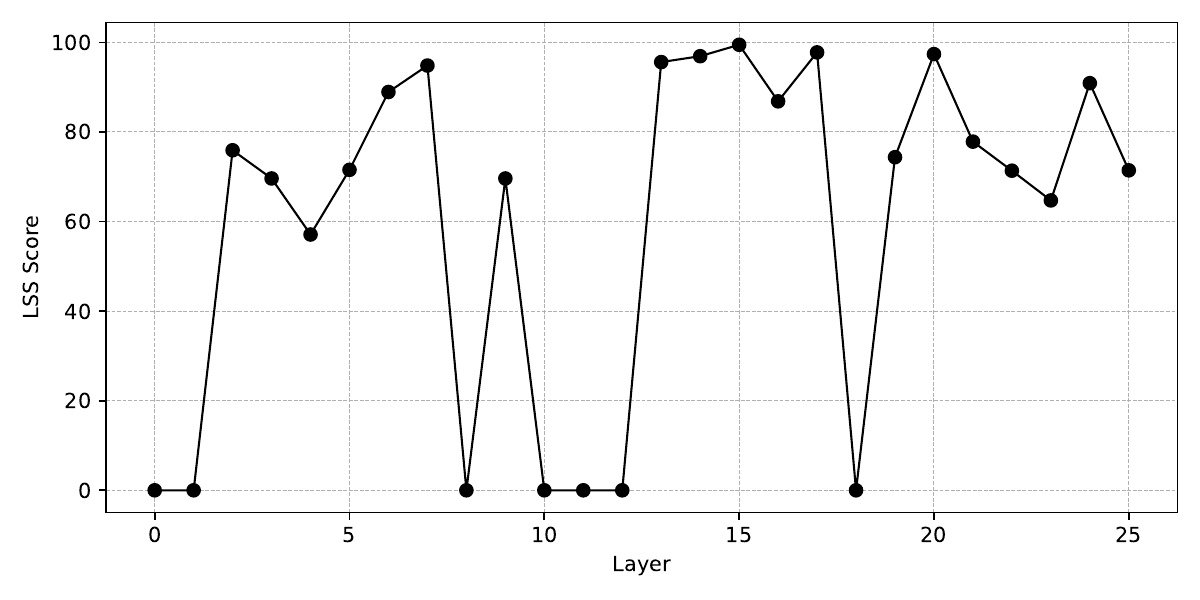}
    \caption{Layer-wise average LSS scores when intervened by \texttt{en} neurons for Gemma2 2B.}
    \label{fig:perlayer-en}
\end{figure}

\begin{figure}[t]
    \centering
    \includegraphics[width=1\linewidth]{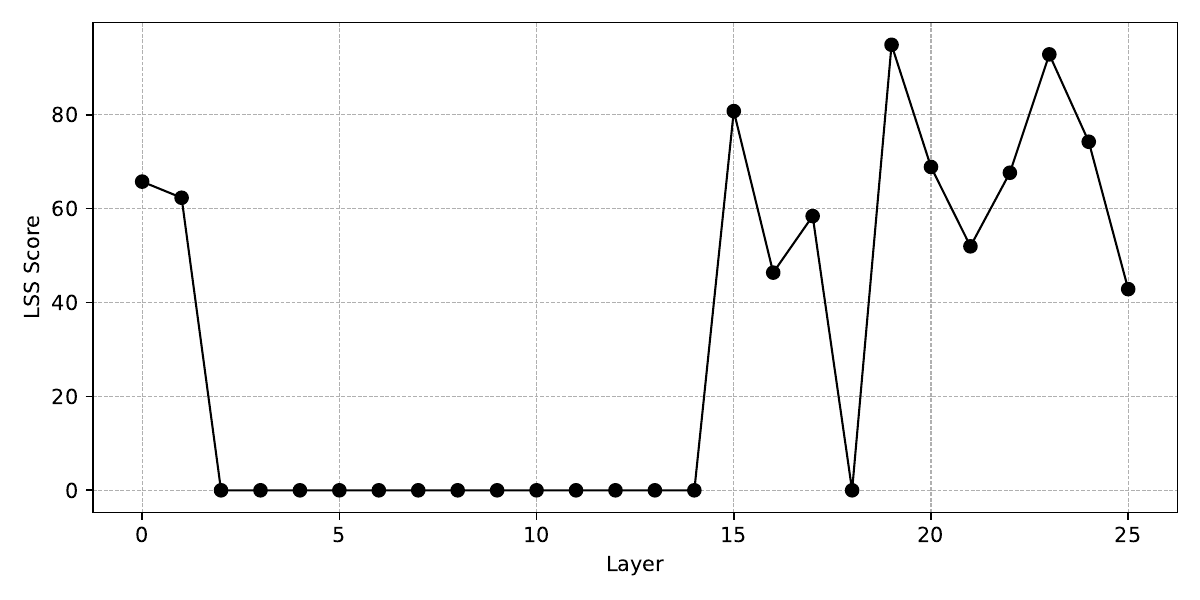}
    \caption{Layer-wise average LSS scores when intervened by \texttt{ms} neurons for Gemma2 2B.}
    \label{fig:perlayer-ms}
\end{figure}

\begin{figure}[t]
    \centering
    \includegraphics[width=1\linewidth]{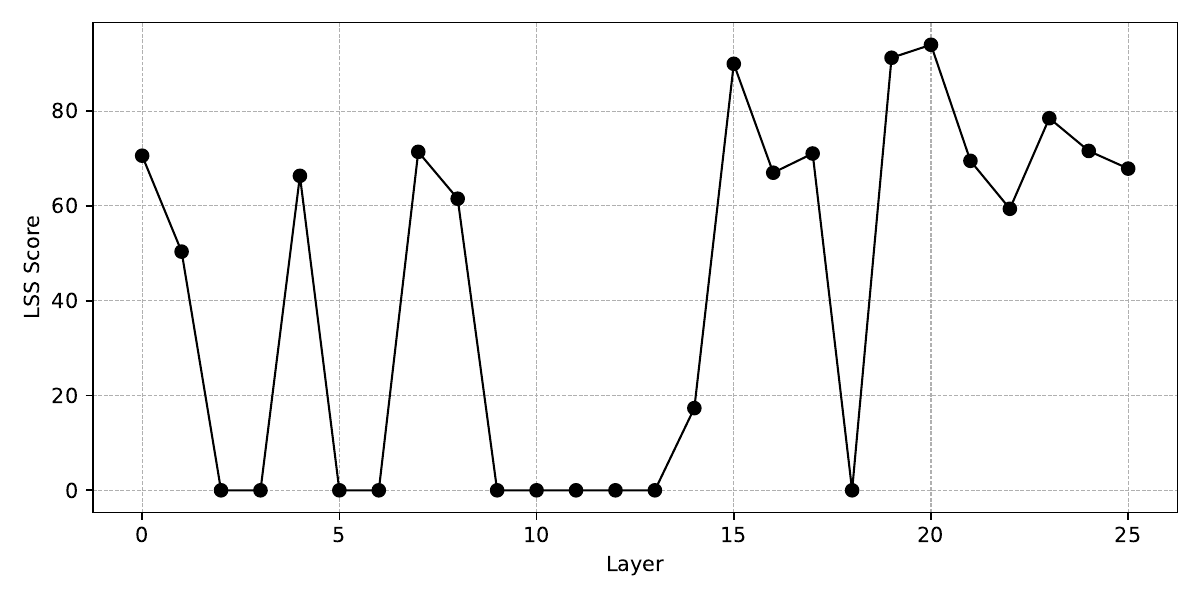}
    \caption{Layer-wise average LSS scores when intervened by \texttt{et} neurons for Gemma2 2B.}
    \label{fig:perlayer-et}
\end{figure}

\begin{figure}[t]
    \centering
    \includegraphics[width=1\linewidth]{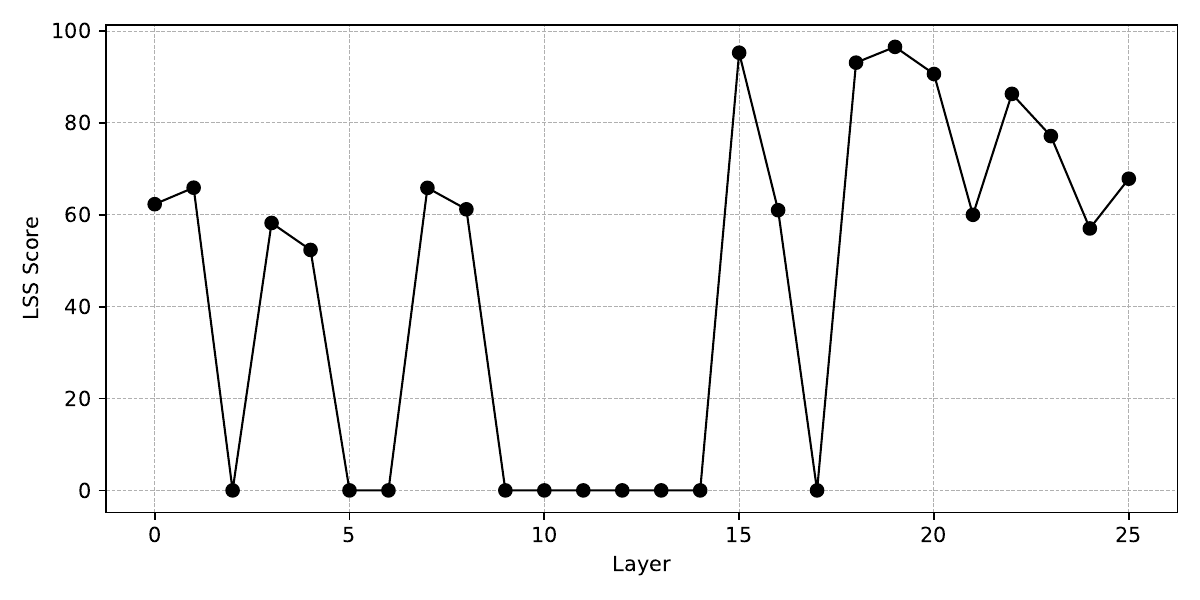}
    \caption{Layer-wise average LSS scores when intervened by \texttt{nl} neurons for Gemma2 2B.}
    \label{fig:perlayer-nl}
\end{figure}

\begin{figure}[t]
    \centering
    \includegraphics[width=1\linewidth]{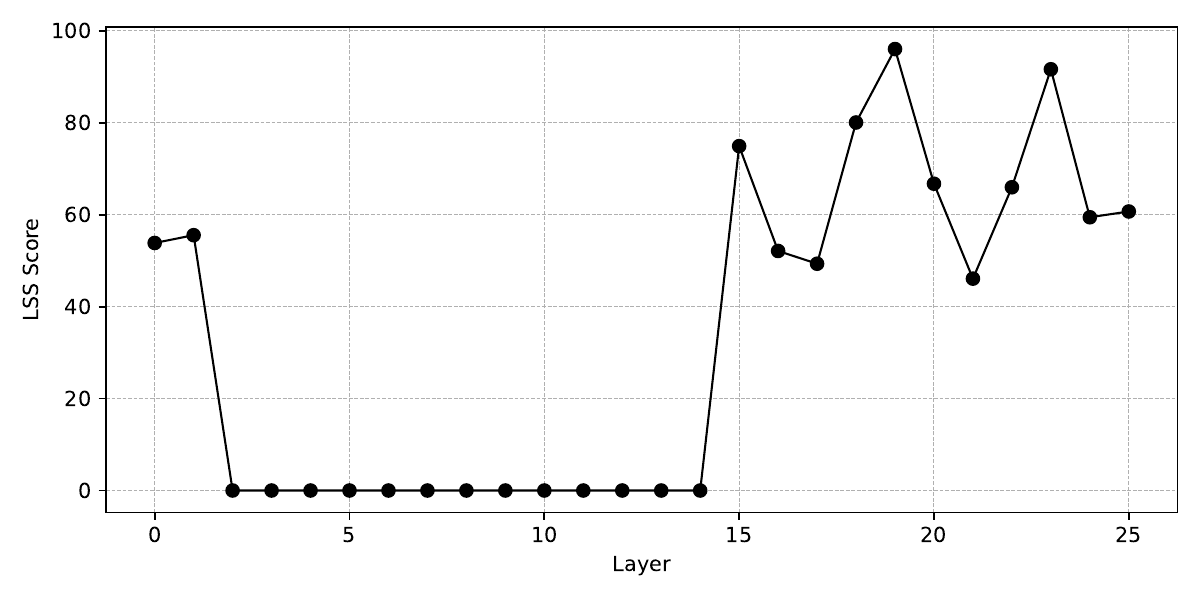}
    \caption{Layer-wise average LSS scores when intervened by \texttt{id} neurons for Gemma2 2B.}
    \label{fig:perlayer-id}
\end{figure}

\begin{figure}[t]
    \centering
    \includegraphics[width=1\linewidth]{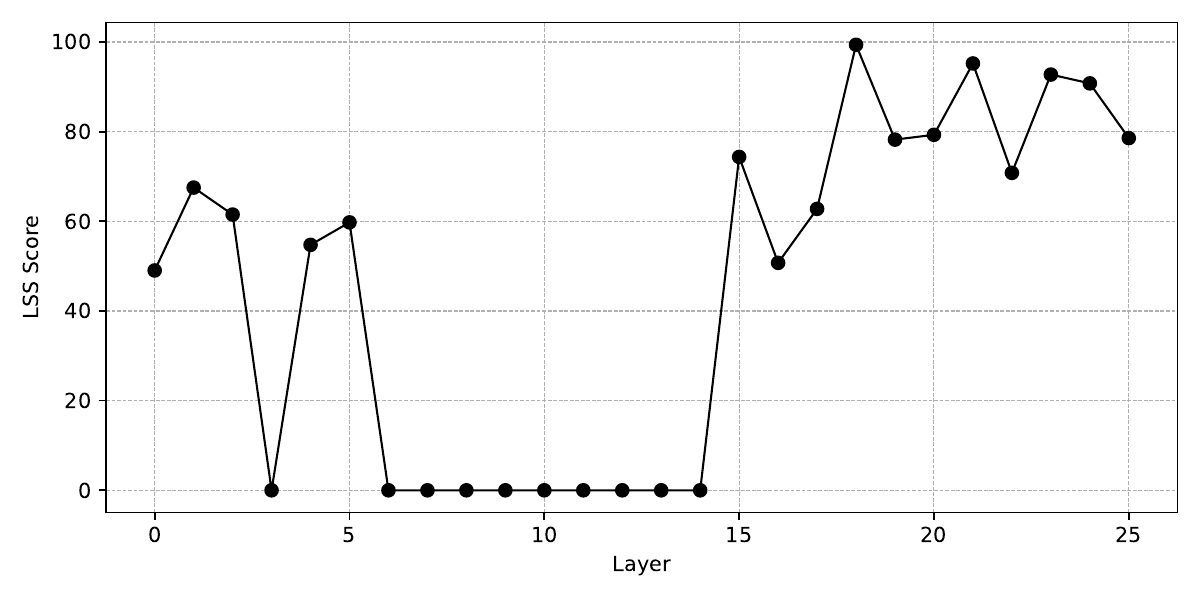}
    \caption{Layer-wise average LSS scores when intervened by \texttt{vi} neurons for Gemma2 2B.}
    \label{fig:perlayer-vi}
\end{figure}

\begin{figure}[t]
    \centering
    \includegraphics[width=1\linewidth]{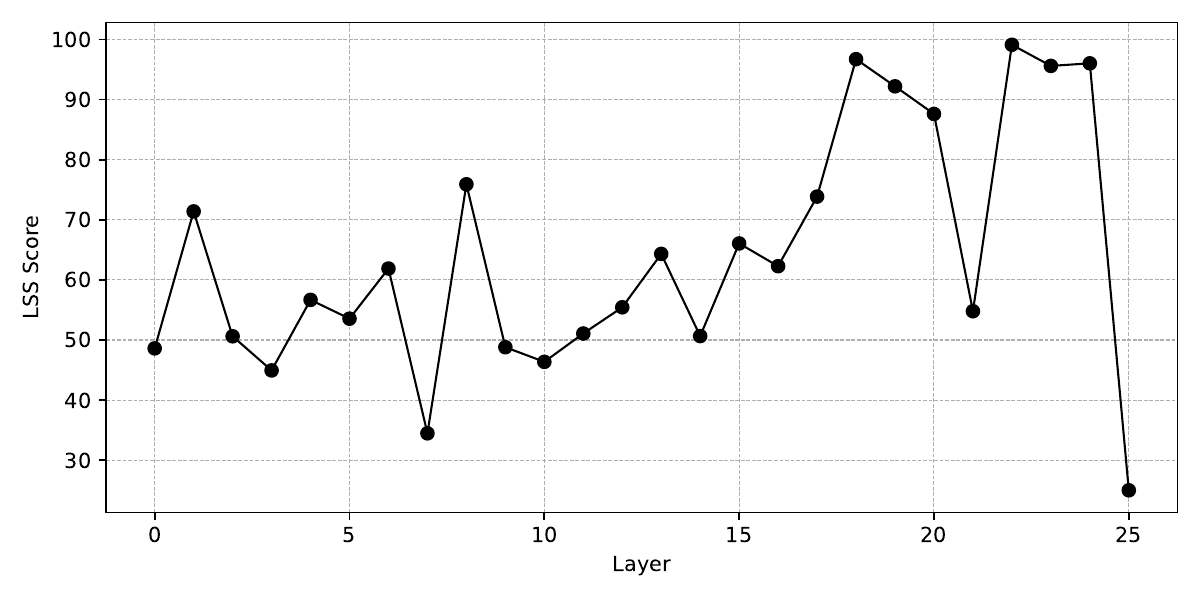}
    \caption{Layer-wise average LSS scores when intervened by \texttt{jp} neurons for Gemma2 2B.}
    \label{fig:perlayer-jp}
\end{figure}

\begin{figure}[t]
    \centering
    \includegraphics[width=1\linewidth]{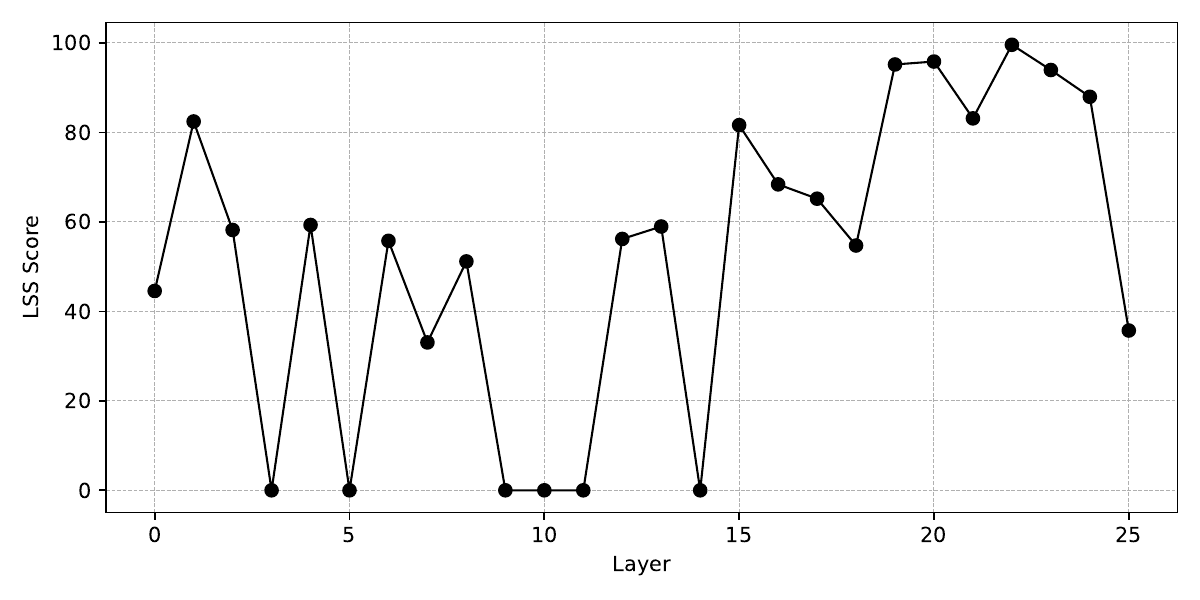}
    \caption{Layer-wise average LSS scores when intervened by \texttt{zh} neurons for Gemma2 2B.}
    \label{fig:perlayer-zh}
\end{figure}

\begin{figure}[t]
    \centering
    \includegraphics[width=1\linewidth]{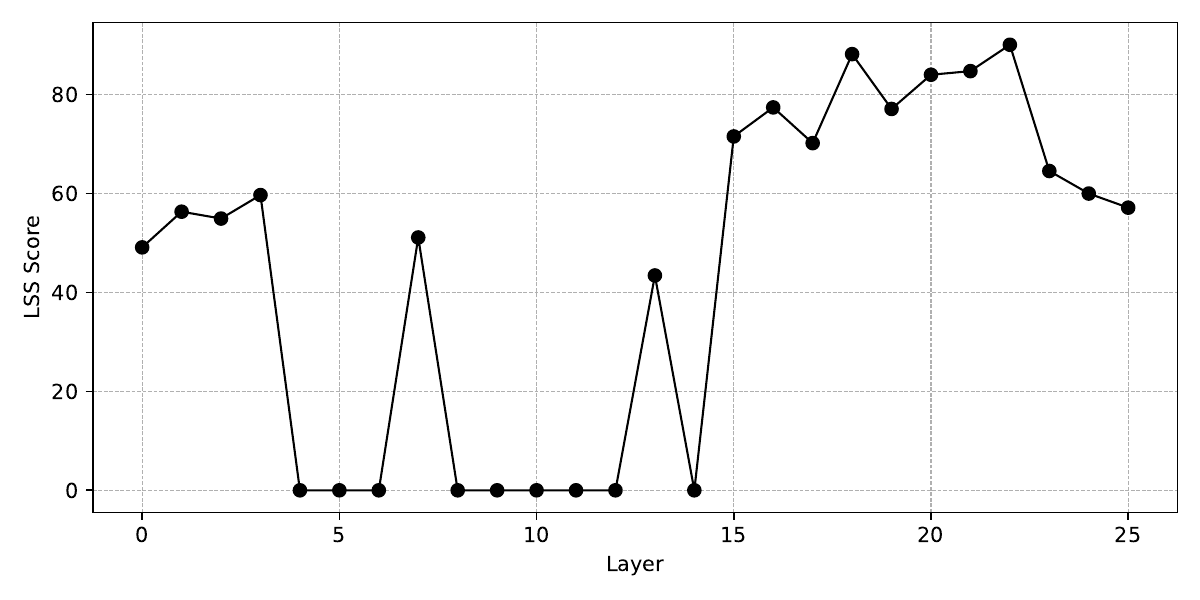}
    \caption{Layer-wise average LSS scores when intervened by \texttt{fr} neurons for Gemma2 2B.}
    \label{fig:perlayer-fr}
\end{figure}

\begin{figure}[t]
    \centering
    \includegraphics[width=1\linewidth]{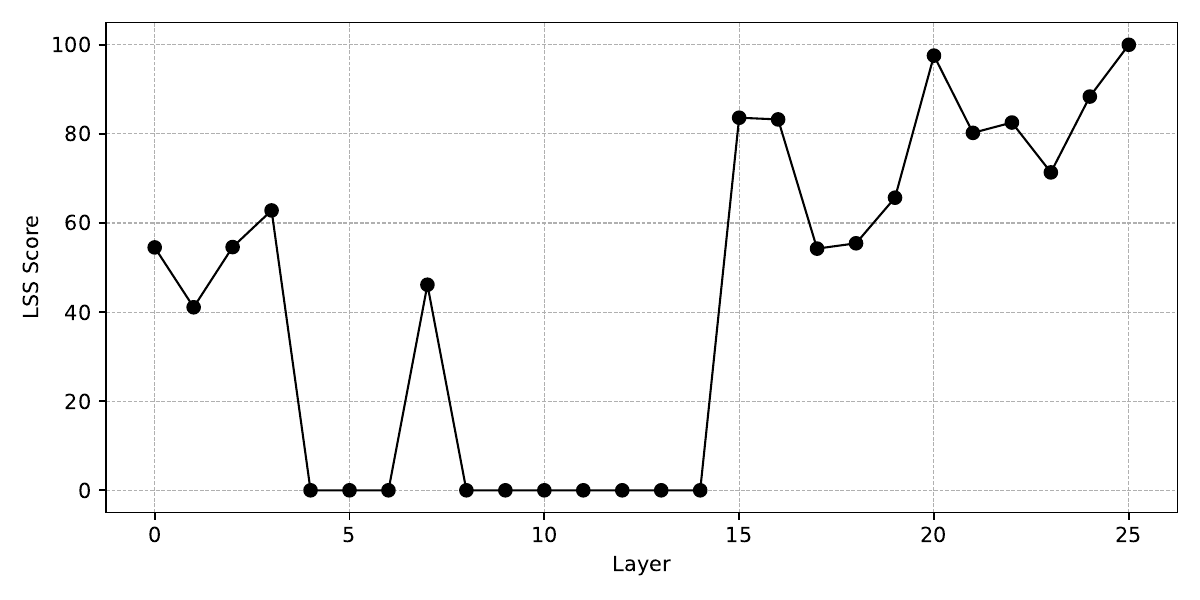}
    \caption{Layer-wise average LSS scores when intervened by \texttt{pt} neurons for Gemma2 2B.}
    \label{fig:perlayer-pt}
\end{figure}

\begin{figure}[t]
    \centering
    \includegraphics[width=1\linewidth]{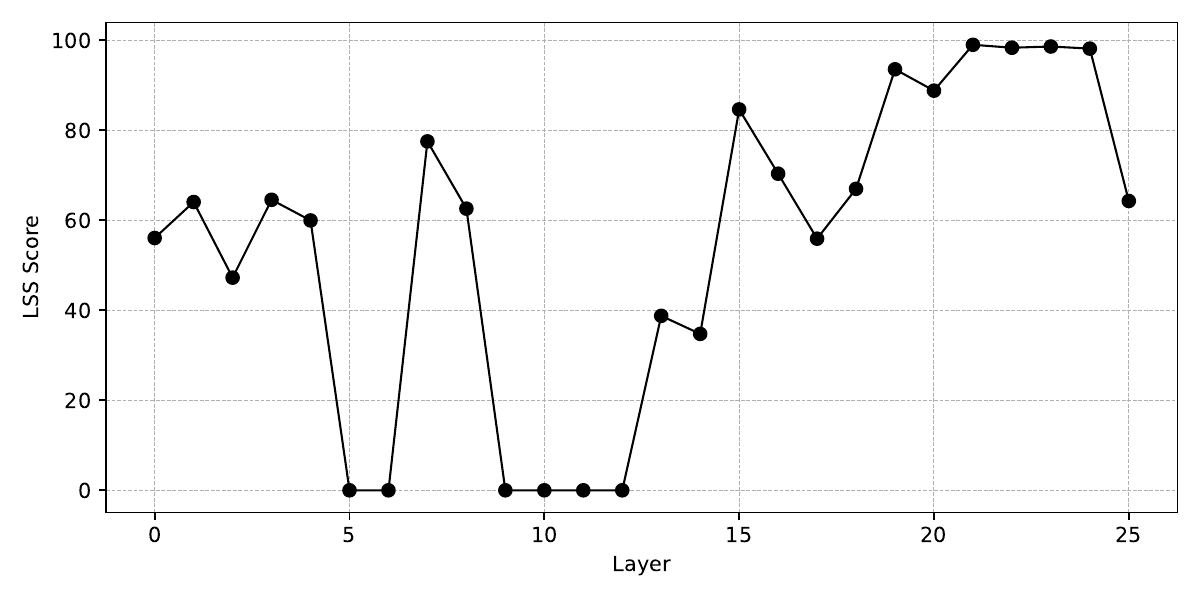}
    \caption{Layer-wise average LSS scores when intervened by \texttt{ru} neurons for Gemma2 2B.}
    \label{fig:perlayer-ru}
\end{figure}

\begin{figure}[t]
    \centering
    \includegraphics[width=1\linewidth]{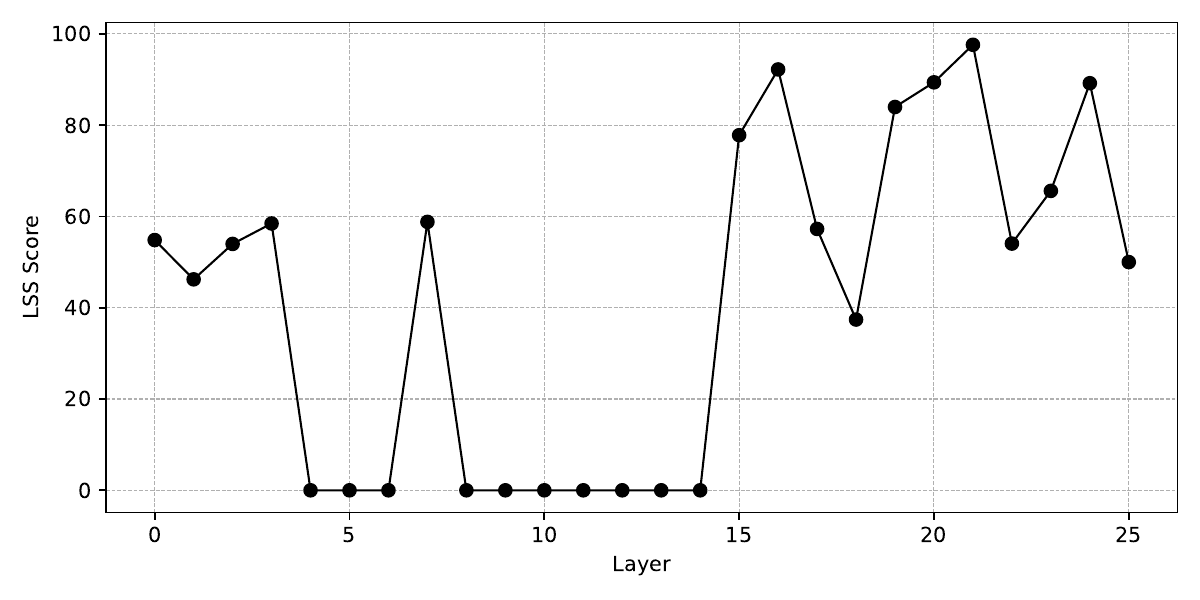}
    \caption{Layer-wise average LSS scores when intervened by \texttt{it} neurons for Gemma2 2B.}
    \label{fig:perlayer-it}
\end{figure}

\begin{figure}[t]
    \centering
    \includegraphics[width=1\linewidth]{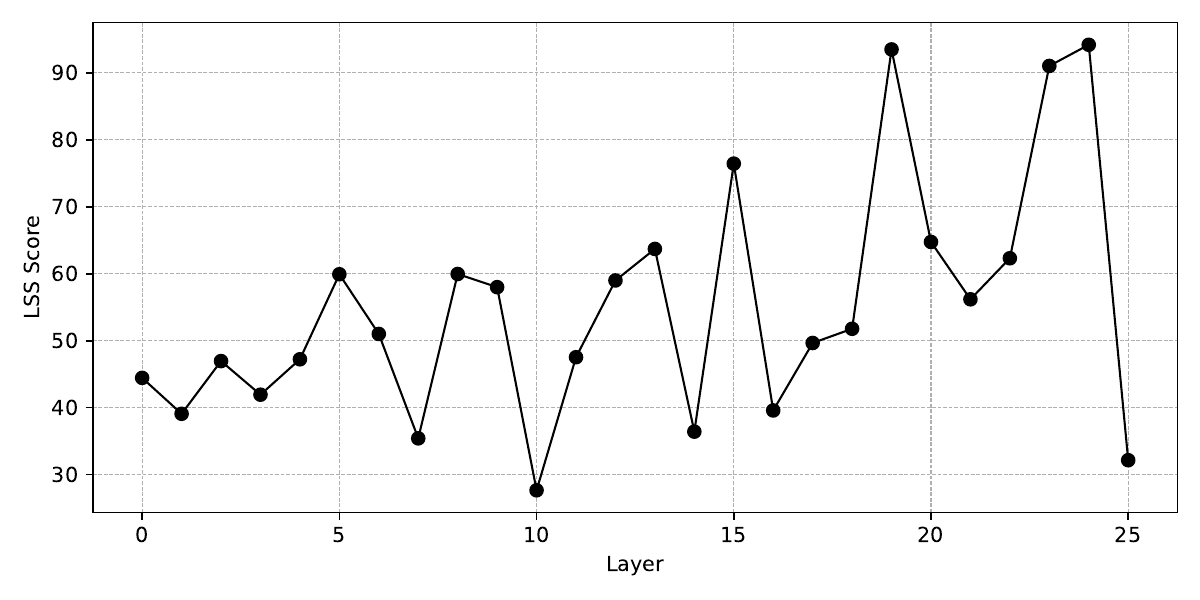}
    \caption{Layer-wise average LSS scores when intervened by \texttt{ta} neurons for Gemma2 2B.}
    \label{fig:perlayer-ta}
\end{figure}

\begin{figure}[t]
    \centering
    \includegraphics[width=1\linewidth]{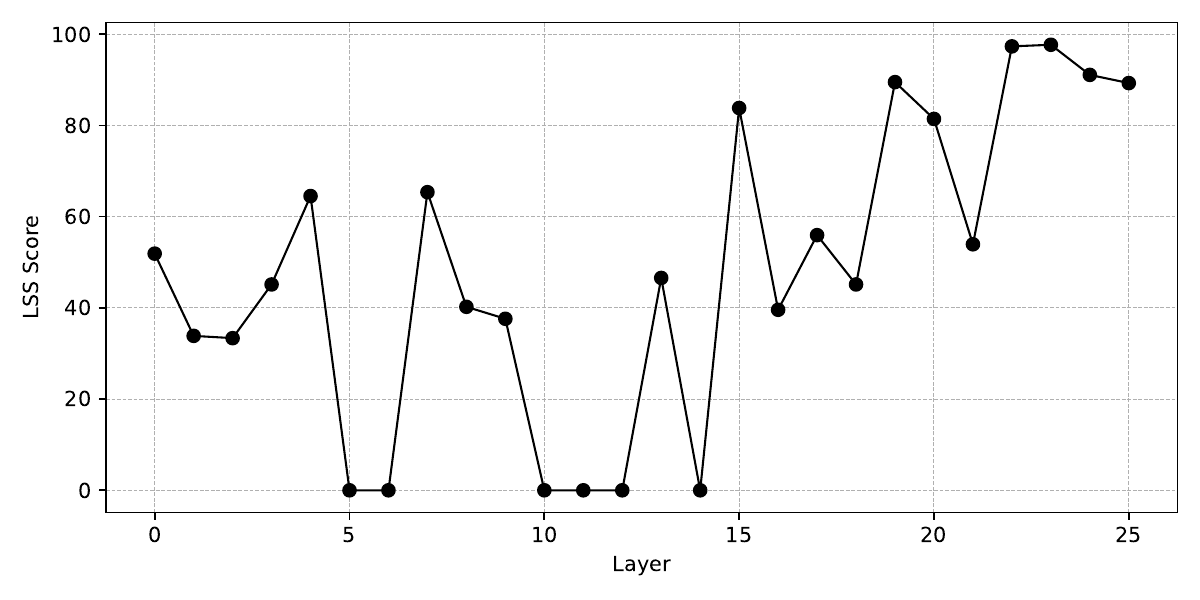}
    \caption{Layer-wise average LSS scores when intervened by \texttt{th} neurons for Gemma2 2B.}
    \label{fig:perlayer-th}
\end{figure}

\begin{figure}[t]
    \centering
    \includegraphics[width=1\linewidth]{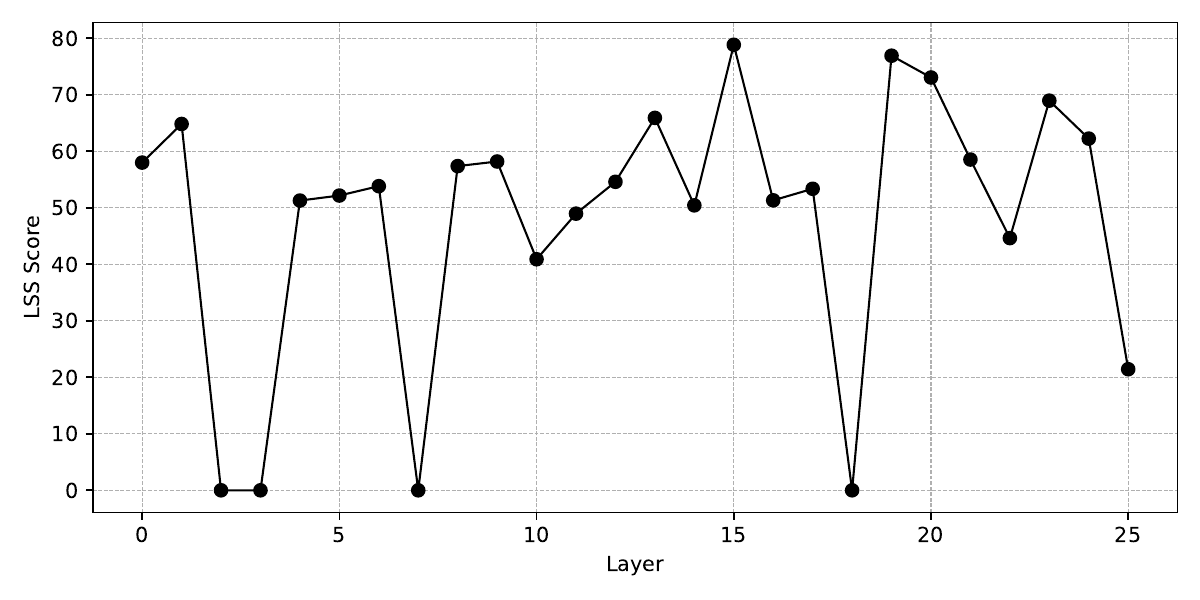}
    \caption{Layer-wise average LSS scores when intervened by \texttt{tr} neurons for Gemma2 2B.}
    \label{fig:perlayer-tr}
\end{figure}

This result aligns with findings from \citet{tang2024languagespecificneuronskeymultilingual} and \citet{kojima-etal-2024-multilingual} as language-specific neurons are mostly found in the first and last layers, excluding the very last one. Results for layer-wise amplification for each language are provided in Figure \ref{fig:perlayer-en} (\texttt{en}), \ref{fig:perlayer-et} (\texttt{et}), \ref{fig:perlayer-nl} (\texttt{nl}), \ref{fig:perlayer-id} (\texttt{id}), \ref{fig:perlayer-ms} (\texttt{ms}), \ref{fig:perlayer-vi} (\texttt{vi}), \ref{fig:perlayer-jp} (\texttt{jp}), \ref{fig:perlayer-zh} (\texttt{zh}), \ref{fig:perlayer-fr} (\texttt{fr}), \ref{fig:perlayer-pt} (\texttt{pt}), \ref{fig:perlayer-ru} (\texttt{ru}), \ref{fig:perlayer-it} (\texttt{it}), \ref{fig:perlayer-ta} (\texttt{ta}), \ref{fig:perlayer-th} (\texttt{th}), \ref{fig:perlayer-tr} (\texttt{tr}).

\subsection{Identifying Language-Specific Neurons from XWinograd to Evaluate XWinograd Performance Under Intervention}
\begin{figure}[t]
\centering
    \includegraphics[width=\linewidth]{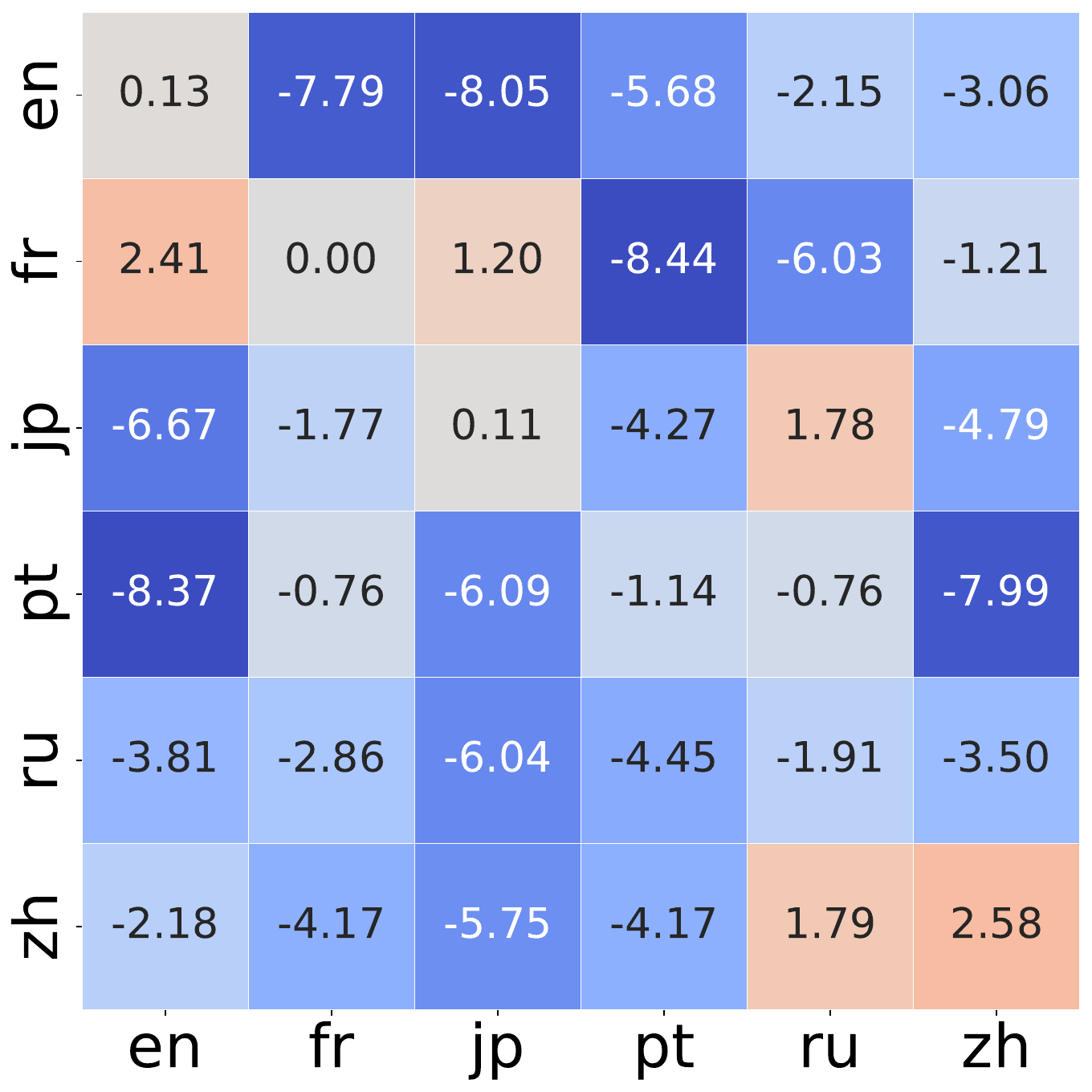}
    \caption{Delta XWinograd accuracy after steering LAPE neurons (obtained from the XWinograd dataset) with \steer{pmax} for Qwen2.5 0.5B.}
    \label{fig:lapexwin-xwin-qwenm}
\end{figure}

In this ablation study, we explore the behavior of amplification when the language-specific neurons are identified from the same dataset used for task evaluation. Instead of using FLORES-200 as in our original approach, we identify the language-specific neurons using the XWinograd dataset. Once identified, we amplify the neurons with \steer{pmax} steering factor to evaluate the exact same XWinograd dataset from which the neurons were identified. The result in Figure \ref{fig:lapexwin-xwin-qwenm} shows that there are no indication of improvement in the self-intervention nor cross-intervention in similar languages, leading to the conclusion that identifying and steering language-specific neurons from the same dataset may not better improve task performance in the respective languages. This result captures less visible patterns, which may be due to the fact that the XWinograd dataset contains significantly fewer tokens than FLORES-200. This scarcity makes LAPE less effective at identifying language-specific neurons, and consequently, it struggles to improve task performance.

\section{Model and Language Details}
\label{sec:appendix}
\subsection{Models and Language Neurons Used}
\label{sec:lang-model}
We use models trained in different primary languages: Qwen2.5 Instruct \cite{qwen2025qwen25technicalreport} 0.5B and 7B (primarily trained in Chinese and English), Gemma2 Instruct \cite{gemmateam2024gemma2improvingopen} 2B and 9B (primarily trained in English), SeaLLMs-v3 Chat \cite{zhang2024seallms3openfoundation} 1.5B and 7B (primarily trained in Southeast Asian languages). Qwen2.5 and SeaLLMs-v3 share the same model architecture, whereas Gemma2 has a different architecture.
We employ neurons specific to 18 languages, English~(\texttt{en}), Dutch~(\texttt{nl}), Indonesian~(\texttt{id}), Malay~(\texttt{ms}), Vietnamese~(\texttt{vi}), Japanese~(\texttt{jp}), Chinese~(\texttt{zh}), French~(\texttt{fr}), Portuguese~(\texttt{pt}), Russian~(\texttt{ru}), Estonian~(\texttt{et}), Haitian Creole~(\texttt{ht}), Italian~(\texttt{it}), Quechua~(\texttt{qu}), Swahili~(\texttt{sw}), Tamil~(\texttt{ta}), Thai~(\texttt{th}), and Turkish~(\texttt{tr}). The selected languages account for different language domains associated with the models and represent both high and low-resource languages to further analyze their behavior.

\subsection{Experimental Settings for Finding the Most Optimal Steering Factors}
\label{sec:appendix-find}
To find the most optimal steering factors, we reconstruct MLAMA dataset \cite{kassner2021multilingual} using 15 languages due to limitation in the dataset. The 15 languages are English~(\texttt{en}), Dutch~(\texttt{nl}), Indonesian~(\texttt{id}), Malay~(\texttt{ms}), Vietnamese~(\texttt{vi}), Japanese~(\texttt{jp}), Chinese~(\texttt{zh}), French~(\texttt{fr}), Portuguese~(\texttt{pt}), Russian~(\texttt{ru}), Estonian~(\texttt{et}), Italian~(\texttt{it}), Tamil~(\texttt{ta}), Thai~(\texttt{th}), and Turkish~(\texttt{tr}). In the reconstructed MLAMA dataset, each question has 15 answers in different languages, with no lexical overlap. 
% For example, if both the Indonesian and Malaysian answers were ``\texttt{Jerman}'' (i.e., Germany in English), that row was discarded.

We evaluate the proposed Language Steering Shift (LSS) scores on the reconstructed MLAMA dataset to measure how much an answer is shifted to the target language after intervened by the steering factors. This evaluation is only performed on the small versions of each model (Qwen2.5 0.5B, Gemma2 2B, and SeaLLMs-v3 1.5B) as they have already resulted in very high LSS scores, showing strong capability in steering to target languages.

\subsection{Experimental Settings for Task Evaluation}
    We evaluate every models (6 in total) mainly using the \steer{pmax} steering factor on several tasks. For translation task and perplexity, we use the \texttt{devtest} split of the FLORES-200 dataset on the same 18 languages the neurons were identified from. The translation evaluation is performed from English to every other 17 languages as English is considered the dominant language due to its widespread use as a multilingual benchmarks. 
    Translation tasks are performed using 2 types of prompt: \texttt{targeted} and \texttt{non-targeted}. Targeted prompts explicitly mention the target language, wherease non-targeted prompts do not. For example, for translation to \texttt{id}, the targeted prompt is as follows,
\begin{verbatim}
    "Translate from English to Indonesian.
    English: {question}
    Indonesian:"
\end{verbatim}
whereas the non-targeted prompt is as follows,
\begin{verbatim}
    "Translate from English into the 
    target language.
    English: {question}
    Target language:" 
\end{verbatim}

  For commonsense reasoning task, we evaluate XCOPA \cite{ponti2020xcopa} and XWinograd \cite{muennighoff2022crosslingual}, each containing 2-option questions. XCOPA dataset aims to choose the more plausible cause or effect of a given premises with a focus on lower-resource languages including Estonian (\texttt{et}), Haitian Creole (\texttt{ht}), Indonesian (\texttt{id}), Italian (\texttt{it}), Quechua (\texttt{qu}), Swahili (\texttt{sw}), Tamil (\texttt{ta}), Thai (\texttt{tah}), Turkish (\texttt{tr}), Vietnamese (\texttt{vi}), and Chinese (\texttt{zh}). XWinograd aims to resolve an ambiguous pronoun using commonsense knowledge, available in English (\texttt{en}), French (\texttt{fr}), Japanese (\texttt{jp}), Portuguese (\texttt{pt}), Russian (\texttt{ru}), and Chinese (\texttt{zh}). 
  
  For knowledge and reasoning task, we evaluate Include-lite dataset \cite{romanou2024include}, containing 4-option multiple-choice-questions (MCQ) extracted from academic and professional exams, covering 57 topics, including regional knowledge. The languages used for this dataset are limited to the available languages in the obtained language-specific neurons set, which are the following: Dutch (\texttt{nl}), Indonesian (\texttt{id}), Malay (\texttt{ms}), Vietnamese (\texttt{vi}), Japanese (\texttt{jp}), Chinese (\texttt{zh}), French (\texttt{fr}), Portuguese (\texttt{pt}), Russian (\texttt{ru}), Estonian (\texttt{et}), Italian (\texttt{it}),  Tamil (\texttt{ta}), and Turkish (\texttt{tr}).

\section{Proportions of LAPE and Baseline neurons}

Throughout all models, the proportions of LAPE neurons are presented in Table \ref{tab:prop-lape}, Baseline neurons in Table \ref{tab:prop-raw}. Proportions of LAPE neurons show a significant variability across languages and models. Some lower-resource languages (e.g. \texttt{qu}, \texttt{sw}, \texttt{ta}) show very high proportions for certain models, while others (e.g. \texttt{en}, \texttt{nl}, \texttt{ms}, \texttt{vi}) generally show lower proportions. This may suggest these languages need more dedicated neurons due to linguistic complexity or limited pre-training exposure. In contrast, \texttt{en} has the fewest neurons across models, consistent with \citet{tang2024languagespecificneuronskeymultilingual}, likely because English's dominance in pre-training leads to more distributed, broadly learned features.

\section{Neuron Overlaps}
\label{sec:overlap}
\begin{figure}
    \centering
    \includegraphics[width=\linewidth, trim=130 70 70 120, clip]{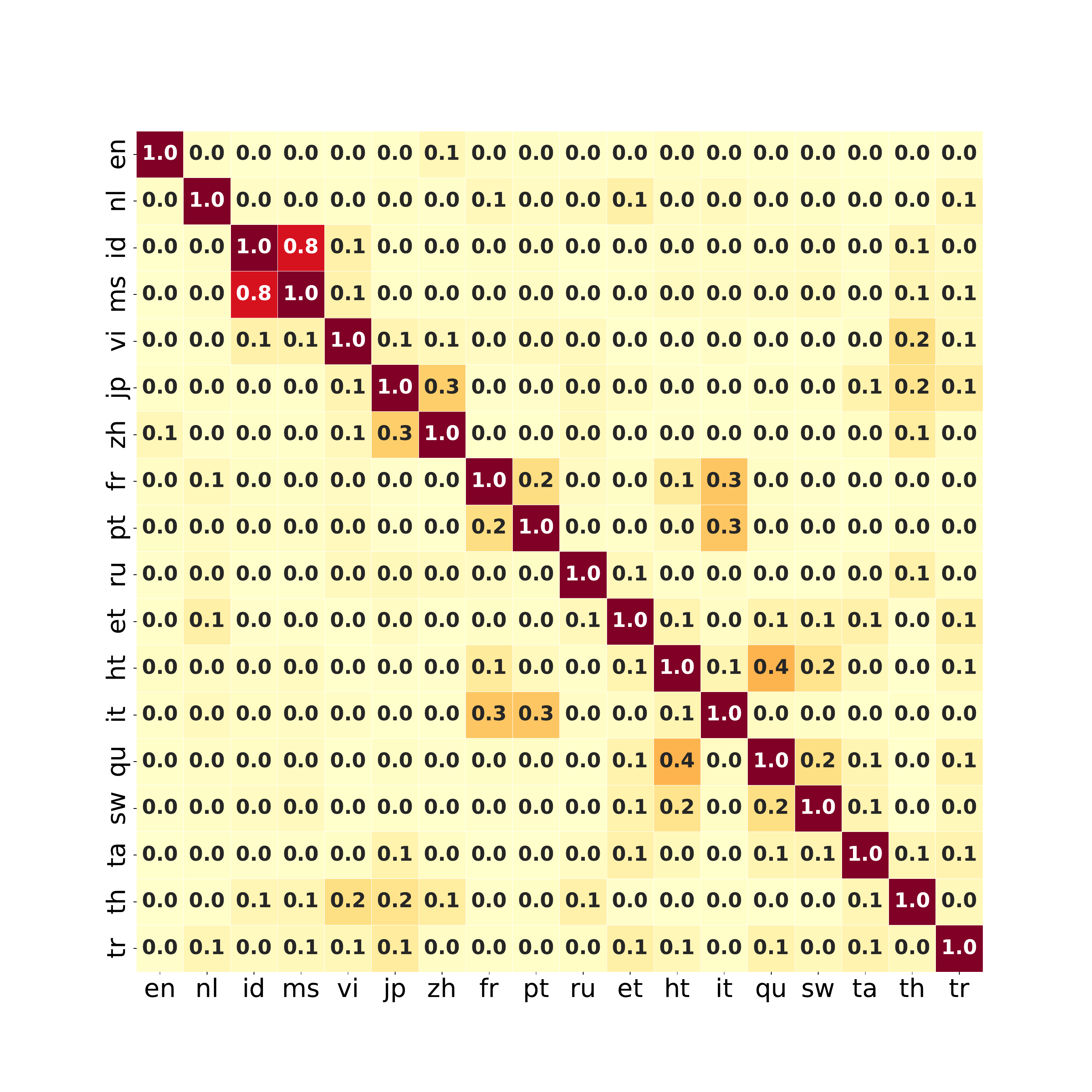}
    \caption{Symmetric heatmaps of overlap neuron proportions in SeaLLMs3 1.5B using jaccard distance for LAPE neurons.}
    \label{fig:ovlape-seam}
\end{figure}

\begin{figure}
    \centering
    \includegraphics[width=\linewidth, trim=130 70 70 120, clip]{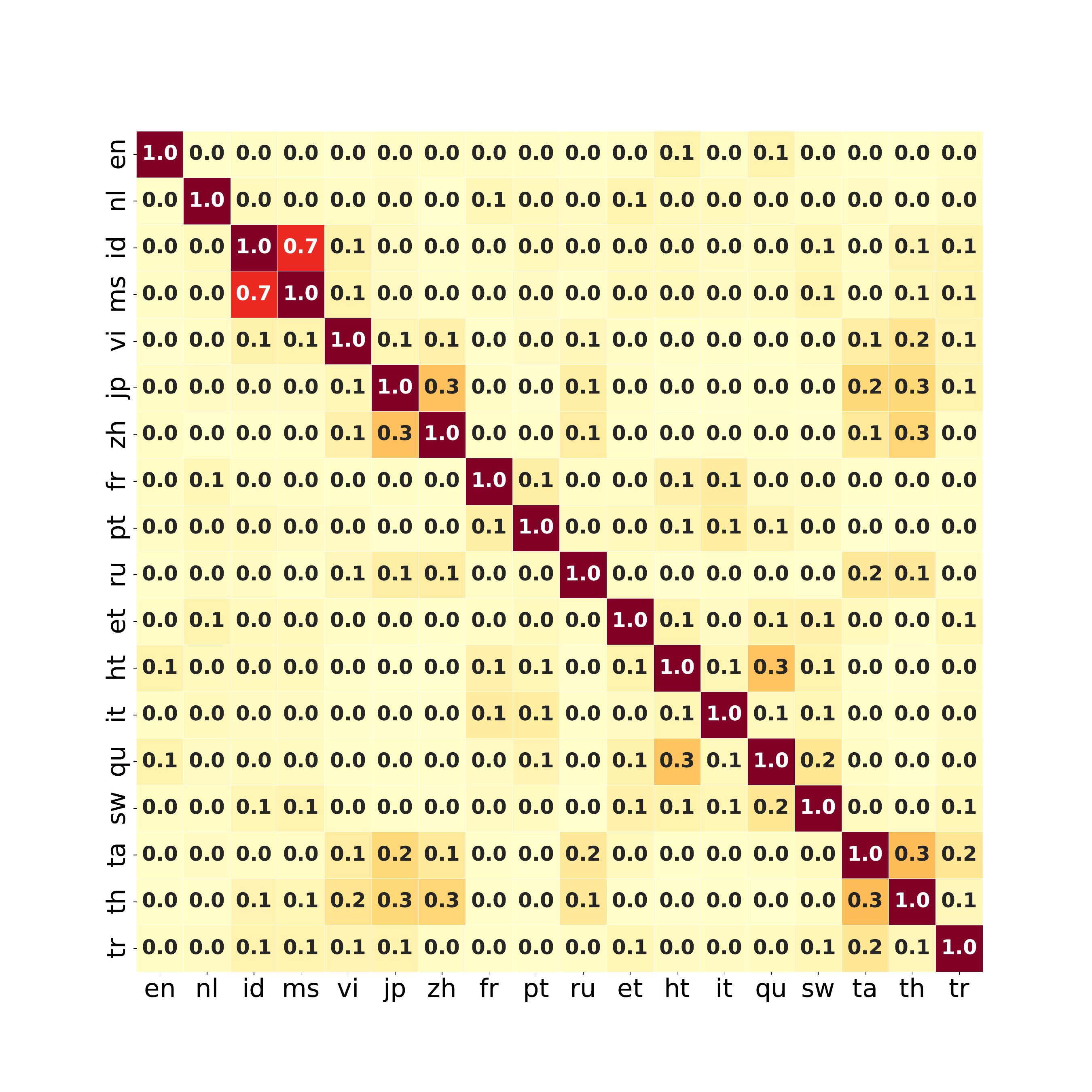}
    \caption{Symmetric heatmaps of overlap neuron proportions in Gemma2 2B using jaccard distance for LAPE neurons.}
    \label{fig:ovlape-gemmam}
\end{figure}

\begin{figure}
    \centering
    \includegraphics[width=\linewidth, trim=130 70 70 120, clip]{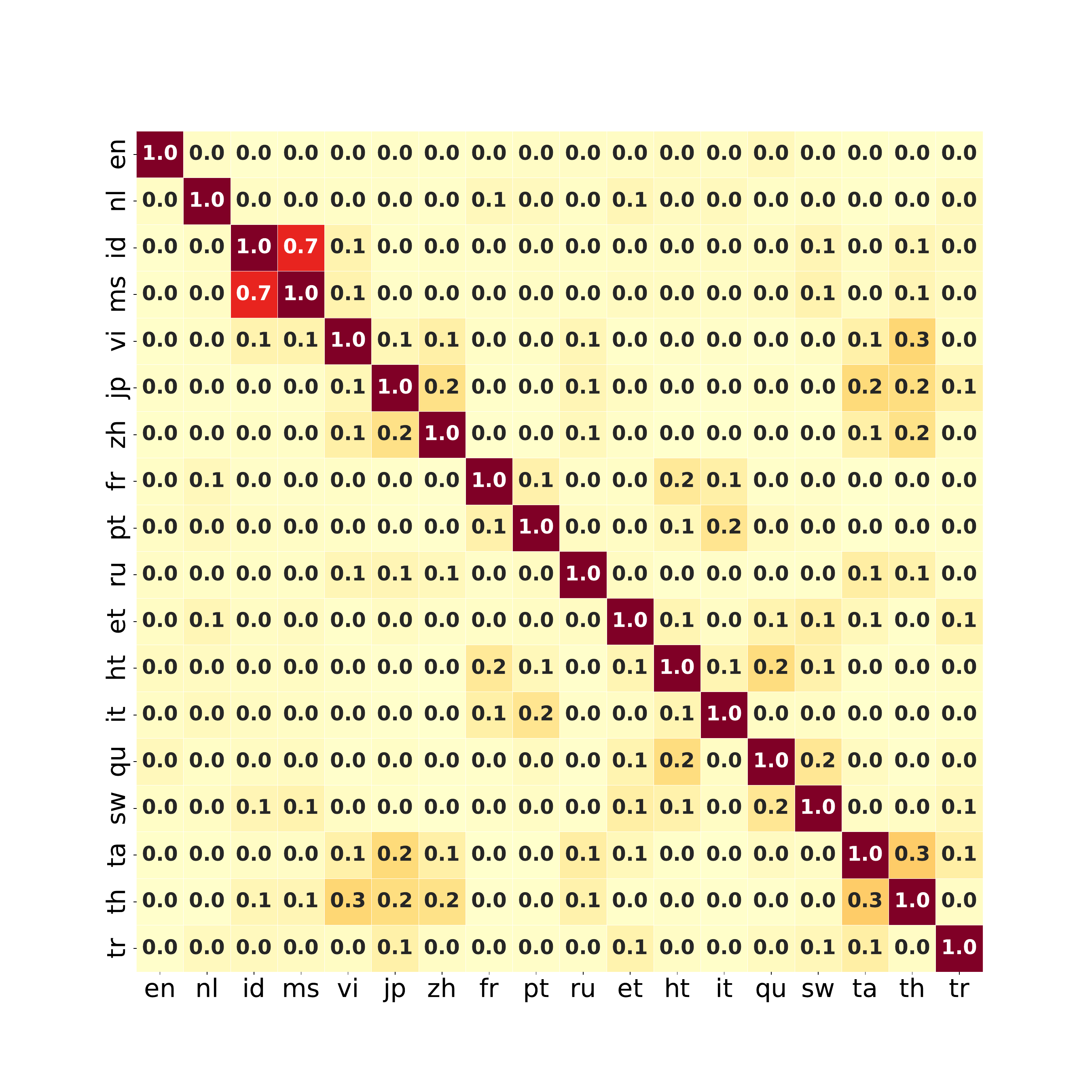}
    \caption{Symmetric heatmaps of overlap neuron proportions in Gemma2 9B using jaccard distance for LAPE neurons.}
    \label{fig:ovlape-gemma}
\end{figure}

\begin{figure}
    \centering
    \includegraphics[width=\linewidth, trim=130 70 70 120, clip]{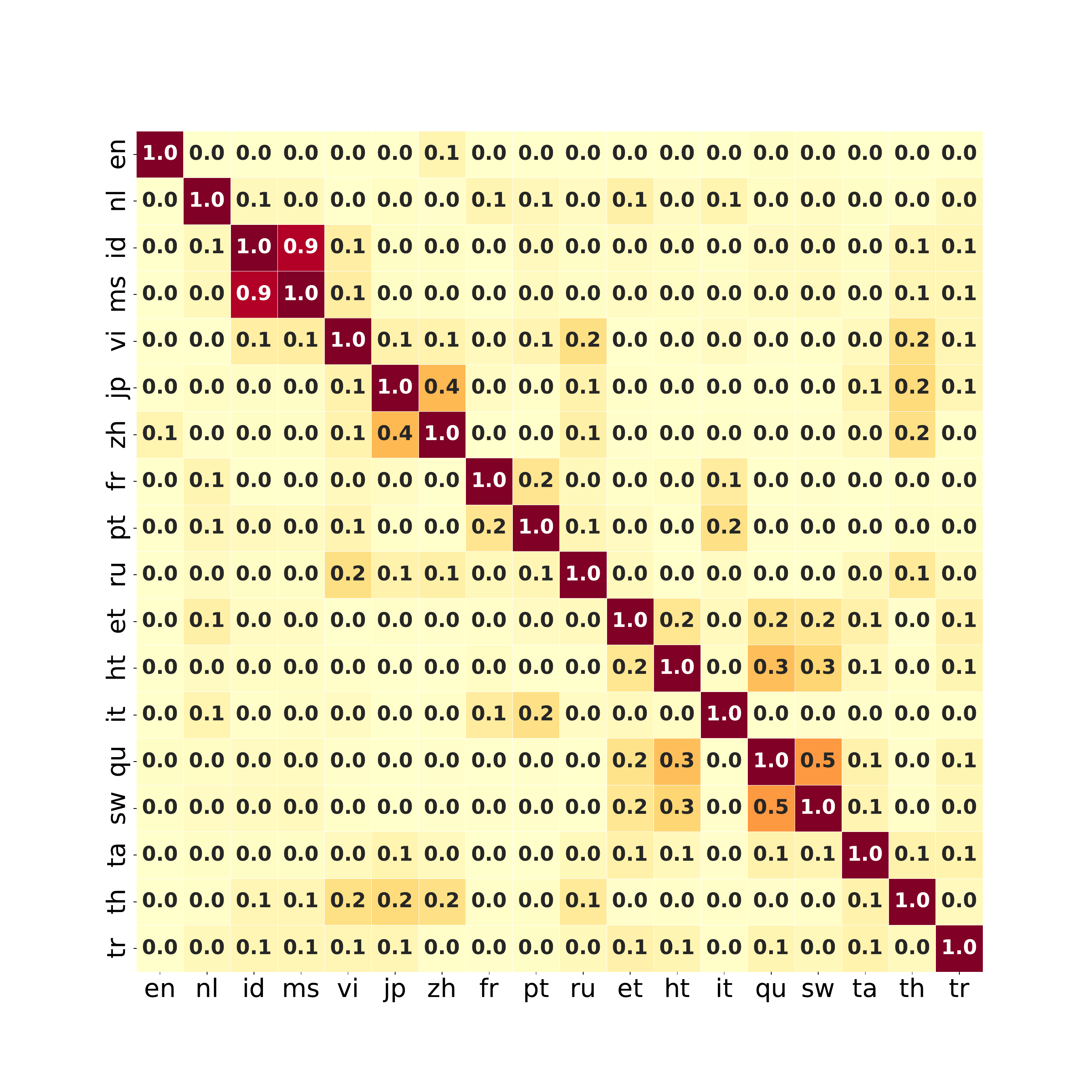}
    \caption{Symmetric heatmaps of overlap neuron proportions in Qwen2.5 0.5B using jaccard distance for LAPE neurons.}
    \label{fig:ovlape-qwenm}
\end{figure}

\begin{figure}
    \centering
    \includegraphics[width=\linewidth, trim=130 70 70 120, clip]{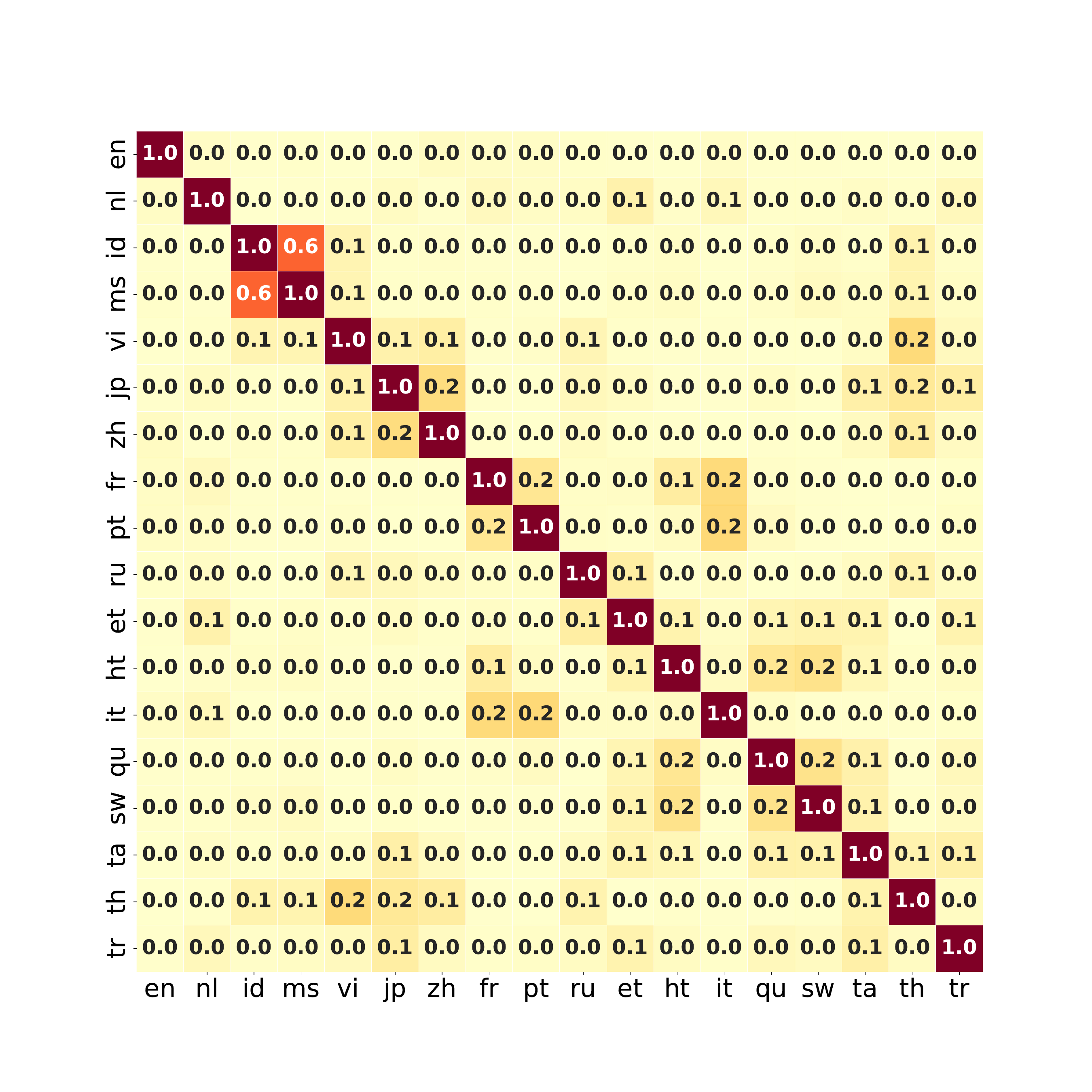}
    \caption{Symmetric heatmaps of overlap neuron proportions in Qwen2.5 7B using jaccard distance for LAPE neurons.}
    \label{fig:ovlape-qwen}
\end{figure}

\begin{figure}
    \centering
    \includegraphics[width=\linewidth, trim=130 70 70 120, clip]{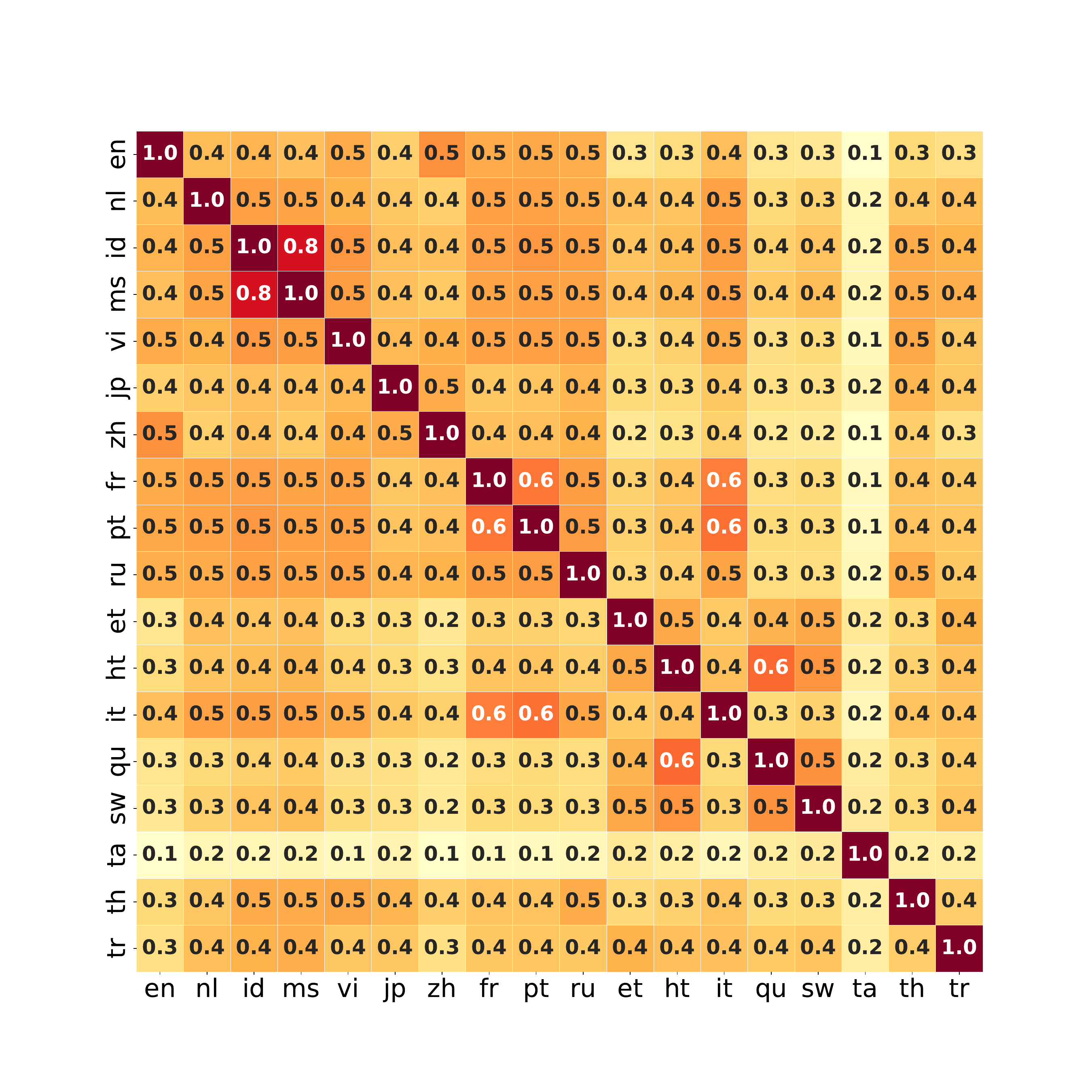}
    \caption{Symmetric heatmaps of overlap neuron proportions in SeaLLMs3 1.5B using jaccard distance for Baseline neurons.}
    \label{fig:ovraw-seam}
\end{figure}

\begin{figure}
    \centering
    \includegraphics[width=\linewidth, trim=130 70 70 120, clip]{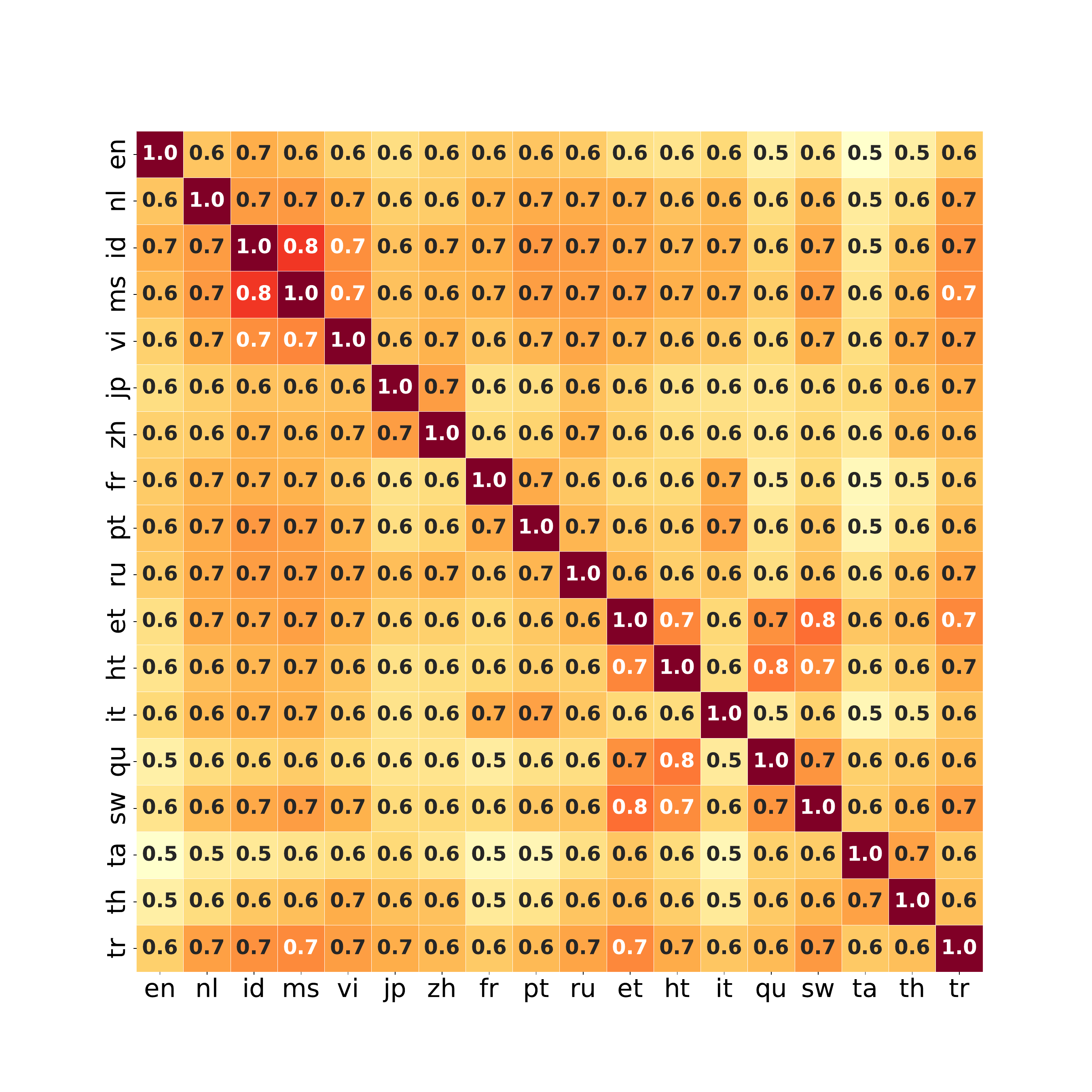}
    \caption{Symmetric heatmaps of overlap neuron proportions in Gemma2 2B using jaccard distance for Baseline neurons.}
    \label{fig:ovraw-gemmam}
\end{figure}

\begin{figure}
    \centering
    \includegraphics[width=\linewidth, trim=130 70 70 120, clip]{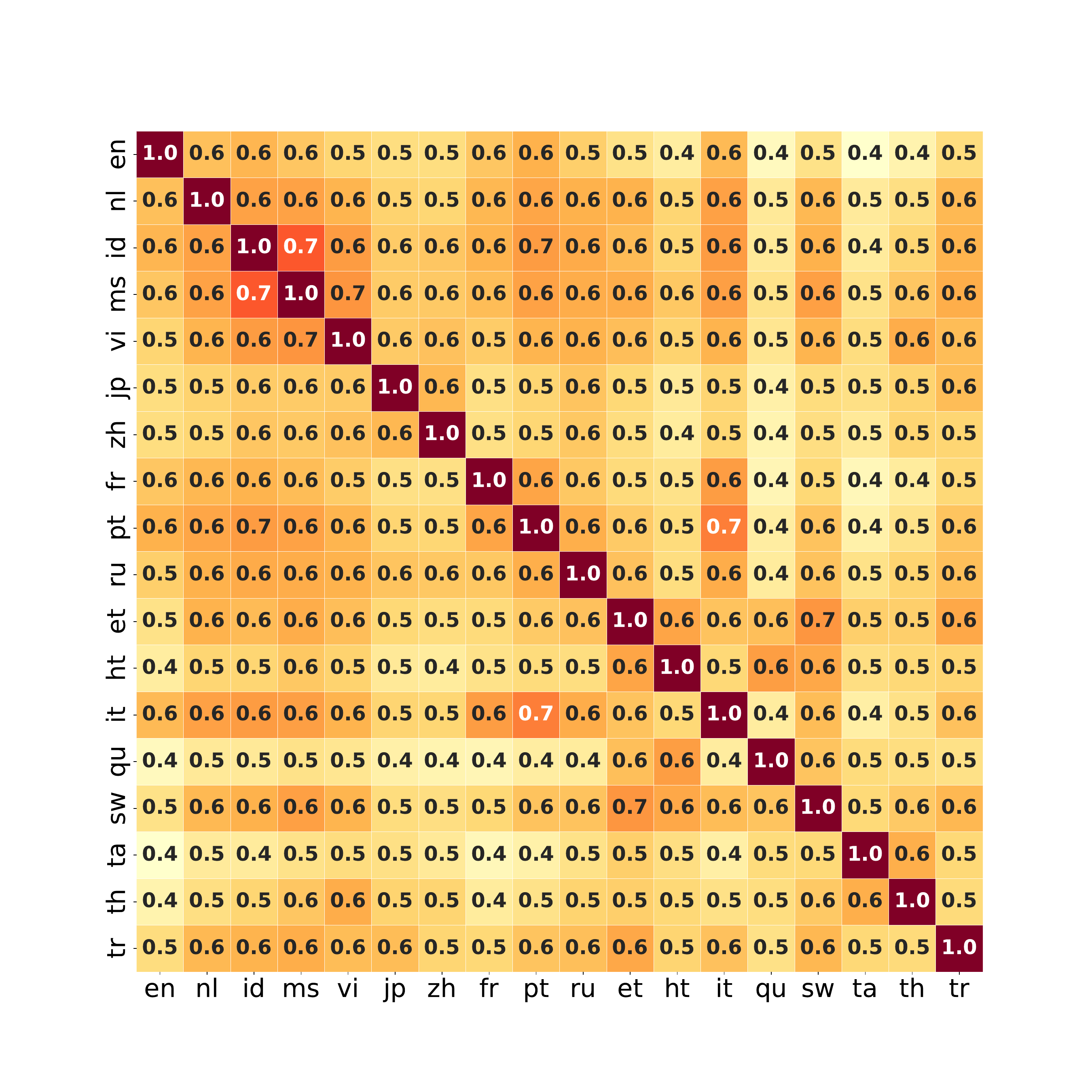}
    \caption{Symmetric heatmaps of overlap neuron proportions in Gemma2 9B using jaccard distance for Baseline neurons.}
    \label{fig:ovraw-gemma}
\end{figure}

\begin{figure}
    \centering
    \includegraphics[width=\linewidth, trim=130 70 70 120, clip]{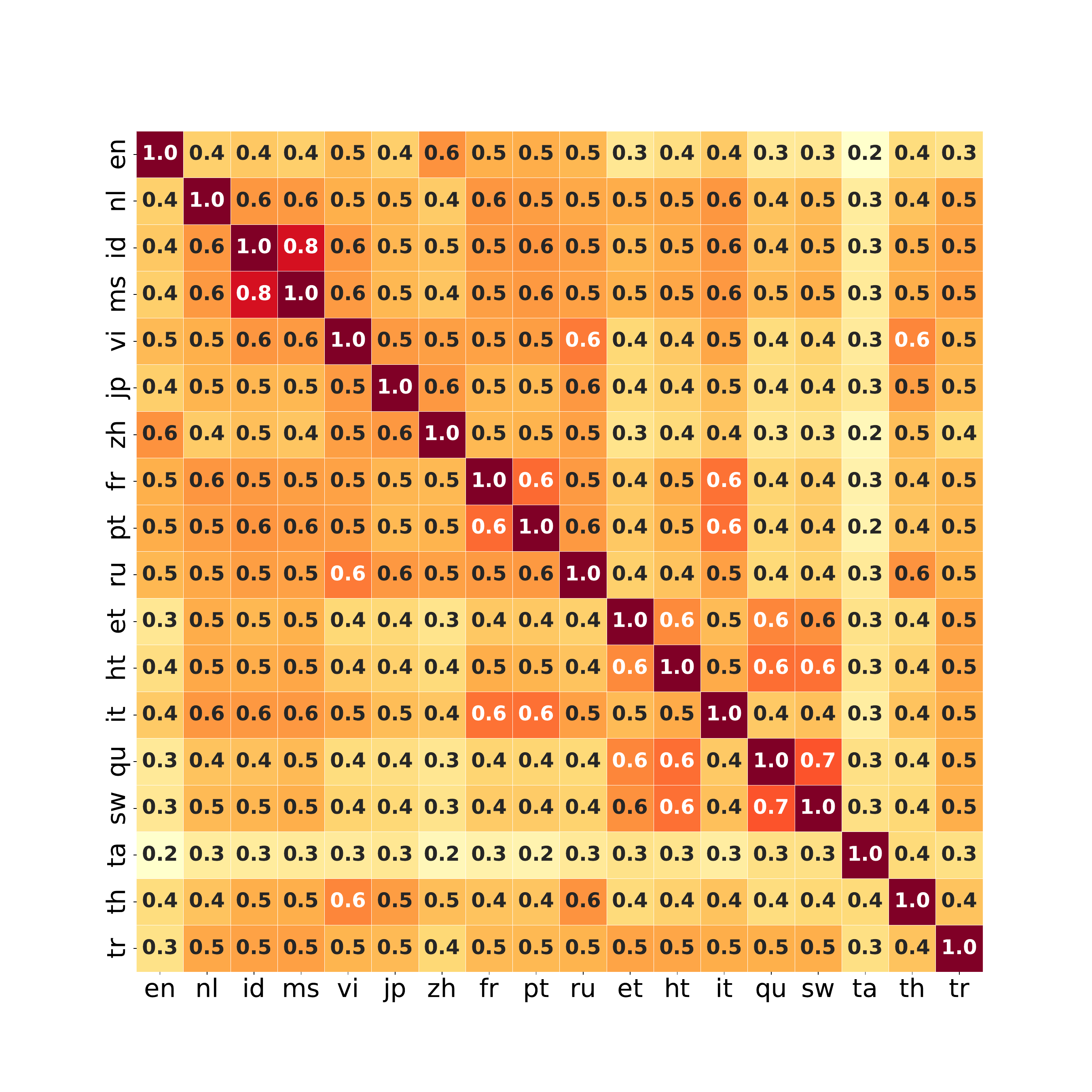}
    \caption{Symmetric heatmaps of overlap neuron proportions in Qwen2.5 0.5B using jaccard distance for Baseline neurons.}
    \label{fig:ovraw-qwenm}
\end{figure}

\begin{figure}
    \centering
    \includegraphics[width=\linewidth, trim=130 70 70 120, clip]{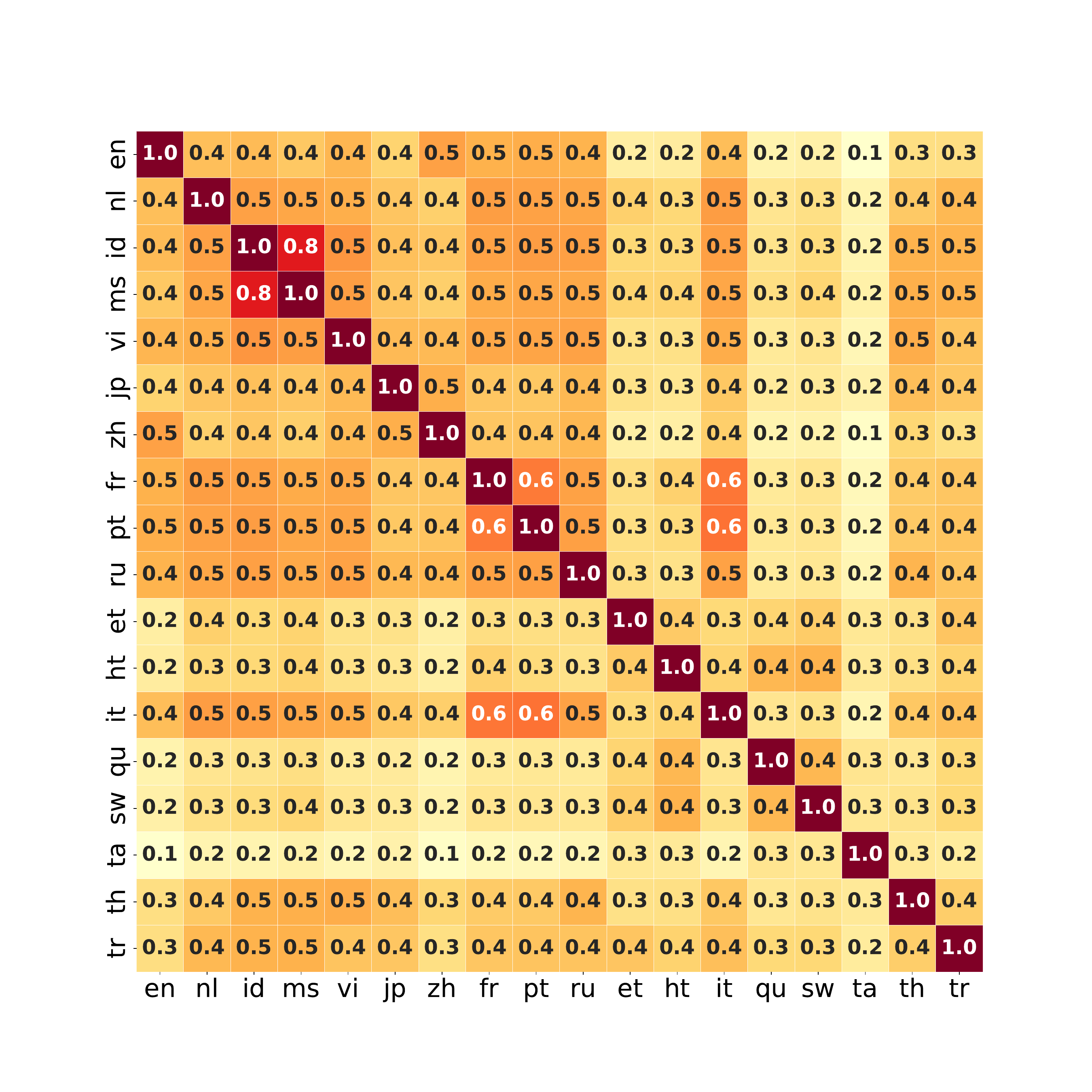}
    \caption{Symmetric heatmaps of overlap neuron proportions in Qwen2.5 7B using jaccard distance for Baseline neurons.}
    \label{fig:ovraw-qwen}
\end{figure}

We calculate the Jaccard distance for neuron overlap between languages for LAPE neurons (Figure \ref{fig:ovlape-seam}, \ref{fig:ovlape-qwenm}, \ref{fig:ovlape-qwen}, \ref{fig:ovlape-gemmam}, \ref{fig:ovlape-gemma}) and Baseline neurons (Figure \ref{fig:ovraw-seam}, \ref{fig:ovraw-qwenm}, \ref{fig:ovraw-qwen}, \ref{fig:ovraw-gemmam}, \ref{fig:ovraw-gemma}).

\section{LSS Scores}
\label{sec:appendix-lss}
\begin{figure*}
    \centering
    \includegraphics[width=\textwidth, trim=50 70 50 40, clip]{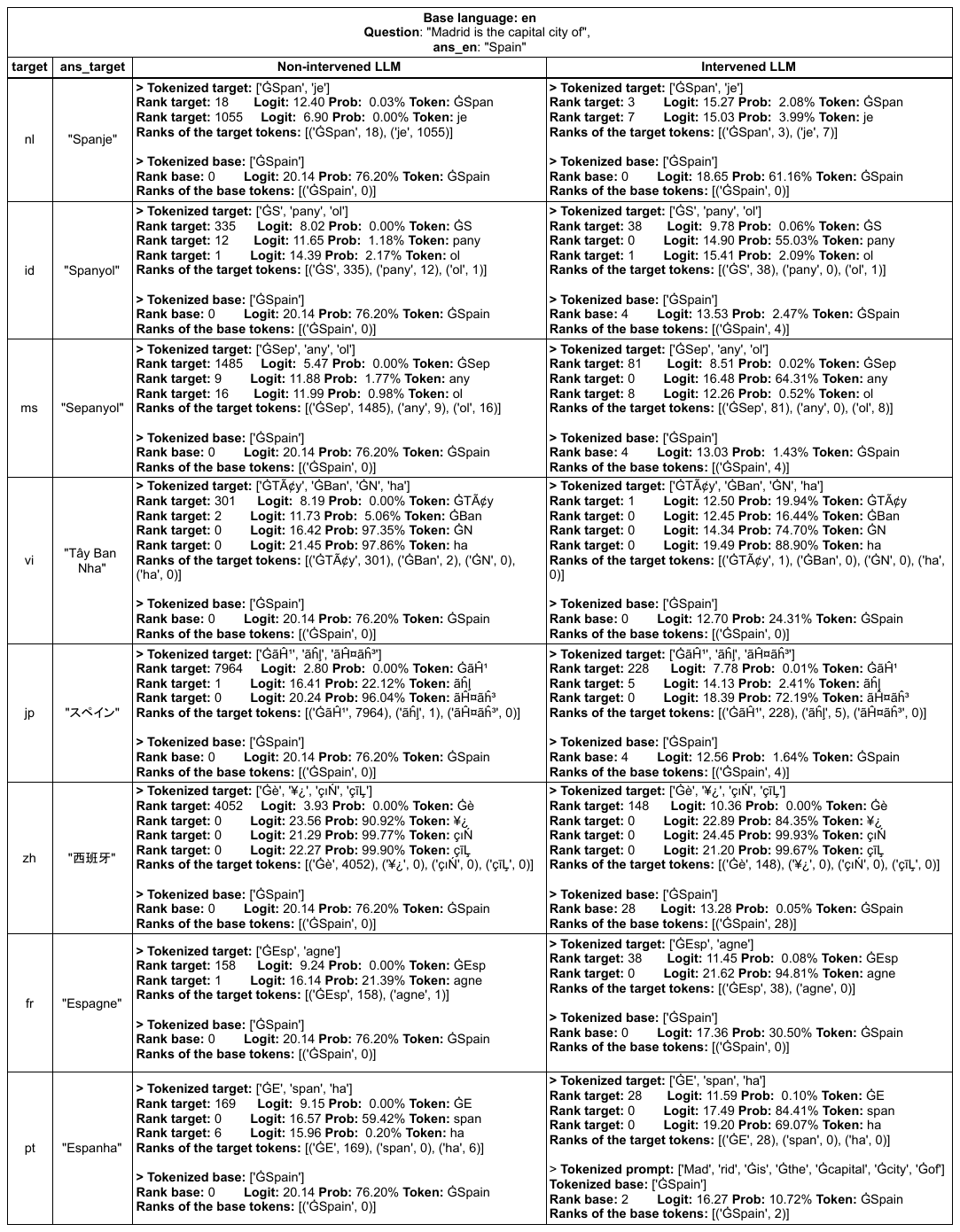}
    \caption{Visualization of token log-prob for \textit{base} and \textit{target} answers. Base answers are in the prompt language (\texttt{en}), target answers are in the language of intervention (\texttt{nl}, \texttt{id},\texttt{ms},\texttt{vi},\texttt{jp},\texttt{zh},\texttt{fr},\texttt{pt}).}
    \label{fig:vis-lss1}
\end{figure*}

\begin{figure*}
    \centering
    \includegraphics[width=\textwidth, trim=50 140 50 40, clip]{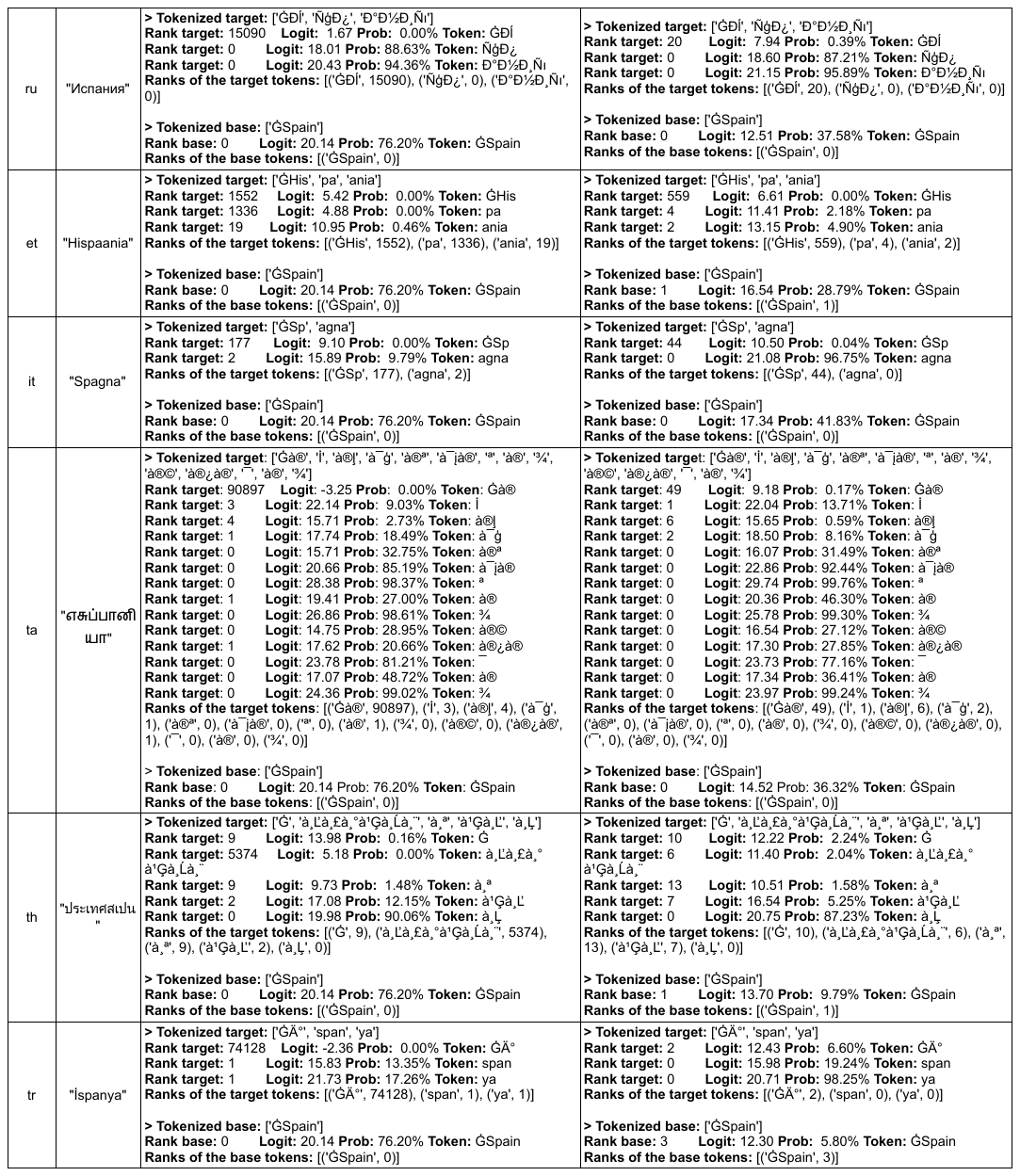}
    \caption{Visualization of token log-prob for \textit{base} and \textit{target} answers in intervened and non-intervened LLMs. Base answers are in the prompt language (\texttt{en}), target answers are in the language of intervention (\texttt{ru}, \texttt{et},\texttt{it},\texttt{ta},\texttt{th},\texttt{tr}).}
    \label{fig:vis-lss2}
\end{figure*}

The complete LSS scores for both Baseline and LAPE neurons are described in Table \ref{tab:en} to Table \ref{tab:zh}. The aggregated LSS scores for each steering factor is presented in Table \ref{tab:mean-scores}. Further statistical testing to determine the significance of the optimal steering factor is provided in \ref{tab:tukey-pairwise}. Visualization of token log-probs for same-meaning answers in different languages are provided in both intervened and non-intervened LLMs in Figure \ref{fig:vis-lss1} and \ref{fig:vis-lss2}.

\section{Perplexity Changes After Amplification}
\label{sec:appendix-ppl}
\begin{figure}[t]
\centering
    \includegraphics[width=\linewidth]{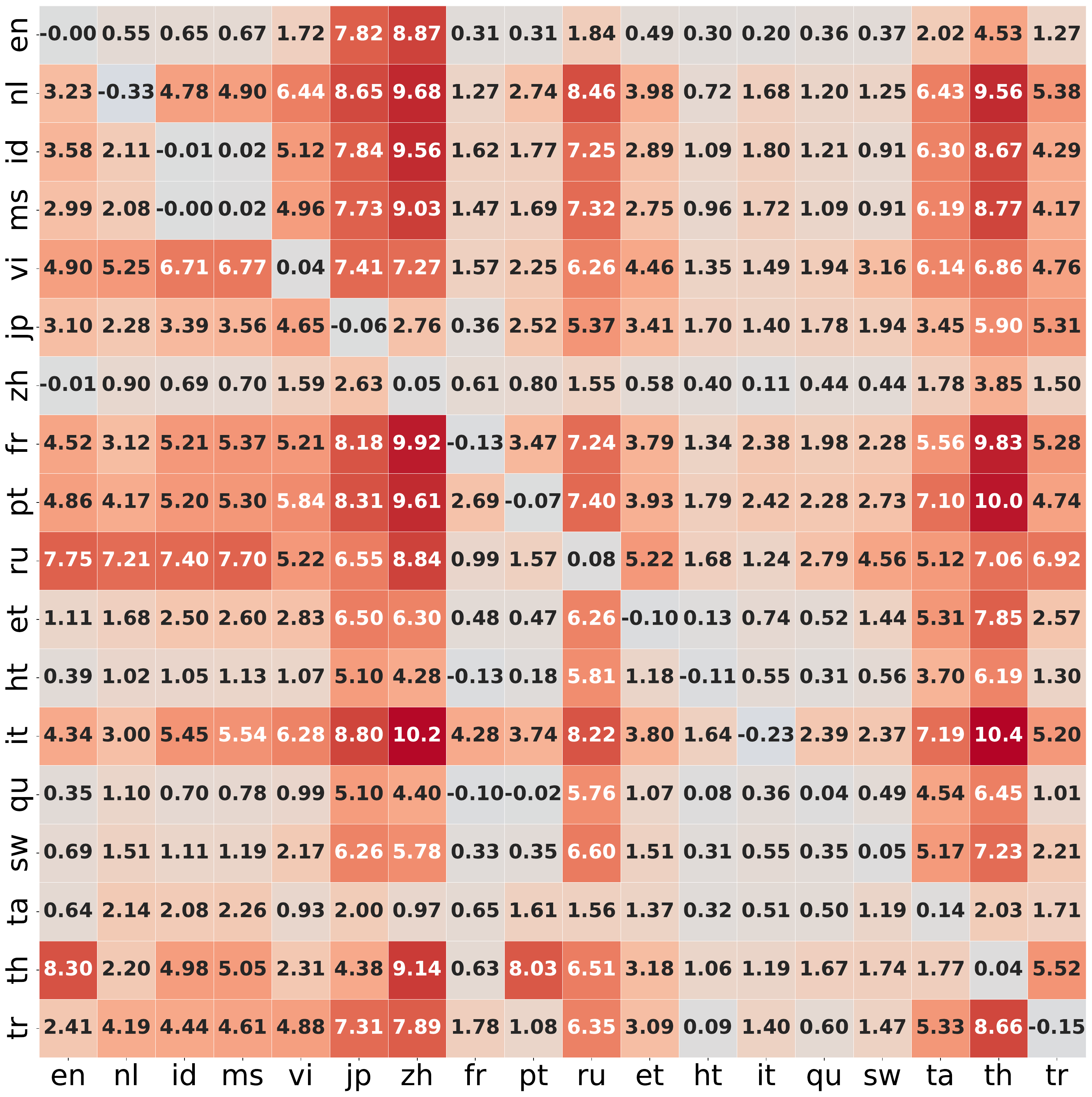}
    \caption{PPL changes after steering LAPE neurons with \steer{pmax} for Qwen2.5 0.5B.}
    \label{fig:ppl-qwenm}
\end{figure}

\begin{figure}[t]
\centering
    \includegraphics[width=\linewidth]{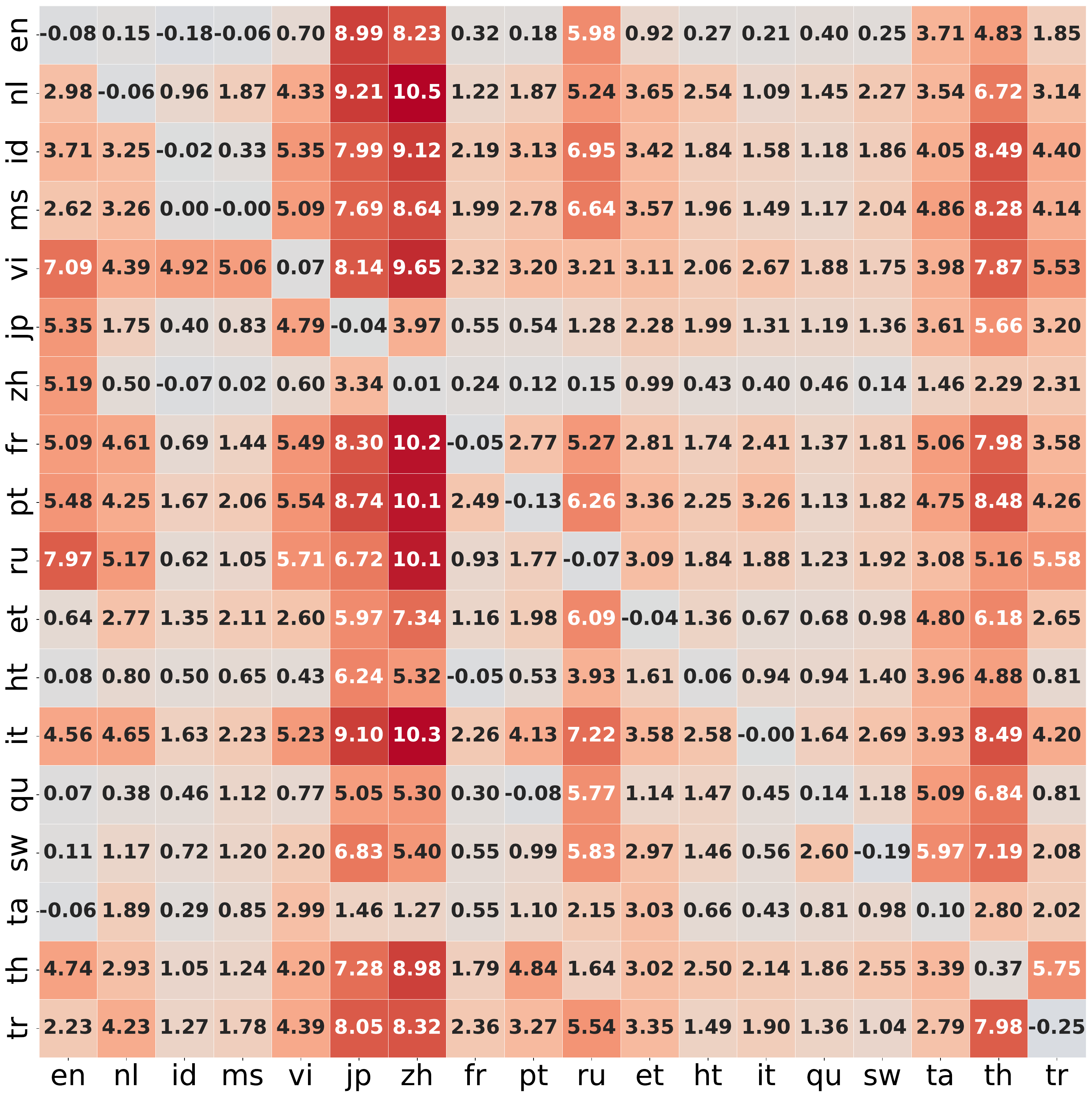}
    \caption{PPL changes after steering LAPE neurons with \steer{pmax} for Qwen2.5 7B.}
    \label{fig:ppl-qwen}
\end{figure}

\begin{figure}[t]
\centering
    \includegraphics[width=\linewidth]{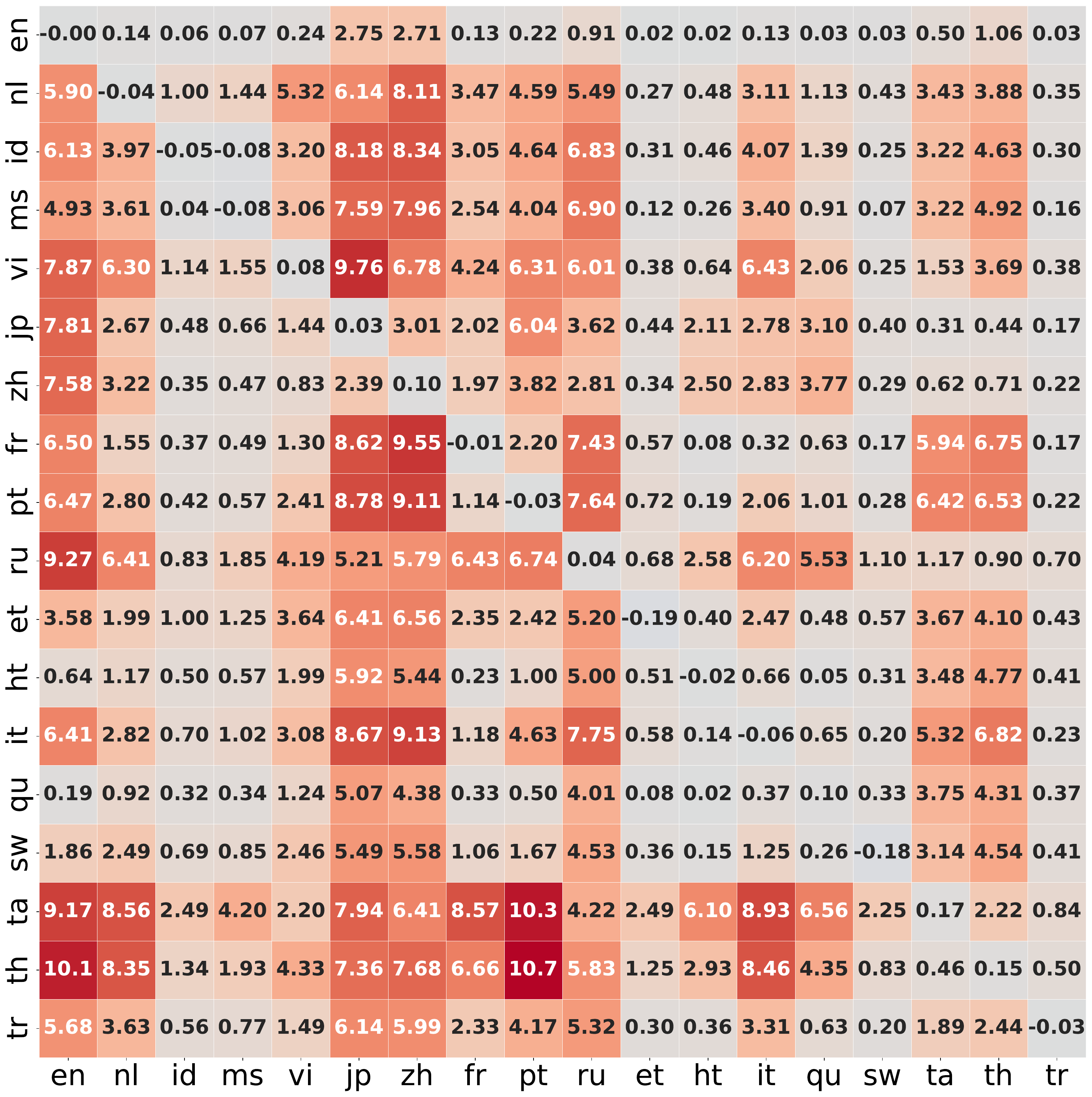}
    \caption{PPL changes after steering LAPE neurons with \steer{pmax} for Gemma2 2B.}
    \label{fig:ppl-gemmam}
\end{figure}

\begin{figure}[t]
\centering
    \includegraphics[width=\linewidth]{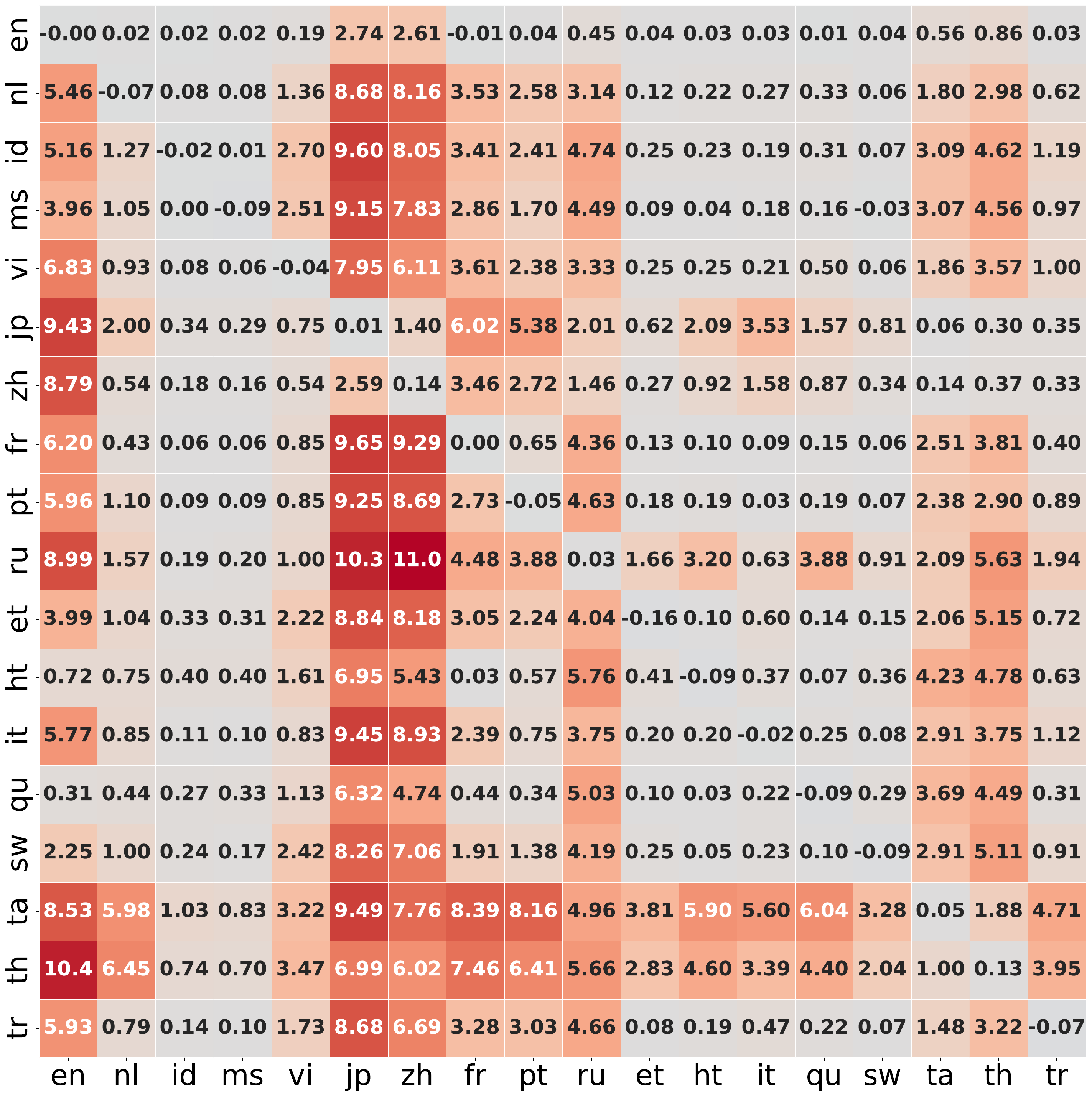}
    \caption{PPL changes after steering LAPE neurons with \steer{pmax} for Gemma2 9.B}
    \label{fig:ppl-gemma}
\end{figure}

\begin{figure}[t]
\centering
    \includegraphics[width=\linewidth]{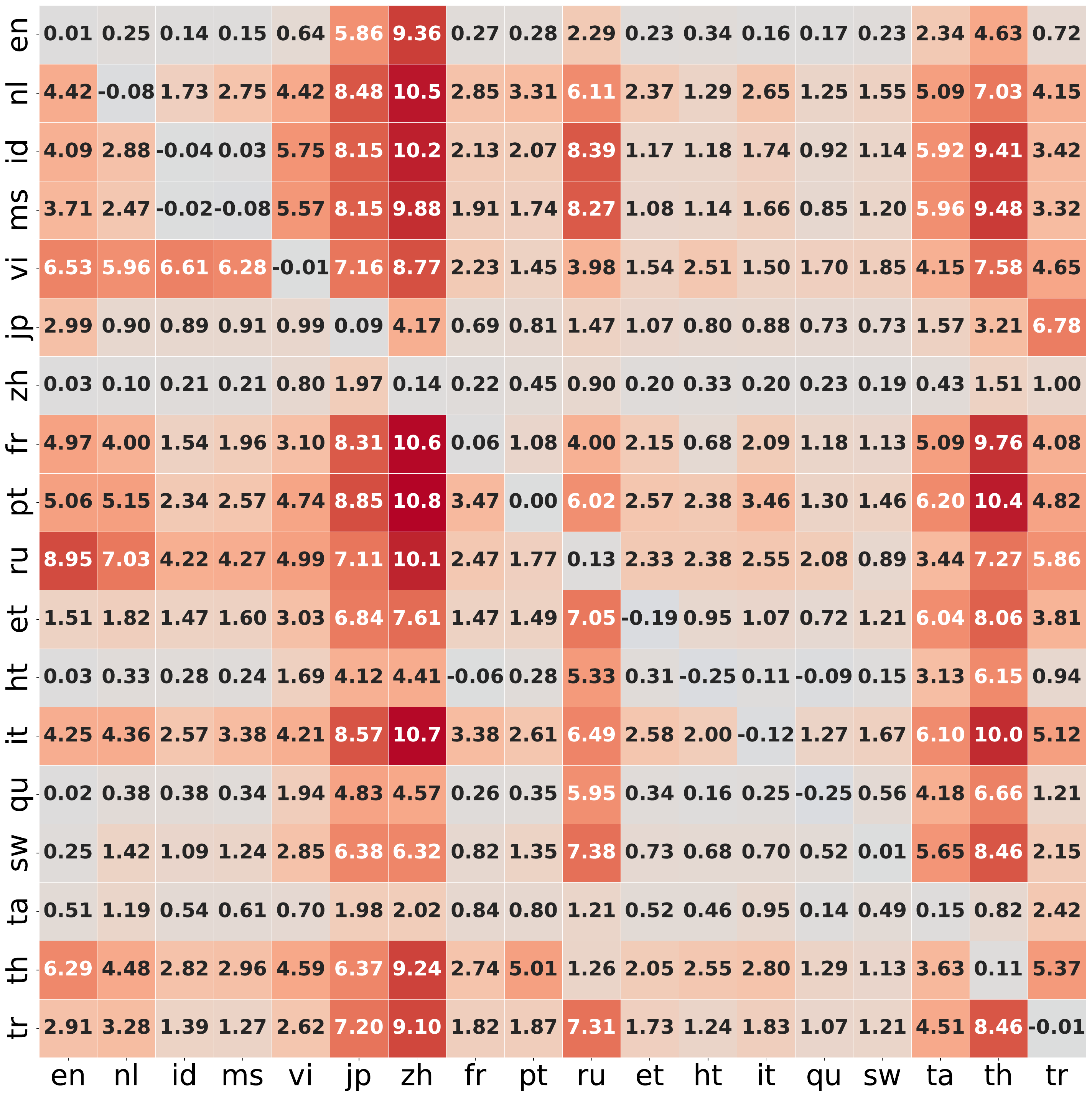}
    \caption{PPL changes after steering LAPE neurons with \steer{pmax} for SeaLLM3 1.5B.}
    \label{fig:ppl-seam}
\end{figure}

\begin{figure}[t]
\centering
    \includegraphics[width=\linewidth]{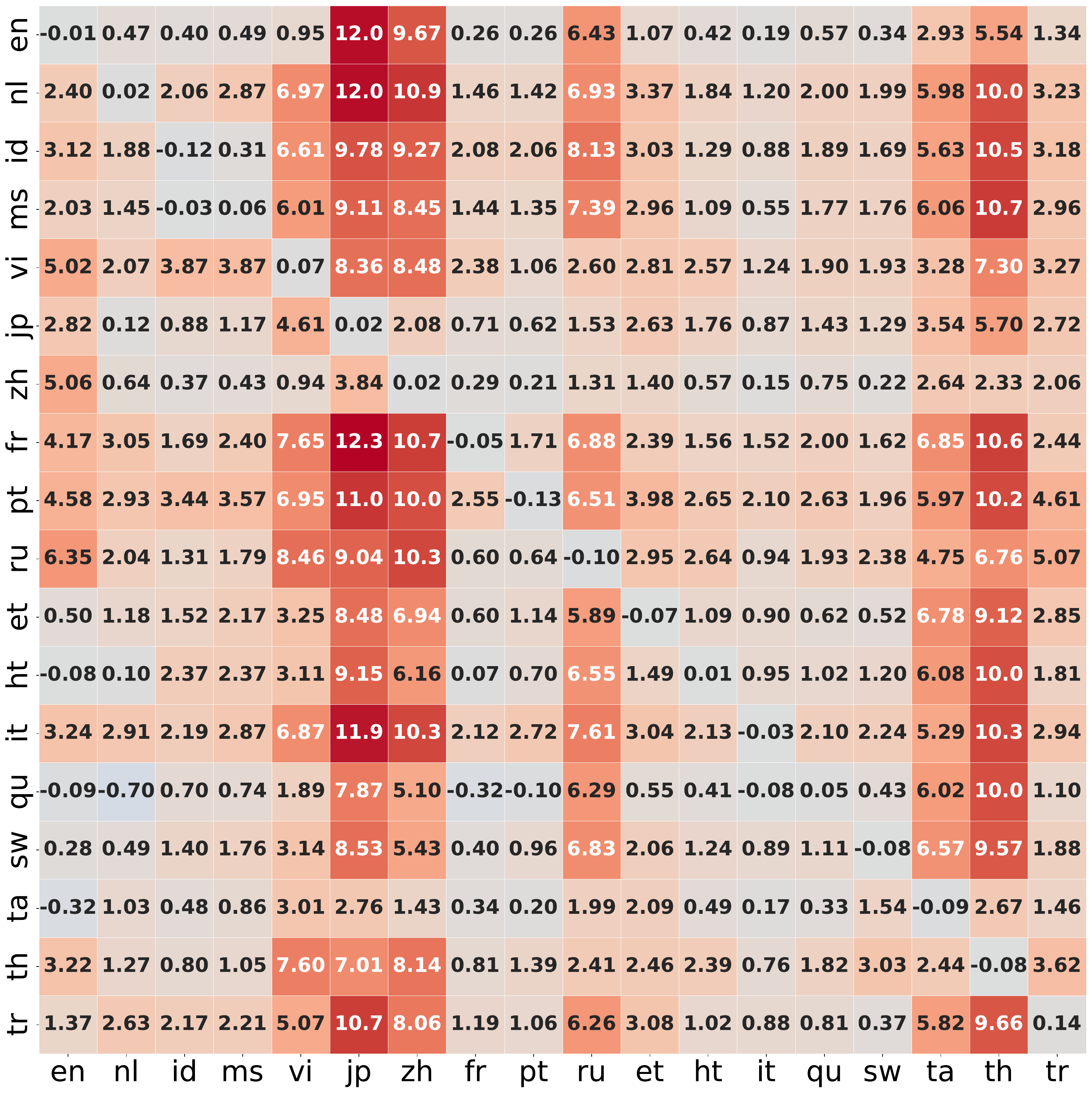}
    \caption{PPL changes after steering LAPE neurons with \steer{pmax} for SeaLLM3 7B.}
    \label{fig:ppl-sea}
\end{figure}

\begin{figure}[t]
\centering
    \includegraphics[width=\linewidth]{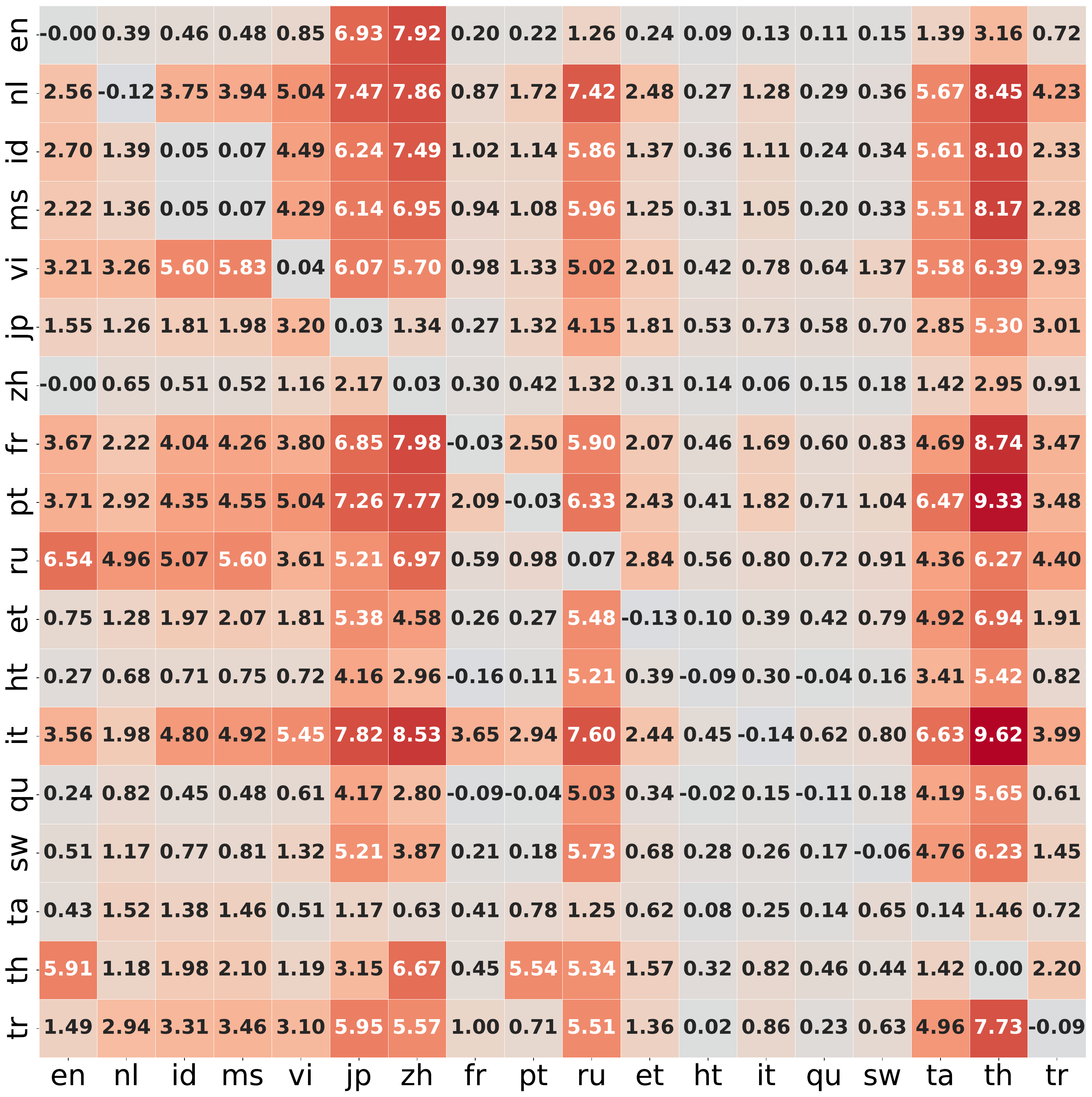}
    \caption{PPL changes after steering LAPE neurons with \steer{pmedian} for Qwen2.5 0.5 B.}
    \label{fig:ppl-med-qwenm}
\end{figure}

\begin{figure}[t]
\centering
    \includegraphics[width=\linewidth]{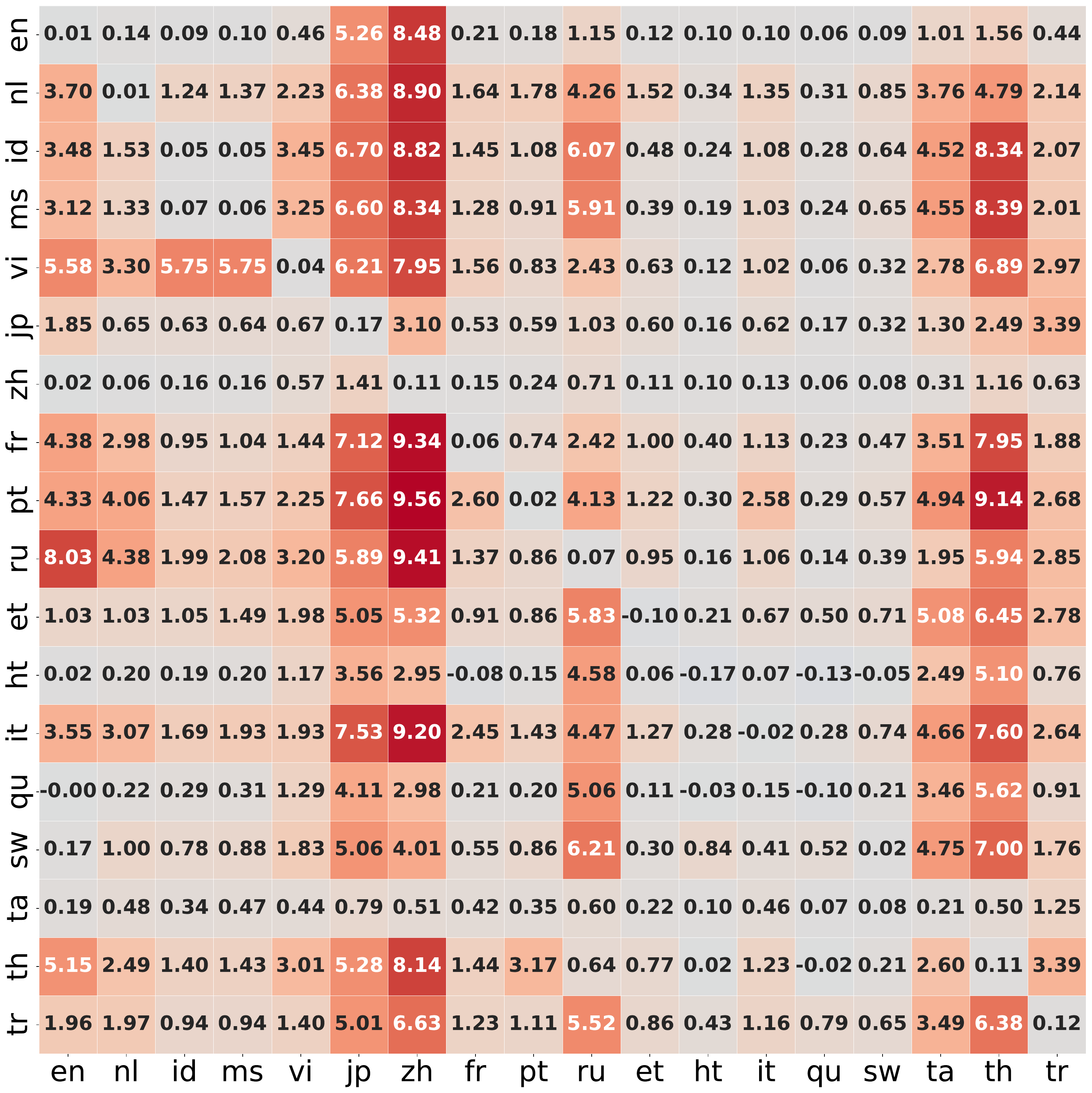}
    \caption{PPL changes after steering LAPE neurons with \steer{pmedian} for SeaLLMs3 1.5 B.}
    \label{fig:ppl-med-seam}
\end{figure}

\begin{figure}[t]
\centering
    \includegraphics[width=\linewidth]{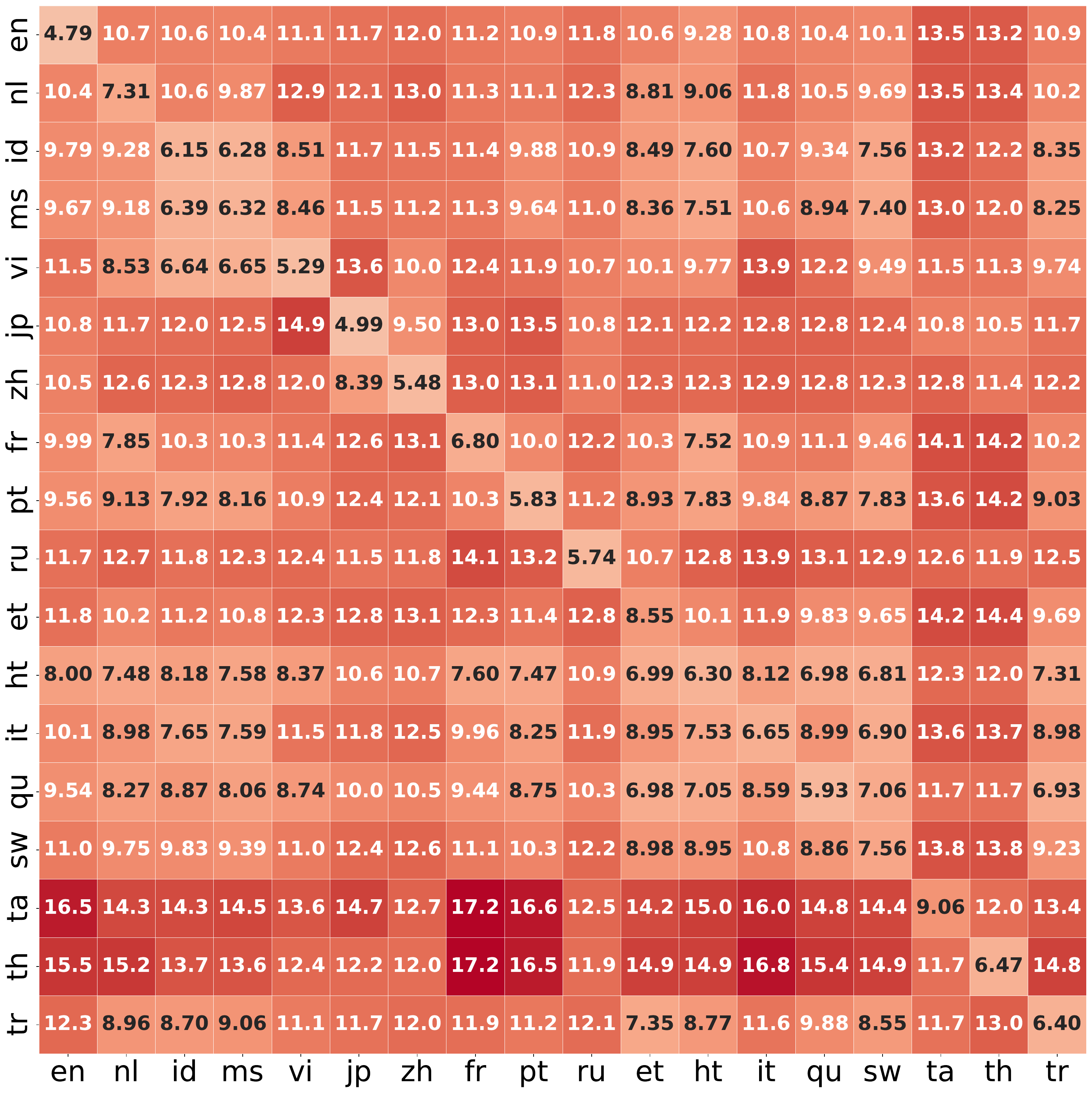}
    \caption{PPL changes after steering LAPE neurons with \steer{pmedian} for Gemma2 2B.}
    \label{fig:ppl-med-gemmam}
\end{figure}

\begin{figure}[t]
\centering
    \includegraphics[width=\linewidth]{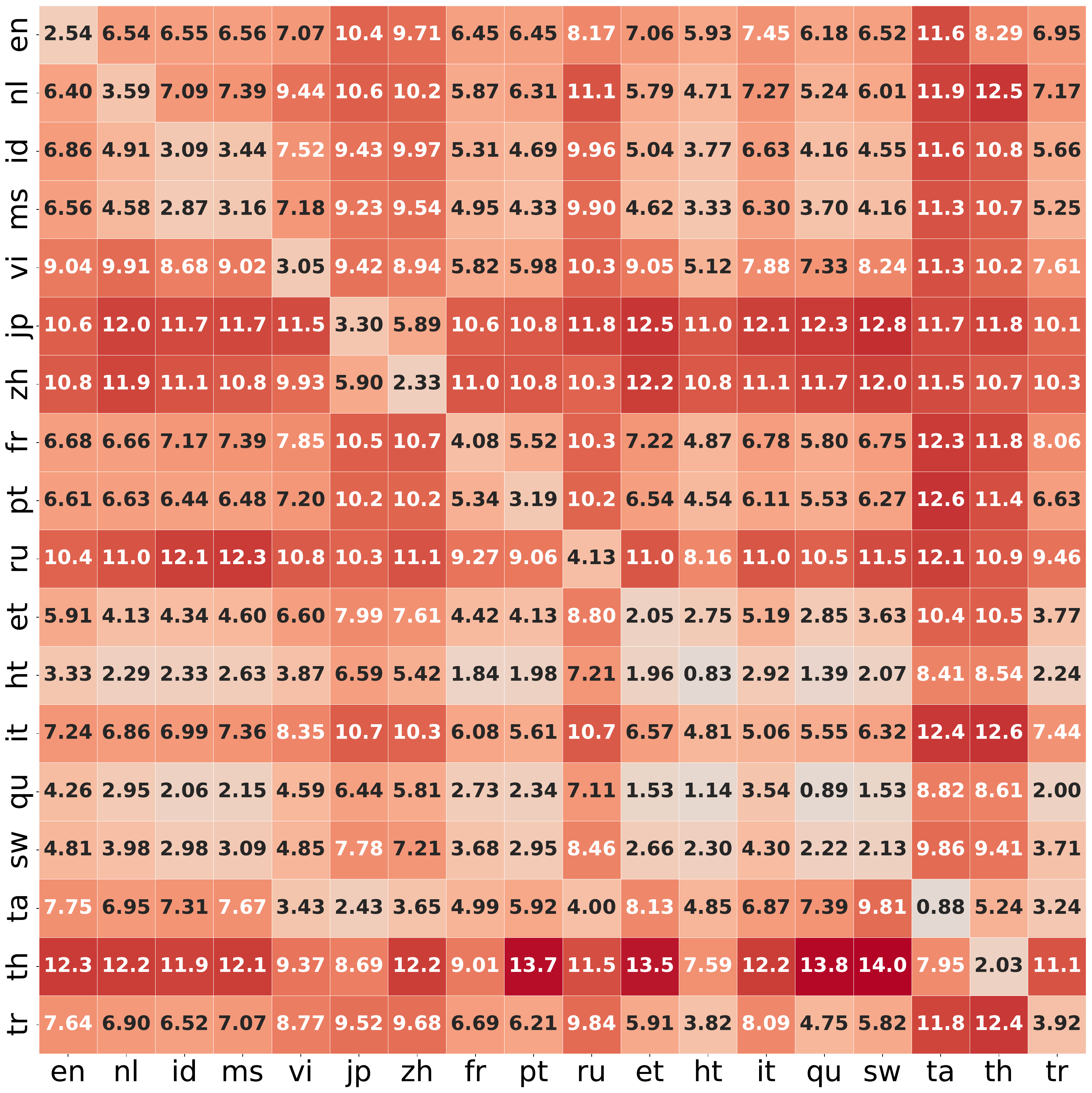}
    \caption{PPL changes after steering Baseline neurons with \steer{pmax} for Qwen2.5 0.5B.}
    \label{fig:ppl-raw-qwenm}
\end{figure}

\begin{figure}[t]
\centering
    \includegraphics[width=\linewidth]{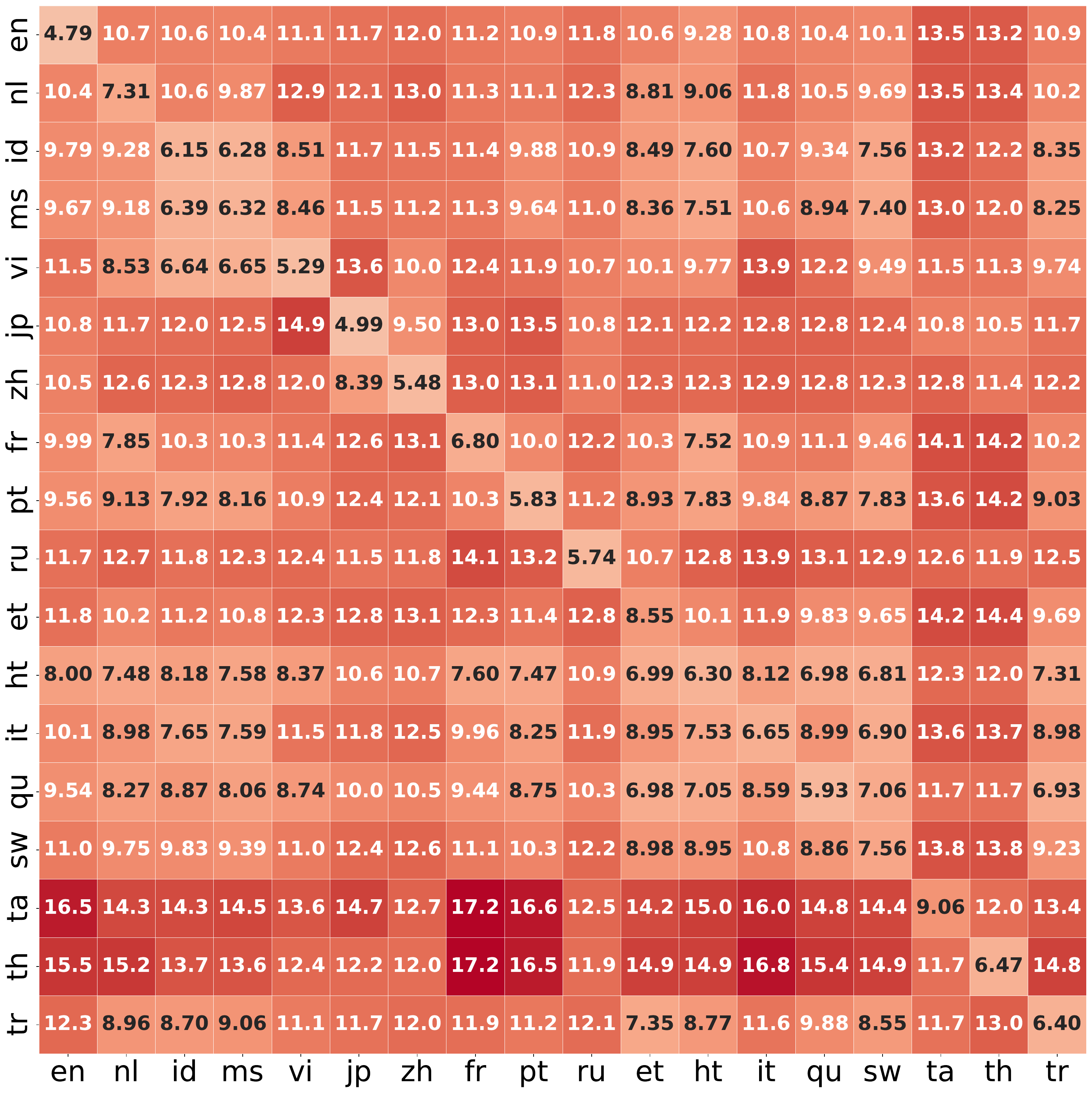}
    \caption{PPL changes after steering Baseline neurons with \steer{pmax} for Gemma2 2B.}
    \label{fig:ppl-raw-gemmam}
\end{figure}

\begin{figure}[t]
\centering
    \includegraphics[width=\linewidth]{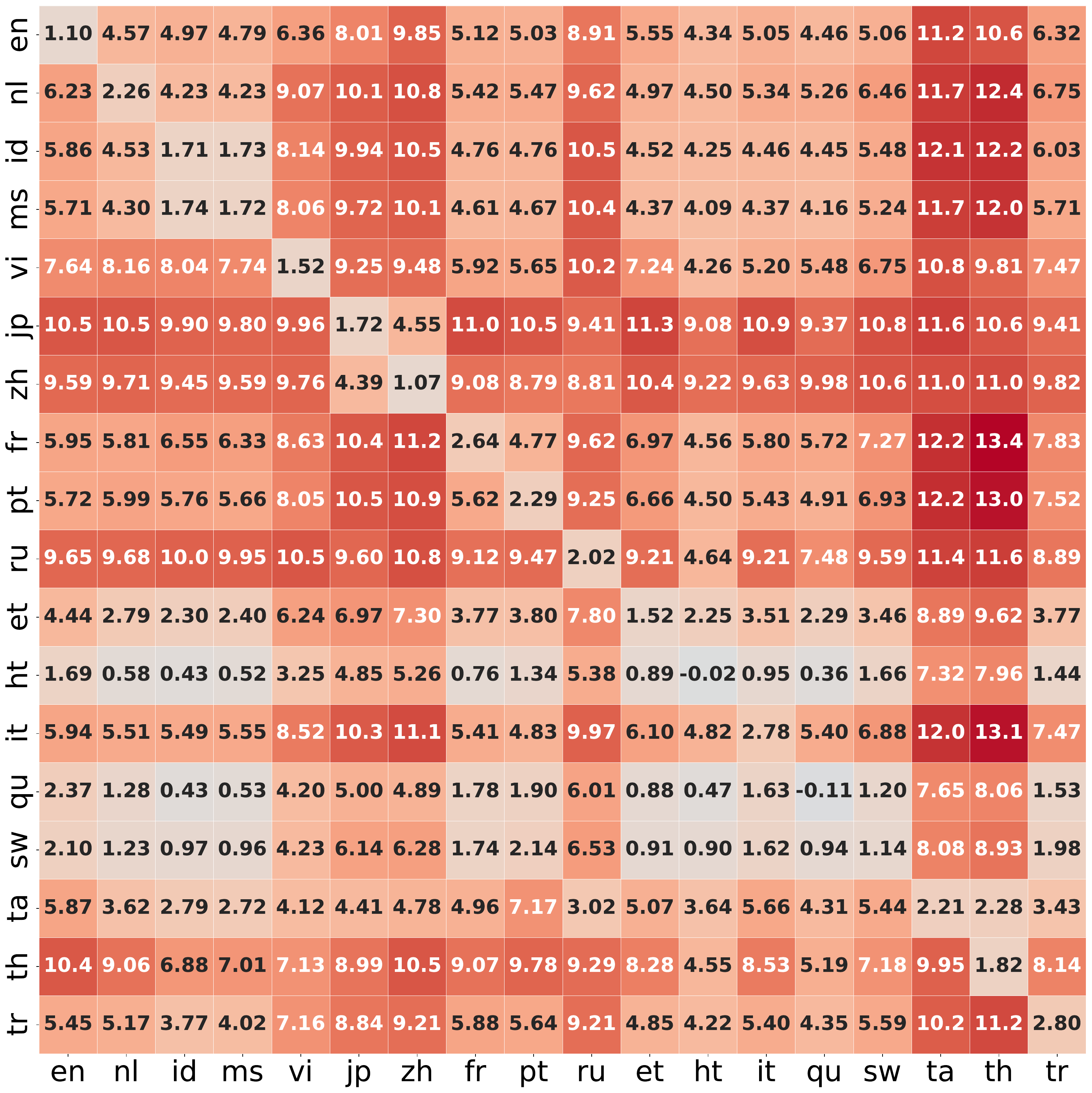}
    \caption{PPL changes after steering baseline neurons with \steer{pmax} for SeaLLM3 1.5B.}
    \label{fig:ppl-raw-seam}
\end{figure}

Perplexity changes for \steer{pmax} intervention of LAPE neurons are provided in Figure \ref{fig:ppl-gemmam}, \ref{fig:ppl-gemma}, \ref{fig:ppl-qwenm}, \ref{fig:ppl-qwen}, \ref{fig:ppl-seam}, and \ref{fig:ppl-sea}; \steer{pmedian} intervention are provided in Figure \ref{fig:ppl-med-gemmam}, \ref{fig:ppl-med-seam}, and \ref{fig:ppl-med-qwenm}. Perplexity changes for \steer{pmax} intervention of Baseline neurons are provided in Figure \ref{fig:ppl-raw-gemmam}, \ref{fig:ppl-raw-qwenm}, and \ref{fig:ppl-raw-seam}.

\section{BLEU Score Changes for Translation Tasks After Amplification}
\label{sec:appendix-translate}
\begin{figure}
    \centering
    \includegraphics[width=1.0\linewidth, trim=10 10 0 15, clip]{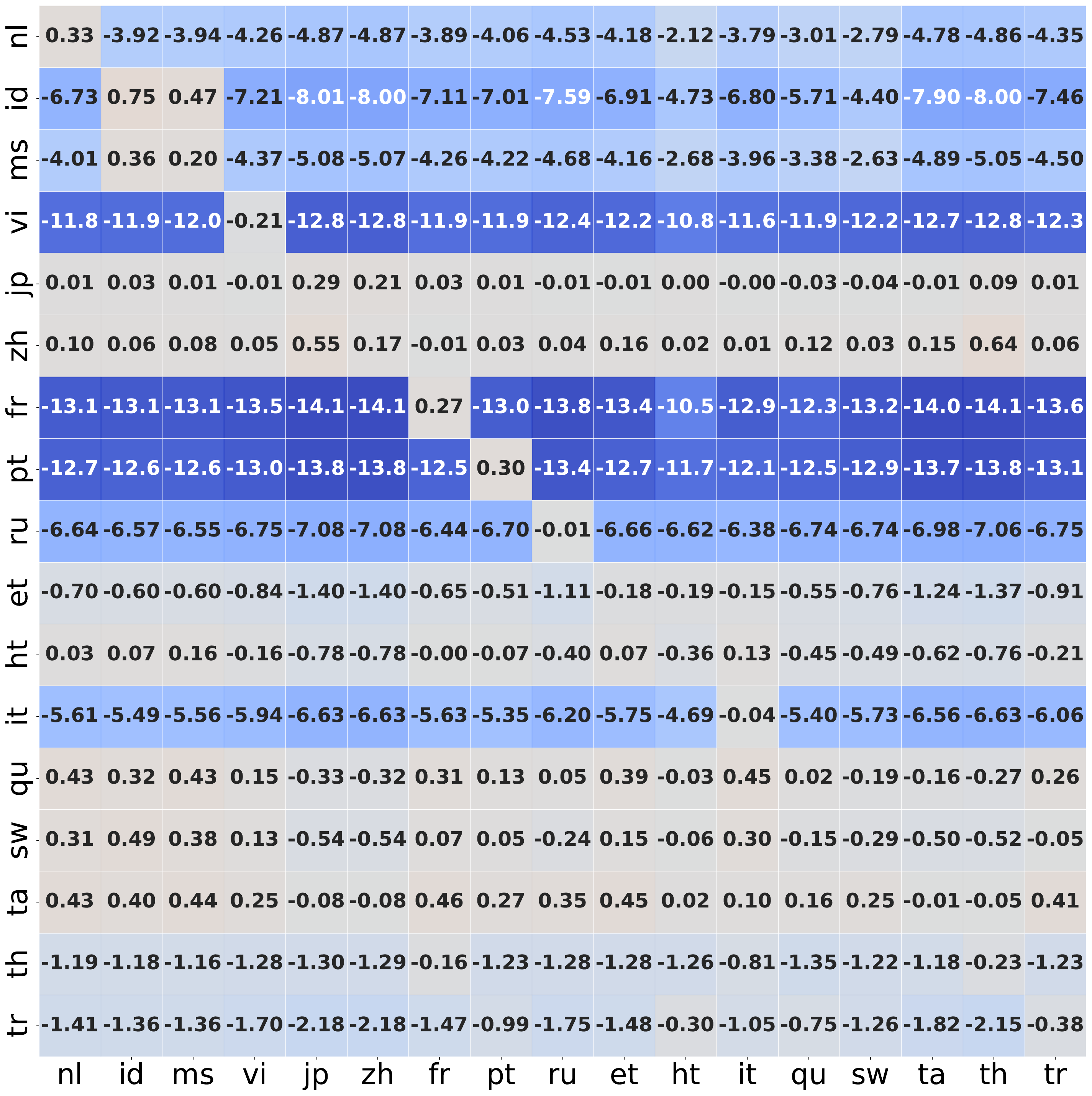}
    \caption{BLEU deltas for \textit{targeted} \steer{pmax} intervention in Qwen2.5 0.5B. Row $i$ is the target language and column $j$ is the intervention language. For example, row $x$ col $y$ represents BLEU score when translating English to language $x$ intervened by $y$ neurons.}
    \label{fig:bleu-cross-targeted}
\end{figure}

\begin{figure*}
    \centering
    \includegraphics[width=\textwidth, trim=50 400 50 40, clip]{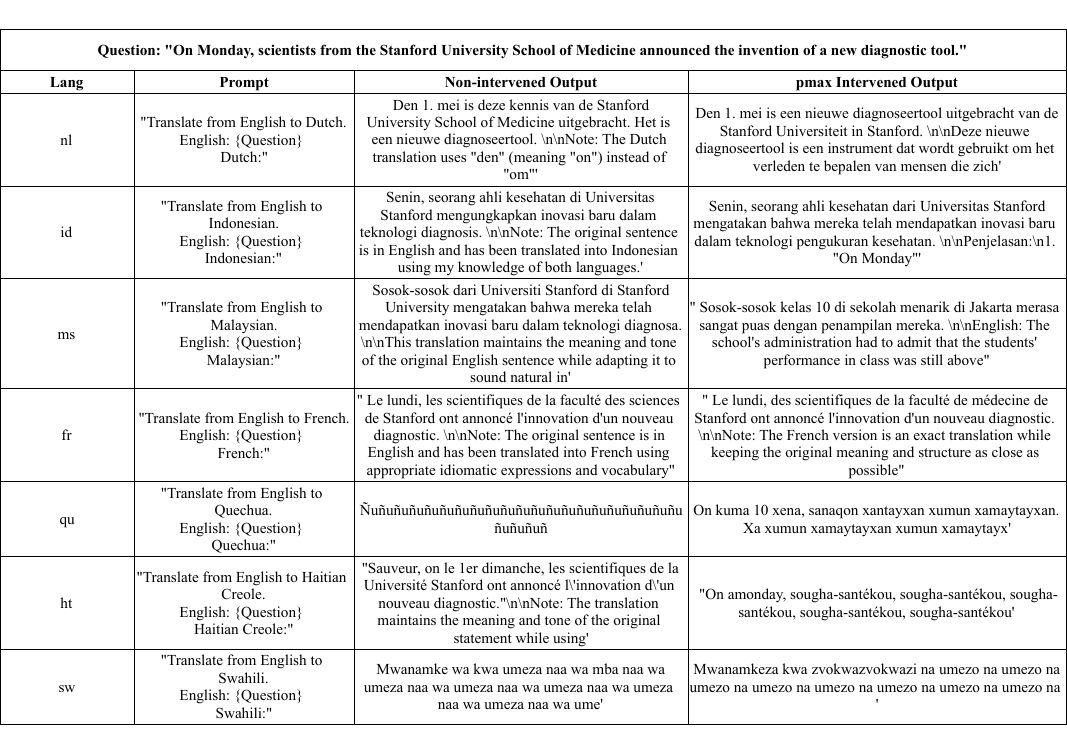}
    \caption{A sample of Qwen2.5 0.5B output for targeted prompt translation of \texttt{en} to \texttt{nl}, \texttt{id},\texttt{ms},\texttt{fr},\texttt{qu},\texttt{ht}, \texttt{sw}.}
    \label{fig:sentence-target}
\end{figure*}

\begin{figure*}
    \centering
    \includegraphics[width=\textwidth, trim=50 470 50 40, clip]{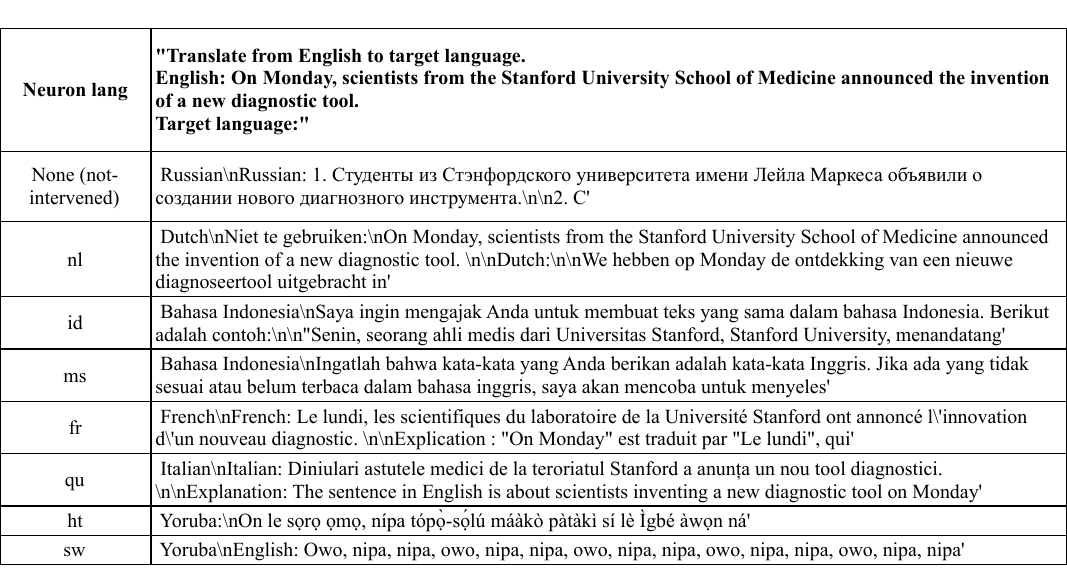}
    \caption{A sample of Qwen2.5 0.5B output for non-targeted prompt translation of \texttt{en} to \texttt{nl}, \texttt{id},\texttt{ms},\texttt{fr},\texttt{qu},\texttt{ht}, \texttt{sw}.}
    \label{fig:sentence-nontarget}
\end{figure*}

For the self-language intervention,
BLEU score changes for \steer{pmax} intervention of LAPE neurons with non-targeted prompt is provided in Table \ref{tab:bleu}. Sample of BLEU delta heatmap (self- and cross-intervention) is provided in Figure \ref{fig:bleu-cross-targeted}. BLEU score changes for \steer{pmedian} intervention of LAPE neurons with targeted prompt is provided in Table \ref{tab:bleu-pmed}. \steer{pmax} intervention on Baseline neurons with targeted prompt is provided in Table \ref{tab:bleu-baseline}. Samples of translation output for both targeted and non-targeted prompts are presented in Figure \ref{fig:sentence-target} and \ref{fig:sentence-nontarget}.

\section{Delta Accuracies of Reasoning and Knowledge Tasks After Amplification}
\label{sec:appendix-acc}
\begin{figure}[t]
\centering

\begin{subfigure}{0.49\columnwidth}
    \includegraphics[width=\linewidth]{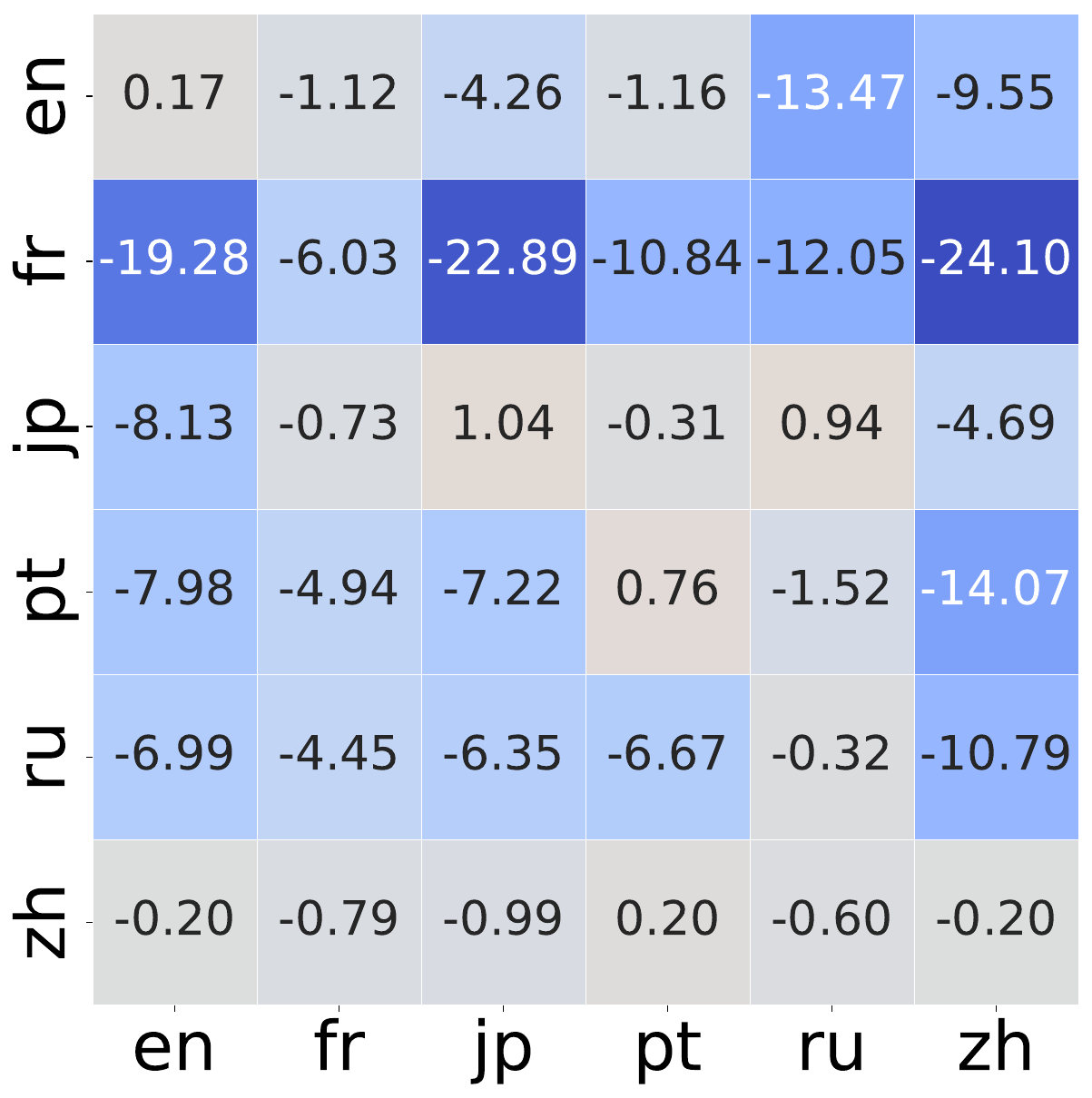}
    \caption{SeaLLMv3 1.5B}
\end{subfigure}
\hfill
\begin{subfigure}{0.49\columnwidth}
    \includegraphics[width=\linewidth]{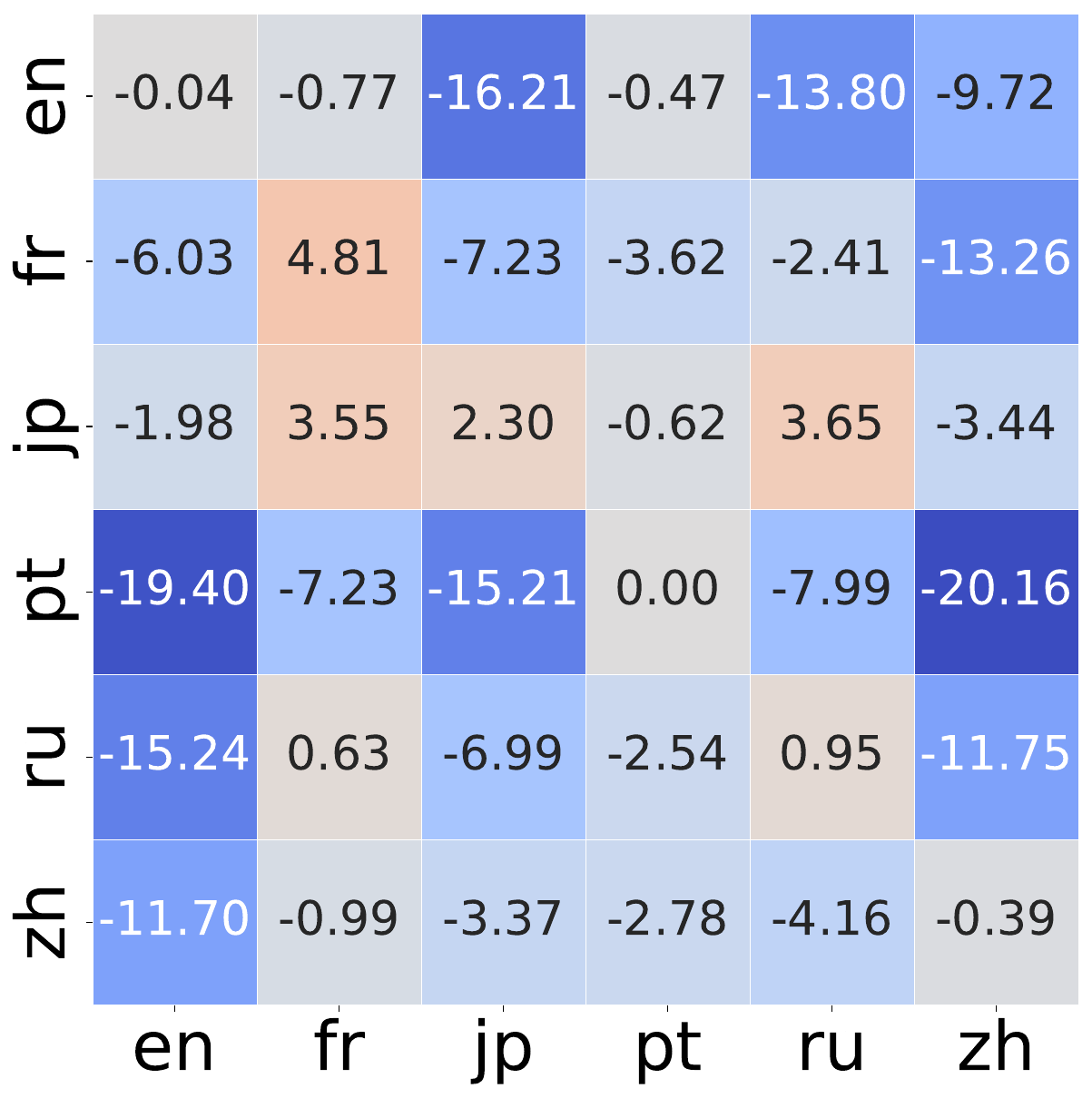}
    \caption{SeaLLMv3 7B}
\end{subfigure}

\caption{Delta XWinograd accuracy after steering LAPE neurons with \steer{pmax} for SeaLLMv3 1.5B and SeaLLMv3 7B. Row \textit{i} is the initial language, column \textit{j} is the language of intervention.}
\label{fig:lape-xwin-sea}
\end{figure}

\begin{figure}[t]
\centering

\begin{subfigure}{0.49\columnwidth}
    \includegraphics[width=\linewidth]{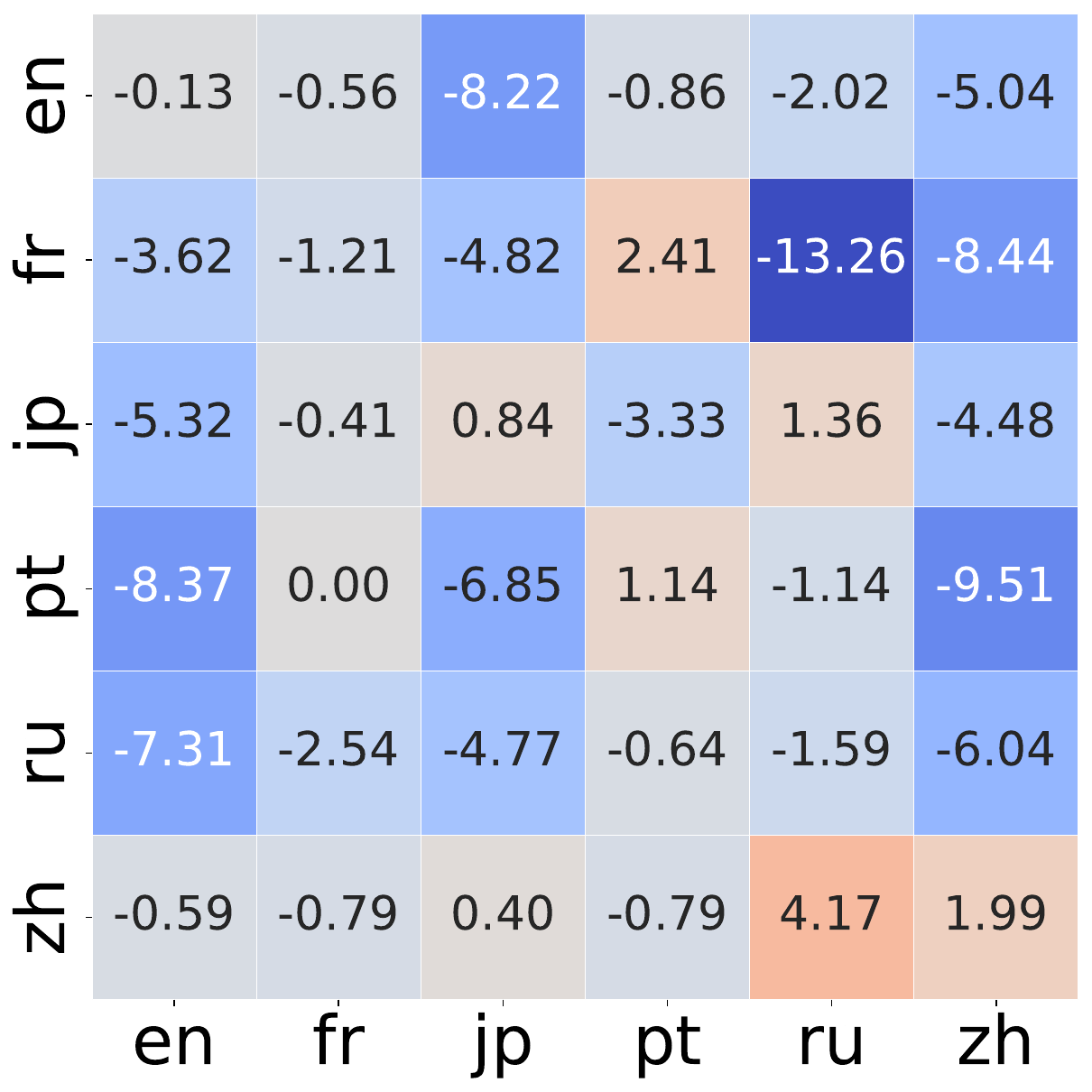}
    \caption{Qwen2.5 0.5B}
\end{subfigure}
\hfill
\begin{subfigure}{0.49\columnwidth}
    \includegraphics[width=\linewidth]{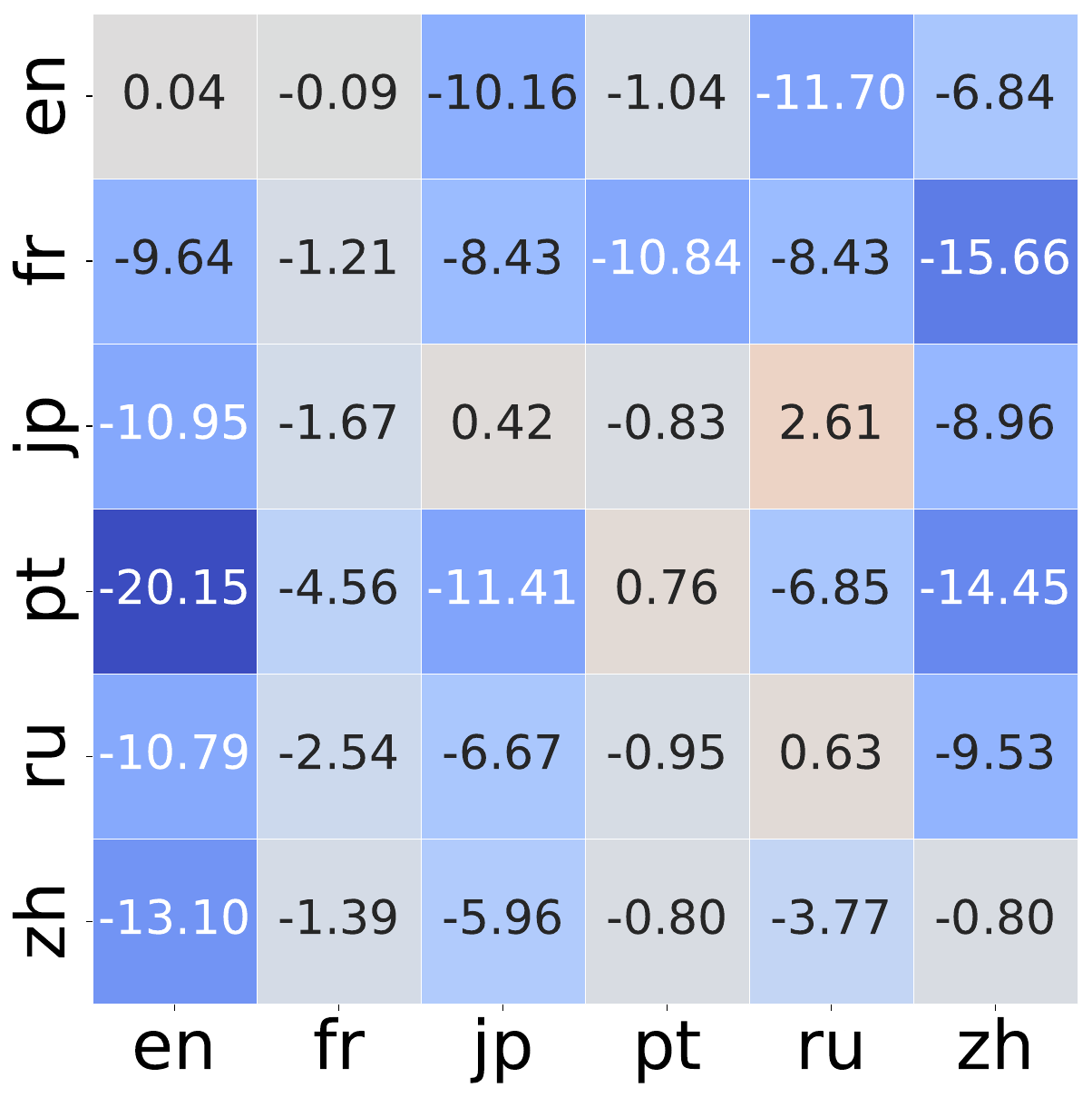}
    \caption{Qwen2.5 7B}
\end{subfigure}

\caption{Delta XWinograd accuracy after steering LAPE neurons with \steer{pmax} for Qwen2.5 0.5B and Qwen2.5 7B. Row \textit{i} is the initial language, column \textit{j} is the language of intervention.}
\label{fig:lape-xwin-qwen}
\end{figure}

\begin{figure}[t]
\centering

\begin{subfigure}{0.49\columnwidth}
    \includegraphics[width=\linewidth]{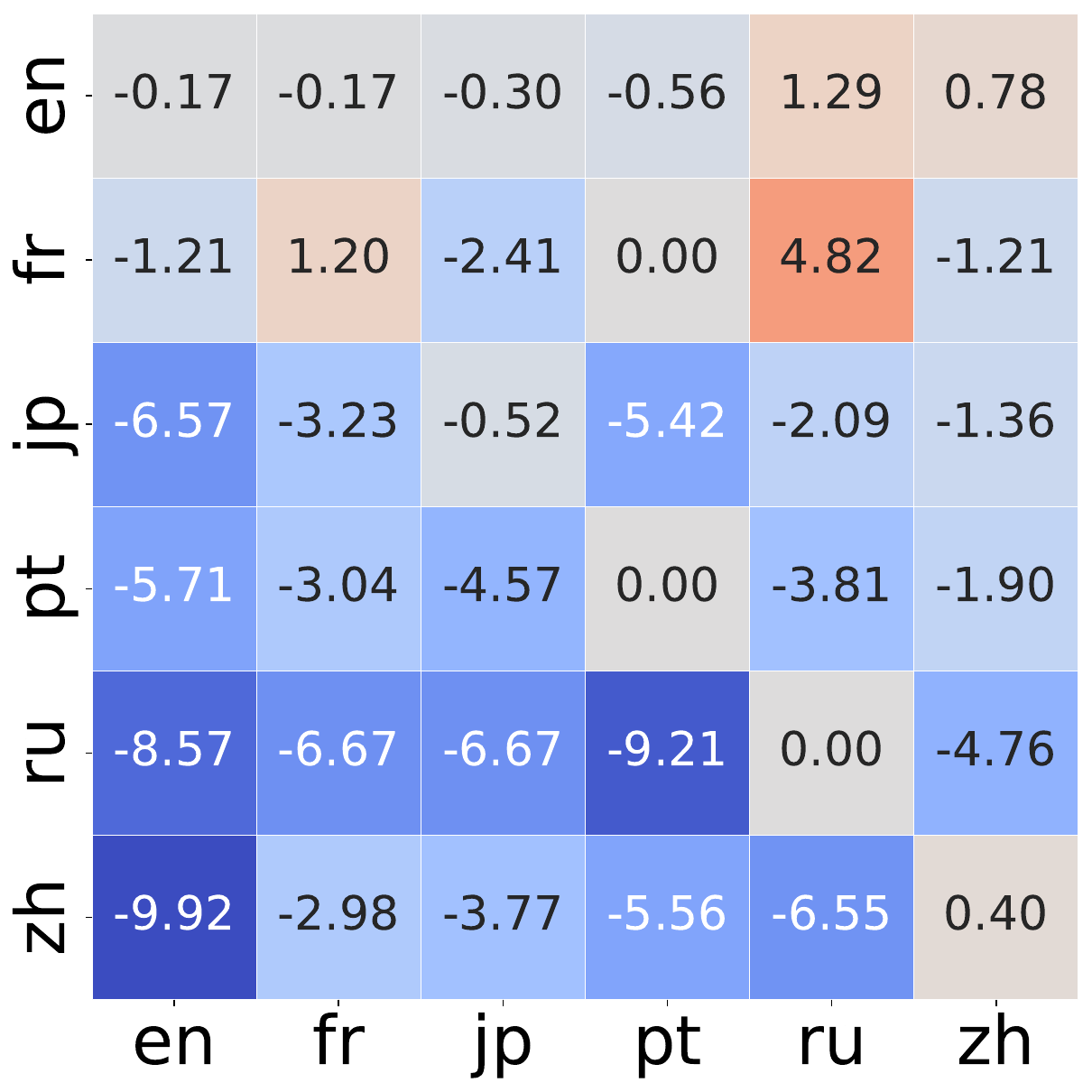}
    \caption{Gemma2 2B}
\end{subfigure}
\hfill
\begin{subfigure}{0.49\columnwidth}
    \includegraphics[width=\linewidth]{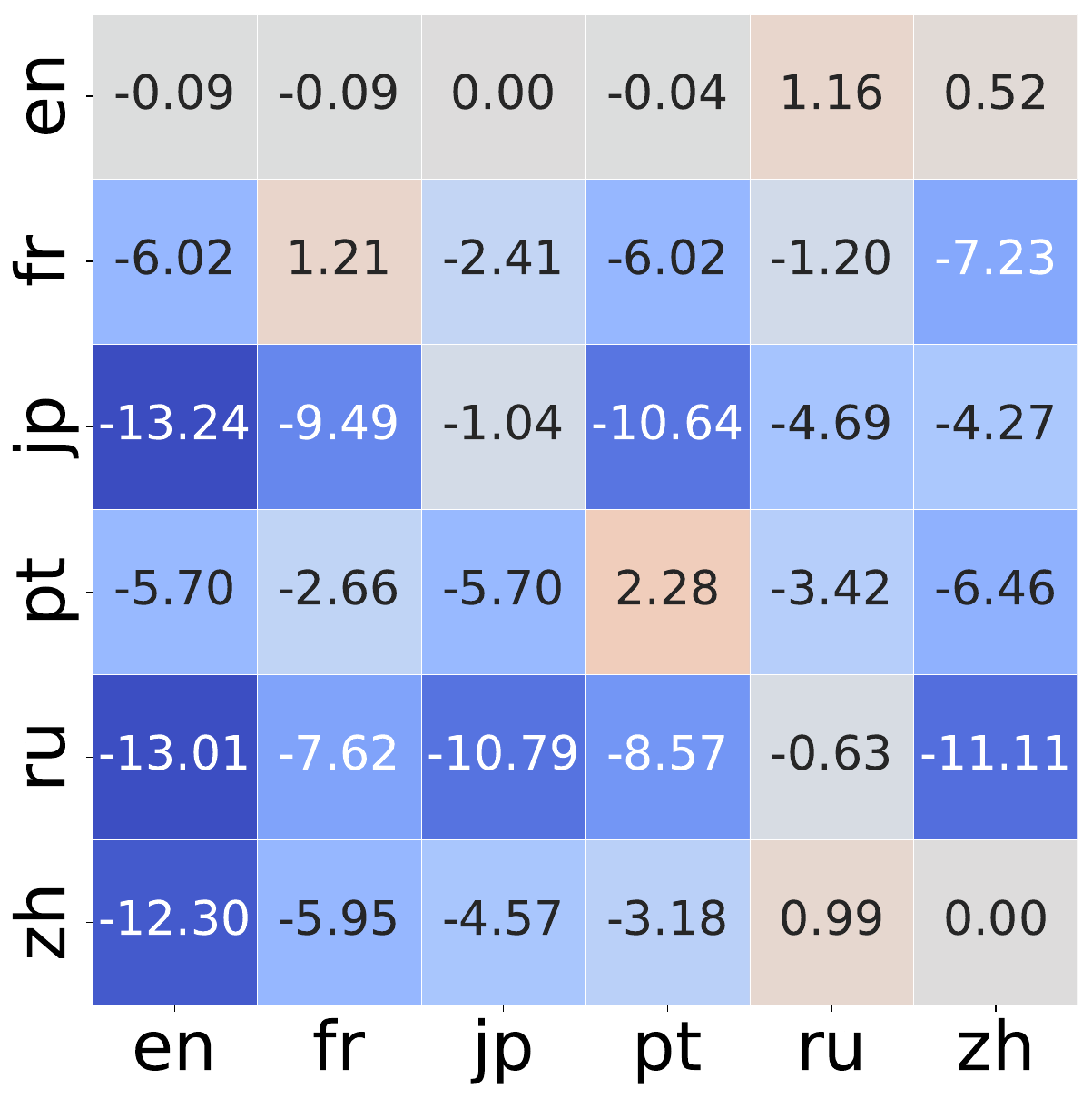}
    \caption{Gemma2 9B}
\end{subfigure}

\caption{Delta XWinograd accuracy after steering LAPE neurons with \steer{pmax} for Gemma2 2B and 9B. Row \textit{i} is the initial language, column \textit{j} is the language of intervention.}
\label{fig:lape-xwin-gemma}
\end{figure}

\begin{figure}
    \centering
    \includegraphics[width=\linewidth]{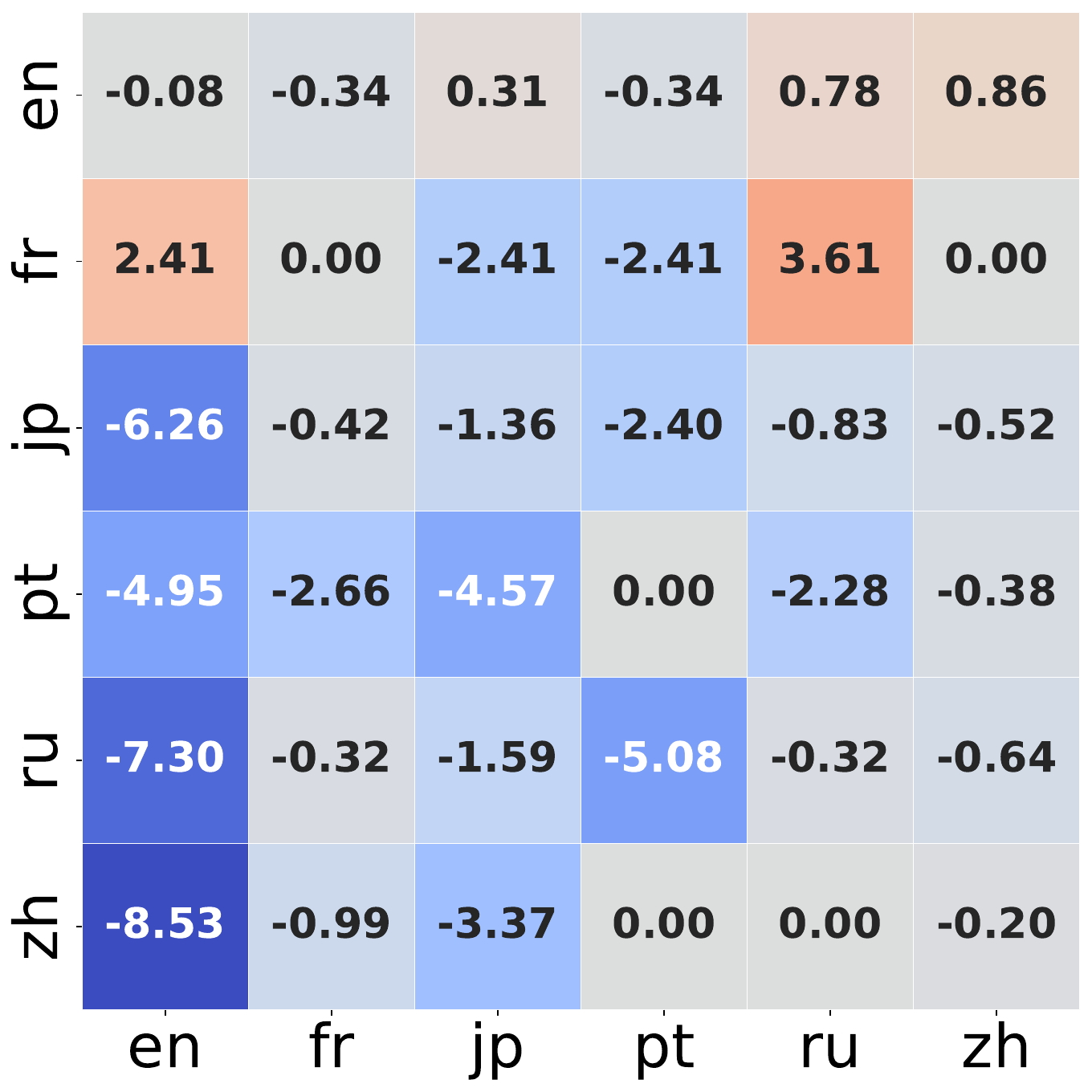}
    \caption{Delta XWinograd accuracy after steering LAPE neurons with \steer{pmedian} for Gemma2 2B. Row \textit{i} is the initial language, column \textit{j} is the language of intervention.}
    \label{fig:lape-xwin-med-gemmam}
\end{figure}

\begin{figure}
    \centering
    \includegraphics[width=\linewidth]{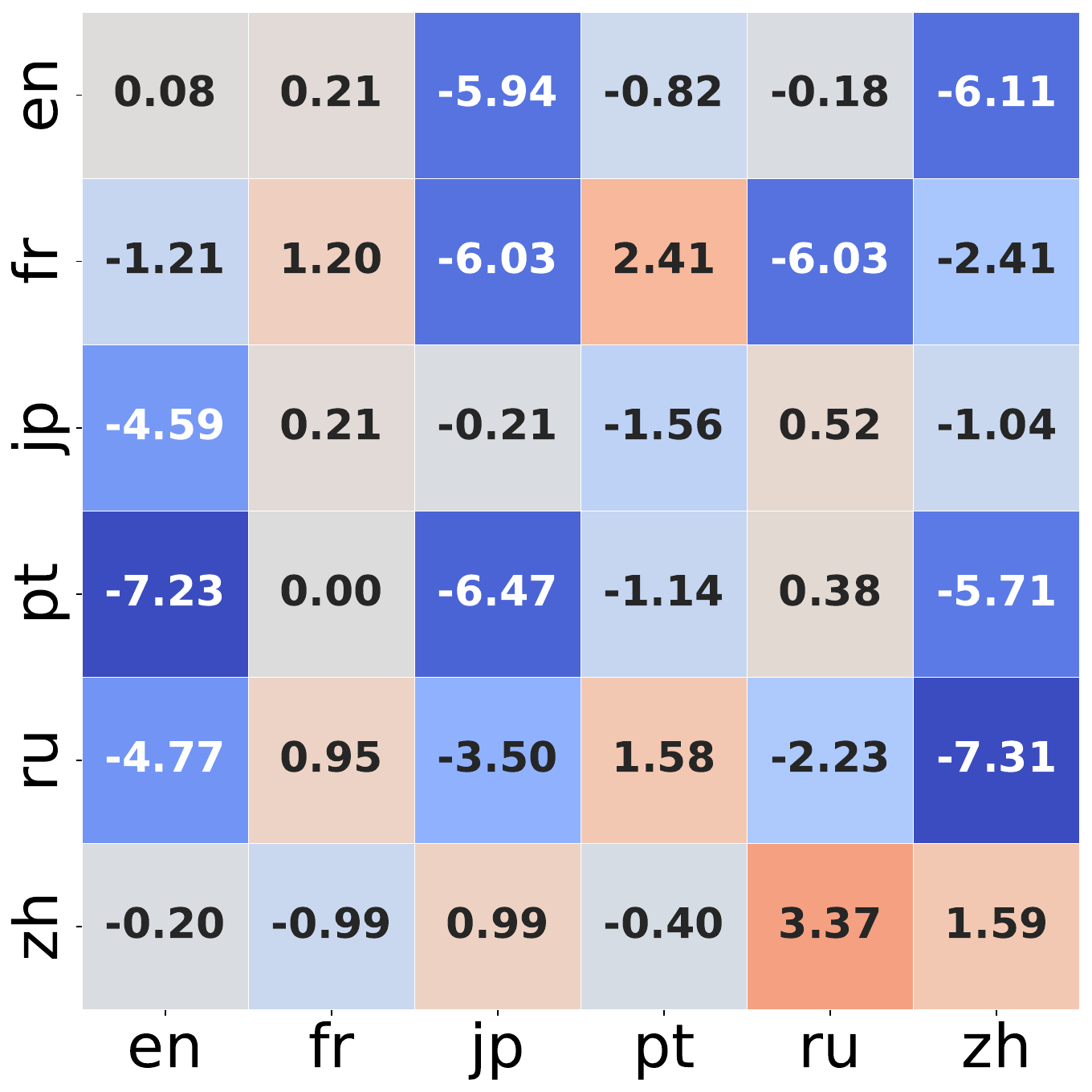}
    \caption{Delta XWinograd accuracy after steering LAPE neurons with \steer{pmedian} for Qwen2.5 0.5B. Row \textit{i} is the initial language, column \textit{j} is the language of intervention.}
    \label{fig:lape-xwin-med-qwenm}
\end{figure}

\begin{figure}
    \centering
    \includegraphics[width=\linewidth]{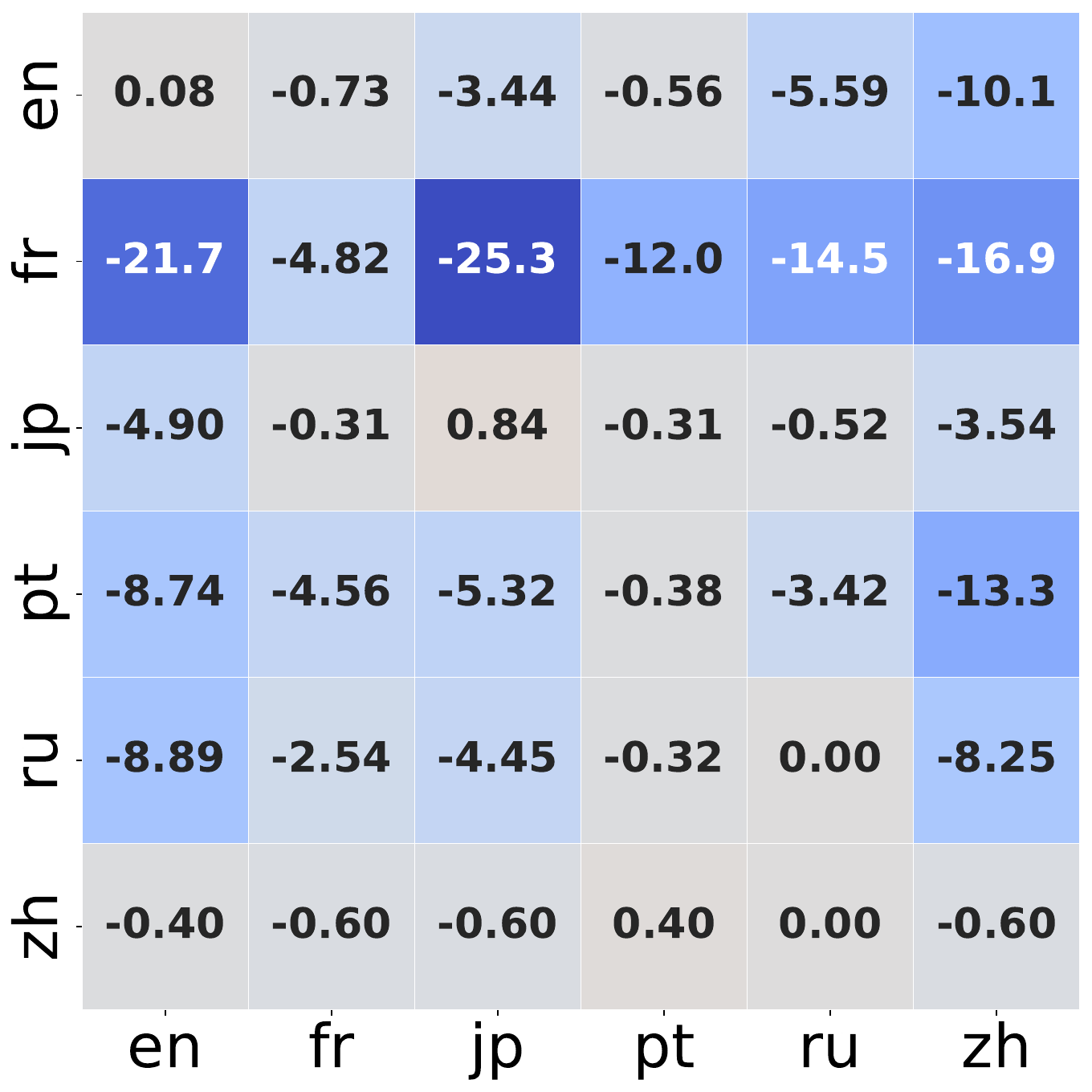}
    \caption{Delta XWinograd accuracy after steering LAPE neurons with \steer{pmedian} for SeaLLMs 1.5B. Row \textit{i} is the initial language, column \textit{j} is the language of intervention.}
    \label{fig:lape-xwin-med-seam}
\end{figure}

\begin{figure}[t]
\centering
\includegraphics[width=\linewidth]{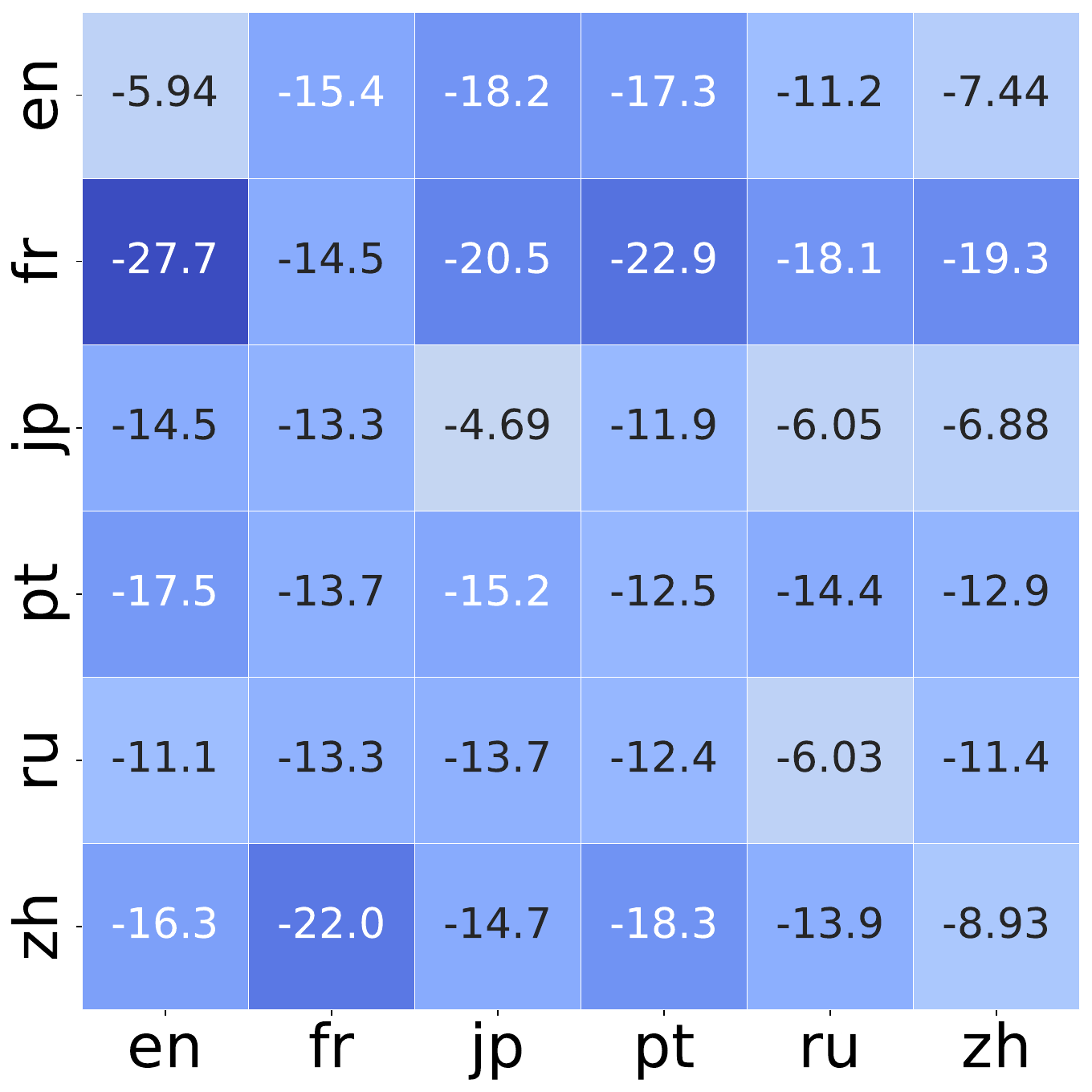}
\caption{Delta XWinograd accuracy after steering baseline neurons with \steer{pmax} for SeaLLMv3 1.5B. Row \textit{i} is the initial language, column \textit{j} is the language of intervention.}
\label{fig:raw-xwin-sea}
\end{figure}

\begin{figure}[t]
\centering

    \includegraphics[width=\linewidth]{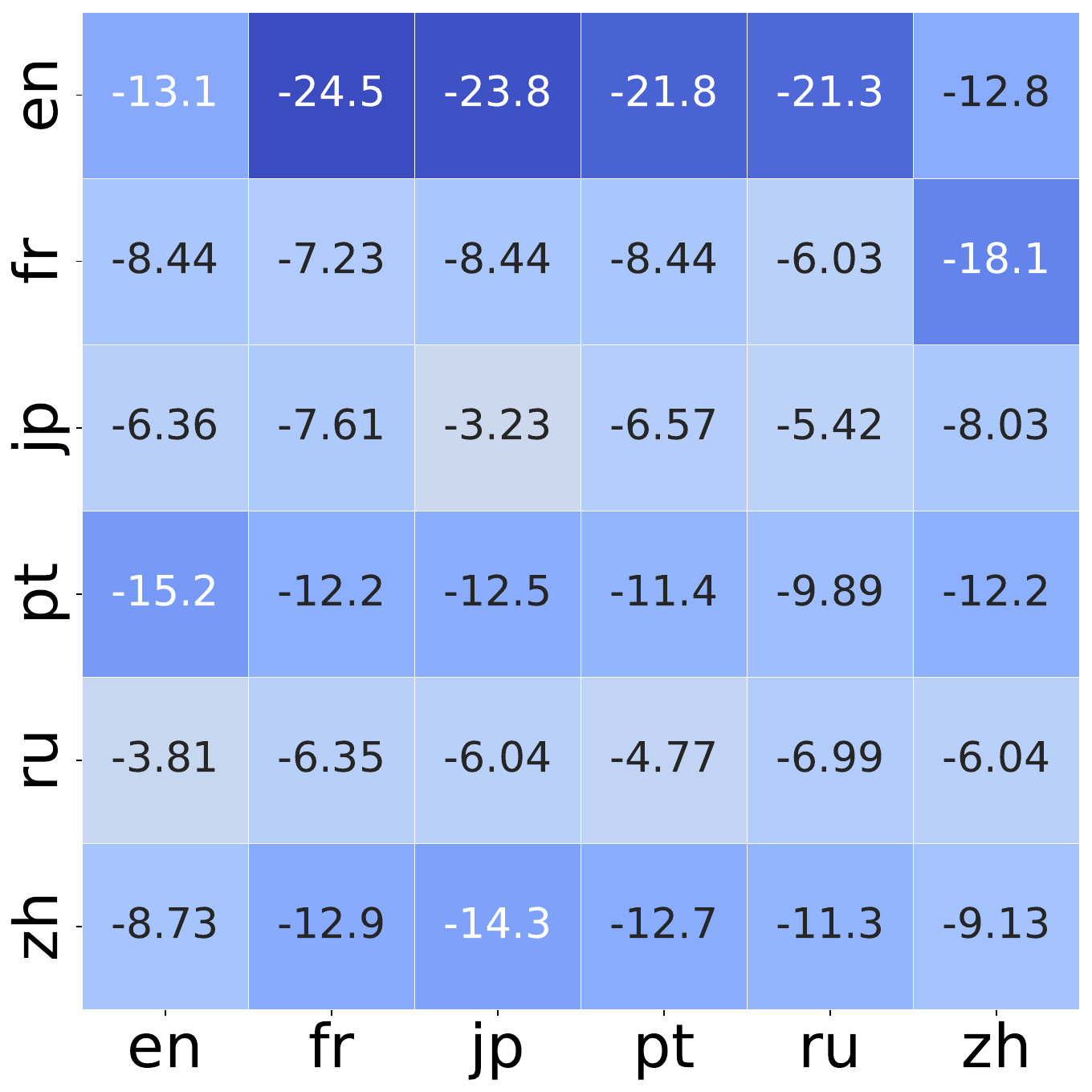}
    % \caption{Qwen2.5 0.5B}

\caption{Delta XWinograd accuracy after steering baseline neurons with \steer{pmax} for Qwen2.5 0.5B. Row \textit{i} is the initial language, column \textit{j} is the language of intervention.}
\label{fig:raw-xwin-qwen}
\end{figure}

\begin{figure}[t]
\centering

    \includegraphics[width=\linewidth]{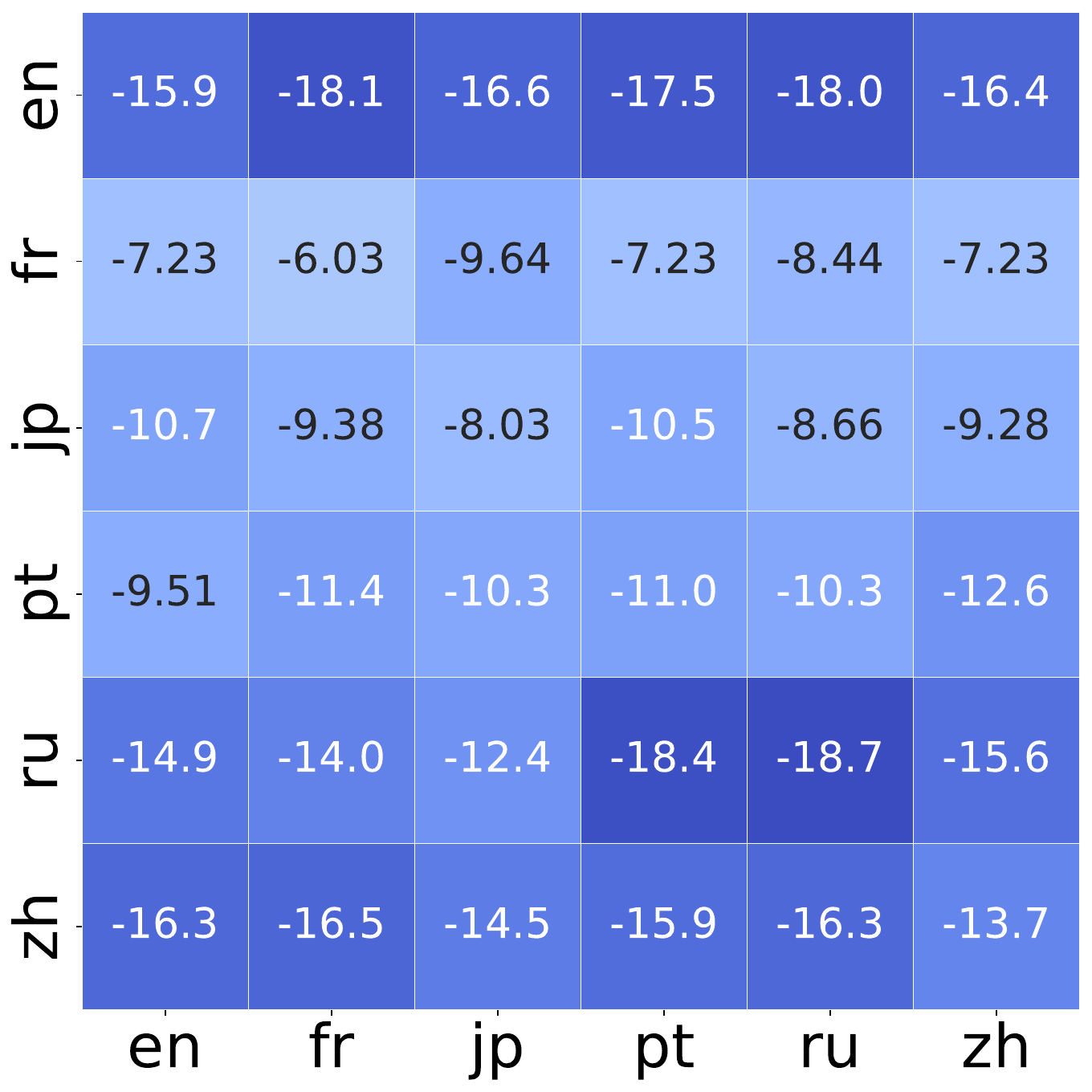}
    % \caption{Gemma2 2B}

\caption{Delta XWinograd accuracy after steering baseline neurons with \steer{pmax} for Gemma2 2B. Row \textit{i} is the initial language, column \textit{j} is the language of intervention.}
\label{fig:raw-xwin-gemma}
\end{figure}

Table \ref{tab:baseline-xwinograd} reveals the baseline (no intervention) XWinograd accuracy. Delta accuracies of XWinograd dataset under LAPE neurons when steered with \steer{pmax} are provided in Figure \ref{fig:lape-xwin-sea}, \ref{fig:lape-xwin-qwen} and \ref{fig:lape-xwin-gemma}; intervention results for \steer{pmedian} are provided in Figure \ref{fig:lape-xwin-med-gemmam}, \ref{fig:lape-xwin-med-qwenm}, and \ref{fig:lape-xwin-med-seam}. Delta XWinograd accuracies of \steer{pmax} intervention on  Baseline neurons are provided in Figure \ref{fig:raw-xwin-sea}, \ref{fig:raw-xwin-qwen}, and \ref{fig:raw-xwin-gemma}. 

\begin{figure}[t]
\centering
    \includegraphics[width=\linewidth]{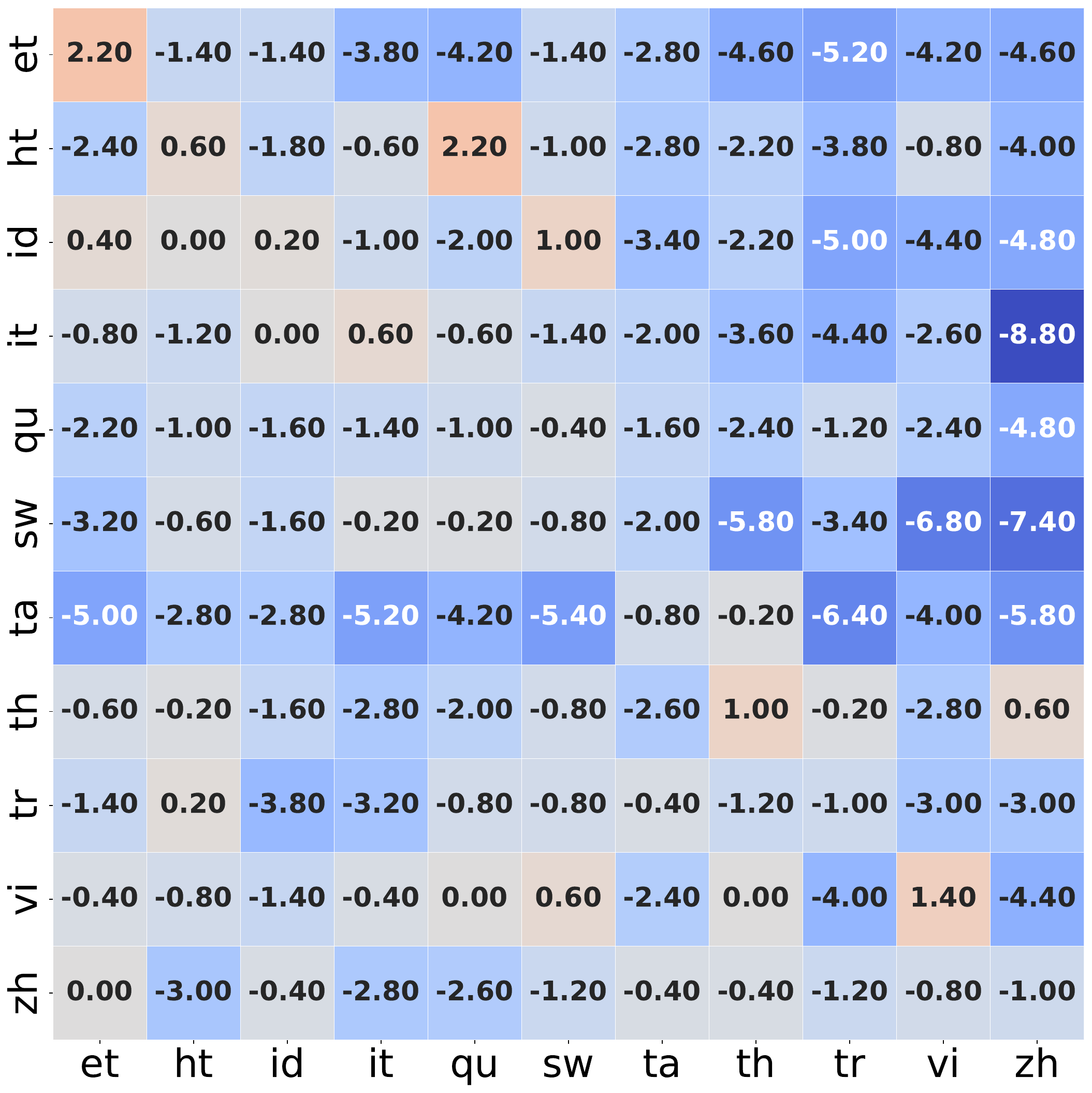}
    \caption{Delta XCOPA accuracy after steering LAPE neurons with \steer{pmax} for Gemma2 9B.}
    \label{fig:lape-xcopa-gemma}
\end{figure}

\begin{figure}[t]

\centering
    \includegraphics[width=\linewidth]{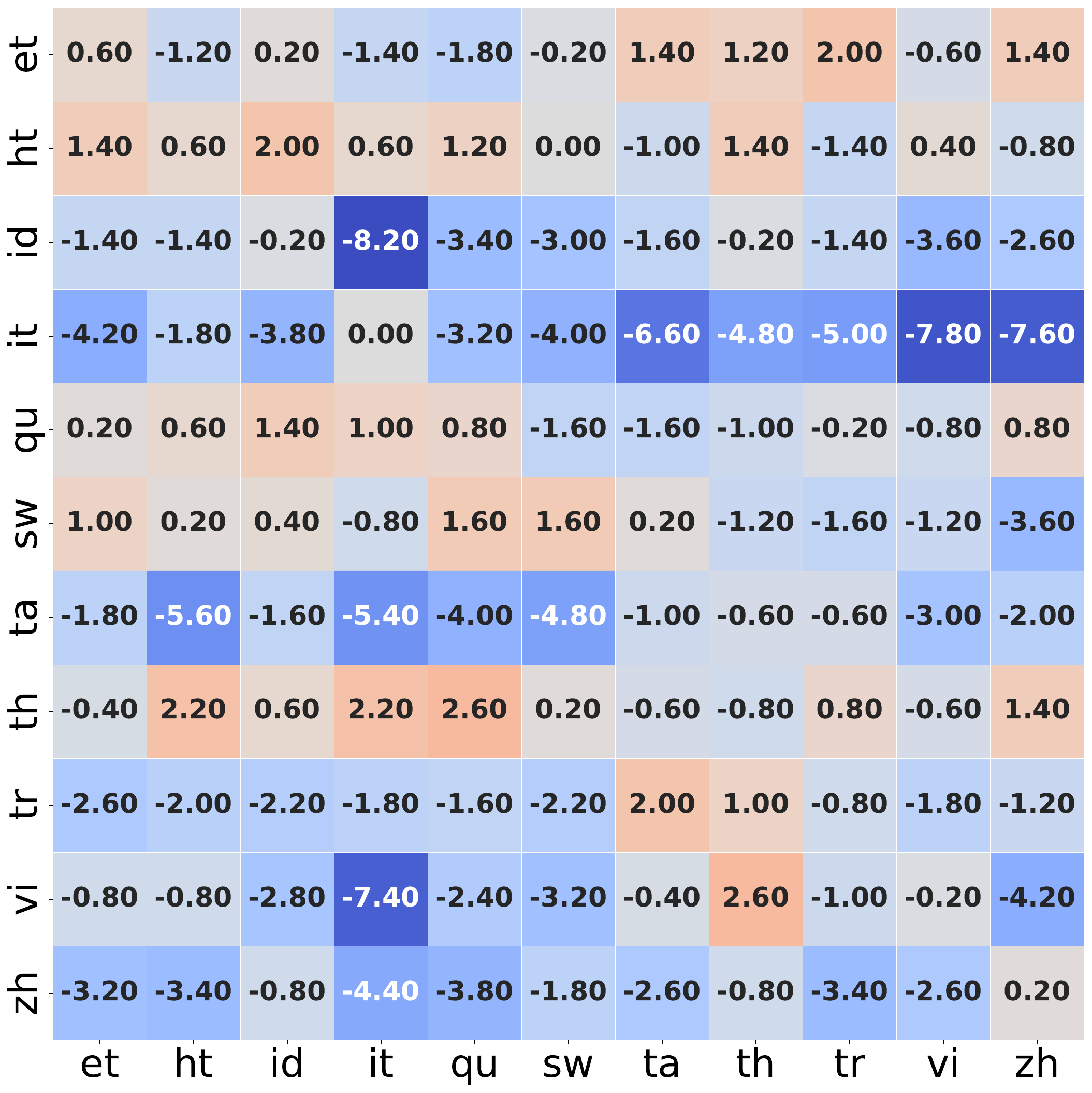}
    \caption{Delta XCOPA accuracy after steering LAPE neurons with \steer{pmax} for Gemma2 2B.}
    \label{fig:lape-xcopa-gemmam}

\end{figure}

\begin{figure}[t]
\centering
    \includegraphics[width=\linewidth]{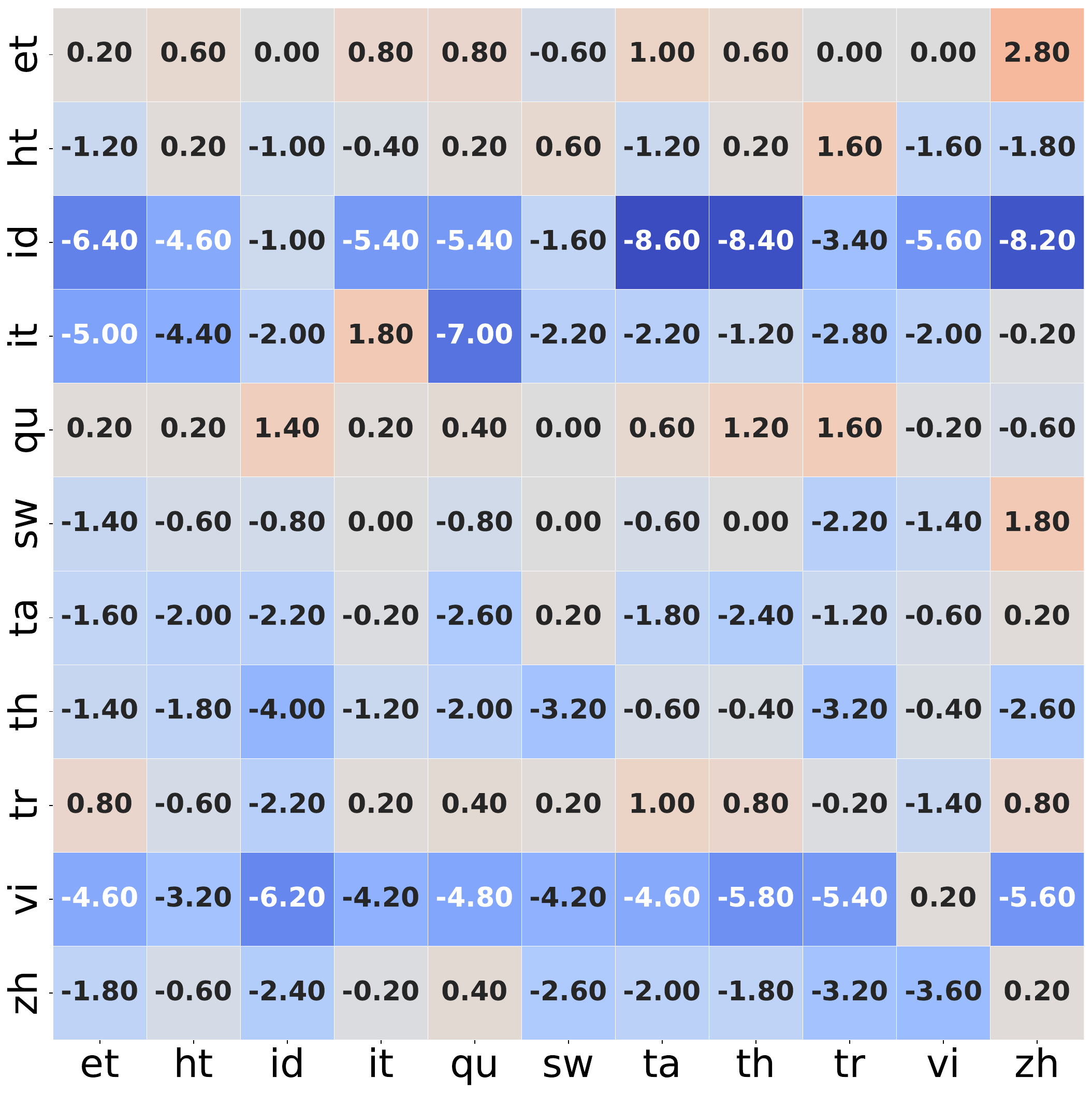}
    \caption{Delta XCOPA accuracy after steering LAPE neurons with \steer{pmax} for Qwen2.5 0.5B.}
    \label{fig:lape-xcopa-qwenm}

\end{figure}

\begin{figure}[t]
\centering
    \includegraphics[width=\linewidth]{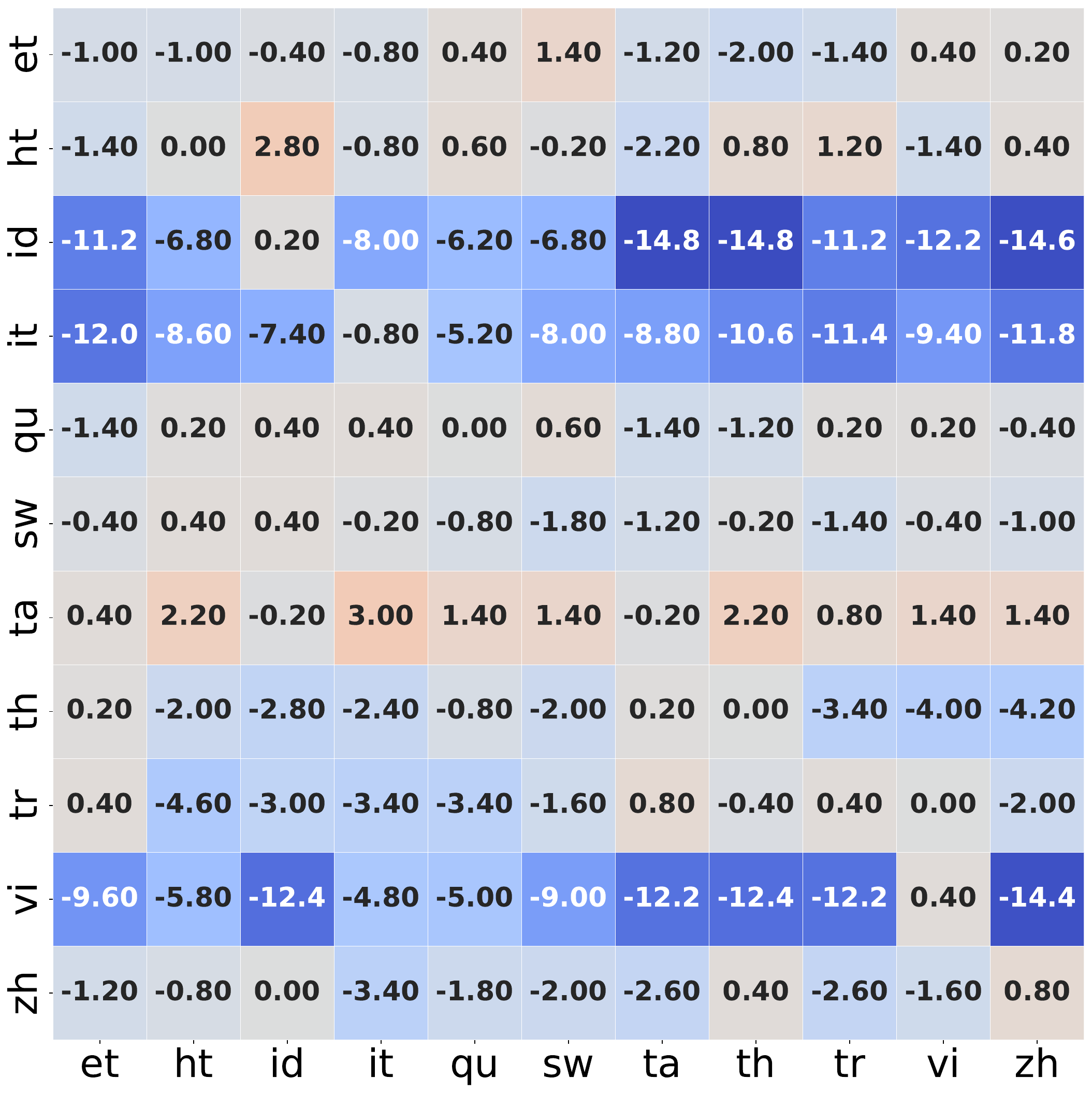}
    \caption{Delta XCOPA accuracy after steering LAPE neurons with \steer{pmax} for SeaLLM3 1.5B.}
    \label{fig:lape-xcopa-seam}
\end{figure}

\begin{figure}
    \centering
    \includegraphics[width=\linewidth]{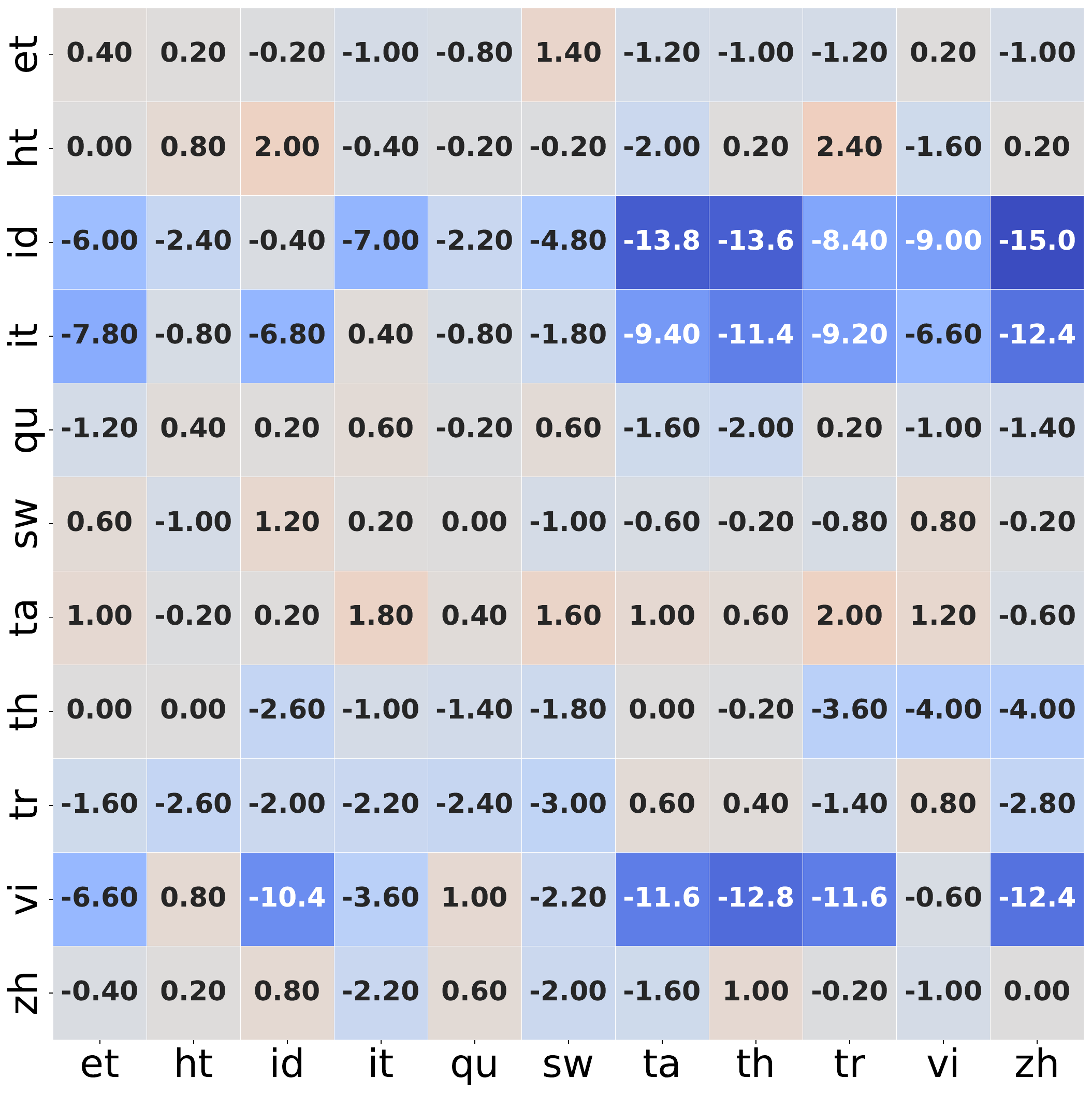}
    \caption{Delta XCOPA accuracy after steering LAPE neurons with \steer{pmedian} for SeaLLMs 1.5B. Row \textit{i} is the initial language, column \textit{j} is the language of intervention.}
    \label{fig:lape-xcopa-med-seam}
\end{figure}

\begin{figure}
    \centering
    \includegraphics[width=\linewidth]{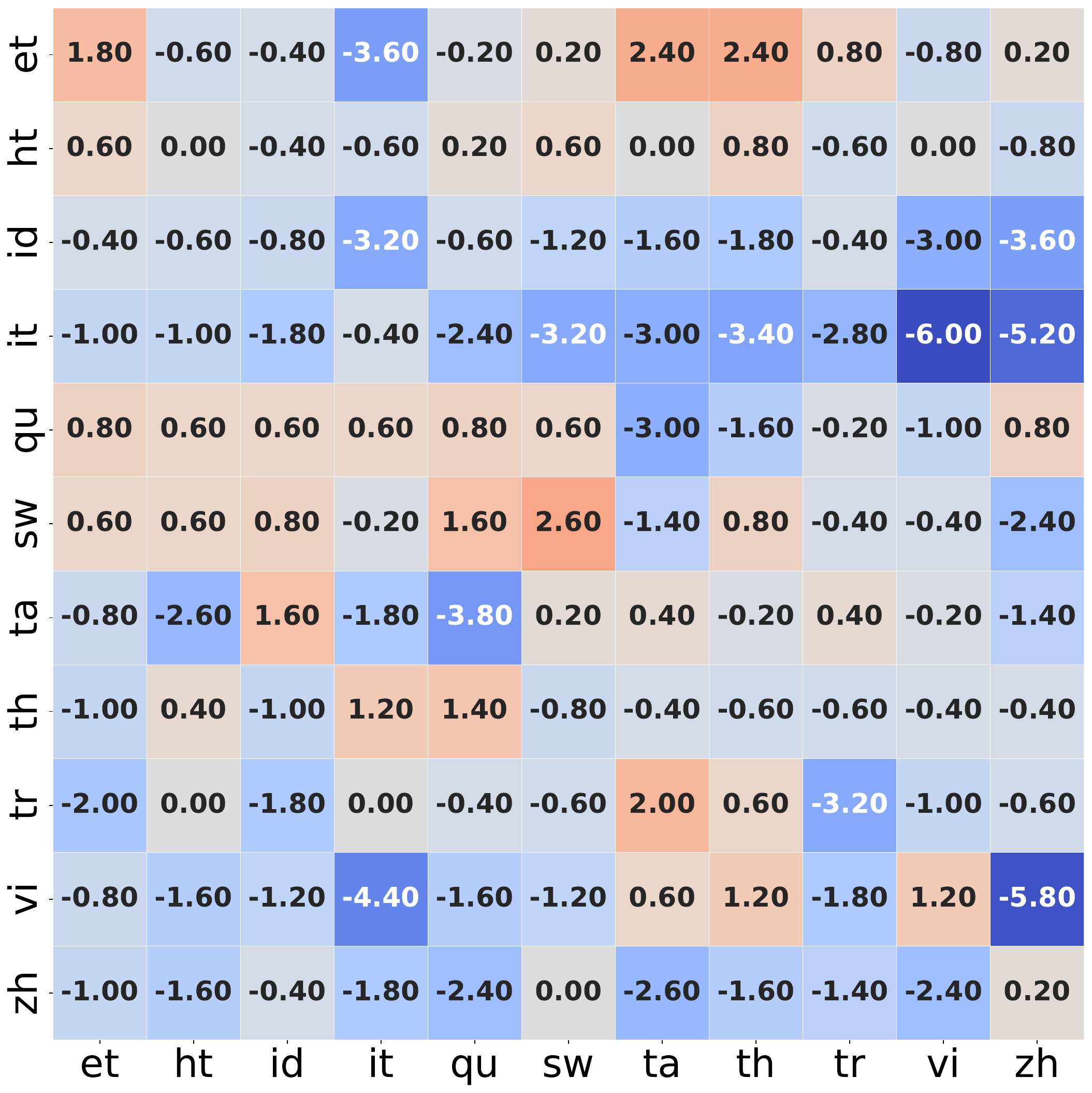}
    \caption{Delta XCOPA accuracy after steering LAPE neurons with \steer{pmedian} for Gemma2 2B. Row \textit{i} is the initial language, column \textit{j} is the language of intervention.}
    \label{fig:lape-xcopa-med-gemmam}
\end{figure}

\begin{figure}
    \centering
    \includegraphics[width=\linewidth]{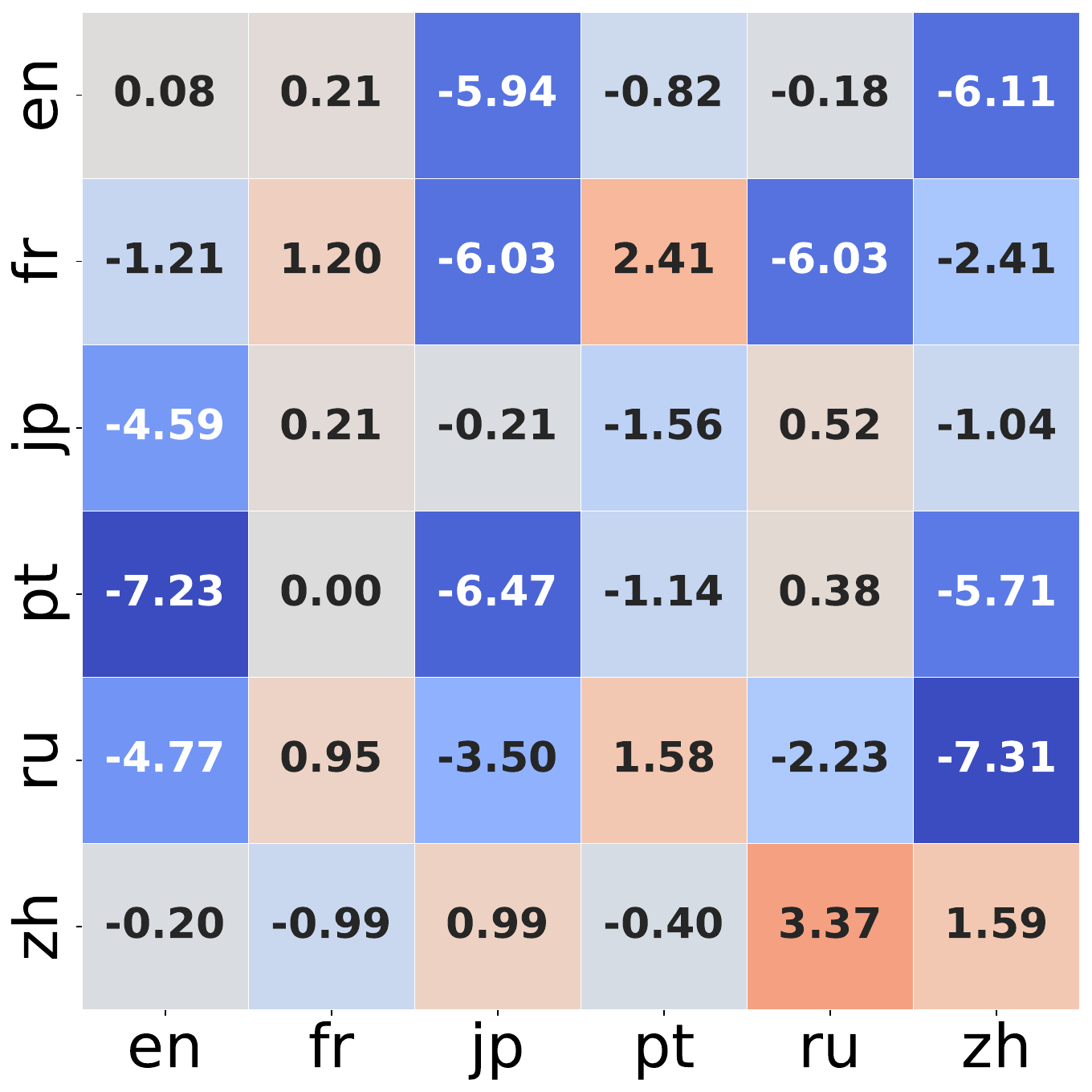}
    \caption{Delta XCOPA accuracy after steering LAPE neurons with \steer{pmedian} for Qwen2.5 0.5B. Row \textit{i} is the initial language, column \textit{j} is the language of intervention.}
    \label{fig:lape-xcopa-med-qwenm}
\end{figure}

\begin{figure}[t]

\centering
    \includegraphics[width=\linewidth]{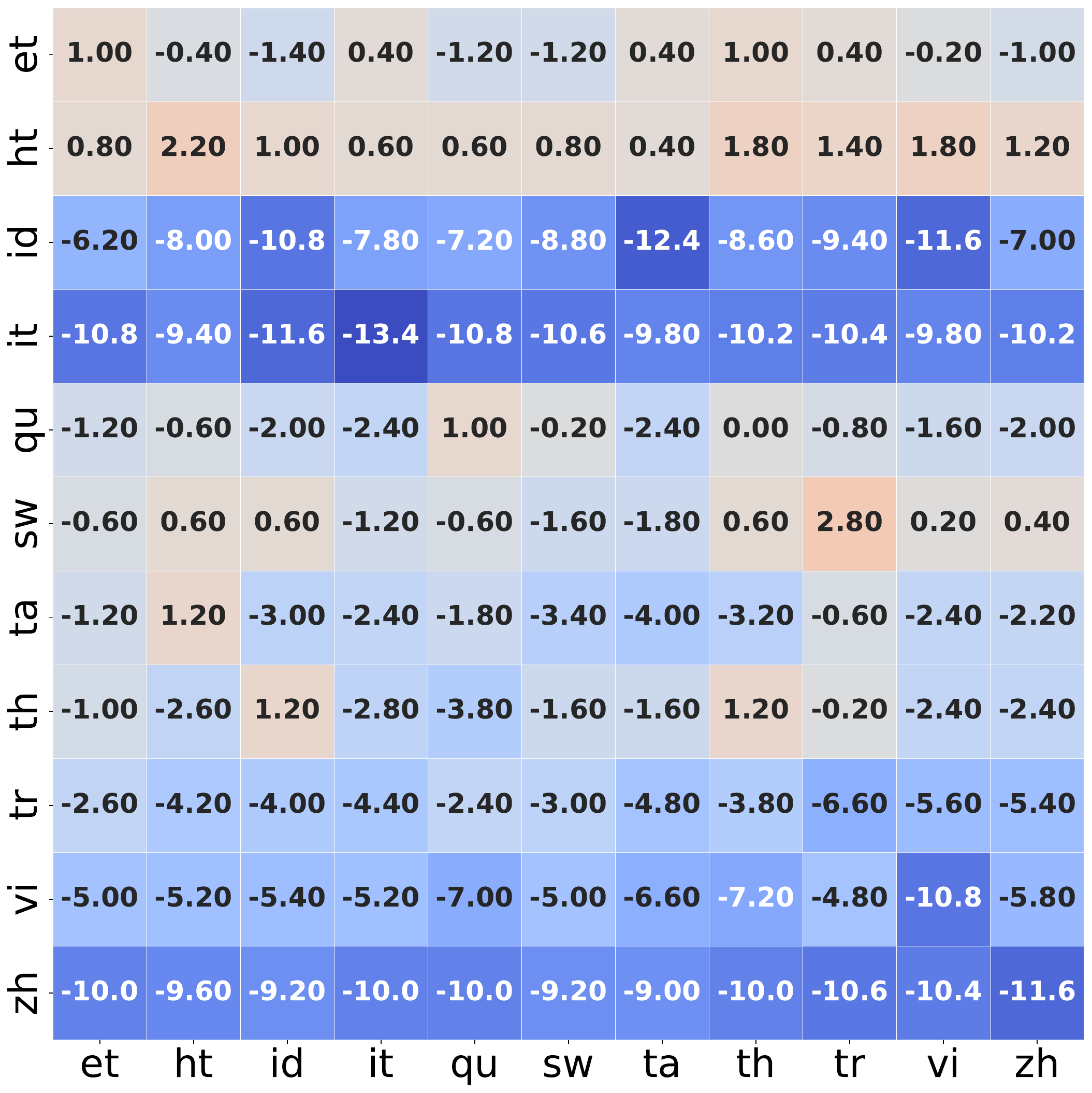}
    \caption{Delta XCOPA accuracy after steering baseline neurons with \steer{pmax} for Gemma2 2B.}
    \label{fig:raw-xcopa-gemmam}
\end{figure}

\begin{figure}[t]
\centering
    \includegraphics[width=\linewidth]{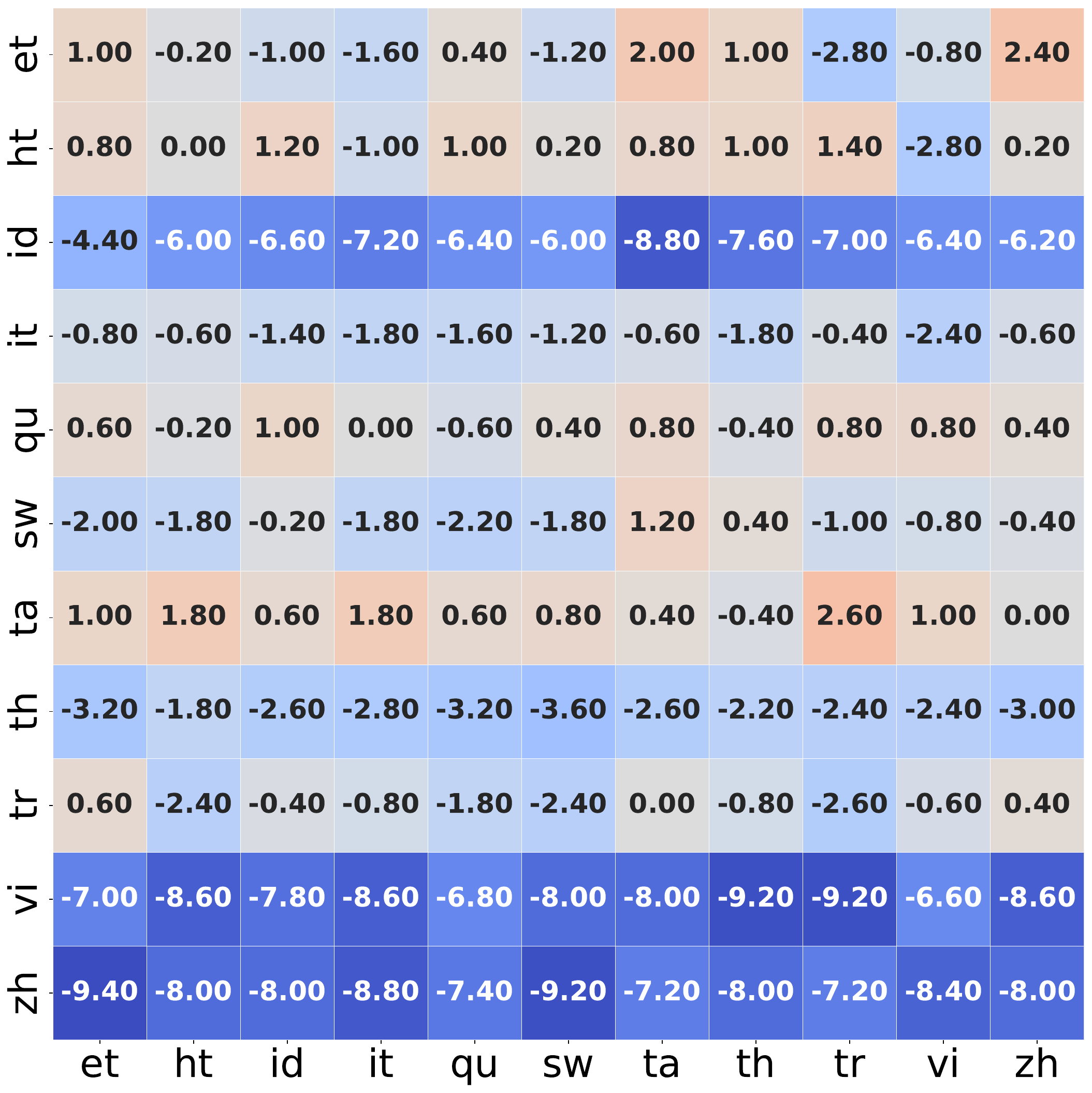}
    \caption{Delta XCOPA accuracy after steering baseline neurons with \steer{pmax} for Qwen2.5 0.5B.}
    \label{fig:raw-xcopa-qwenm}
\end{figure}

\begin{figure}[t]
\centering
    \includegraphics[width=\linewidth]{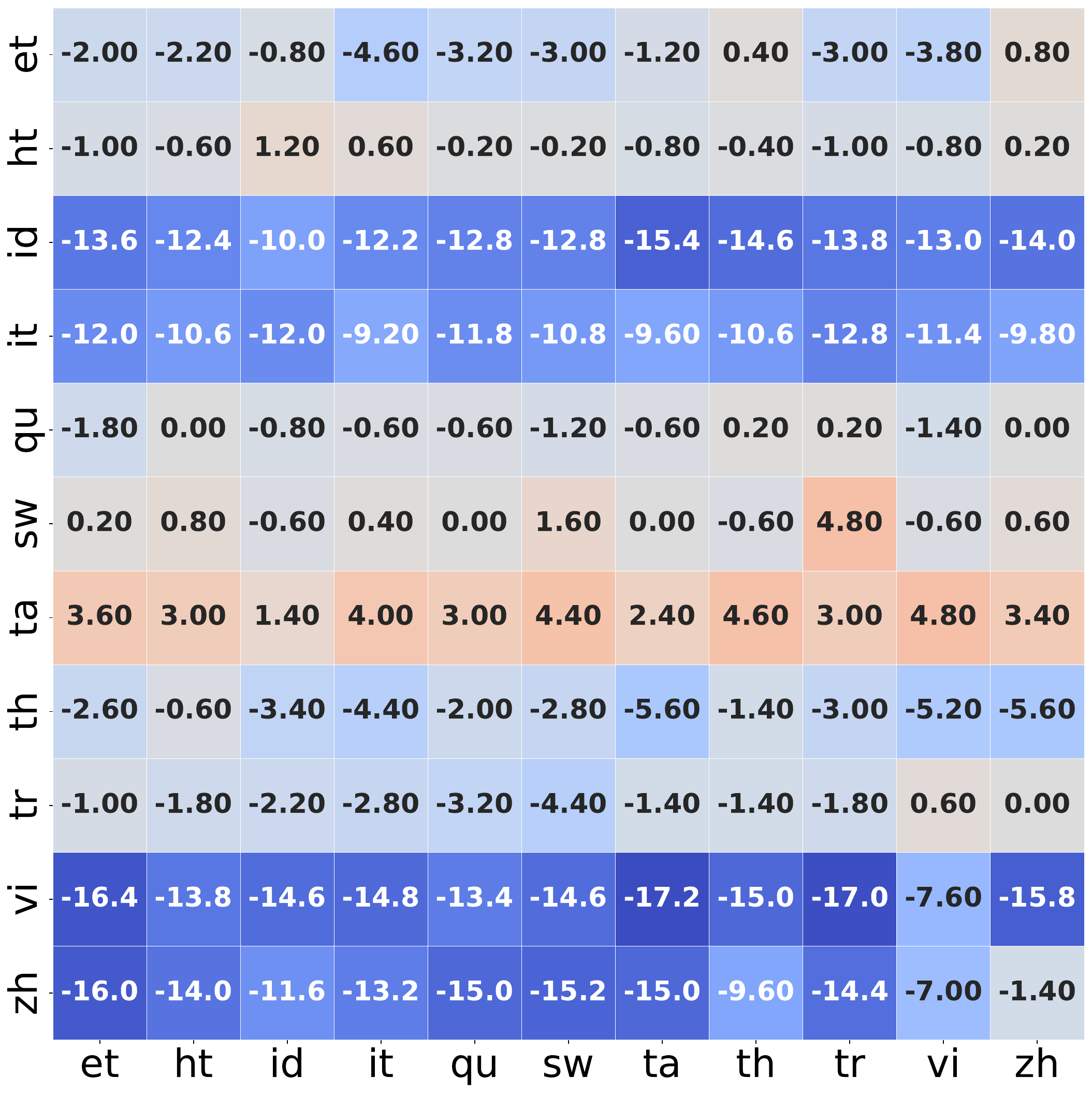}
    \caption{Delta XCOPA accuracy after steering baseline neurons with \steer{pmax} for SeaLLM3 1.5B.}
    \label{fig:raw-xcopa-seam}
\end{figure}

Table \ref{tab:xcopa-baseline} reveals the baseline (no intervention) XCOPA accuracy. Delta accuracies of XCOPA dataset under LAPE neurons when steered with \steer{pmax} are shown in Figure \ref{fig:lape-xcopa-gemma}, \ref{fig:lape-xcopa-gemmam}, \ref{fig:lape-xcopa-qwenm}, and \ref{fig:lape-xcopa-seam}; intervention results for \steer{pmedian} are shown in Figure \ref{fig:lape-xcopa-med-gemmam}, \ref{fig:lape-xcopa-med-seam}, and \ref{fig:lape-xcopa-med-qwenm}. Delta XCOPA accuracies of \steer{pmax} intervention on Baseline neurons are provided in Figure \ref{fig:raw-xcopa-gemmam}, \ref{fig:raw-xcopa-qwenm}, and \ref{fig:raw-xcopa-seam}.

\begin{figure}[t]
\centering
    \includegraphics[width=\linewidth]{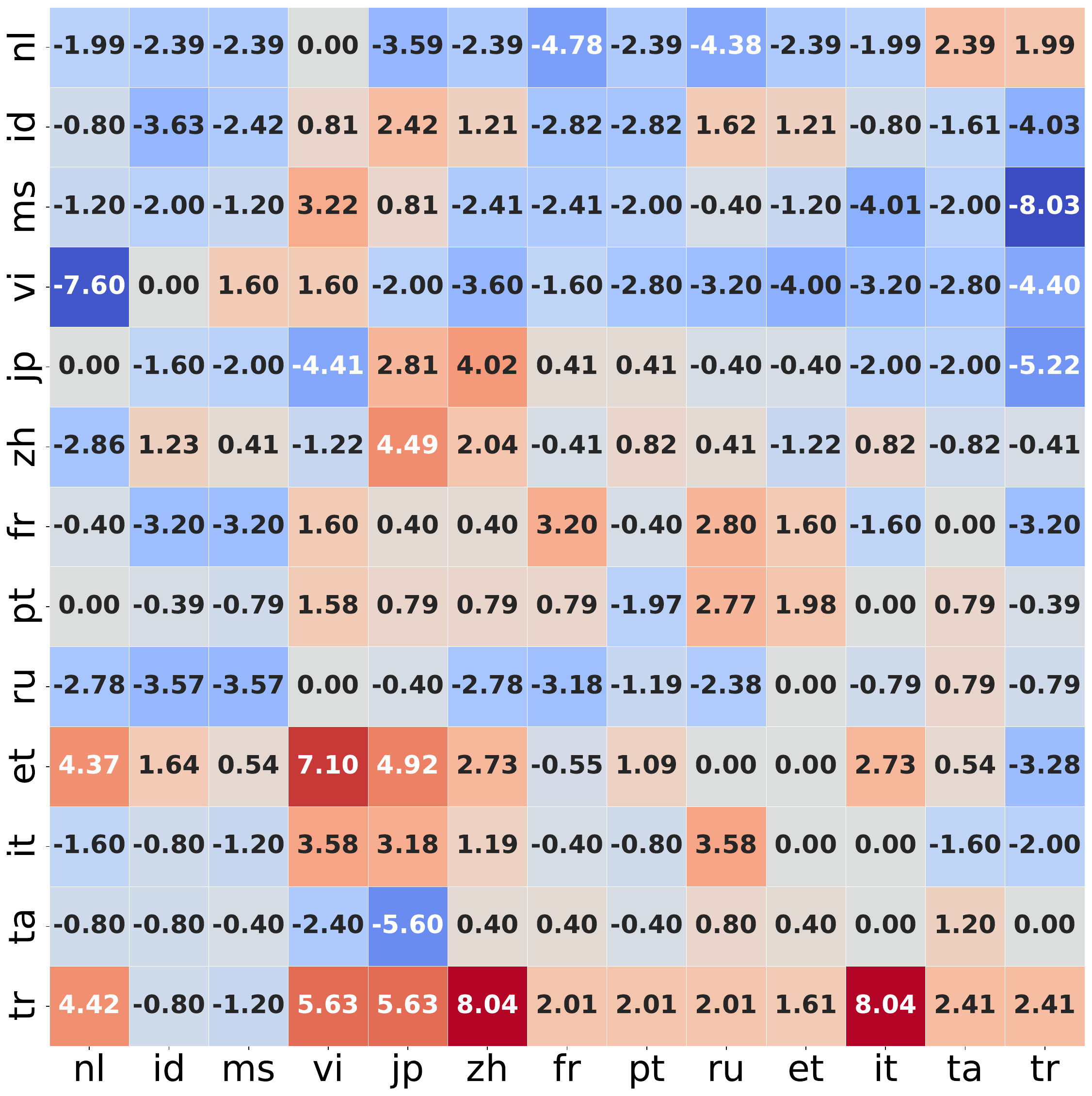}
    \caption{Delta Include-lite accuracy after steering LAPE neurons with \steer{pmax} for Qwen2.5 0.5B.}
    \label{fig:lape-include-qwenm}
\end{figure}

\begin{figure}[t]
\centering
    \includegraphics[width=\linewidth]{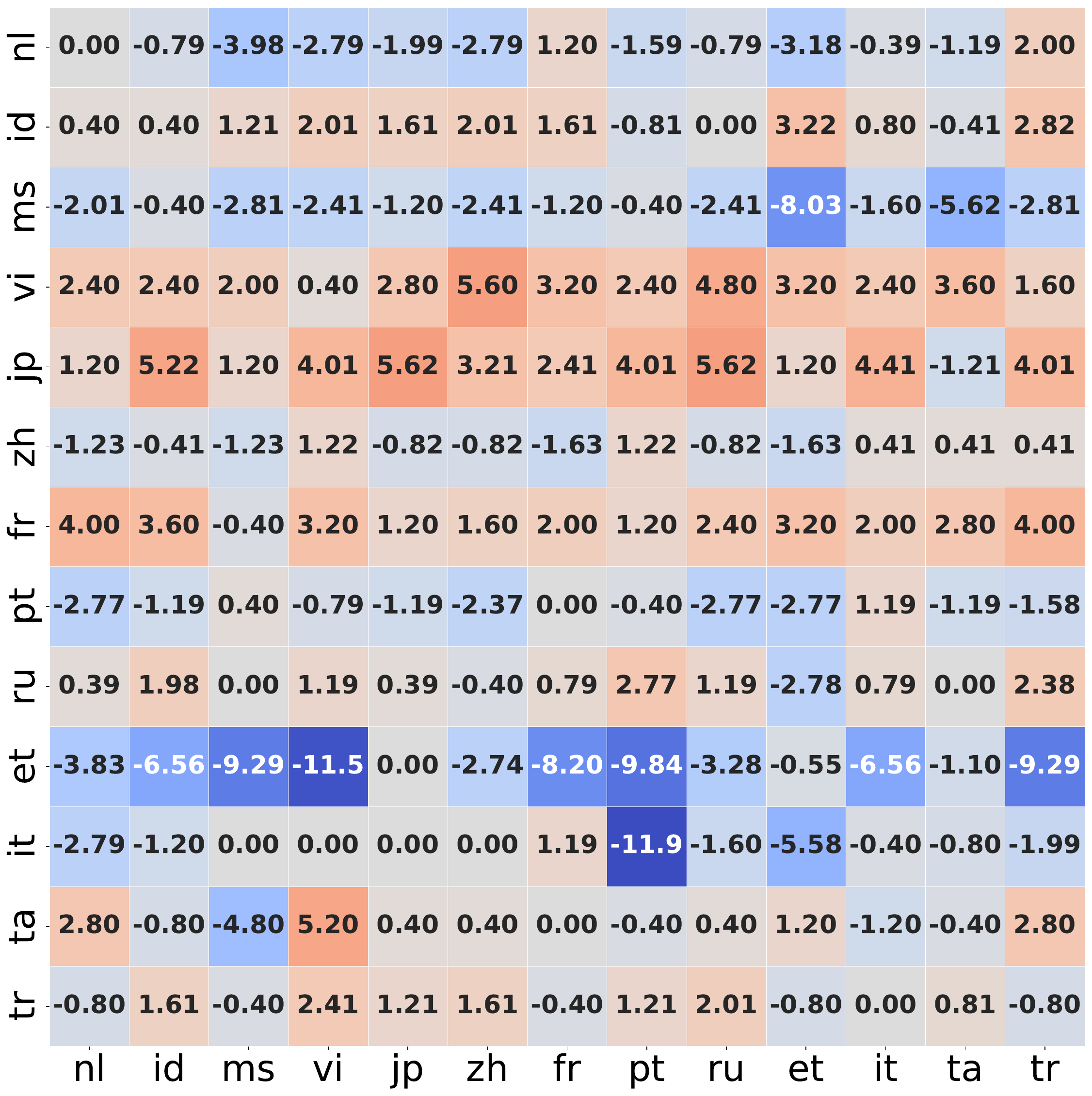}
    \caption{Delta Include-lite accuracy after steering LAPE neurons with \steer{pmax} for Qwen2.5 7B.}
    \label{fig:lape-include-qwen}
\end{figure}

\begin{figure}[t]
\centering
    \includegraphics[width=\linewidth]{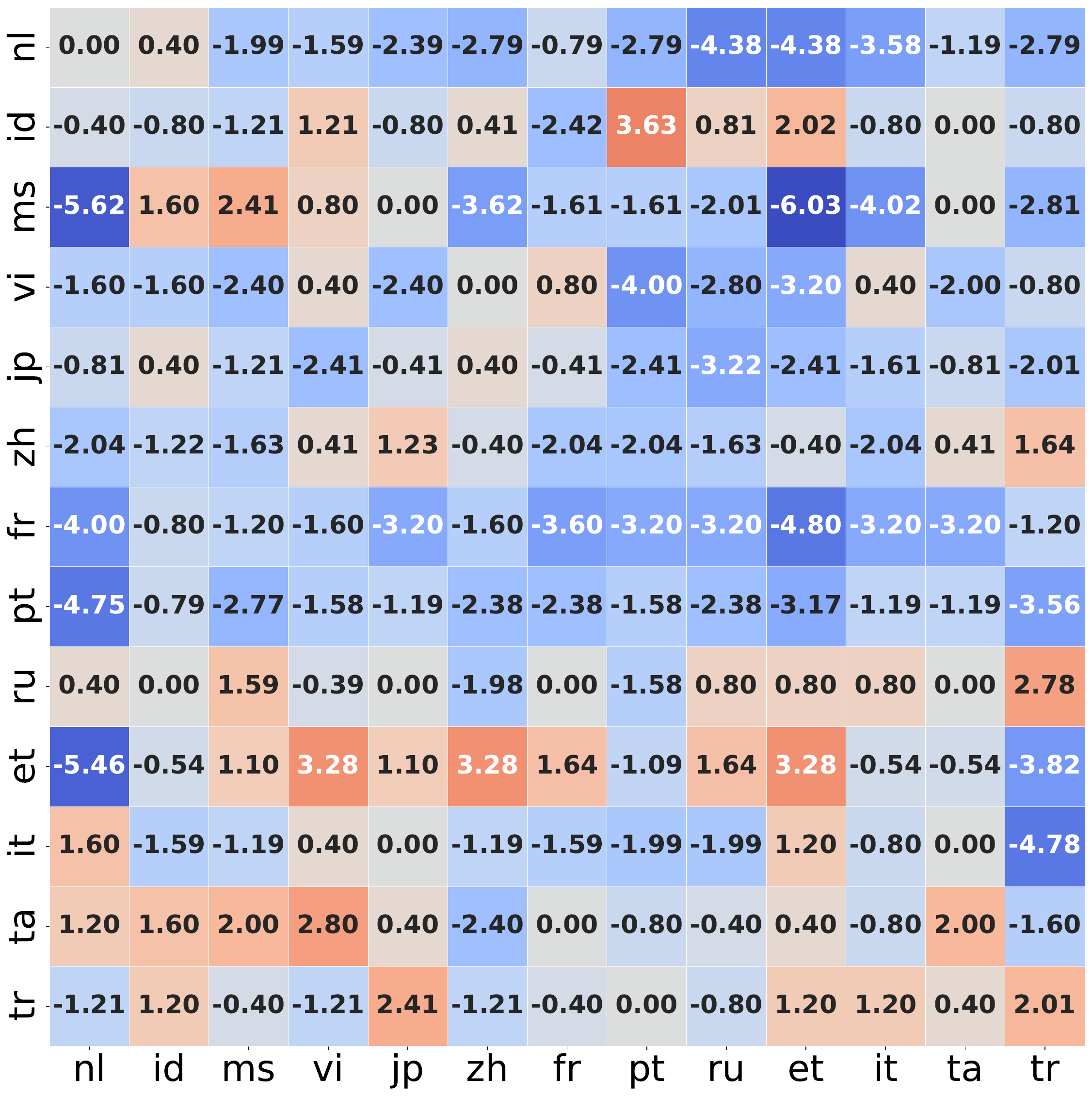}
    \caption{Delta Include-lite accuracy after steering LAPE neurons with \steer{pmax} for Gemma2 2B.}
    \label{fig:lape-include-gemmam}
\end{figure}

\begin{figure}[t]
\centering
    \includegraphics[width=\linewidth]{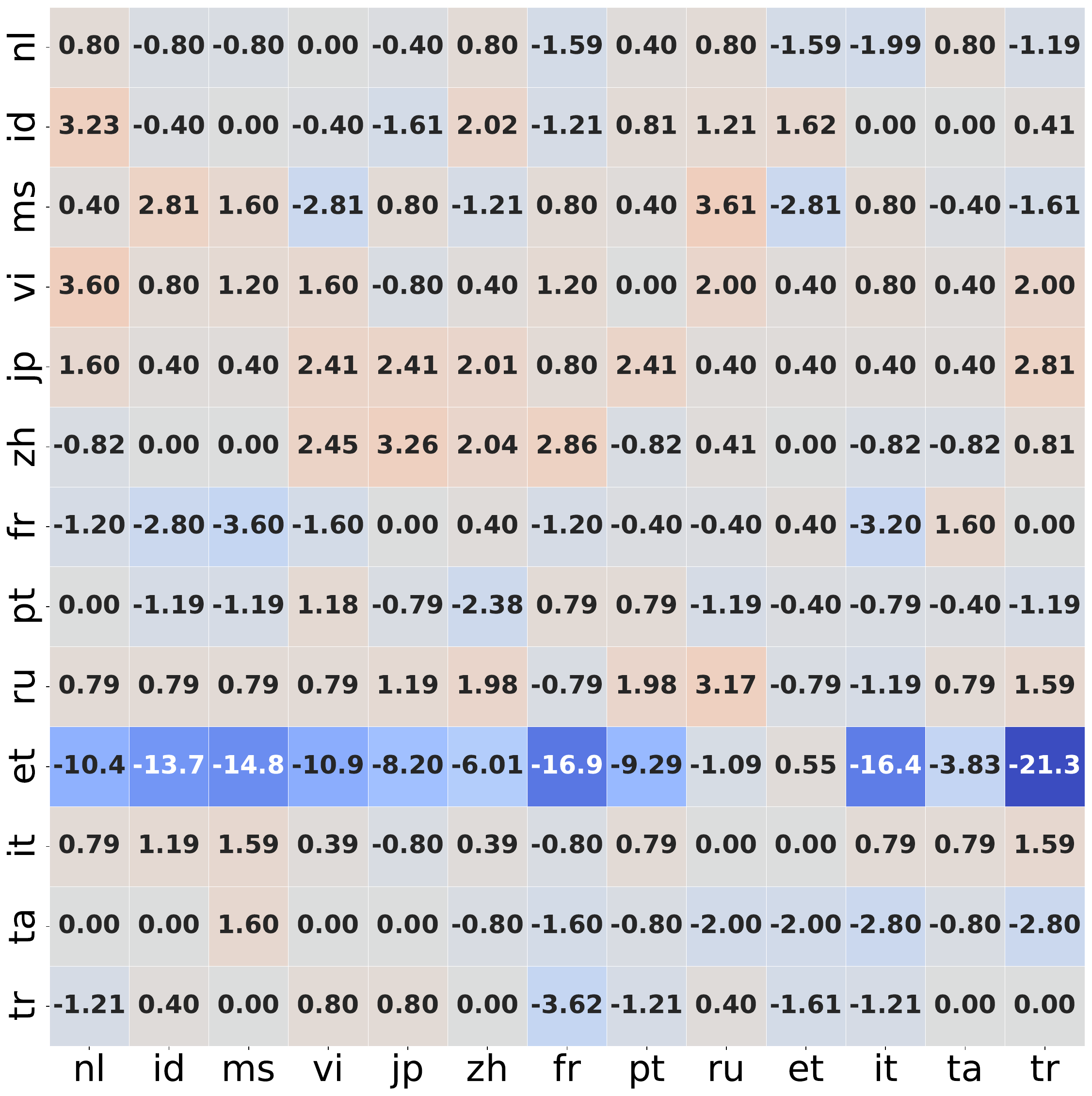}
    \caption{Delta Include-lite accuracy after steering LAPE neurons with \steer{pmax} for Gemma2 9B.}
    \label{fig:lape-include-gemma}
\end{figure}

\begin{figure}[t]
\centering
    \includegraphics[width=\linewidth]{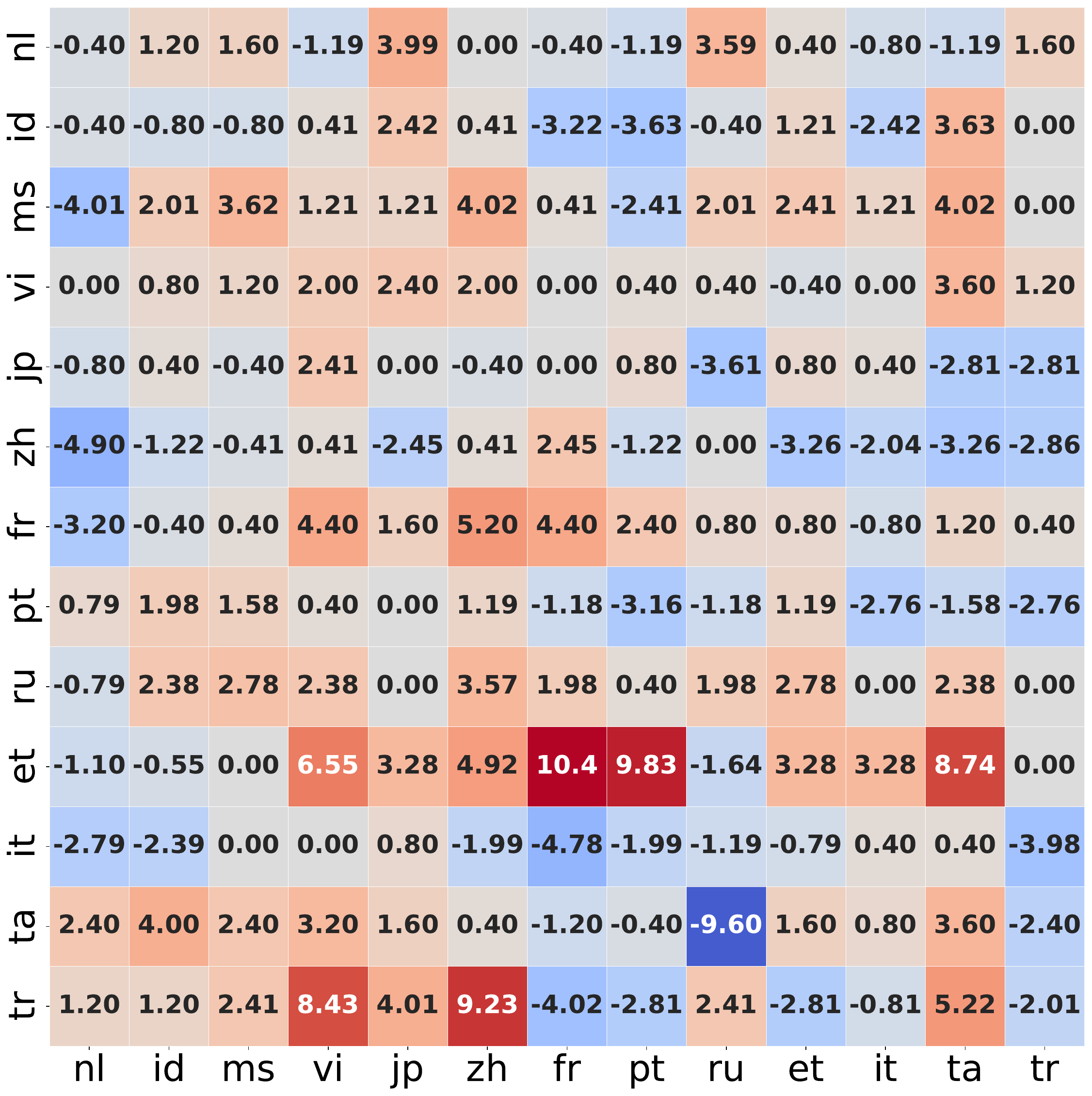}
    \caption{Delta Include-lite accuracy after steering LAPE neurons with \steer{pmax} for SeaLLMv3 1.5B.}
    \label{fig:lape-include-seam}
\end{figure}

\begin{figure}[t]
\centering
    \includegraphics[width=\linewidth]{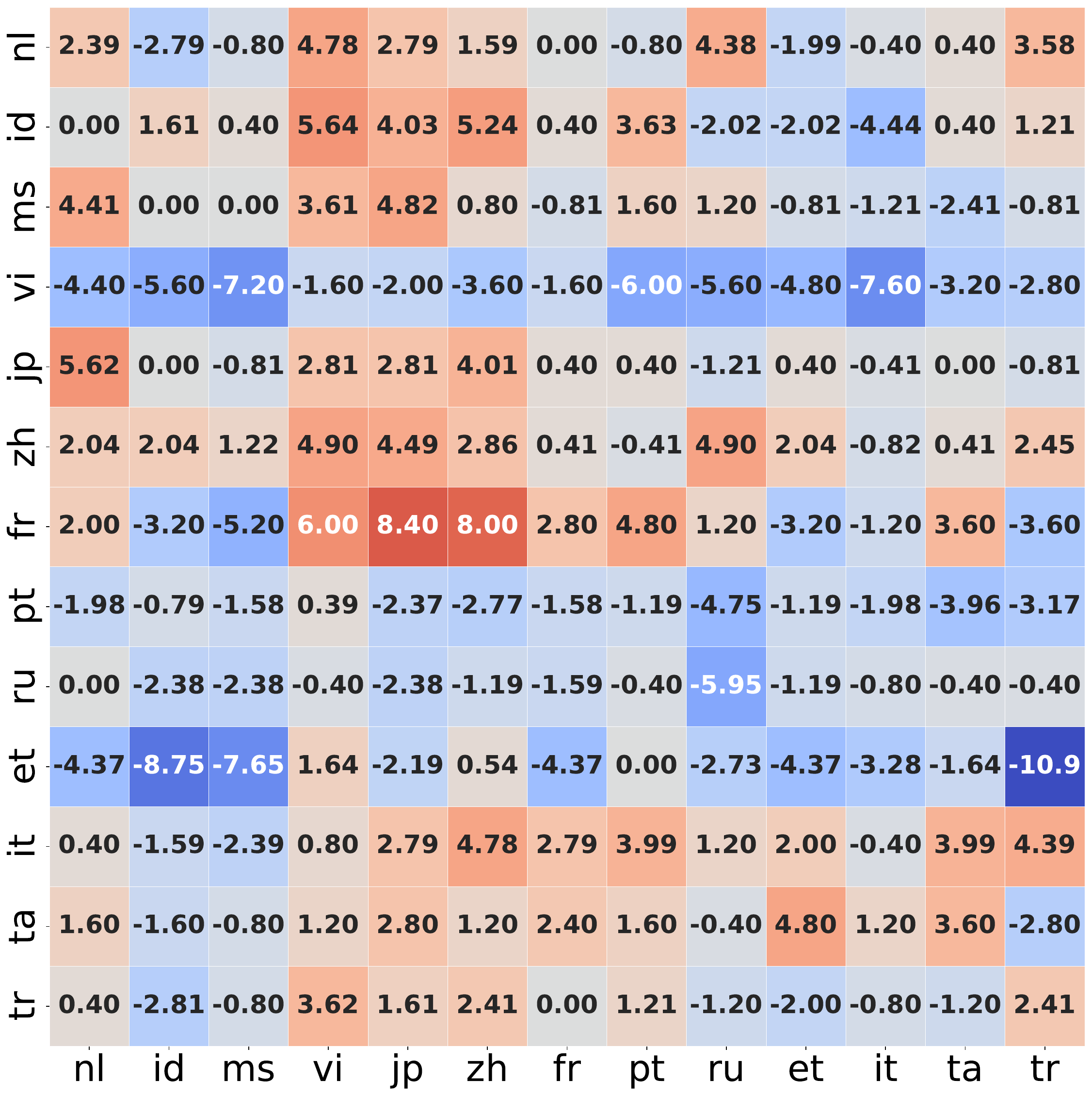}
    \caption{Delta Include-lite accuracy after steering LAPE neurons with \steer{pmax} for SeaLLMv3 7B.}
    \label{fig:lape-include-sea}
\end{figure}

\begin{figure}[t]
\centering
    \includegraphics[width=\linewidth]{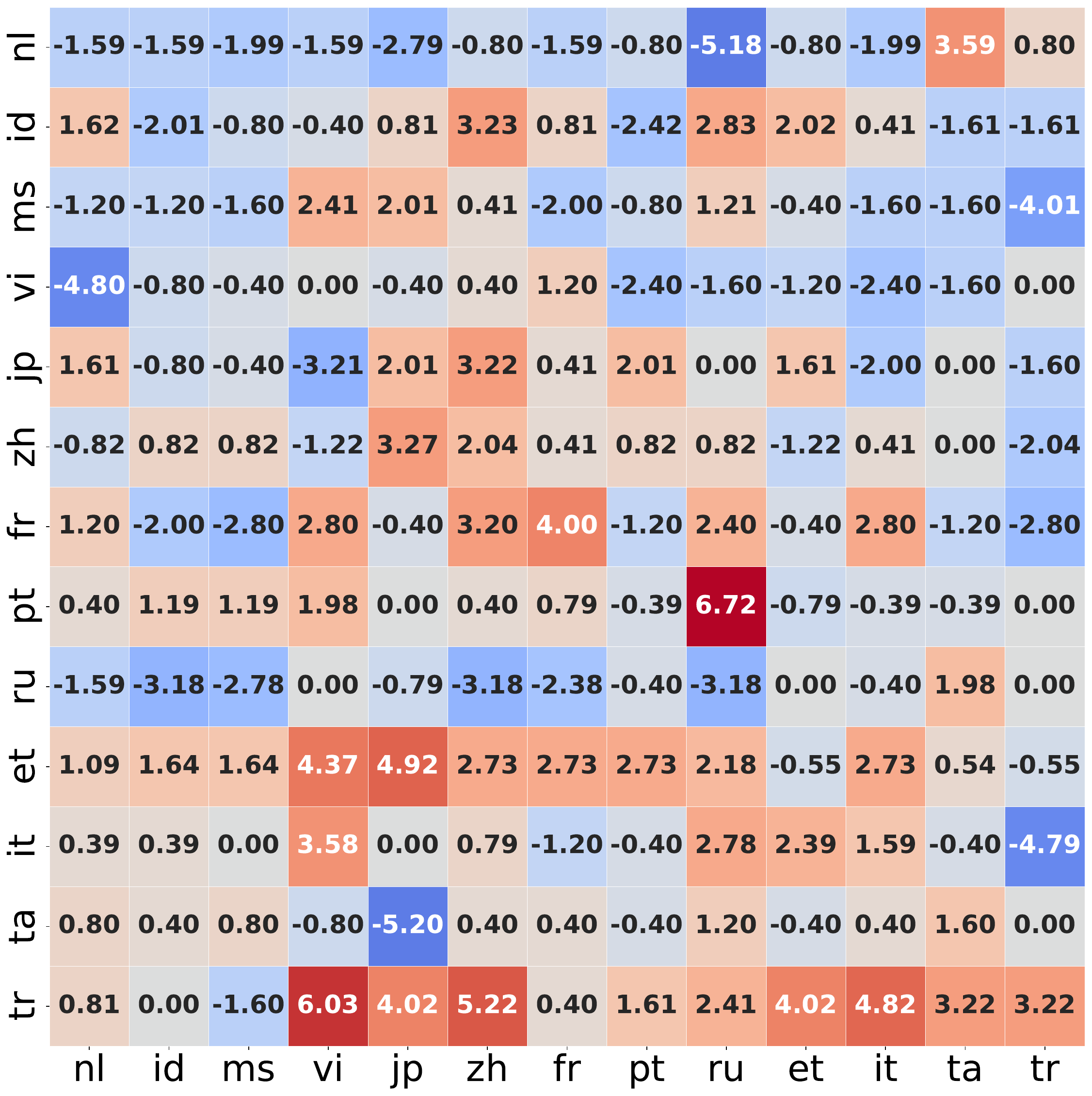}
    \caption{Delta Include-lite accuracy after steering LAPE neurons with \steer{pmedian} for Qwen2.5 0.5B.}
    \label{fig:lape-include-med-qwenm}
\end{figure}

\begin{figure}[t]
\centering
    \includegraphics[width=\linewidth]{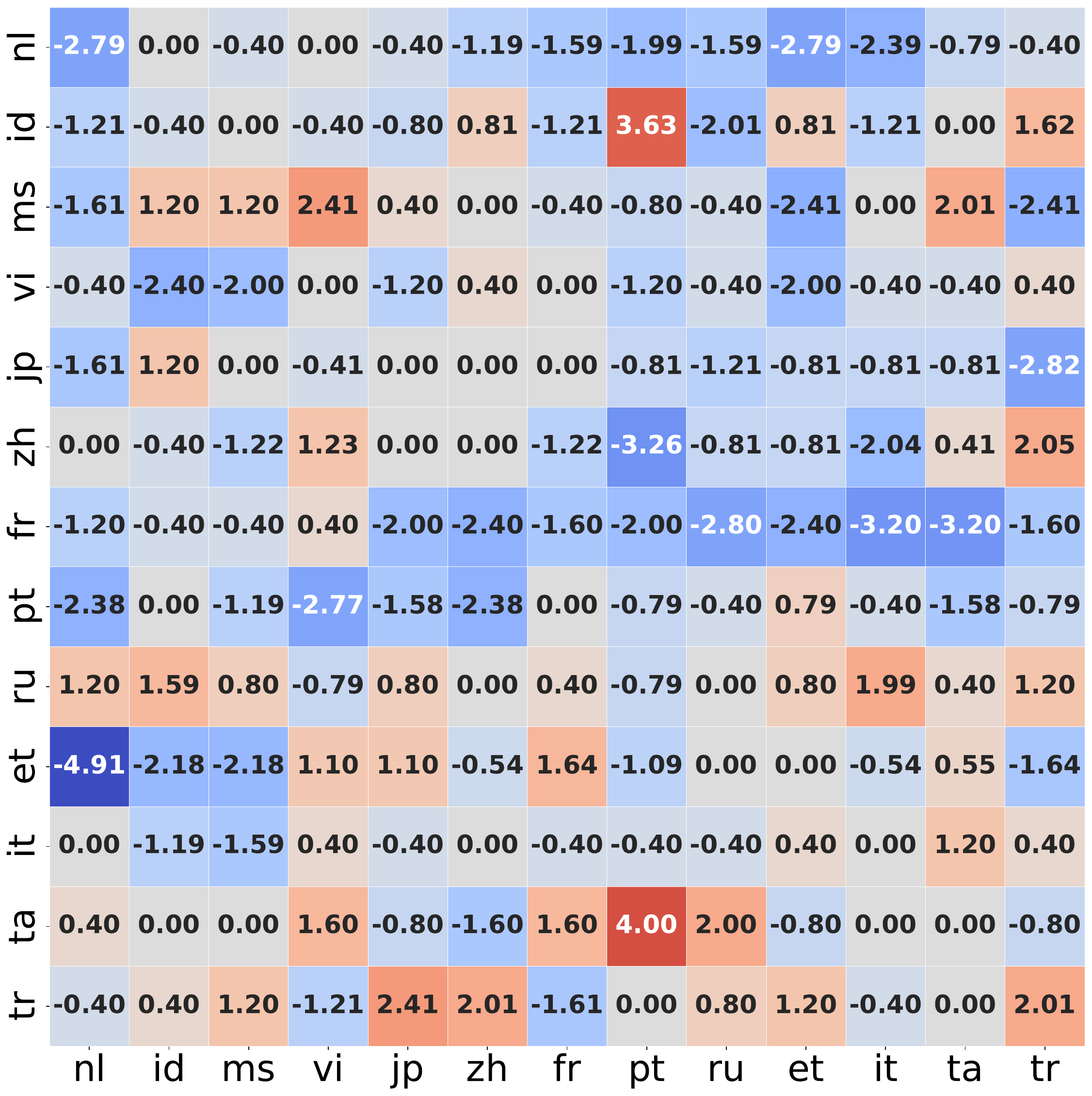}
    \caption{Delta Include-lite accuracy after steering LAPE neurons with \steer{pmedian} for Gemma2 2B.}
    \label{fig:lape-include-med-gemmam}
\end{figure}

\begin{figure}[t]
\centering
    \includegraphics[width=\linewidth]{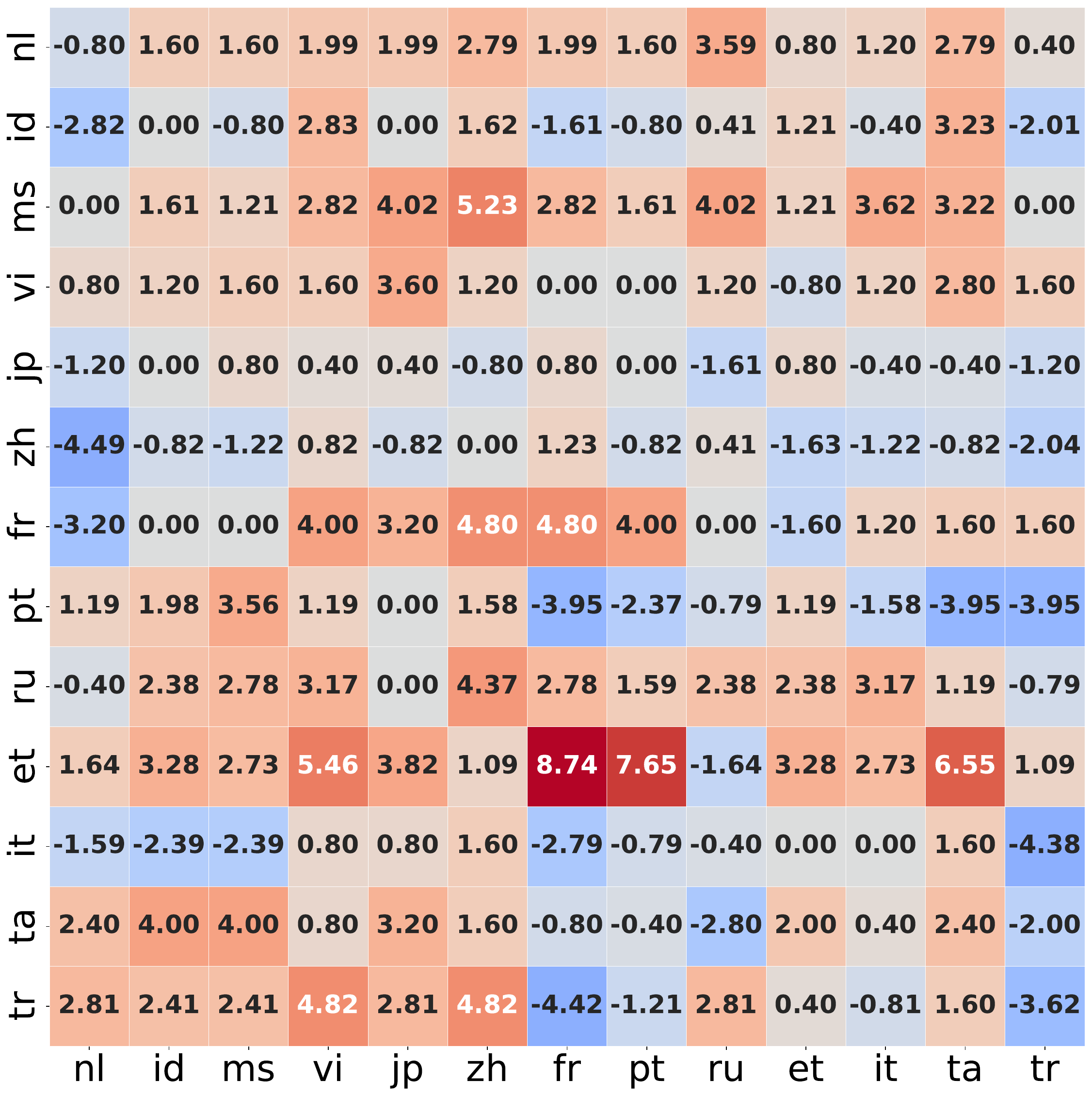}
    \caption{Delta Include-lite accuracy after steering LAPE neurons with \steer{pmedian} for SeaLLMv3 1.5B.}
    \label{fig:lape-include-med-seam}
\end{figure}

\begin{figure}[t]
\centering
    \includegraphics[width=\linewidth]{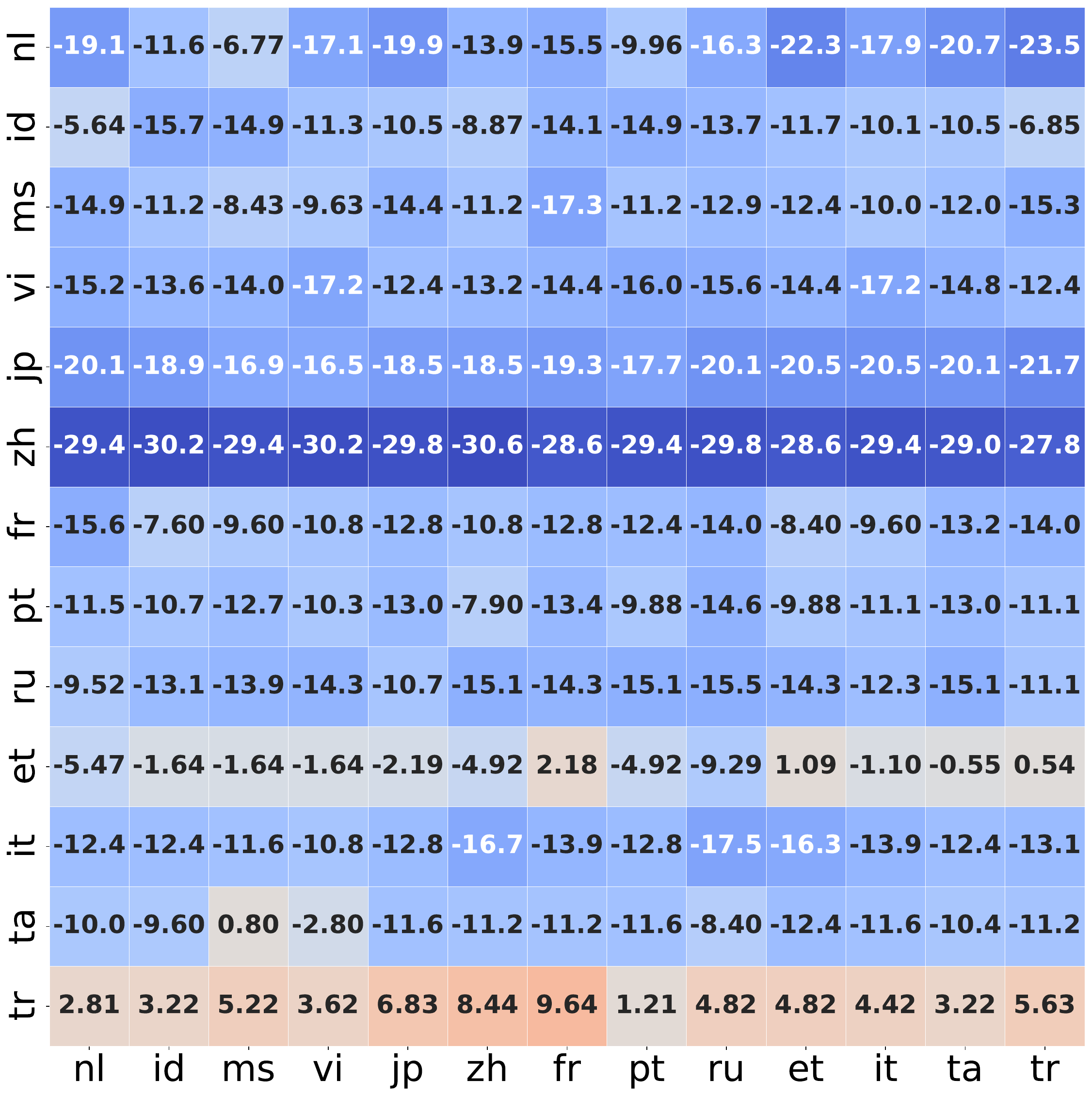}
    \caption{Delta Include-lite accuracy after steering Baseline neurons with \steer{pmax} for Qwen2.5 0.5B.}
    \label{fig:raw-include-qwenm}
\end{figure}

\begin{figure}[t]
\centering
    \includegraphics[width=\linewidth]{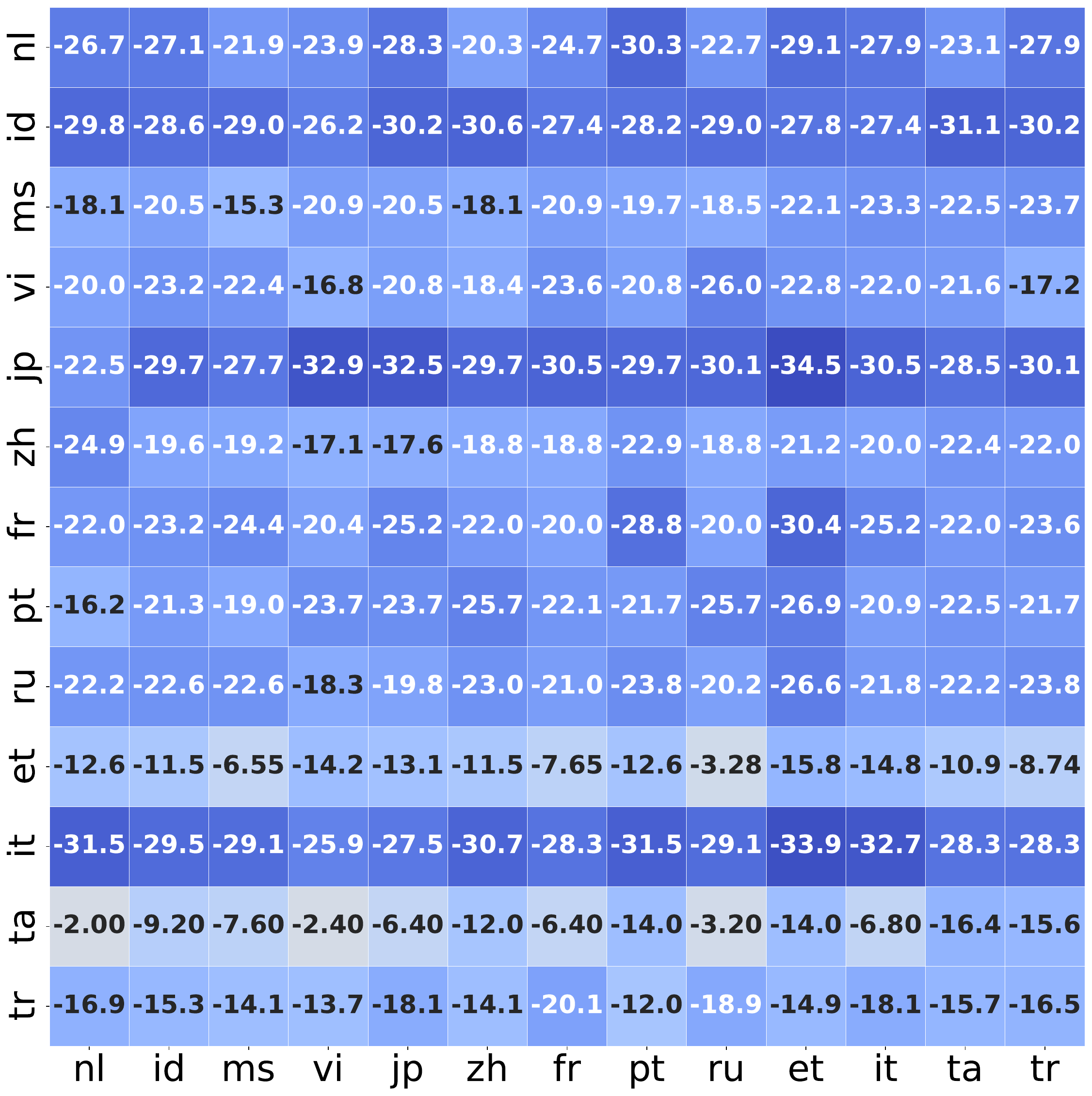}
    \caption{Delta Include-lite accuracy after steering Baseline neurons with \steer{pmax} for Gemma2 2B.}
    \label{fig:raw-include-gemmam}
\end{figure}

\begin{figure}[t]
\centering
    \includegraphics[width=\linewidth]{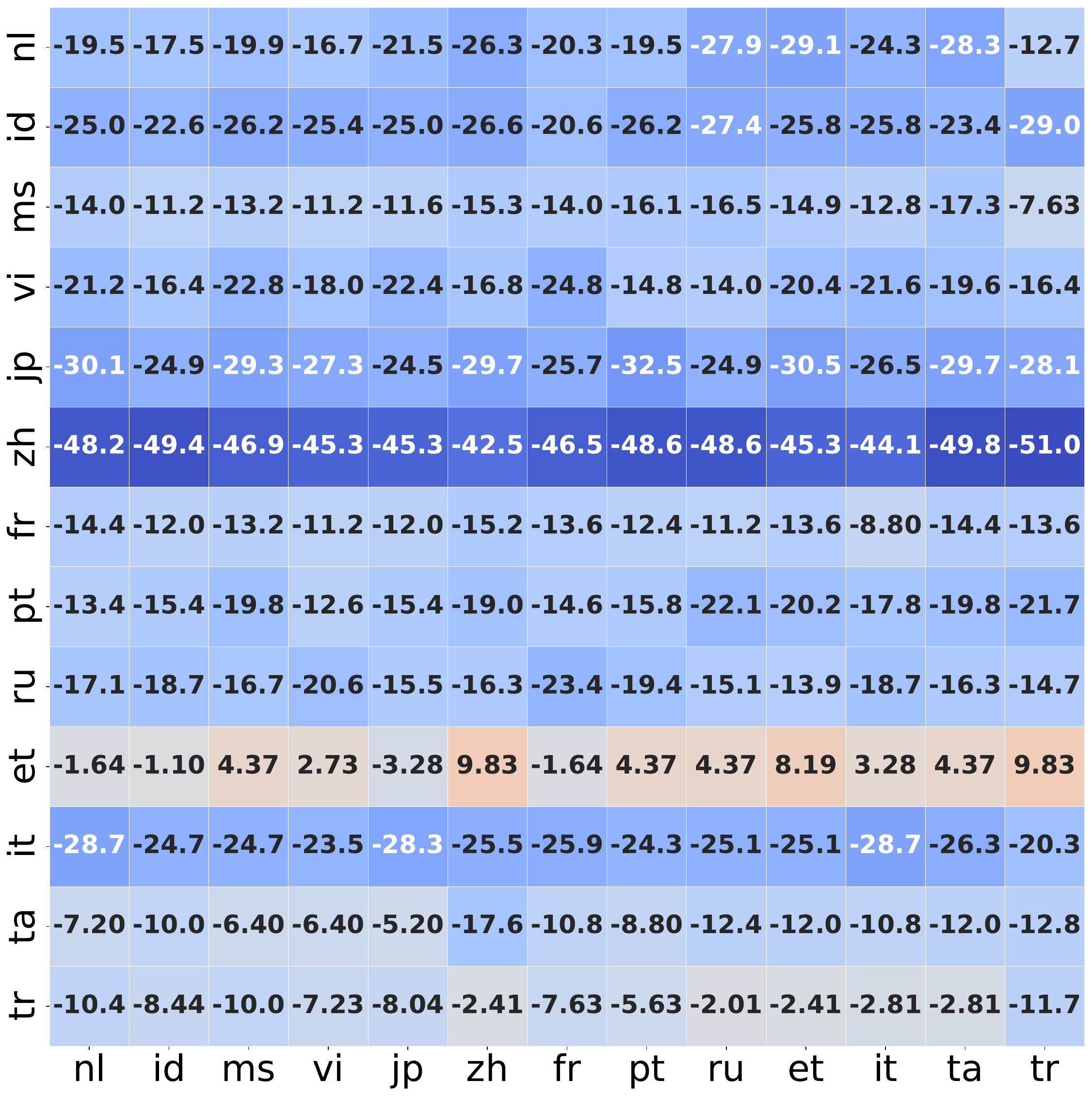}
    \caption{Delta Include-lite accuracy after steering Baseline neurons with \steer{pmax} for SeaLLmv3 1.5B.}
    \label{fig:raw-include-seam}
\end{figure}

Table \ref{tab:include-baseline} reveals the baseline (no intervention) Include-lite accuracy. Delta accuracies of Include-lite dataset under LAPE neurons when steered with \steer{pmax} are shown in Figure \ref{fig:lape-include-gemma}, \ref{fig:lape-include-gemmam}, \ref{fig:lape-include-qwen}, \ref{fig:lape-include-qwenm}, \ref{fig:lape-include-sea}, and \ref{fig:lape-include-seam}, intervention results for \steer{pmedian} are provided in Figure \ref{fig:lape-include-med-gemmam}, \ref{fig:lape-include-med-qwenm}, and \ref{fig:lape-include-med-seam}. Delta Include-lite accuracies of \steer{pmax} intervention on Baseline neurons are provided in Figure \ref{fig:raw-include-gemmam}, \ref{fig:raw-include-qwenm}, and \ref{fig:raw-include-seam}.

\begin{table*}[ht]
\centering
\small
\renewcommand{\arraystretch}{1.2}
\setlength{\tabcolsep}{2pt}

\resizebox{\textwidth}{!}{
\begin{tabular}{
|l
|*{6}{c}|
*{6}{c}|
*{6}{c}|
}
\hline
\multirow{2}{*}{} & 
\multicolumn{6}{c|}{\textbf{Qwen}} & 
\multicolumn{6}{c|}{\textbf{Gemma}} & 
\multicolumn{6}{c|}{\textbf{SeaLLM}} \\
\cline{2-19}
& \textbf{\scriptsize pmax\,$\uparrow$} & \textbf{\scriptsize pmed\,$\uparrow$} & \textbf{\scriptsize =max\,$\uparrow$}  & \textbf{\scriptsize +max\,$\uparrow$}  & \textbf{\scriptsize =0\,$\downarrow$}  & \textbf{\scriptsize =10p\,$\downarrow$}
& \textbf{\scriptsize pmax\,$\uparrow$} & \textbf{\scriptsize pmed\,$\uparrow$} & \textbf{\scriptsize =max\,$\uparrow$}  & \textbf{\scriptsize +max\,$\uparrow$}  & \textbf{\scriptsize =0\,$\downarrow$}  & \textbf{\scriptsize =10p\,$\downarrow$}
& \textbf{\scriptsize pmax\,$\uparrow$} & \textbf{\scriptsize pmed\,$\uparrow$} & \textbf{\scriptsize =max\,$\uparrow$}  & \textbf{\scriptsize +max\,$\uparrow$}  & \textbf{\scriptsize =0\,$\downarrow$}  & \textbf{\scriptsize =10p\,$\downarrow$} \\
\hline
& \multicolumn{18}{c|}{\textbf{Baseline}} \\
\hline
en & 96.22 & \textbf{97.84} & 95.14 & 94.05 & 92.43 & 92.97 & \textbf{99.46} & \textbf{99.46} & 79.46 & 78.38 & 31.35 & 81.62 & 99.46 & \textbf{100.0} & 95.14 & 94.05 & 51.89 & 60.54 \\ nl & \textbf{90.81} & 88.11 & 80.54 & 81.08 & 71.89 & 78.92 & \textbf{71.89} & \DD62.7 & 42.16 & 44.32 & 31.35 & 58.92 & 94.05 & \textbf{97.3} & 76.22 & 77.84 & \DD42.7 & 43.78 \\ ms & 40.54 & \textbf{46.49} & 31.89 & 30.81 & 29.73 & 34.05 & 29.73 & 25.95 & 31.35 & 31.35 & \textbf{41.08} & \DD40.0 & \DD40.0 & \textbf{42.16} & 31.89 & 32.43 & 36.22 & 23.78 \\ vi & 89.73 & \textbf{90.81} & 76.22 & 76.22 & 54.59 & 59.46 & \textbf{81.08} & 72.97 & 31.35 & 29.73 & \DD9.19 & 14.59 & 98.38 & \textbf{100.0} & 72.43 & 70.81 & 12.97 & 16.22 \\ jp & 87.57 & \textbf{88.11} & 63.78 & 68.11 & 29.73 & 28.11 & 79.46 & \textbf{91.89} & \DD42.7 & 41.62 & 21.62 & 38.38 & \textbf{97.3} & \textbf{97.3} & \DD42.7 & 36.76 & 10.81 & \DD5.41 \\
\hline

\end{tabular}
}
\caption{Language Steering Shift (LSS) scores from \texttt{id} to languages listed for amplifying (\steer{pmax}, \steer{pmedian}, \steer{=max}, \steer{+max}) and deactivating factors (\steer{=0}, \steer{=10p}) across models. Bolded texts denote the highest values within a model.}
\label{tab:id}
\vspace{-10pt}
\end{table*}

\begin{table*}[htbp]
\centering
\small
\renewcommand{\arraystretch}{1.2}
\setlength{\tabcolsep}{2pt}

\resizebox{\textwidth}{!}{
\begin{tabular}{
|l
|*{6}{c}|
*{6}{c}|
*{6}{c}|
}
\hline
\multirow{2}{*}{} & 
\multicolumn{6}{c|}{\textbf{Qwen}} & 
\multicolumn{6}{c|}{\textbf{Gemma}} & 
\multicolumn{6}{c|}{\textbf{SeaLLM}} \\
\cline{2-19}
& \textbf{\scriptsize pmax\,$\uparrow$} & \textbf{\scriptsize pmed\,$\uparrow$} & \textbf{\scriptsize =max\,$\uparrow$}  & \textbf{\scriptsize +max\,$\uparrow$}  & \textbf{\scriptsize =0\,$\downarrow$}  & \textbf{\scriptsize =10p\,$\downarrow$}
& \textbf{\scriptsize pmax\,$\uparrow$} & \textbf{\scriptsize pmed\,$\uparrow$} & \textbf{\scriptsize =max\,$\uparrow$}  & \textbf{\scriptsize +max\,$\uparrow$}  & \textbf{\scriptsize =0\,$\downarrow$}  & \textbf{\scriptsize =10p\,$\downarrow$}
& \textbf{\scriptsize pmax\,$\uparrow$} & \textbf{\scriptsize pmed\,$\uparrow$} & \textbf{\scriptsize =max\,$\uparrow$}  & \textbf{\scriptsize +max\,$\uparrow$}  & \textbf{\scriptsize =0\,$\downarrow$}  & \textbf{\scriptsize =10p\,$\downarrow$} \\
\hline
& \multicolumn{18}{c|}{\textbf{Baseline}} \\
\hline 
nl & 63.24 & \textbf{68.65} & 30.81 & 32.97 & \DD5.41 & 10.27 & 63.78 & \textbf{70.81} & 37.84 & 40.54 & 29.19 & 11.89 & 90.27 & \textbf{97.3} & \DD17.3 & 19.46 & 10.27 & \DD8.11 \\ id & 67.03 & \textbf{69.73} & 24.32 & 21.08 & 11.35 & 12.97 & \DD77.3 & \textbf{77.84} & 65.41 & 65.95 & \DD60.0 & 52.43 & 87.03 & \textbf{96.22} & 15.68 & 16.22 & \DD5.95 & \DD4.86 \\ ms & \textbf{47.57} & \textbf{47.57} & 24.86 & 23.78 & \DD8.65 & \DD6.49 & \textbf{54.59} & \DD42.7 & 28.11 & 28.65 & 28.65 & 24.86 & 77.84 & \textbf{87.03} & 12.43 & \DD7.03 & \DD5.41 & \DD3.78 \\ vi & \textbf{74.59} & 67.57 & 28.11 & 25.41 & \DD7.57 & \DD7.57 & \textbf{73.51} & 70.27 & 39.46 & 41.62 & \DD9.19 & \DD5.95 & \textbf{95.14} & 94.05 & 10.81 & 10.27 & \DD4.86 & \DD5.95 \\ jp & 34.59 & \textbf{45.95} & \DD7.03 & \DD6.49 & \DD1.08 & \DD1.08 & 57.84 & \textbf{66.49} & 29.19 & 28.11 & 27.03 & \DD7.03 & 84.86 & \textbf{97.3} & \DD0.54 & \DD0.54 & \DD1.62 & \DD0.54 \\
\hline

\end{tabular}
}

\caption{Language Steering Shift (LSS) scores from \texttt{en} to languages listed for different steering factors across models. Bolded texts denote the highest values within a model.}
\label{tab:en}
\end{table*}

\begin{table*}[htbp]
\centering
\small
\renewcommand{\arraystretch}{1.2}
\setlength{\tabcolsep}{2pt}

\resizebox{\textwidth}{!}{
\begin{tabular}{
|l
|*{6}{c}|
*{6}{c}|
*{6}{c}|
}
\hline
\multirow{2}{*}{} & 
\multicolumn{6}{c|}{\textbf{Qwen}} & 
\multicolumn{6}{c|}{\textbf{Gemma}} & 
\multicolumn{6}{c|}{\textbf{SeaLLM}} \\
\cline{2-19}
& \textbf{\scriptsize pmax\,$\uparrow$} & \textbf{\scriptsize pmed\,$\uparrow$} & \textbf{\scriptsize =max\,$\uparrow$}  & \textbf{\scriptsize +max\,$\uparrow$}  & \textbf{\scriptsize =0\,$\downarrow$}  & \textbf{\scriptsize =10p\,$\downarrow$}
& \textbf{\scriptsize pmax\,$\uparrow$} & \textbf{\scriptsize pmed\,$\uparrow$} & \textbf{\scriptsize =max\,$\uparrow$}  & \textbf{\scriptsize +max\,$\uparrow$}  & \textbf{\scriptsize =0\,$\downarrow$}  & \textbf{\scriptsize =10p\,$\downarrow$}
& \textbf{\scriptsize pmax\,$\uparrow$} & \textbf{\scriptsize pmed\,$\uparrow$} & \textbf{\scriptsize =max\,$\uparrow$}  & \textbf{\scriptsize +max\,$\uparrow$}  & \textbf{\scriptsize =0\,$\downarrow$}  & \textbf{\scriptsize =10p\,$\downarrow$} \\
\hline
& \multicolumn{18}{c|}{\textbf{Baseline}} \\
\hline 
en & \textbf{100.0} & 99.46 & 95.68 & 92.97 & 78.92 & 88.65 & 99.46 & 99.46 & \textbf{100.0} & \textbf{100.0} & 84.86 & 96.76 & \textbf{100.0} & 93.51 & 99.46 & 97.84 & 78.38 & 79.46 \\ nl & 69.19 & 65.95 & 66.49 & 69.19 & 65.41 & \textbf{70.81} & \textbf{91.35} & \textbf{91.35} & 78.92 & 84.32 & 74.59 & 81.62 & \textbf{83.78} & 71.35 & 74.59 & 74.05 & 58.38 & 69.19 \\ id & 63.78 & 64.86 & \textbf{75.68} & 62.16 & \DD60.0 & 67.57 & \textbf{96.76} & 95.68 & 94.05 & 94.59 & 84.32 & 92.97 & 70.81 & 67.57 & 71.89 & \textbf{72.97} & 59.46 & 63.78 \\ ms & \textbf{52.43} & 49.19 & 38.38 & 34.59 & 31.89 & 41.62 & 84.32 & 84.32 & 69.19 & 69.19 & 66.49 & \textbf{84.86} & 45.95 & \textbf{50.27} & 37.84 & 37.84 & 25.95 & 27.03 \\ vi & \textbf{89.73} & \textbf{89.73} & 63.24 & 66.49 & 37.84 & 54.59 & \textbf{98.92} & 98.38 & 86.49 & 86.49 & 30.81 & 71.89 & \textbf{99.46} & 98.92 & 76.76 & \DD80.0 & 27.03 & 29.19 \\
\hline

\end{tabular}
}
\caption{Language Steering Shift (LSS) scores from \texttt{et} to languages listed for different steering factors across models. Bolded texts denote the highest values within a model.}
\label{tab:et}
\end{table*}

\begin{table*}[htbp]
\centering
\small
\renewcommand{\arraystretch}{1.2}
\setlength{\tabcolsep}{2pt}

\resizebox{\textwidth}{!}{
\begin{tabular}{
|l
|*{6}{c}|
*{6}{c}|
*{6}{c}|
}
\hline
\multirow{2}{*}{} & 
\multicolumn{6}{c|}{\textbf{Qwen}} & 
\multicolumn{6}{c|}{\textbf{Gemma}} & 
\multicolumn{6}{c|}{\textbf{SeaLLM}} \\
\cline{2-19}
& \textbf{\scriptsize pmax\,$\uparrow$} & \textbf{\scriptsize pmed\,$\uparrow$} & \textbf{\scriptsize =max\,$\uparrow$}  & \textbf{\scriptsize +max\,$\uparrow$}  & \textbf{\scriptsize =0\,$\downarrow$}  & \textbf{\scriptsize =10p\,$\downarrow$}
& \textbf{\scriptsize pmax\,$\uparrow$} & \textbf{\scriptsize pmed\,$\uparrow$} & \textbf{\scriptsize =max\,$\uparrow$}  & \textbf{\scriptsize +max\,$\uparrow$}  & \textbf{\scriptsize =0\,$\downarrow$}  & \textbf{\scriptsize =10p\,$\downarrow$}
& \textbf{\scriptsize pmax\,$\uparrow$} & \textbf{\scriptsize pmed\,$\uparrow$} & \textbf{\scriptsize =max\,$\uparrow$}  & \textbf{\scriptsize +max\,$\uparrow$}  & \textbf{\scriptsize =0\,$\downarrow$}  & \textbf{\scriptsize =10p\,$\downarrow$} \\
\hline
& \multicolumn{18}{c|}{\textbf{Baseline}} \\
\hline
en & \textbf{98.38} & 97.84 & 92.43 & 95.14 & 43.24 & 69.73 & 98.92 & \textbf{99.46} & 58.92 & 58.92 & 49.73 & 94.59 & \textbf{98.38} & \textbf{98.38} & 72.97 & 73.51 & 54.05 & 51.89 \\ nl & 76.76 & \textbf{78.92} & 56.22 & 57.84 & 35.14 & 47.03 & 44.32 & 32.97 & 43.24 & 45.95 & 57.84 & \textbf{70.81} & 88.11 & \textbf{96.22} & 58.38 & 56.22 & 25.41 & 27.03 \\ id & 75.14 & \textbf{78.92} & 61.08 & 61.62 & 34.59 & 49.73 & \textbf{88.65} & 85.95 & 81.08 & \DD80.0 & 68.65 & 84.86 & 95.68 & \textbf{98.38} & 57.84 & 52.97 & \DD17.3 & 15.68 \\ ms & \textbf{69.19} & \textbf{69.19} & \DD57.3 & 56.22 & 31.89 & 39.46 & 47.57 & \DD37.3 & 38.92 & \DD40.0 & 40.54 & \textbf{58.38} & 75.14 & \textbf{93.51} & \DD42.7 & 42.16 & 14.05 & 14.59 \\ vi & 69.73 & \textbf{78.92} & 51.35 & 53.51 & 30.27 & 40.54 & \textbf{81.08} & 74.05 & 47.03 & 43.78 & \DD8.11 & 14.05 & \textbf{97.84} & \DD97.3 & 45.95 & 48.11 & \DD7.03 & 11.89 \\
\hline

\end{tabular}
}
\caption{Language Steering Shift (LSS) scores from \texttt{fr} to languages listed for different steering factors across models. Bolded texts denote the highest values within a model.}
\label{tab:fr}
\end{table*}

\begin{table*}[htbp]
\centering
\small
\renewcommand{\arraystretch}{1.2}
\setlength{\tabcolsep}{2pt}

\resizebox{\textwidth}{!}{
\begin{tabular}{
|l
|*{6}{c}|
*{6}{c}|
*{6}{c}|
}
\hline
\multirow{2}{*}{} & 
\multicolumn{6}{c|}{\textbf{Qwen}} & 
\multicolumn{6}{c|}{\textbf{Gemma}} & 
\multicolumn{6}{c|}{\textbf{SeaLLM}} \\
\cline{2-19}
& \textbf{\scriptsize pmax\,$\uparrow$} & \textbf{\scriptsize pmed\,$\uparrow$} & \textbf{\scriptsize =max\,$\uparrow$}  & \textbf{\scriptsize +max\,$\uparrow$}  & \textbf{\scriptsize =0\,$\downarrow$}  & \textbf{\scriptsize =10p\,$\downarrow$}
& \textbf{\scriptsize pmax\,$\uparrow$} & \textbf{\scriptsize pmed\,$\uparrow$} & \textbf{\scriptsize =max\,$\uparrow$}  & \textbf{\scriptsize +max\,$\uparrow$}  & \textbf{\scriptsize =0\,$\downarrow$}  & \textbf{\scriptsize =10p\,$\downarrow$}
& \textbf{\scriptsize pmax\,$\uparrow$} & \textbf{\scriptsize pmed\,$\uparrow$} & \textbf{\scriptsize =max\,$\uparrow$}  & \textbf{\scriptsize +max\,$\uparrow$}  & \textbf{\scriptsize =0\,$\downarrow$}  & \textbf{\scriptsize =10p\,$\downarrow$} \\
\hline 
& \multicolumn{18}{c|}{\textbf{Baseline}} \\
\hline 
en & \textbf{96.22} & 95.14 & 94.05 & 93.51 & 75.68 & 88.65 & 98.92 & \textbf{99.46} & 64.86 & 66.49 & 46.49 & 85.95 & 98.38 & \textbf{99.46} & 91.89 & 92.43 & 55.14 & 57.84 \\ nl & 90.27 & \textbf{94.05} & 78.38 & 80.54 & 56.22 & 79.46 & 40.54 & 38.38 & 50.27 & 48.11 & 47.57 & \textbf{68.65} & 94.59 & \textbf{97.3} & \DD80.0 & 77.84 & 25.95 & 23.78 \\ id & \textbf{90.81} & 89.73 & 74.59 & 70.27 & 48.65 & 71.89 & 73.51 & 71.89 & 76.22 & 76.76 & 76.76 & \textbf{87.03} & 96.22 & \textbf{98.38} & 67.03 & 67.03 & 22.16 & 15.14 \\ ms & 72.43 & \textbf{74.59} & 53.51 & 54.05 & 38.38 & 39.46 & 34.59 & 29.19 & \DD40.0 & 39.46 & 38.92 & \textbf{54.05} & 87.03 & \textbf{92.97} & 36.22 & \DD40.0 & 14.59 & 11.89 \\ vi & 90.27 & \textbf{93.51} & 68.11 & 68.11 & 44.32 & 66.49 & \textbf{80.54} & 73.51 & 52.97 & 54.05 & 12.43 & 21.62 & \textbf{99.46} & 98.92 & 70.81 & 69.73 & \DD7.03 & \DD9.19 \\
\hline 

\end{tabular}
}
\caption{Language Steering Shift (LSS) scores from \texttt{it} to languages listed for different steering factors across models. Bolded texts denote the highest values within a model.}
\label{tab:it}
\end{table*}

\begin{table*}[htbp]
\centering
\small
\renewcommand{\arraystretch}{1.2}
\setlength{\tabcolsep}{2pt}

\resizebox{\textwidth}{!}{
\begin{tabular}{
|l
|*{6}{c}|
*{6}{c}|
*{6}{c}|
}
\hline
\multirow{2}{*}{} & 
\multicolumn{6}{c|}{\textbf{Qwen}} & 
\multicolumn{6}{c|}{\textbf{Gemma}} & 
\multicolumn{6}{c|}{\textbf{SeaLLM}} \\
\cline{2-19}
& \textbf{\scriptsize pmax\,$\uparrow$} & \textbf{\scriptsize pmed\,$\uparrow$} & \textbf{\scriptsize =max\,$\uparrow$}  & \textbf{\scriptsize +max\,$\uparrow$}  & \textbf{\scriptsize =0\,$\downarrow$}  & \textbf{\scriptsize =10p\,$\downarrow$}
& \textbf{\scriptsize pmax\,$\uparrow$} & \textbf{\scriptsize pmed\,$\uparrow$} & \textbf{\scriptsize =max\,$\uparrow$}  & \textbf{\scriptsize +max\,$\uparrow$}  & \textbf{\scriptsize =0\,$\downarrow$}  & \textbf{\scriptsize =10p\,$\downarrow$}
& \textbf{\scriptsize pmax\,$\uparrow$} & \textbf{\scriptsize pmed\,$\uparrow$} & \textbf{\scriptsize =max\,$\uparrow$}  & \textbf{\scriptsize +max\,$\uparrow$}  & \textbf{\scriptsize =0\,$\downarrow$}  & \textbf{\scriptsize =10p\,$\downarrow$} \\
\hline 
& \multicolumn{18}{c|}{\textbf{Baseline}} \\
\hline 
en & \textbf{100.0} & \textbf{100.0} & \textbf{100.0} & \textbf{100.0} & 95.14 & 96.22 & \textbf{100.0} & 98.92 & 97.84 & \DD97.3 & 58.92 & 91.35 & \textbf{100.0} & \textbf{100.0} & \textbf{100.0} & 99.46 & 94.05 & 91.35 \\ nl & \textbf{99.46} & 98.92 & 96.22 & \DD97.3 & 60.54 & 72.43 & 89.73 & \textbf{92.43} & 69.19 & 73.51 & 52.97 & 65.41 & \textbf{100.0} & \textbf{100.0} & 92.43 & 92.97 & 14.59 & 17.84 \\ id & \textbf{100.0} & 99.46 & 92.43 & 94.05 & 38.38 & 67.57 & \textbf{95.68} & 92.97 & 95.14 & 93.51 & 58.38 & 79.46 & \textbf{100.0} & \textbf{100.0} & 90.27 & 90.27 & 16.76 & 16.22 \\ ms & \textbf{99.46} & \textbf{99.46} & 75.14 & 75.68 & 32.97 & \DD40.0 & \textbf{76.76} & 71.35 & 63.78 & 62.16 & 28.65 & 52.97 & \textbf{100.0} & 99.46 & 75.14 & 75.68 & 12.97 & \DD9.73 \\ vi & \textbf{100.0} & \textbf{100.0} & \DD97.3 & 97.84 & 78.38 & 83.24 & \textbf{95.68} & 94.05 & 76.76 & 76.76 & \DD6.49 & 21.08 & \textbf{100.0} & \textbf{100.0} & 87.03 & 89.73 & 16.76 & 18.92 \\
\hline 

\end{tabular}
}

\caption{Language Steering Shift (LSS) scores from \texttt{jp} to languages listed for different steering factors across models. Bolded texts denote the highest values within a model.}
\label{tab:jp}
\end{table*}

\begin{table*}[htbp]
\centering
\small
\renewcommand{\arraystretch}{1.2}
\setlength{\tabcolsep}{2pt}

\resizebox{\textwidth}{!}{
\begin{tabular}{
|l
|*{6}{c}|
*{6}{c}|
*{6}{c}|
}
\hline
\multirow{2}{*}{} & 
\multicolumn{6}{c|}{\textbf{Qwen}} & 
\multicolumn{6}{c|}{\textbf{Gemma}} & 
\multicolumn{6}{c|}{\textbf{SeaLLM}} \\
\cline{2-19}
& \textbf{\scriptsize pmax\,$\uparrow$} & \textbf{\scriptsize pmed\,$\uparrow$} & \textbf{\scriptsize =max\,$\uparrow$}  & \textbf{\scriptsize +max\,$\uparrow$}  & \textbf{\scriptsize =0\,$\downarrow$}  & \textbf{\scriptsize =10p\,$\downarrow$}
& \textbf{\scriptsize pmax\,$\uparrow$} & \textbf{\scriptsize pmed\,$\uparrow$} & \textbf{\scriptsize =max\,$\uparrow$}  & \textbf{\scriptsize +max\,$\uparrow$}  & \textbf{\scriptsize =0\,$\downarrow$}  & \textbf{\scriptsize =10p\,$\downarrow$}
& \textbf{\scriptsize pmax\,$\uparrow$} & \textbf{\scriptsize pmed\,$\uparrow$} & \textbf{\scriptsize =max\,$\uparrow$}  & \textbf{\scriptsize +max\,$\uparrow$}  & \textbf{\scriptsize =0\,$\downarrow$}  & \textbf{\scriptsize =10p\,$\downarrow$} \\
\hline 
& \multicolumn{18}{c|}{\textbf{Baseline}} \\
\hline 
en & \textbf{100.0} & \textbf{100.0} & 98.92 & 98.92 & 93.51 & 96.76 & 96.22 & \textbf{97.84} & 92.97 & 93.51 & 41.08 & 67.03 & \textbf{100.0} & 99.46 & 98.38 & 98.92 & 52.43 & 51.89 \\ nl & \textbf{96.22} & \textbf{96.22} & 83.78 & 83.78 & 81.62 & \DD77.3 & \textbf{82.16} & 77.84 & 73.51 & 74.05 & \DD37.3 & 55.68 & \DD97.3 & \textbf{99.46} & 81.62 & 84.86 & 42.16 & 48.11 \\ id & 61.08 & 46.49 & 67.57 & 67.03 & 70.81 & \textbf{71.35} & 71.89 & \textbf{74.05} & 69.73 & 69.73 & 57.84 & 62.16 & 65.41 & 49.19 & 66.49 & 67.03 & 64.32 & \textbf{69.19} \\ vi & 96.76 & \textbf{98.38} & 87.57 & 88.11 & 72.43 & 71.35 & \textbf{79.46} & 76.76 & 54.05 & 57.84 & 16.76 & 43.24 & \textbf{99.46} & \textbf{99.46} & 77.84 & \DD80.0 & 18.92 & 34.59 \\ jp & 92.97 & \textbf{94.59} & 70.81 & 72.97 & 35.68 & 35.68 & 85.41 & \textbf{88.65} & 68.65 & 69.73 & 29.19 & 38.92 & \textbf{100.0} & \textbf{100.0} & \DD57.3 & 55.68 & \DD6.49 & \DD5.95 \\
\hline 

\end{tabular}
}

\caption{Language Steering Shift (LSS) scores from \texttt{ms} to languages listed for different steering factors across models. Bolded texts denote the highest values within a model.}
\label{tab:ms}
\end{table*}

\begin{table*}[htbp]
\centering
\small
\renewcommand{\arraystretch}{1.2}
\setlength{\tabcolsep}{2pt}

\resizebox{\textwidth}{!}{
\begin{tabular}{
|l
|*{6}{c}|
*{6}{c}|
*{6}{c}|
}
\hline
\multirow{2}{*}{} & 
\multicolumn{6}{c|}{\textbf{Qwen}} & 
\multicolumn{6}{c|}{\textbf{Gemma}} & 
\multicolumn{6}{c|}{\textbf{SeaLLM}} \\
\cline{2-19}
& \textbf{\scriptsize pmax\,$\uparrow$} & \textbf{\scriptsize pmed\,$\uparrow$} & \textbf{\scriptsize =max\,$\uparrow$}  & \textbf{\scriptsize +max\,$\uparrow$}  & \textbf{\scriptsize =0\,$\downarrow$}  & \textbf{\scriptsize =10p\,$\downarrow$}
& \textbf{\scriptsize pmax\,$\uparrow$} & \textbf{\scriptsize pmed\,$\uparrow$} & \textbf{\scriptsize =max\,$\uparrow$}  & \textbf{\scriptsize +max\,$\uparrow$}  & \textbf{\scriptsize =0\,$\downarrow$}  & \textbf{\scriptsize =10p\,$\downarrow$}
& \textbf{\scriptsize pmax\,$\uparrow$} & \textbf{\scriptsize pmed\,$\uparrow$} & \textbf{\scriptsize =max\,$\uparrow$}  & \textbf{\scriptsize +max\,$\uparrow$}  & \textbf{\scriptsize =0\,$\downarrow$}  & \textbf{\scriptsize =10p\,$\downarrow$} \\
\hline 
& \multicolumn{18}{c|}{\textbf{Baseline}} \\
\hline 
en & 99.46 & 99.46 & 99.46 & \textbf{100.0} & 84.32 & 96.76 & 98.38 & \textbf{99.46} & 81.08 & \DD82.7 & \DD57.3 & \DD82.7 & \textbf{100.0} & 98.92 & 99.46 & 98.92 & 55.14 & 75.68 \\ id & 94.05 & \textbf{95.68} & 56.22 & 57.84 & 54.59 & \DD57.3 & \textbf{94.05} & 90.81 & 89.73 & 90.81 & 68.65 & 87.03 & 93.51 & \textbf{95.68} & 53.51 & 43.78 & 31.89 & 35.14 \\ ms & \DD80.0 & \textbf{87.57} & 55.68 & \DD57.3 & \DD40.0 & 49.19 & \textbf{68.65} & 60.54 & 57.84 & 56.22 & 54.59 & 59.46 & \DD62.7 & \textbf{86.49} & 45.95 & \DD42.7 & 18.38 & 21.62 \\ vi & 88.11 & \textbf{90.27} & 80.54 & 78.92 & 60.54 & 67.03 & \textbf{91.89} & 86.49 & 65.41 & 70.81 & 10.27 & 24.86 & \textbf{100.0} & 99.46 & 71.89 & 71.89 & 11.35 & 16.76 \\ jp & 94.59 & \textbf{95.14} & 42.16 & 39.46 & 18.92 & 16.76 & 89.19 & \textbf{89.73} & 55.68 & \DD57.3 & 31.35 & 44.32 & \textbf{98.38} & \textbf{98.38} & 22.16 & 21.62 & \DD2.7 & \DD1.62 \\
\hline 

\end{tabular}
}
\caption{Language Steering Shift (LSS) scores from \texttt{nl} to languages listed for different steering factors across models. Bolded texts denote the highest values within a model.}
\label{tab:nl}
\end{table*}

\begin{table*}[htbp]
\centering
\small
\renewcommand{\arraystretch}{1.2}
\setlength{\tabcolsep}{2pt}

\resizebox{\textwidth}{!}{
\begin{tabular}{
|l
|*{6}{c}|
*{6}{c}|
*{6}{c}|
}
\hline
\multirow{2}{*}{} & 
\multicolumn{6}{c|}{\textbf{Qwen}} & 
\multicolumn{6}{c|}{\textbf{Gemma}} & 
\multicolumn{6}{c|}{\textbf{SeaLLM}} \\
\cline{2-19}
& \textbf{\scriptsize pmax\,$\uparrow$} & \textbf{\scriptsize pmed\,$\uparrow$} & \textbf{\scriptsize =max\,$\uparrow$}  & \textbf{\scriptsize +max\,$\uparrow$}  & \textbf{\scriptsize =0\,$\downarrow$}  & \textbf{\scriptsize =10p\,$\downarrow$}
& \textbf{\scriptsize pmax\,$\uparrow$} & \textbf{\scriptsize pmed\,$\uparrow$} & \textbf{\scriptsize =max\,$\uparrow$}  & \textbf{\scriptsize +max\,$\uparrow$}  & \textbf{\scriptsize =0\,$\downarrow$}  & \textbf{\scriptsize =10p\,$\downarrow$}
& \textbf{\scriptsize pmax\,$\uparrow$} & \textbf{\scriptsize pmed\,$\uparrow$} & \textbf{\scriptsize =max\,$\uparrow$}  & \textbf{\scriptsize +max\,$\uparrow$}  & \textbf{\scriptsize =0\,$\downarrow$}  & \textbf{\scriptsize =10p\,$\downarrow$} \\
\hline 
& \multicolumn{18}{c|}{\textbf{Baseline}} \\
\hline 
en & 99.46 & 99.46 & \textbf{100.0} & \textbf{100.0} & 72.97 & 95.14 & \textbf{96.76} & 96.22 & 68.65 & 65.95 & 52.43 & 84.86 & \textbf{100.0} & \textbf{100.0} & 99.46 & 98.92 & 65.41 & 65.41 \\ nl & \textbf{97.84} & \DD97.3 & 87.57 & 88.11 & 63.24 & 69.73 & 54.05 & \DD40.0 & 48.11 & 48.65 & 61.62 & \textbf{70.27} & 98.92 & \textbf{100.0} & 80.54 & \DD80.0 & 47.03 & 47.03 \\ id & 97.84 & \textbf{99.46} & 71.35 & 68.65 & 56.22 & 59.46 & 76.22 & 58.38 & 72.97 & 72.97 & 71.35 & \textbf{88.65} & \textbf{100.0} & \textbf{100.0} & 72.43 & 70.81 & 38.38 & 29.19 \\ ms & 81.08 & \textbf{96.22} & 56.76 & 56.76 & 45.95 & 47.03 & 36.22 & 26.49 & 38.92 & 37.84 & 37.84 & \textbf{55.68} & 91.89 & \textbf{98.92} & 47.57 & 47.03 & 23.24 & \DD20.0 \\ vi & 88.65 & \textbf{93.51} & 58.38 & 58.38 & 51.89 & 53.51 & \textbf{81.08} & 75.68 & 49.19 & 50.27 & \DD8.11 & 16.22 & \textbf{100.0} & \textbf{100.0} & \DD77.3 & 78.38 & \DD22.7 & \DD22.7 \\
\hline

\end{tabular}
}
\caption{Language Steering Shift (LSS) scores from \texttt{pt} to languages listed for different steering factors across models. Bolded texts denote the highest values within a model.}
\label{tab:pt}
\end{table*}

\begin{table*}[htbp]
\centering
\small
\renewcommand{\arraystretch}{1.2}
\setlength{\tabcolsep}{2pt}

\resizebox{\textwidth}{!}{
\begin{tabular}{
|l
|*{6}{c}|
*{6}{c}|
*{6}{c}|
}
\hline
\multirow{2}{*}{} & 
\multicolumn{6}{c|}{\textbf{Qwen}} & 
\multicolumn{6}{c|}{\textbf{Gemma}} & 
\multicolumn{6}{c|}{\textbf{SeaLLM}} \\
\cline{2-19}
& \textbf{\scriptsize pmax\,$\uparrow$} & \textbf{\scriptsize pmed\,$\uparrow$} & \textbf{\scriptsize =max\,$\uparrow$}  & \textbf{\scriptsize +max\,$\uparrow$}  & \textbf{\scriptsize =0\,$\downarrow$}  & \textbf{\scriptsize =10p\,$\downarrow$}
& \textbf{\scriptsize pmax\,$\uparrow$} & \textbf{\scriptsize pmed\,$\uparrow$} & \textbf{\scriptsize =max\,$\uparrow$}  & \textbf{\scriptsize +max\,$\uparrow$}  & \textbf{\scriptsize =0\,$\downarrow$}  & \textbf{\scriptsize =10p\,$\downarrow$}
& \textbf{\scriptsize pmax\,$\uparrow$} & \textbf{\scriptsize pmed\,$\uparrow$} & \textbf{\scriptsize =max\,$\uparrow$}  & \textbf{\scriptsize +max\,$\uparrow$}  & \textbf{\scriptsize =0\,$\downarrow$}  & \textbf{\scriptsize =10p\,$\downarrow$} \\
\hline 
& \multicolumn{18}{c|}{\textbf{Baseline}} \\
\hline 
en & \textbf{100.0} & \textbf{100.0} & \textbf{100.0} & \textbf{100.0} & \DD80.0 & 94.05 & \textbf{100.0} & \textbf{100.0} & 97.84 & 97.84 & 53.51 & 75.14 & \textbf{100.0} & \textbf{100.0} & 99.46 & \textbf{100.0} & 70.81 & 54.59 \\ nl & 99.46 & \textbf{100.0} & 95.14 & 94.59 & 30.81 & 65.95 & \textbf{96.22} & 94.59 & 84.86 & 86.49 & 49.19 & 67.03 & \textbf{100.0} & \textbf{100.0} & 93.51 & 92.97 & 19.46 & 19.46 \\ id & \textbf{100.0} & \textbf{100.0} & 96.76 & 96.76 & 30.81 & 55.68 & \textbf{96.76} & 94.05 & 94.59 & 93.51 & 64.86 & 87.57 & \textbf{100.0} & \textbf{100.0} & 90.81 & 90.81 & 21.62 & 15.68 \\ ms & \textbf{100.0} & \textbf{100.0} & 78.38 & \DD77.3 & 24.86 & 47.57 & \textbf{75.68} & 68.65 & 68.11 & 68.11 & 39.46 & 66.49 & \textbf{100.0} & \textbf{100.0} & 71.89 & 71.35 & 12.97 & \DD9.73 \\ vi & \textbf{99.46} & 98.38 & 89.73 & 91.89 & 49.73 & 90.27 & \textbf{91.35} & 88.65 & 68.65 & 69.73 & 15.68 & 67.03 & \textbf{100.0} & \textbf{100.0} & 87.57 & 86.49 & 25.95 & 25.95 \\
\hline

\end{tabular}
}
\caption{Language Steering Shift (LSS) scores from \texttt{ru} to languages listed for different steering factors across models. Bolded texts denote the highest values within a model.}
\label{tab:ru}
\end{table*}

\begin{table*}[htbp]
\centering
\small
\renewcommand{\arraystretch}{1.2}
\setlength{\tabcolsep}{2pt}

\resizebox{\textwidth}{!}{
\begin{tabular}{
|l
|*{6}{c}|
*{6}{c}|
*{6}{c}|
}
\hline
\multirow{2}{*}{} & 
\multicolumn{6}{c|}{\textbf{Qwen}} & 
\multicolumn{6}{c|}{\textbf{Gemma}} & 
\multicolumn{6}{c|}{\textbf{SeaLLM}} \\
\cline{2-19}
& \textbf{\scriptsize pmax\,$\uparrow$} & \textbf{\scriptsize pmed\,$\uparrow$} & \textbf{\scriptsize =max\,$\uparrow$}  & \textbf{\scriptsize +max\,$\uparrow$}  & \textbf{\scriptsize =0\,$\downarrow$}  & \textbf{\scriptsize =10p\,$\downarrow$}
& \textbf{\scriptsize pmax\,$\uparrow$} & \textbf{\scriptsize pmed\,$\uparrow$} & \textbf{\scriptsize =max\,$\uparrow$}  & \textbf{\scriptsize +max\,$\uparrow$}  & \textbf{\scriptsize =0\,$\downarrow$}  & \textbf{\scriptsize =10p\,$\downarrow$}
& \textbf{\scriptsize pmax\,$\uparrow$} & \textbf{\scriptsize pmed\,$\uparrow$} & \textbf{\scriptsize =max\,$\uparrow$}  & \textbf{\scriptsize +max\,$\uparrow$}  & \textbf{\scriptsize =0\,$\downarrow$}  & \textbf{\scriptsize =10p\,$\downarrow$} \\
\hline 
& \multicolumn{18}{c|}{\textbf{Baseline}} \\
\hline 
en & \textbf{100.0} & \textbf{100.0} & \textbf{100.0} & \textbf{100.0} & \textbf{100.0} & \textbf{100.0} & \textbf{100.0} & \textbf{100.0} & \textbf{100.0} & \textbf{100.0} & 97.84 & \textbf{100.0} & \textbf{100.0} & \textbf{100.0} & \textbf{100.0} & \textbf{100.0} & \DD97.3 & 96.76 \\ nl & \textbf{100.0} & \textbf{100.0} & \textbf{100.0} & \textbf{100.0} & 95.68 & \textbf{100.0} & \textbf{100.0} & \textbf{100.0} & 98.92 & 98.92 & 88.65 & \textbf{100.0} & \textbf{100.0} & \textbf{100.0} & \textbf{100.0} & \textbf{100.0} & 84.86 & 81.08 \\ id & \textbf{100.0} & \textbf{100.0} & \textbf{100.0} & \textbf{100.0} & 93.51 & 99.46 & \textbf{100.0} & \textbf{100.0} & 99.46 & 98.92 & 94.59 & 99.46 & \textbf{100.0} & 98.38 & \textbf{100.0} & \textbf{100.0} & \DD80.0 & 84.86 \\
\hline

\end{tabular}
}
\caption{Language Steering Shift (LSS) scores from \texttt{ta} to languages listed for different steering factors across models. Bolded texts denote the highest values within a model.}
\label{tab:ta}
\end{table*}
\begin{table*}[htbp]
\centering
\small
\renewcommand{\arraystretch}{1.2}
\setlength{\tabcolsep}{2pt}

\resizebox{\textwidth}{!}{
\begin{tabular}{
|l
|*{6}{c}|
*{6}{c}|
*{6}{c}|
}
\hline
\multirow{2}{*}{} & 
\multicolumn{6}{c|}{\textbf{Qwen}} & 
\multicolumn{6}{c|}{\textbf{Gemma}} & 
\multicolumn{6}{c|}{\textbf{SeaLLM}} \\
\cline{2-19}
& \textbf{\scriptsize pmax\,$\uparrow$} & \textbf{\scriptsize pmed\,$\uparrow$} & \textbf{\scriptsize =max\,$\uparrow$}  & \textbf{\scriptsize +max\,$\uparrow$}  & \textbf{\scriptsize =0\,$\downarrow$}  & \textbf{\scriptsize =10p\,$\downarrow$}
& \textbf{\scriptsize pmax\,$\uparrow$} & \textbf{\scriptsize pmed\,$\uparrow$} & \textbf{\scriptsize =max\,$\uparrow$}  & \textbf{\scriptsize +max\,$\uparrow$}  & \textbf{\scriptsize =0\,$\downarrow$}  & \textbf{\scriptsize =10p\,$\downarrow$}
& \textbf{\scriptsize pmax\,$\uparrow$} & \textbf{\scriptsize pmed\,$\uparrow$} & \textbf{\scriptsize =max\,$\uparrow$}  & \textbf{\scriptsize +max\,$\uparrow$}  & \textbf{\scriptsize =0\,$\downarrow$}  & \textbf{\scriptsize =10p\,$\downarrow$} \\
\hline 
& \multicolumn{18}{c|}{\textbf{Baseline}} \\
\hline
en & \textbf{100.0} & \textbf{100.0} & \textbf{100.0} & \textbf{100.0} & 98.92 & 99.46 & \textbf{100.0} & \textbf{100.0} & 99.46 & \textbf{100.0} & 97.84 & \textbf{100.0} & \textbf{100.0} & \textbf{100.0} & \textbf{100.0} & \textbf{100.0} & 65.95 & 74.05 \\ nl & \textbf{100.0} & \textbf{100.0} & 99.46 & 99.46 & 88.11 & 92.43 & \textbf{100.0} & 97.84 & 95.68 & 94.59 & 81.08 & 91.89 & \textbf{100.0} & \textbf{100.0} & 98.92 & 98.92 & 32.97 & 31.89 \\ id & \textbf{100.0} & \textbf{100.0} & 99.46 & 99.46 & 75.68 & 94.59 & \textbf{99.46} & 95.14 & \textbf{99.46} & \textbf{99.46} & 95.68 & \textbf{99.46} & \textbf{100.0} & \textbf{100.0} & 99.46 & 98.38 & 27.03 & 30.81 \\ ms & \textbf{100.0} & \textbf{100.0} & 97.84 & 97.84 & 52.43 & 70.81 & \textbf{99.46} & 78.92 & 94.05 & 96.22 & 93.51 & \textbf{99.46} & \textbf{100.0} & 99.46 & 97.84 & 95.68 & 22.16 & 33.51 \\
\hline

\end{tabular}
}
\caption{Language Steering Shift (LSS) scores from \texttt{th} to languages listed for different steering factors across models. Bolded texts denote the highest values within a model.}
\label{tab:th}
\end{table*}

\begin{table*}[htbp]
\centering
\small
\renewcommand{\arraystretch}{1.2}
\setlength{\tabcolsep}{2pt}

\resizebox{\textwidth}{!}{
\begin{tabular}{
|l
|*{6}{c}|
*{6}{c}|
*{6}{c}|
}
\hline
\multirow{2}{*}{} & 
\multicolumn{6}{c|}{\textbf{Qwen}} & 
\multicolumn{6}{c|}{\textbf{Gemma}} & 
\multicolumn{6}{c|}{\textbf{SeaLLM}} \\
\cline{2-19}
& \textbf{\scriptsize pmax\,$\uparrow$} & \textbf{\scriptsize pmed\,$\uparrow$} & \textbf{\scriptsize =max\,$\uparrow$}  & \textbf{\scriptsize +max\,$\uparrow$}  & \textbf{\scriptsize =0\,$\downarrow$}  & \textbf{\scriptsize =10p\,$\downarrow$}
& \textbf{\scriptsize pmax\,$\uparrow$} & \textbf{\scriptsize pmed\,$\uparrow$} & \textbf{\scriptsize =max\,$\uparrow$}  & \textbf{\scriptsize +max\,$\uparrow$}  & \textbf{\scriptsize =0\,$\downarrow$}  & \textbf{\scriptsize =10p\,$\downarrow$}
& \textbf{\scriptsize pmax\,$\uparrow$} & \textbf{\scriptsize pmed\,$\uparrow$} & \textbf{\scriptsize =max\,$\uparrow$}  & \textbf{\scriptsize +max\,$\uparrow$}  & \textbf{\scriptsize =0\,$\downarrow$}  & \textbf{\scriptsize =10p\,$\downarrow$} \\
\hline 
& \multicolumn{18}{c|}{\textbf{Baseline}} \\
\hline 
en & \textbf{100.0} & \textbf{100.0} & \textbf{100.0} & \textbf{100.0} & 91.35 & \textbf{100.0} & \textbf{100.0} & \textbf{100.0} & 97.84 & \DD97.3 & 29.19 & 77.84 & \textbf{100.0} & \textbf{100.0} & \textbf{100.0} & \textbf{100.0} & 79.46 & 64.86 \\ nl & \textbf{99.46} & 98.92 & 84.86 & 87.03 & 69.73 & \DD77.3 & \textbf{89.73} & 81.62 & 83.78 & 86.49 & 24.86 & \DD77.3 & \textbf{100.0} & 99.46 & 84.86 & 81.62 & \DD60.0 & 53.51 \\ id & \textbf{95.68} & \textbf{95.68} & 83.24 & 85.95 & 77.84 & 79.46 & \textbf{95.14} & 94.05 & 90.81 & 89.19 & 49.73 & 87.03 & \textbf{97.3} & \textbf{97.3} & 76.76 & 78.92 & 38.92 & 34.59 \\ ms & 84.32 & \textbf{84.86} & 75.14 & 74.05 & 44.32 & 54.05 & 66.49 & 63.24 & 69.19 & 69.19 & 37.84 & \textbf{70.27} & 89.73 & \textbf{91.35} & 65.41 & 64.86 & 23.78 & 28.11 \\ vi & 96.76 & \textbf{97.3} & 86.49 & 87.03 & 63.24 & 67.03 & \textbf{97.3} & 91.35 & 82.16 & \DD82.7 & \DD8.65 & 47.03 & 99.46 & \textbf{100.0} & 79.46 & \DD80.0 & 23.78 & 28.11 \\
\hline 

\end{tabular}
}
\caption{Language Steering Shift (LSS) scores from \texttt{tr} to languages listed for different steering factors across models. Bolded texts denote the highest values within a model.}
\label{tab:tr}
\end{table*}

\begin{table*}[htbp]
\centering
\small
\renewcommand{\arraystretch}{1.2}
\setlength{\tabcolsep}{2pt}

\resizebox{\textwidth}{!}{
\begin{tabular}{
|l
|*{6}{c}|
*{6}{c}|
*{6}{c}|
}
\hline
\multirow{2}{*}{} & 
\multicolumn{6}{c|}{\textbf{Qwen}} & 
\multicolumn{6}{c|}{\textbf{Gemma}} & 
\multicolumn{6}{c|}{\textbf{SeaLLM}} \\
\cline{2-19}
& \textbf{\scriptsize pmax\,$\uparrow$} & \textbf{\scriptsize pmed\,$\uparrow$} & \textbf{\scriptsize =max\,$\uparrow$}  & \textbf{\scriptsize +max\,$\uparrow$}  & \textbf{\scriptsize =0\,$\downarrow$}  & \textbf{\scriptsize =10p\,$\downarrow$}
& \textbf{\scriptsize pmax\,$\uparrow$} & \textbf{\scriptsize pmed\,$\uparrow$} & \textbf{\scriptsize =max\,$\uparrow$}  & \textbf{\scriptsize +max\,$\uparrow$}  & \textbf{\scriptsize =0\,$\downarrow$}  & \textbf{\scriptsize =10p\,$\downarrow$}
& \textbf{\scriptsize pmax\,$\uparrow$} & \textbf{\scriptsize pmed\,$\uparrow$} & \textbf{\scriptsize =max\,$\uparrow$}  & \textbf{\scriptsize +max\,$\uparrow$}  & \textbf{\scriptsize =0\,$\downarrow$}  & \textbf{\scriptsize =10p\,$\downarrow$} \\
\hline 
& \multicolumn{18}{c|}{\textbf{Baseline}} \\
\hline 
en & 99.46 & \textbf{100.0} & 99.46 & 99.46 & 93.51 & 95.68 & \textbf{100.0} & \textbf{100.0} & 94.59 & 94.59 & 69.19 & 95.68 & \textbf{100.0} & \textbf{100.0} & 99.46 & 99.46 & 77.84 & 62.16 \\ nl & \textbf{100.0} & \textbf{100.0} & 95.14 & 96.76 & 64.32 & 63.78 & \textbf{88.11} & 78.92 & 85.95 & 84.32 & 59.46 & 76.22 & \textbf{100.0} & \textbf{100.0} & 75.14 & 79.46 & 40.54 & 55.14 \\ id & 98.38 & \textbf{100.0} & 67.57 & 71.89 & 58.38 & 61.62 & 90.27 & 87.03 & 89.73 & 90.81 & 78.92 & \textbf{94.05} & \textbf{100.0} & \textbf{100.0} & 76.76 & 78.92 & 51.35 & 59.46 \\ ms & 87.57 & \textbf{94.59} & 63.24 & 61.62 & \DD37.3 & 38.38 & 63.78 & 56.22 & 56.76 & 57.84 & 53.51 & \textbf{72.97} & 98.92 & \textbf{100.0} & 48.65 & 50.27 & 28.65 & 38.38 \\ jp & 89.19 & \textbf{89.73} & 52.43 & 54.05 & 30.81 & 17.84 & \textbf{99.46} & 98.92 & 78.92 & 78.92 & 44.32 & 81.62 & 99.46 & \textbf{100.0} & 40.54 & 34.59 & 17.84 & 15.14 \\
\hline

\end{tabular}
}
\caption{Language Steering Shift (LSS) scores from \texttt{vi} to languages listed for different steering factors across models. Bolded texts denote the highest values within a model.}
\label{tab:vi}
\end{table*}

\begin{table*}[htbp]
\centering
\small
\renewcommand{\arraystretch}{1.2}
\setlength{\tabcolsep}{2pt}

\resizebox{\textwidth}{!}{
\begin{tabular}{
|l
|*{6}{c}|
*{6}{c}|
*{6}{c}|
}
\hline
\multirow{2}{*}{} & 
\multicolumn{6}{c|}{\textbf{Qwen}} & 
\multicolumn{6}{c|}{\textbf{Gemma}} & 
\multicolumn{6}{c|}{\textbf{SeaLLM}} \\
\cline{2-19}
& \textbf{\scriptsize pmax\,$\uparrow$} & \textbf{\scriptsize pmed\,$\uparrow$} & \textbf{\scriptsize =max\,$\uparrow$}  & \textbf{\scriptsize +max\,$\uparrow$}  & \textbf{\scriptsize =0\,$\downarrow$}  & \textbf{\scriptsize =10p\,$\downarrow$}
& \textbf{\scriptsize pmax\,$\uparrow$} & \textbf{\scriptsize pmed\,$\uparrow$} & \textbf{\scriptsize =max\,$\uparrow$}  & \textbf{\scriptsize +max\,$\uparrow$}  & \textbf{\scriptsize =0\,$\downarrow$}  & \textbf{\scriptsize =10p\,$\downarrow$}
& \textbf{\scriptsize pmax\,$\uparrow$} & \textbf{\scriptsize pmed\,$\uparrow$} & \textbf{\scriptsize =max\,$\uparrow$}  & \textbf{\scriptsize +max\,$\uparrow$}  & \textbf{\scriptsize =0\,$\downarrow$}  & \textbf{\scriptsize =10p\,$\downarrow$} \\
\hline 
& \multicolumn{18}{c|}{\textbf{Baseline}} \\
\hline 
en & \textbf{100.0} & \textbf{100.0} & \textbf{100.0} & \textbf{100.0} & 84.86 & 96.76 & \textbf{100.0} & \textbf{100.0} & 95.14 & 95.14 & 51.35 & 71.35 & \textbf{100.0} & \textbf{100.0} & \textbf{100.0} & \textbf{100.0} & 58.92 & 43.24 \\ nl & \textbf{100.0} & \textbf{100.0} & 96.76 & 95.14 & 38.92 & 67.57 & 90.27 & \textbf{92.43} & 81.08 & 79.46 & 35.68 & 42.16 & \textbf{100.0} & \textbf{100.0} & 93.51 & 95.14 & \DD5.95 & \DD5.41 \\ id & \textbf{100.0} & \textbf{100.0} & 95.68 & 93.51 & 31.35 & 56.22 & \textbf{97.3} & 96.22 & 95.68 & 95.68 & 48.11 & 75.14 & \textbf{100.0} & \textbf{100.0} & 91.35 & 91.35 & 11.89 & \DD5.95 \\ ms & \textbf{100.0} & 99.46 & 77.84 & 76.22 & 24.86 & 45.41 & \textbf{76.76} & 71.89 & 70.27 & 69.73 & 27.03 & 54.59 & \textbf{100.0} & \textbf{100.0} & 69.19 & 67.57 & \DD5.95 & \DD4.86 \\
\hline 

\end{tabular}
}
\caption{Language Steering Shift (LSS) scores from \texttt{zh} to languages listed for different steering factors across models. Bolded texts denote the highest values within a model.}
\label{tab:zh}
\end{table*}

\begin{table}[h]
\centering
\small
\begin{tabular}{c|c|c|c|c}
% \hline
\toprule

Factor 1 & Factor 2 & $\Delta$ Mean & \textit{p}-value & Sig. \\
\midrule
\steer{pmax} & \steer{pmed} & -0.0109 & 0.9813 & No \\
\steer{pmax} & \steer{=max} & -0.1789 & 0.0000 & Yes \\
\steer{pmax} & \steer{+max} & -0.1946 & 0.0000 & Yes \\
\steer{pmed} & \steer{=max} & -0.1680 & 0.0000 & Yes \\
\steer{pmed} & \steer{+max} & -0.1837 & 0.0000 & Yes \\
\steer{=max} & \steer{+max} & -0.0157 & 0.9117 & No \\
\bottomrule
\end{tabular}
\caption{Tukey HSD pairwise comparisons between LSS scores of steering factors on LAPE neurons. \textit{Sig.} denotes the significance of Factor2 compared to Factor1.}

\label{tab:tukey-pairwise}
\end{table}

% \documentclass{article}
% \usepackage{booktabs}
% \usepackage{tabularx}
% \usepackage{geometry}
% \usepackage{adjustbox}
% \geometry{margin=1in}
% =
\begin{table*}[htbp]
\centering
\small
\resizebox{0.6\textwidth}{!}{%
\begin{tabular}{llllll}
\toprule
\textbf{Lang} & \textbf{SeaLLMs3-1.5B} & \textbf{Qwen2.5-0.5B} & \textbf{Gemma2-2B} \\
\midrule
et & 0.87 $\to$ \textbf{1.50} & 0.49 $\to$ \textbf{0.68} & 0.45 $\to$ \textbf{0.46} \\
fr & 1.43 $\to$ \textbf{18.49} & 0.52 $\to$ \textbf{12.94} & 0.76 $\to$ \textbf{2.96} \\
ht & \textbf{1.01} $\to$ 0.12 & 0.55 $\to$ \textbf{0.89} & 0.42 $\to$ \textbf{0.45} \\
id & 1.92 $\to$ \textbf{15.95} & 0.65 $\to$ \textbf{5.52} & \textbf{0.55} $\to$ 0.55 \\
it & 1.15 $\to$ \textbf{10.53} & 0.63 $\to$ \textbf{1.26} & 0.48 $\to$ \textbf{1.69} \\
jp & 0.11 $\to$ \textbf{0.26} & 0.06 $\to$ \textbf{0.15} & 0.02 $\to$ \textbf{0.02} \\
ms & 1.54 $\to$ \textbf{9.70} & 0.61 $\to$ \textbf{3.83} & 0.49 $\to$ \textbf{0.60} \\
nl & 1.20 $\to$ \textbf{2.25} & 0.55 $\to$ \textbf{2.79} & 0.45 $\to$ \textbf{0.62} \\
pt & 1.47 $\to$ \textbf{15.57} & 0.61 $\to$ \textbf{8.36} & 0.61 $\to$ \textbf{2.42} \\
qu & \textbf{0.93} $\to$ 0.24 & 0.46 $\to$ \textbf{0.76} & 0.44 $\to$ \textbf{0.46} \\
ru & 0.75 $\to$ \textbf{10.06} & \textbf{5.22} $\to$ 4.76 & 0.38 $\to$ \textbf{0.48} \\
sw & \textbf{1.07} $\to$ 0.17 & 0.51 $\to$ \textbf{0.89} & 0.45 $\to$ \textbf{0.46} \\
ta & \textbf{0.64} $\to$ 0.51 & \textbf{0.39} $\to$ 0.18 & \textbf{0.40} $\to$ 0.38 \\
th & 0.56 $\to$ \textbf{2.02} & 0.25 $\to$ \textbf{1.44} & 0.17 $\to$ \textbf{0.23} \\
tr & 1.02 $\to$ \textbf{3.44} & 0.48 $\to$ \textbf{1.62} & \textbf{0.45} $\to$ 0.44 \\
vi & 1.65 $\to$ \textbf{18.38} & 0.52 $\to$ \textbf{10.72} & 0.33 $\to$ \textbf{0.66} \\
zh & 0.32 $\to$ \textbf{0.33} & 0.20 $\to$ \textbf{0.22} & 0.09 $\to$ \textbf{0.11} \\
\bottomrule
\end{tabular}
}
\caption{BLEU score changes of self-language intervention with \steer{pmedian} from \texttt{en} to target listed on FLORES-200 dataset. The translation prompts are \textit{non-targeted}.}
\label{tab:bleu-pmed}
\end{table*}
% \documentclass{article}
% \usepackage{booktabs}
% \usepackage{tabularx}
% \usepackage{geometry}
% \usepackage{adjustbox}
% \geometry{margin=1in}
% =
\begin{table*}[htbp]
\centering
\small
\resizebox{\textwidth}{!}{%
\begin{tabular}{lllllll}
\toprule
\textbf{Lang} & \textbf{Qwen2.5-7B} & \textbf{Qwen2.5-0.5B} & \textbf{SeaLLMs3-7B} & \textbf{SeaLLMsv3-1.5B} & \textbf{Gemma2-2B} & \textbf{Gemma2-9b} \\
\midrule
et & 0.76 $\to$ \textbf{2.54} & 0.49 $\to$ \textbf{0.89} & 0.43 $\to$ \textbf{2.99} & 0.87 $\to$ \textbf{1.87} & 0.45 $\to$ \textbf{0.52} & \textbf{0.24} $\to$ 0.23 \\
fr & 13.75 $\to$ \textbf{28.15} & 0.52 $\to$ \textbf{12.90} & 0.32 $\to$ \textbf{25.01} & 1.43 $\to$ \textbf{19.82} & 0.76 $\to$ \textbf{9.08} & 0.57 $\to$ \textbf{4.54} \\
ht & 0.92 $\to$ \textbf{1.68} & 0.55 $\to$ \textbf{1.02} & 0.34 $\to$ \textbf{1.35} & \textbf{1.01} $\to$ 0.89 & 0.42 $\to$ \textbf{0.50} & 0.24 $\to$ \textbf{0.26} \\
id & 1.13 $\to$ \textbf{25.01} & 0.65 $\to$ \textbf{6.37} & 0.52 $\to$ \textbf{26.88} & 1.92 $\to$ \textbf{15.99} & 0.55 $\to$ \textbf{1.64} & 0.29 $\to$ \textbf{0.50} \\
it & 1.16 $\to$ \textbf{14.60} & 0.63 $\to$ \textbf{2.37} & 0.44 $\to$ \textbf{15.99} & 1.15 $\to$ \textbf{10.77} & 0.48 $\to$ \textbf{4.14} & 0.25 $\to$ \textbf{0.45} \\
jp & 0.09 $\to$ \textbf{0.15} & 0.06 $\to$ \textbf{0.15} & 0.08 $\to$ \textbf{0.14} & 0.11 $\to$ \textbf{0.35} & 0.02 $\to$ \textbf{0.11} & \textbf{0.03} $\to$ 0.01 \\
ms & \textbf{1.01} $\to$ 0.63 & 0.61 $\to$ \textbf{3.93} & 0.50 $\to$ \textbf{19.14} & 1.54 $\to$ \textbf{10.59} & 0.49 $\to$ \textbf{1.18} & 0.31 $\to$ \textbf{0.39} \\
\bottomrule
\end{tabular}
}
\caption{BLEU score changes of self-language intervention from \texttt{en} to targets listed on FLORES-200 dataset. The translation prompts are \textit{non-targeted}.}
\label{tab:bleu}
\end{table*}
% \documentclass{article}
% \usepackage{booktabs}
% \usepackage{tabularx}
% \usepackage{geometry}
% \usepackage{adjustbox}
% \geometry{margin=1in}
% =
\begin{table*}[htbp]
\centering
\small
\resizebox{0.6\textwidth}{!}{%
\begin{tabular}{llllll}
\toprule
\textbf{Lang} & \textbf{SeaLLMs3-1.5B} & \textbf{Qwen2.5-0.5B} & \textbf{Gemma2-2B} \\
\midrule
et & \textbf{2.86} $\to$ 0.35 & \textbf{1.46} $\to$ 0.18 & \textbf{0.45} $\to$ 0.01 \\
fr & \textbf{23.58} $\to$ 0.16 & \textbf{14.02} $\to$ 0.13 & \textbf{0.76} $\to$ 0.15 \\
ht & \textbf{1.28} $\to$ 0.12 & \textbf{0.82} $\to$ 0.09 & \textbf{0.42} $\to$ 0.13 \\
id & \textbf{22.06} $\to$ 0.14 & \textbf{7.93} $\to$ 0.03 & \textbf{0.55} $\to$ 0.14 \\
it & \textbf{12.44} $\to$ 0.18 & \textbf{6.51} $\to$ 0.09 & \textbf{0.48} $\to$ 0.09 \\
jp & \textbf{0.13} $\to$ 0.00 & \textbf{0.07} $\to$ 0.01 & \textbf{0.02} $\to$ 0.00 \\
ms & \textbf{17.05} $\to$ 0.13 & \textbf{5.04} $\to$ 0.08 & \textbf{0.49} $\to$ 0.14 \\
nl & \textbf{10.72} $\to$ 0.23 & \textbf{4.77} $\to$ 0.12 & \textbf{0.45} $\to$ 0.12 \\
pt & \textbf{21.99} $\to$ 0.24 & \textbf{13.76} $\to$ 0.07 & \textbf{0.61} $\to$ 0.14 \\
qu & \textbf{1.18} $\to$ 0.01 & \textbf{0.39} $\to$ 0.05 & \textbf{0.44} $\to$ 0.06 \\
ru & \textbf{13.48} $\to$ 0.32 & \textbf{6.96} $\to$ 0.09 & \textbf{0.38} $\to$ 0.17 \\
sw & \textbf{1.67} $\to$ 0.14 & \textbf{0.61} $\to$ 0.00 & \textbf{0.45} $\to$ 0.06 \\
ta & \textbf{0.33} $\to$ 0.00 & \textbf{0.12} $\to$ 0.00 & \textbf{0.40} $\to$ 0.02 \\
th & \textbf{3.59} $\to$ 0.00 & \textbf{1.54} $\to$ 0.03 & \textbf{0.17} $\to$ 0.00 \\
tr & \textbf{4.56} $\to$ 0.15 & \textbf{2.21} $\to$ 0.05 & \textbf{0.45} $\to$ 0.14 \\
vi & \textbf{22.37} $\to$ 1.42 & \textbf{12.74} $\to$ 0.06 & \textbf{0.33} $\to$ 0.14 \\
zh & \textbf{0.66} $\to$ 0.00 & \textbf{0.12} $\to$ 0.09 & \textbf{0.09} $\to$ 0.02 \\
\bottomrule
\end{tabular}
}
\caption{BLEU score changes of self-language \steer{pmax} intervention on Baseline neurons from \texttt{en} to target listed on FLORES-200 dataset. The translation prompts are \textit{targeted}.}
\label{tab:bleu-baseline}
\end{table*}

\begin{table}[htbp]
    \centering
    \setlength{\tabcolsep}{4pt}
    \begin{tabular}{lcc|cc|cc}
\toprule
\textbf{Lang} & \multicolumn{2}{c|}{\textbf{Gemma2}} & \multicolumn{2}{c|}{\textbf{SeaLLMs3}} & \multicolumn{2}{c}{\textbf{Qwen2.5}} \\
 & 9B & 2B & 7B & 1.5B & 7B & 0.5B \\
\midrule
\texttt{en} & 70.58 & 68.34 & {86.88} & 82.67 & 85.51 & 75.96 \\
\texttt{fr} & 65.06 & 56.63 & 66.27 & {77.11} & 72.29 & 61.45 \\
\texttt{jp} & 66.11 & 59.02 & 68.82 & 64.44 & {72.99} & 57.14 \\
\texttt{pt} & 66.54 & 61.60 & 77.57 & 68.44 & {79.85} & 62.36 \\
\texttt{ru} & 71.11 & 65.08 & {73.02} & 67.62 & 70.16 & 58.10 \\
\texttt{zh} & 70.24 & 66.27 & 80.95 & 76.39 & {81.75} & 64.88 \\
\bottomrule
\end{tabular}
    \caption{Baseline performance of XWinograd across models.}
    \label{tab:baseline-xwinograd}
\end{table}

\begin{table}[htbp]
    \centering
    \setlength{\tabcolsep}{4pt}
    \begin{tabular}{lcc|cc|cc}
        \toprule
        \textbf{Lang} & \multicolumn{2}{c|}{\textbf{Gemma2}} & \multicolumn{2}{c|}{\textbf{SeaLLMsv3}} & \multicolumn{2}{c}{\textbf{Qwen2.5}} \\
                     & 9B & 2B & 7B & 1.5B & 7B & 0.5B \\
\midrule
et  & 57.8 & 49.4 & 50.8 & 51.2 & 50.4 & 48.8 \\
ht  & 52.4 & 46.8 & 50.4 & 51.4 & 51.8 & 51.2 \\
id  & 69.4 & 62.0 & 73.6 & 65.4 & 73.4 & 57.2 \\
it  & 69.6 & 63.6 & 72.0 & 62.0 & 76.6 & 52.2 \\
qu  & 51.0 & 50.2 & 50.0 & 51.2 & 49.0 & 50.4 \\
sw  & 61.8 & 54.0 & 53.4 & 53.4 & 53.4 & 54.4 \\
ta  & 57.8 & 54.2 & 57.6 & 53.2 & 56.2 & 56.6 \\
th  & 56.8 & 53.0 & 59.0 & 57.4 & 61.6 & 55.6 \\
tr  & 64.2 & 54.4 & 61.2 & 54.8 & 61.4 & 53.6 \\
vi  & 67.6 & 59.2 & 74.4 & 65.8 & {78.0} & 58.2 \\
zh  & 72.6 & 67.2 & 78.0 & 70.0 & {79.8} & 63.6 \\
\bottomrule
\end{tabular}
    \caption{Baseline performance of XCOPA across models.}
    \label{tab:xcopa-baseline}
\end{table}

\begin{table}[htbp]
    \centering
    \setlength{\tabcolsep}{4pt}
    \begin{tabular}{lcc|cc|cc}
        \toprule
        \textbf{Lang} & \multicolumn{2}{c|}{\textbf{Gemma2}} & \multicolumn{2}{c|}{\textbf{SeaLLMsv3}} & \multicolumn{2}{c}{\textbf{Qwen2.5}} \\
                     & 9B & 2B & 7B & 1.5B & 7B & 0.5B \\
\midrule
et  & 69.4 & 33.3 & 38.8 & 23.5 & 42.1 & 32.8 \\
fr  & 66.4 & 47.6 & 53.6 & 40.0 & 57.6 & 41.6 \\
id  & 65.3 & 51.6 & 57.3 & 50.4 & 59.7 & 41.1 \\
it  & 70.9 & 54.6 & 61.8 & 51.8 & 72.9 & 41.4 \\
jp  & 71.9 & 53.8 & 63.9 & 55.0 & 69.1 & 48.2 \\
ms  & 59.4 & 44.6 & 49.0 & 38.6 & 53.0 & 33.7 \\
nl  & 72.1 & 54.2 & 60.6 & 44.6 & 64.1 & 32.7 \\
pt  & 63.6 & 47.8 & 56.9 & 42.3 & 61.7 & 38.3 \\
ru  & 55.2 & 45.6 & 52.8 & 42.5 & 55.6 & 40.1 \\
ta  & 45.6 & 32.8 & 30.0 & 33.6 & 32.4 & 28.0 \\
tr  & 57.0 & 39.8 & 45.8 & 31.7 & 51.0 & 30.5 \\
vi  & 62.4 & 46.0 & 58.0 & 43.6 & 56.8 & 40.8 \\
zh  & 55.9 & 45.7 & 73.5 & 73.1 & 77.9 & 55.5 \\
\bottomrule
\end{tabular}

    \caption{Baseline performance of Include-lite across models.}
    \label{tab:include-baseline}
\end{table}

\begin{table}[htbp]
\centering
\begin{tabular}{l c}
\toprule
\textbf{Factor} & \textbf{Mean LSS} \\
\midrule
\steer{pmax} & 0.9204 \\
\steer{pmedian} & 0.9095 \\
\steer{=max} & 0.7415 \\
\steer{+max} & 0.7258 \\
\steer{=0} & 0.4109 \\
\steer{=10p} & 0.1640 \\
\bottomrule
\end{tabular}
\caption{Mean LSS throughout all languages and models for each steering factor.}
\label{tab:mean-scores}
\end{table}
\label{sec:appendx}

\begin{table}
    \centering
    \setlength{\tabcolsep}{4pt}
\begin{tabular}{lcccccc}
\toprule
\textbf{Lang} & \multicolumn{2}{c}{\textbf{Gemma2}} & \multicolumn{2}{c}{\textbf{SeaLLMs3}} & \multicolumn{2}{c}{\textbf{Qwen2.5}} \\
             & 9B & 2B & 7B & 1.5B & 7B & 0.5B \\
\midrule
en  & 9   & 65   & 49   & 97   & 30   & 114  \\
nl  & 61  & 223  & 352  & 672  & 126  & 238  \\
id  & 43  & 126  & 108  & 382  & 56   & 137  \\
ms  & 48  & 178  & 131  & 377  & 71   & 203  \\
vi  & 52  & 214  & 216  & 435  & 86   & 224  \\
jp  & 139 & 495  & 351  & 690  & 309  & 573  \\
zh  & 72  & 362  & 299  & 887  & 196  & 436  \\
fr  & 42  & 204  & 343  & 897  & 109  & 220  \\
pt  & 33  & 189  & 354  & 633  & 75   & 195  \\
ru  & 64  & 262  & 226  & 444  & 135  & 256  \\
et  & 156 & 638  & 176  & 504  & 355  & 806  \\
ht  & 71  & 968  & 73   & 383  & 210  & 832  \\
it  & 45  & 185  & 327  & 637  & 107  & 203  \\
qu  & 172 & 1644 & 110  & 434  & 333  & 1426 \\
sw  & 177 & 1231 & 184  & 551  & 435  & 1297 \\
ta  & 550 & 1127 & 325  & 457  & 1085 & 1053 \\
th  & 155 & 311  & 267  & 507  & 164  & 292  \\
tr  & 132 & 286  & 159  & 668  & 219  & 355  \\
\bottomrule
\end{tabular}

    \caption{Proportions of LAPE neurons obtained from FLORES-200 dataset.}
    \label{tab:prop-lape}
\end{table}
\begin{table}
    \centering
    \setlength{\tabcolsep}{4pt}
\begin{tabular}{lcccccc}
\toprule
\textbf{Lang} & \multicolumn{2}{c}{\textbf{Gemma2}} & \multicolumn{2}{c}{\textbf{SeaLLMs3}} & \multicolumn{2}{c}{\textbf{Qwen2.5}} \\
             & 9B & 2B & 7B & 1.5B & 7B & 0.5B \\
\midrule
en  & 6   & 3995  & 8450  & 13838 & 2759  & 3653  \\
nl  & 32  & 7051  & 10085 & 17462 & 5214  & 6570  \\
id  & 22  & 6730  & 9724  & 16126 & 4864  & 6318  \\
ms  & 27  & 7393  & 10188 & 17815 & 5324  & 6847  \\
vi  & 19  & 6271  & 10828 & 20036 & 4303  & 5878  \\
jp  & 62  & 7119  & 11032 & 19553 & 4996  & 6624  \\
zh  & 38  & 4268  & 10246 & 19029 & 2935  & 3799  \\
fr  & 29  & 6592  & 10608 & 17814 & 4683  & 6153  \\
pt  & 21  & 6407  & 10692 & 17069 & 4547  & 6036  \\
ru  & 32  & 5903  & 10554 & 17496 & 4231  & 5420  \\
et  & 58  & 12607 & 11542 & 19546 & 7718  & 11544 \\
ht  & 16  & 12497 & 11896 & 21920 & 6583  & 11363 \\
it  & 28  & 7173  & 10945 & 17064 & 5352  & 6615  \\
qu  & 35  & 14061 & 13129 & 25239 & 7579  & 12479 \\
sw  & 47  & 14225 & 11856 & 19813 & 8511  & 13050 \\
ta  & 120 & 21817 & 14465 & 27403 & 19088 & 19088 \\
th  & 33  & 8352  & 12832 & 24646 & 5573  & 7692  \\
tr  & 68  & 9139  & 10740 & 19227 & 6905  & 8502  \\
\bottomrule
\end{tabular}

    \caption{Proportions of baseline neurons obtained from FLORES-200 dataset.}
    \label{tab:prop-raw}
\end{table}

\FloatBarrier

% % XWinograd

\begin{figure}
    \centering
    \includegraphics[width=\linewidth]{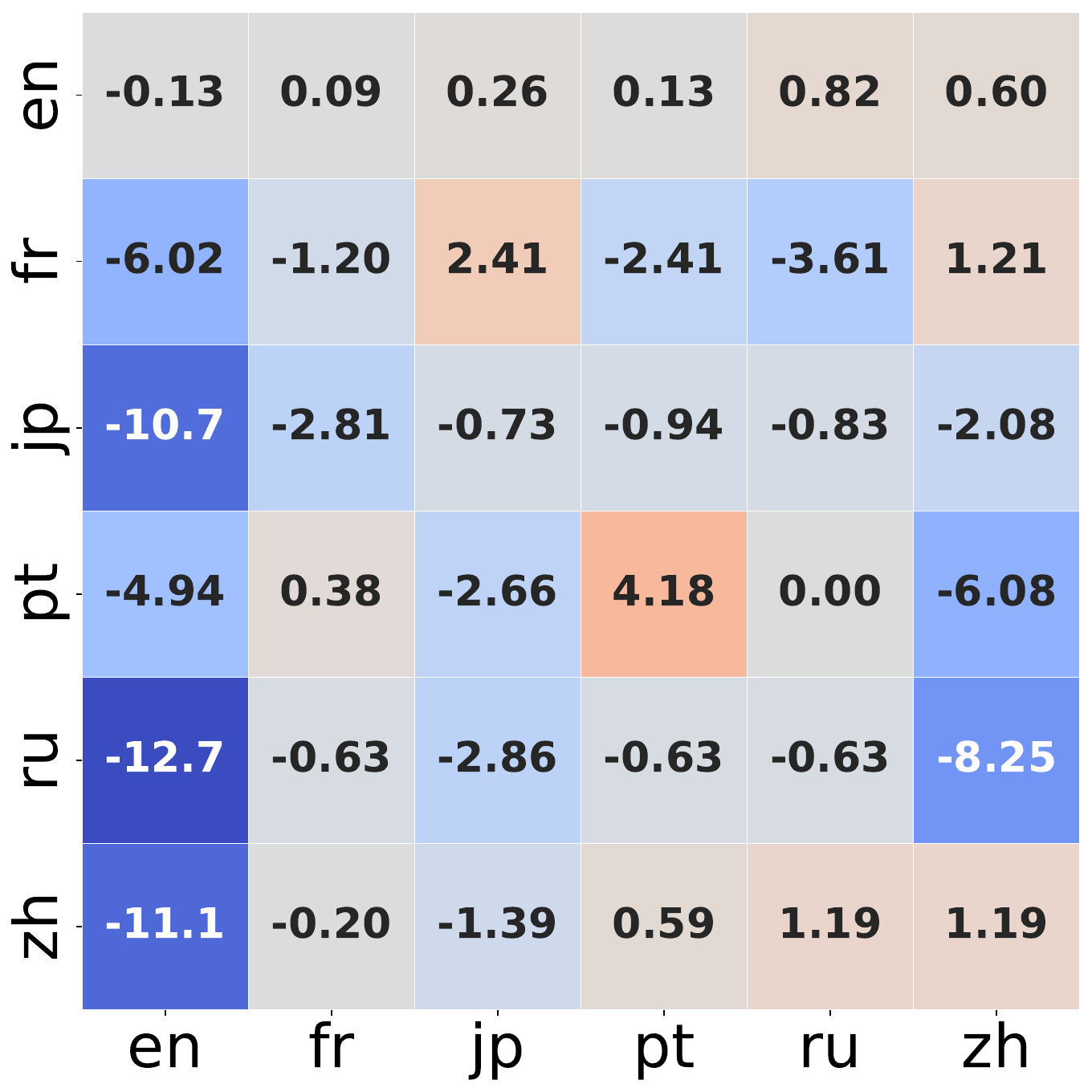}
    \caption{Delta XWinograd accuracy after steering LAPE neurons with \steer{pmedian} for Gemma2 9B. Row \textit{i} is the initial language, column \textit{j} is the language of intervention.}
    \label{fig:lape-xwin-med-gemma}
\end{figure}

\begin{figure}
    \centering
    \includegraphics[width=\linewidth]{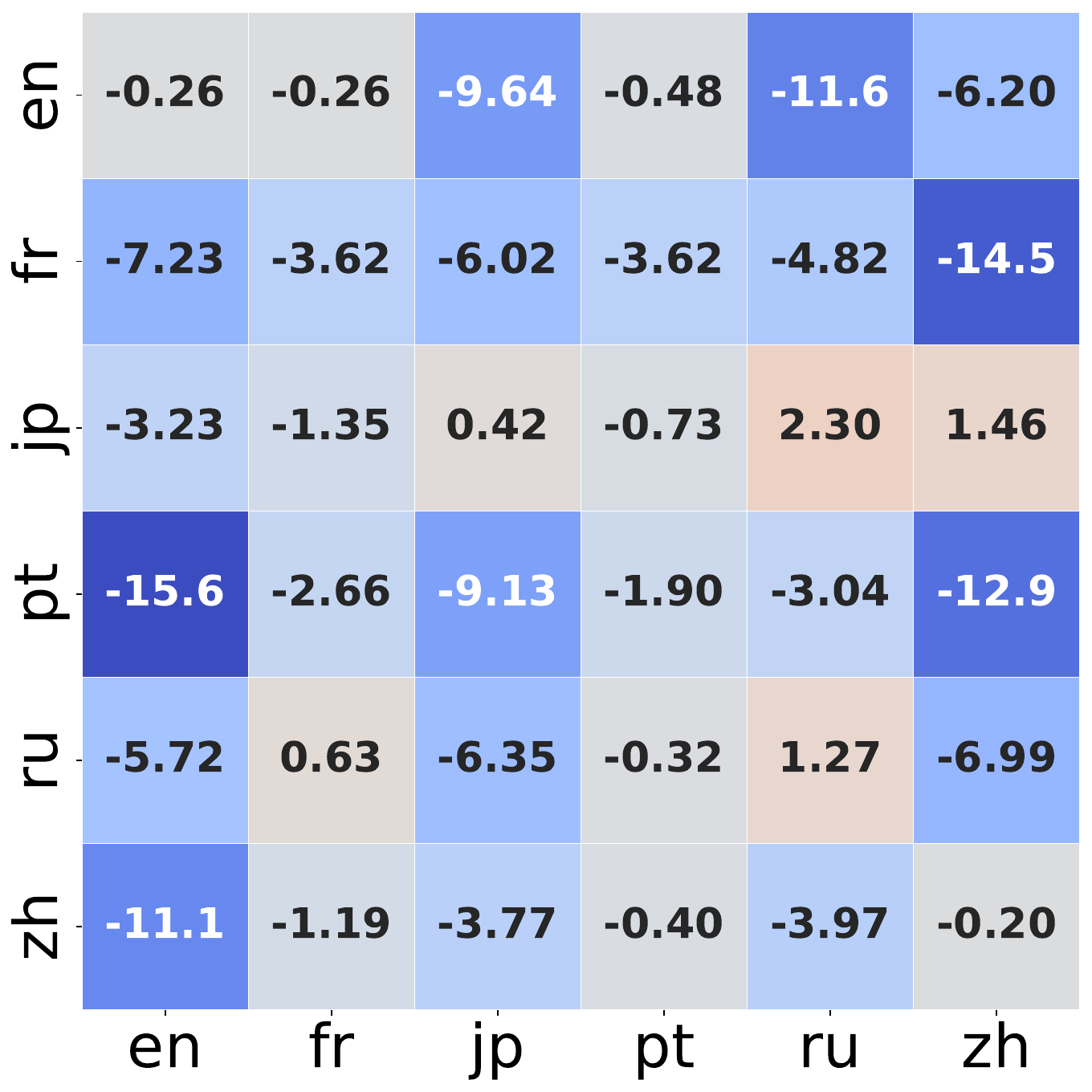}
    \caption{Delta XWinograd accuracy after steering LAPE neurons with \steer{pmedian} for Qwen2.5 7B. Row \textit{i} is the initial language, column \textit{j} is the language of intervention.}
    \label{fig:lape-xwin-med-qwen}
\end{figure}

\FloatBarrier

% XCOPA
\begin{figure}[t]
\centering
    \includegraphics[width=\linewidth]{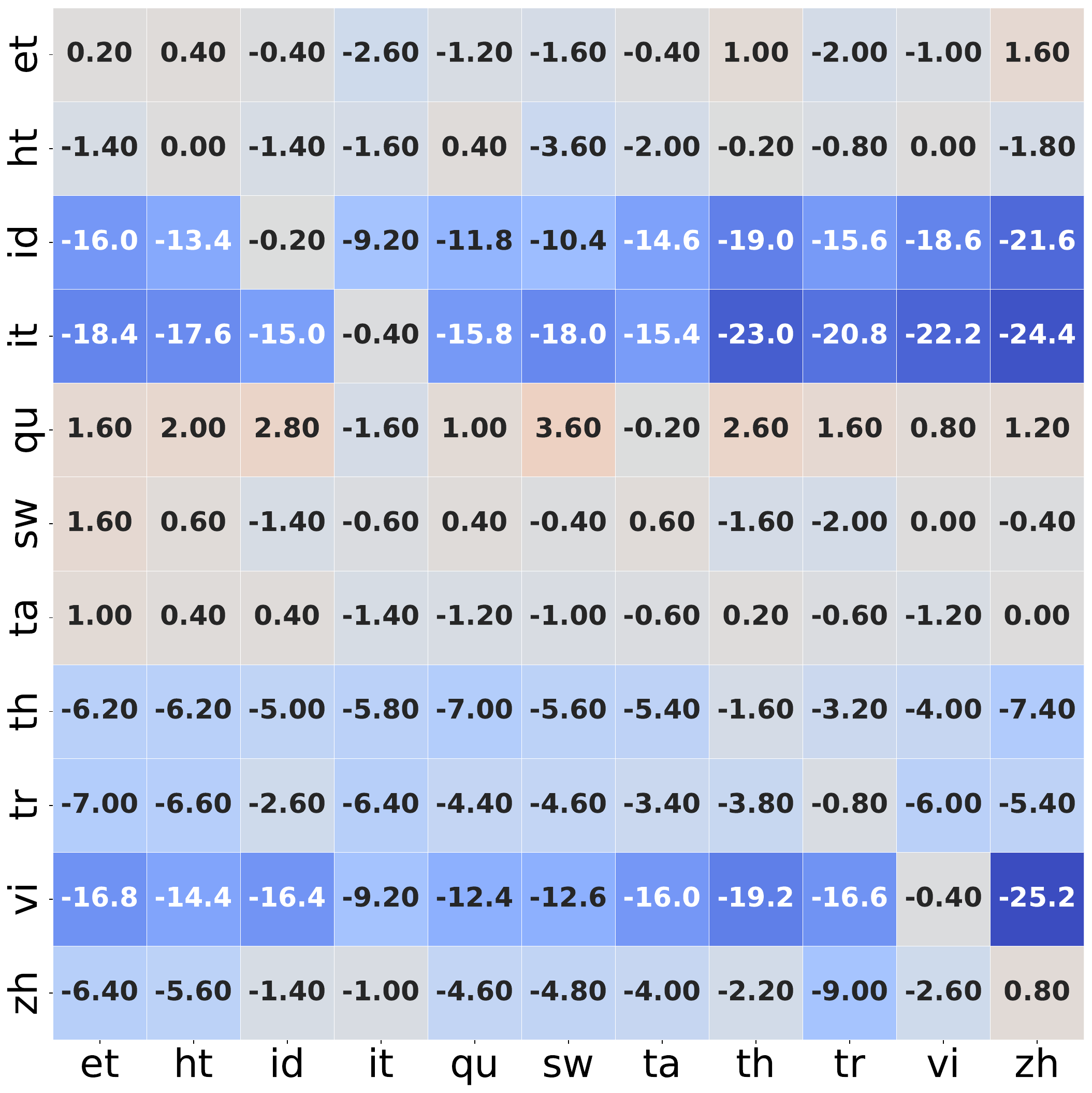}
    \caption{Delta XCOPA accuracy after steering LAPE neurons with \steer{pmax} for Qwen2.5 7B.}
    \label{fig:lape-xcopa-qwen}

\end{figure}

\begin{figure}[t]
\centering
    \includegraphics[width=\linewidth]{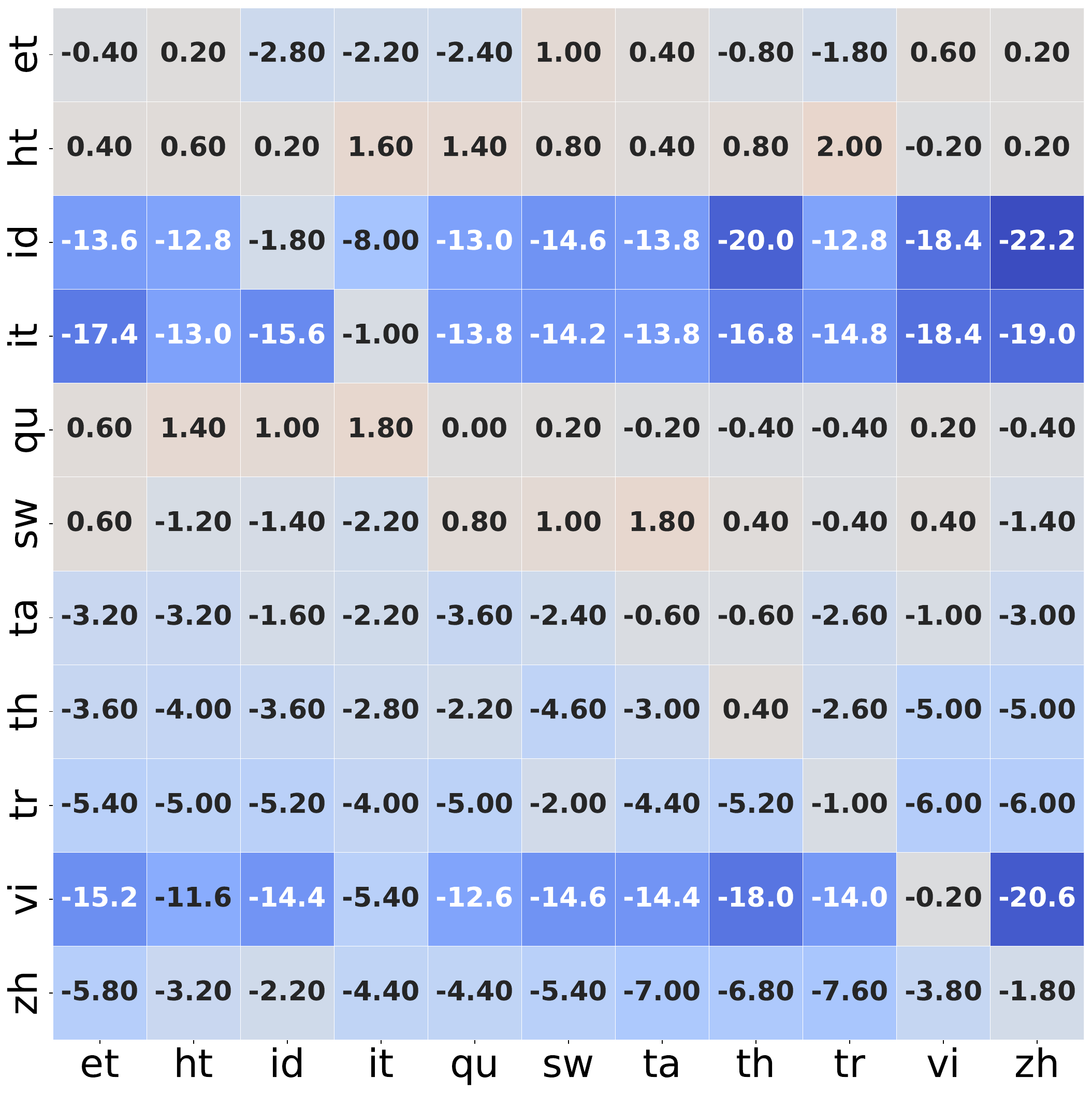}
    \caption{Delta XCOPA accuracy after steering LAPE neurons with \steer{pmax} for SeaLLM3 7B.}
    \label{fig:lape-xcopa-sea}
\end{figure}


\FloatBarrier

% INCLUDE

\FloatBarrier

% % PPL FLORES
% % PMAX


% PMEDIAN

% PPL FLORES BASELINE

% per layer amplification
% \input{figs/perlayer}
% \FloatBarrier

\FloatBarrier

\end{document}